\pdfoutput=1
\documentclass[runningheads]{llncs}

\usepackage{graphicx}
\usepackage{amsmath}
\usepackage{amssymb}
\usepackage{color}
\DeclareMathOperator{\sign}{sign}
\usepackage[binary-units=true]{siunitx}
\DeclareSIUnit\pixel{P}
\usepackage[super]{nth}

\usepackage{listings}
\usepackage{comment}

\usepackage[nolist]{acronym}
\usepackage{csquotes}
\usepackage{subcaption}
\usepackage{color,soul}
\usepackage[edges]{forest}
\usepackage{blindtext}
\usepackage{wrapfig}
\usepackage{mwe}
\usepackage{nicefrac}
\usepackage{bm}
\usepackage{mathtools}
\usepackage{latexsym}
\usepackage{tablefootnote}
\usepackage{threeparttable}
\usepackage{adjustbox}
\usepackage{enumitem} 
\usepackage{pifont}
\usepackage{xfrac}
\usepackage{tabularx}
\usepackage{multirow}
\usepackage{array}
\usepackage{booktabs}
\usepackage{todonotes}
\usepackage{import}
\usepackage{placeins}

\usepackage{layouts}
\usetikzlibrary{shadows.blur}
\usetikzlibrary{positioning}
\usetikzlibrary{arrows.meta}
\usetikzlibrary{positioning}
\usetikzlibrary{spy}
\usepackage[skins]{tcolorbox}
\usepackage{pgfplots}
\DeclareUnicodeCharacter{2212}{−}
\usepgfplotslibrary{groupplots,dateplot}
\usetikzlibrary{patterns,shapes.arrows}
\pgfplotsset{compat=newest}

\usepackage{footnote}
\makesavenoteenv{tabular}

\makeatletter
\newcommand{\printfnsymbol}[1]{%
  \textsuperscript{\@fnsymbol{#1}}%
}
\makeatother

\usepackage[pagebackref,breaklinks,colorlinks]{hyperref}
\makeatletter
\DeclareRobustCommand\onedot{\futurelet\@let@token\@onedot}
\def\@onedot{\ifx\@let@token.\else.\null\fi\xspace}
\def\etal{\textit{et~al}\onedot}
\def\eg{e.g\onedot, } 
\def\ie{i.e\onedot, }

\def\vs{vs\onedot}
\makeatother

\def\clap#1{\hbox to 0pt{\hss #1\hss}}%
\def\initials#1{\protect\clap{\smash{\raisebox{1.4ex}{\tiny{\textsf{\textit{#1}}}}}}}%
\makeatletter
\newcommand{\EDIT}[4][]{\protect\@ifundefined{hidecomments}{%
  \strut{\color{#3}{\hspace{0pt}\initials{#2}\protect\sout{#1}{~#4}}}%
  }{}}
\newcommand{\NOTEboxed}[3]{\protect\@ifundefined{hidecomments}{%
  {\begin{center}\fbox{\parbox{0.97\linewidth}{\protect\EDIT{#1}{#2}{#3}}}\end{center}}
  }{}}
\makeatletter
\newcommand{\DefAuthor}[2] 
{%
  \expandafter\newcommand\csname #1edit\endcsname[2][]{\protect\EDIT[##1]{#1}{#2}{##2}}
  \expandafter\newcommand\csname #1\endcsname[1]{\protect\csname #1edit\endcsname{[##1]}}
  \expandafter\newcommand\csname #1boxed\endcsname[1]{\NOTEboxed{#1}{#2}{##1}}
}
\makeatother

\DeclareMathOperator*{\argmax}{arg\,max}

\def\numx#1e#2{{#1}\mathrm{e}{#2}}

\newcommand{\WISE}{\textit{WISE}\xspace}

\usepackage[capitalize]{cleveref}
\Crefname{figure}{Fig.}{Figs.}
\Crefname{section}{Sec.}{Secs.}
\Crefname{table}{Tab.}{Tabs.}
\Crefname{equation}{Eq.}{Eqs.}

\newlength\secmargin
\newlength\paramargin
\newlength\figmargin
\newlength\capmargin
 \newlength\Iwidth\newlength\Iheight\newsavebox\IBox
\savebox\IBox{\includegraphics{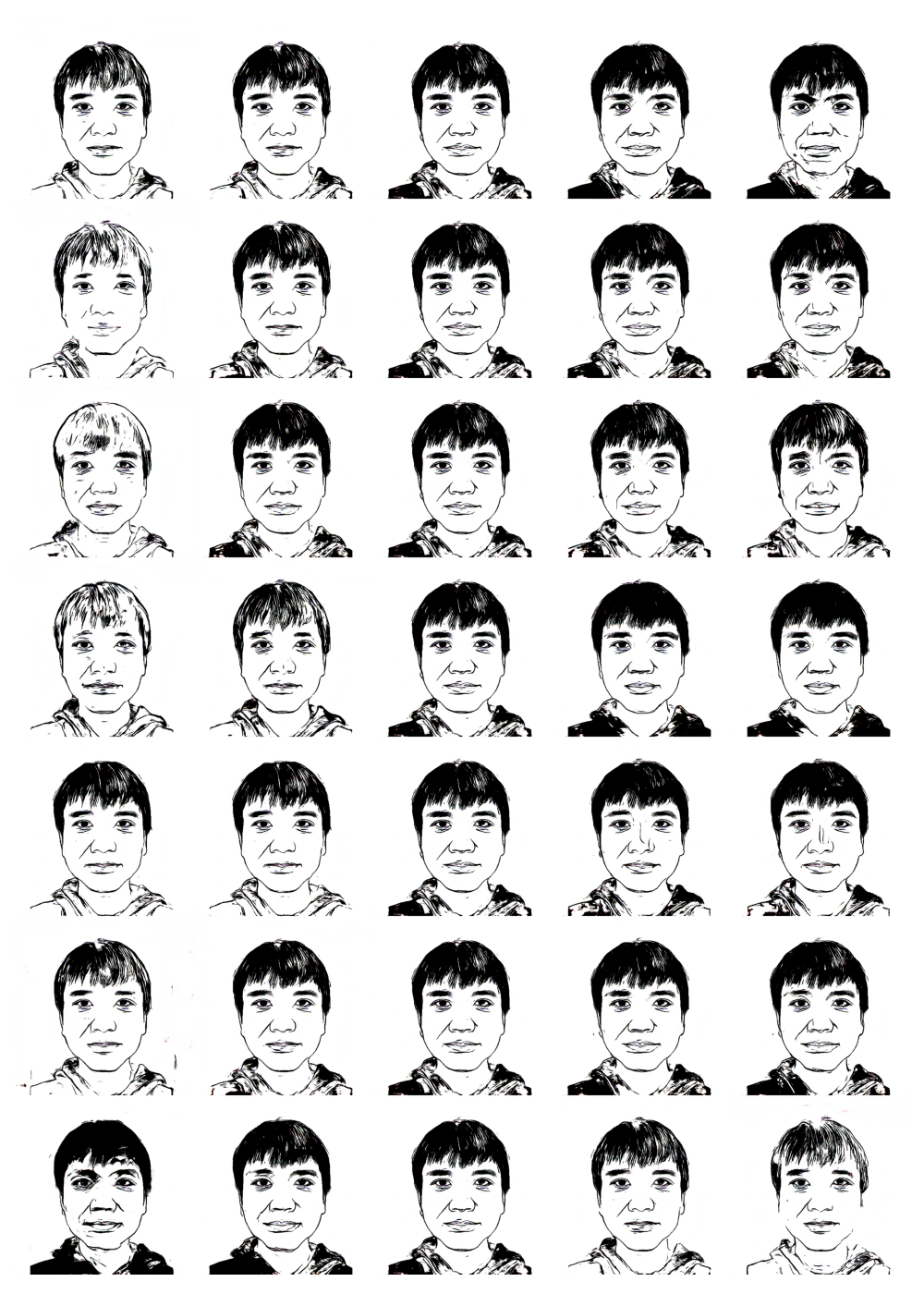}}
\newlength{\tempdima}
\newcommand{\rowname}[1]
{\rotatebox{90}{\makebox[\tempdima][c]{\text{#1}}}}

\usepackage{xr}

\begin{document}

\pagestyle{headings}
\mainmatter
\def\ECCVSubNumber{6063}  

\title{WISE: Whitebox Image Stylization by Example-based Learning}

\titlerunning{WISE: Whitebox Image Stylization by Example-based Learning}
\author{
Winfried Lötzsch\inst{1}\thanks{Both authors contributed equally to this work.}
\and
Max Reimann\inst{1}\printfnsymbol{1}
\and
Martin Büssemeyer\inst{1}
\and
Amir Semmo\inst{2}
\and 
Jürgen Döllner\inst{1}
\and
Matthias Trapp\inst{1}
}

\authorrunning{Lötzsch and Reimann et al.}
\institute{Hasso Plattner Institute, University of Potsdam, Germany  \and
Digital Masterpieces GmbH, Germany}

\maketitle
	
\acused{CPU}
\acused{FPS}
\acused{K}
\acused{M}
\acused{TFLOPS}
\acused{MB}

\newcommand{\aequal}[0]{$\pm$ 0\%}
\newcommand{\abest}[1]{\underline{#1}}

\begin{abstract}
Image-based artistic rendering can synthesize a variety of expressive styles using algorithmic image filtering.
In contrast to deep learning-based methods, these heuristics-based filtering techniques can operate on high-resolution images, are interpretable, and can be parameterized according to various design aspects.
However, adapting or extending these techniques to produce new styles is often a tedious and error-prone task that requires expert knowledge.
We propose a new paradigm to alleviate this problem: implementing algorithmic image filtering techniques as differentiable operations that can learn parametrizations aligned to certain reference styles.
To this end, we present WISE, an example-based image-processing system that can handle a multitude of stylization techniques, such as watercolor, oil or cartoon stylization, within a common framework.
By training parameter prediction networks for global and local filter parameterizations, we can simultaneously adapt effects to reference styles and image content, e.g., to enhance facial features.
Our method can be optimized in a style-transfer framework or learned in a generative-adversarial setting for image-to-image translation.
We demonstrate that jointly training an XDoG filter and a CNN for postprocessing can achieve comparable results to a state-of-the-art GAN-based \nolinebreak method. \href{https://github.com/winfried-loetzsch/wise}{https://github.com/winfried-loetzsch/wise}
\end{abstract}

\section{Introduction}
\label{sec:Introduction}
Image stylization has become a major part of visual communication, with millions of edited and stylized photos shared every day. 
At this, a large body of research in \ac{NPR} has been dedicated to imitating hand-drawn artistic styles \cite{kyprianidis2012state,rosin2012image}. 
Traditionally, such \textit{heuristics-based algorithms} \cite{semmo2017neural} for image-based artistic rendering emulate a certain artistic style using a series of specifically developed \textit{algorithmic} image processing operations. 
Thus, creating new styles is often a time-consuming process that requires the knowledge of domain experts.

Recently, deep learning-based techniques for stylization and image-to-image translation have gained popularity by enabling the learning of stylistic abstractions from example data. 
In particular, \ac{NST} \cite{johnson2016perceptual,jing2019neural} that transfers the artistic style of a reference image 
and \ac{GAN}-based \cite{goodfellow2014generative,isola2017image} methods for fitting style distributions
have achieved impressive results and are increasingly used in commercial applications \cite{Photoshopfilters}.

\begin{figure}[t]
\centering
\begin{subfigure}{0.329\linewidth}%
\includegraphics[width=\linewidth]{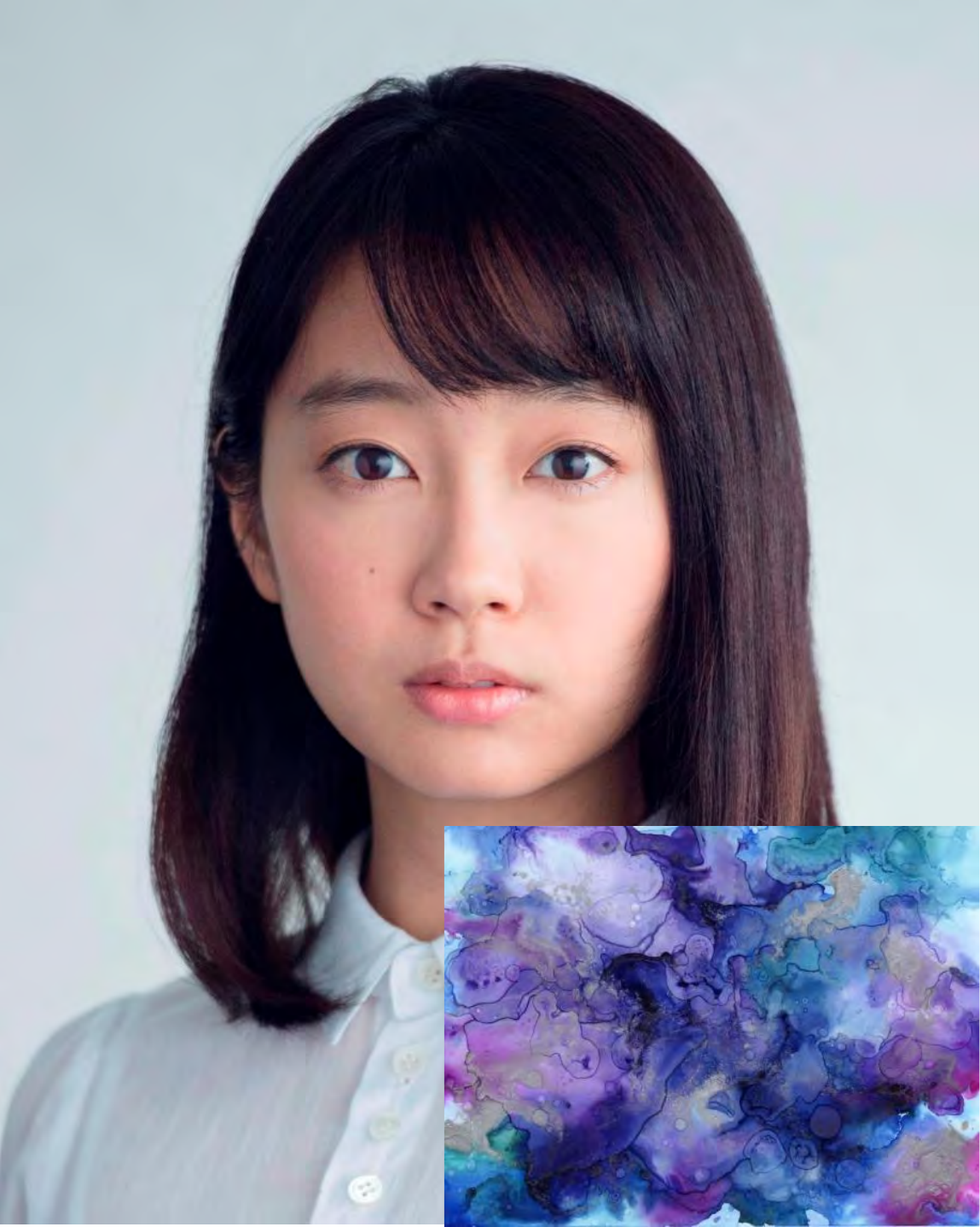}%
\end{subfigure}\hfill
\begin{subfigure}{0.329\linewidth}%
\includegraphics[width=\linewidth]{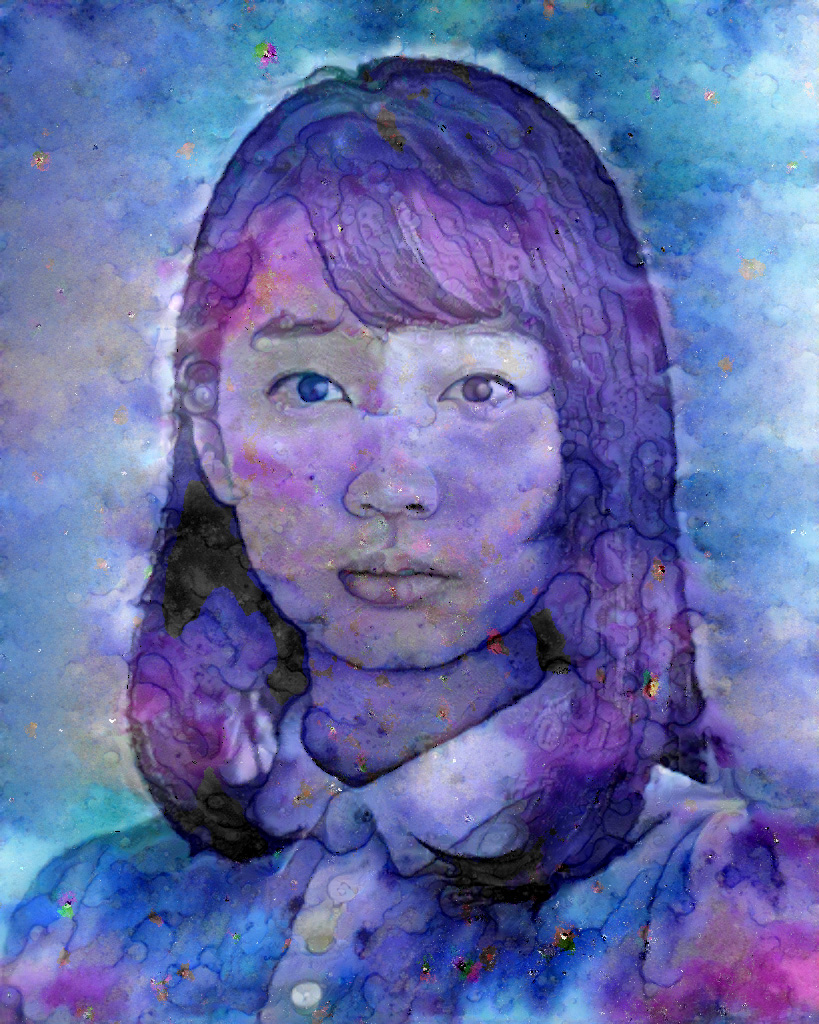}%
\end{subfigure}\hfill
\begin{subfigure}{0.329\linewidth}%
\includegraphics[width=\linewidth]{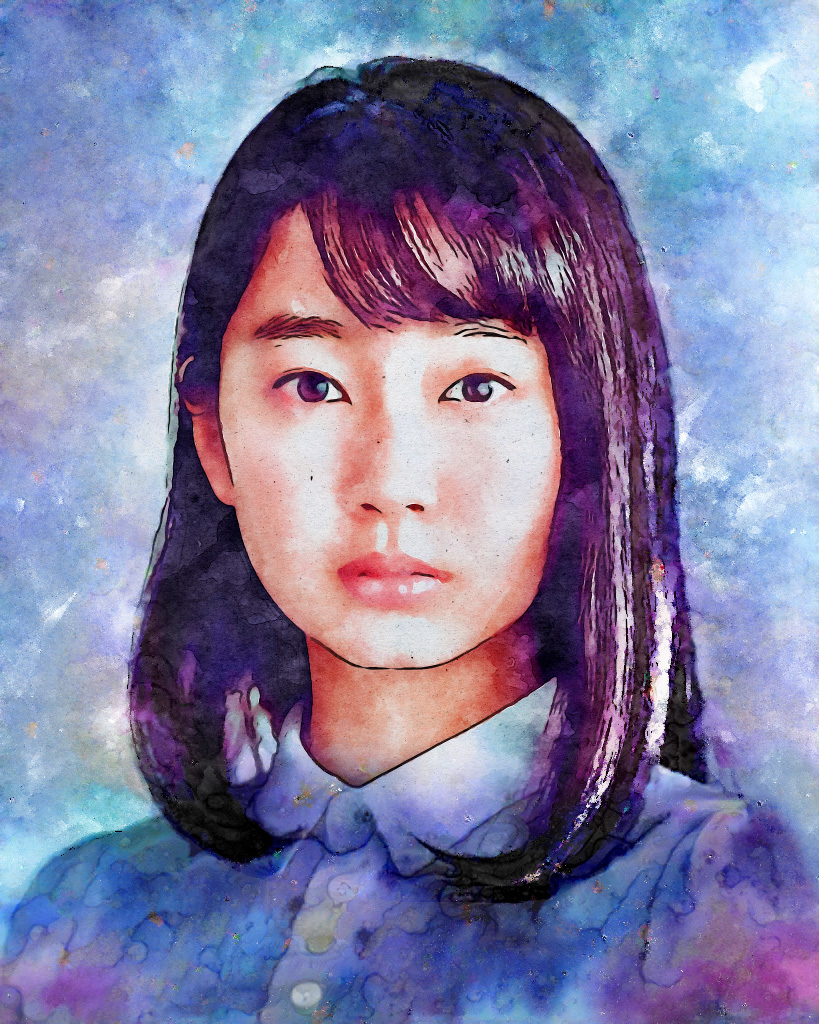}%
\end{subfigure}%

\begin{subfigure}{0.329\linewidth}%
\includegraphics[width=\linewidth]{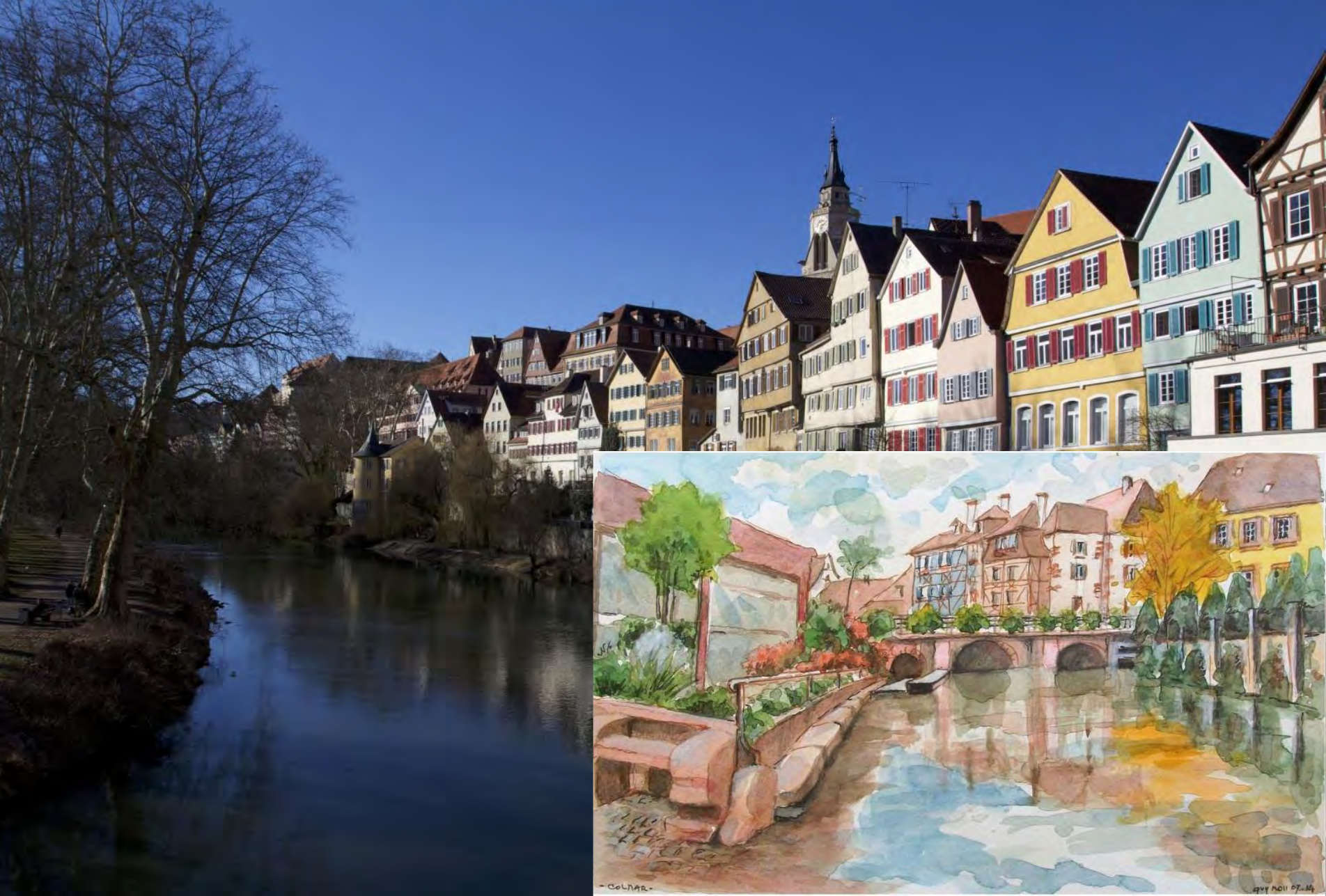}%
\caption{Content/Style}
\label{fig:teaser:content_style}%
\end{subfigure}\hfill
\begin{subfigure}{0.329\linewidth}%
\includegraphics[width=\linewidth]{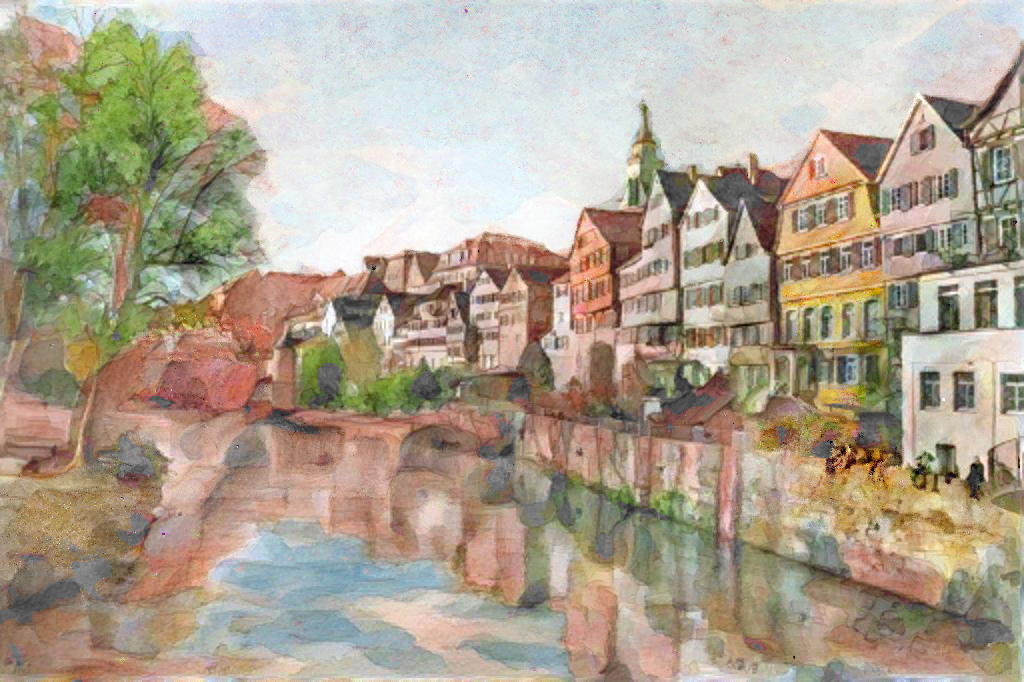}%
\caption{\WISE~/ without edits}%
\label{fig:teaser:result}%
\end{subfigure}\hfill
\begin{subfigure}{0.329\linewidth}%
\includegraphics[width=\linewidth]{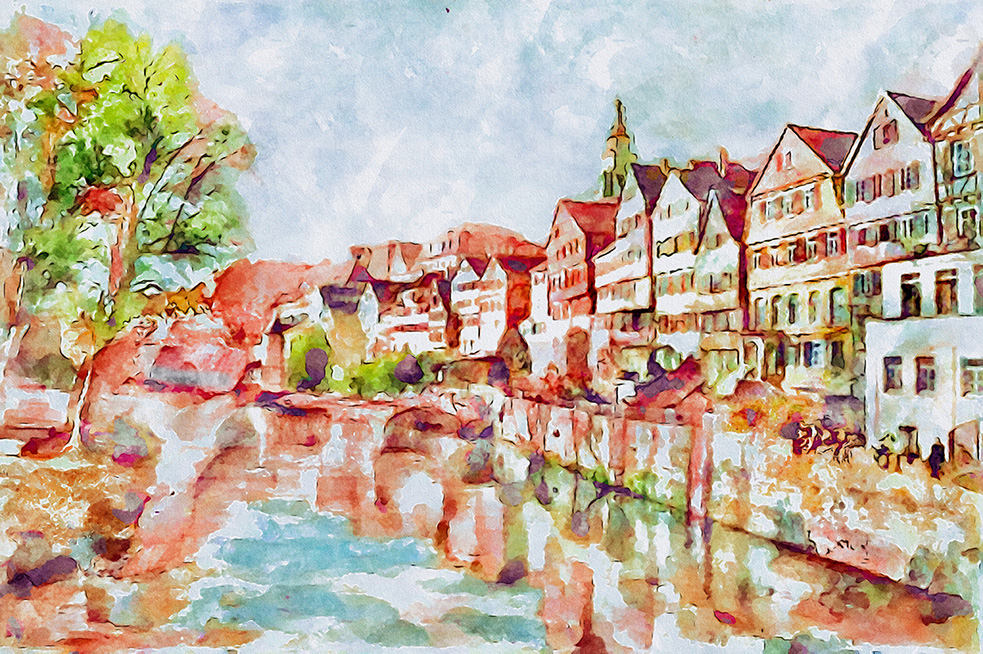}%
\caption{\WISE~/ with local edits}%
\label{fig:teaser:adjusted}%
\end{subfigure}%
\caption{\textbf{Example-based effect adjustment.} \WISE optimizes effect parameters, \eg of a watercolor stylization effect, to match stylized outputs to a reference style \protect\subref{fig:teaser:content_style}. The results  \protect\subref{fig:teaser:result} can then be interactively adjusted by tuning the obtained parameters globally and locally for increased artistic control \protect\subref{fig:teaser:adjusted}\protect\footnotemark. 
}
\label{fig:teaseer}
\end{figure}

\footnotetext{View examples of editing in our supplemental video: \href{https://youtu.be/wIndN7cr0PE}{https://youtu.be/wIndN7cr0PE}}

Classical heuristics-based filters and filter-based image stylization pipelines, such as the \ac{XDoG} filter~\cite{winnemoller2011xdog}, cartoon effect~\cite{winnemoller2006real}, or watercolor effect~\cite{bousseau2006interactive,wang2014towards}, expose a range of parameters to the user that enable fine-grained global and local control 
over artistic aspects of the stylized output.
By contrast, learning-based techniques are commonly limited in their modes of control, \ie  \ac{NST} \cite{gatys2016image} only offers control over a general content-style tradeoff. 
Furthermore, their learned representations are generally not interpretable as a set of design aspects and configurations. 
Thus, these approaches often do not meet the requirements of interactive image editing tasks that go beyond one-shot global stylizations towards editing with high-level and low-level artistic control  
~\cite{isenberg2016tools,hertzmann2020computers,Gooch2010npr}. 
Additionally, deep network-based methods are often computationally expensive in both training and inference on high image resolutions \cite{gatys2016image,karras2018progressive,karras2019style}. 
This further limits their applicability in interactive or mobile applications~\cite{gharbi2017deep} and their capability to simulate fine-grained (pigment-based) local effects and phenomena of artistic media such as watercolor and oil paint.

\begin{figure}[!t]
\centering
\includegraphics[width=\linewidth, trim={4.5cm 4.75cm 4cm 6cm},clip]{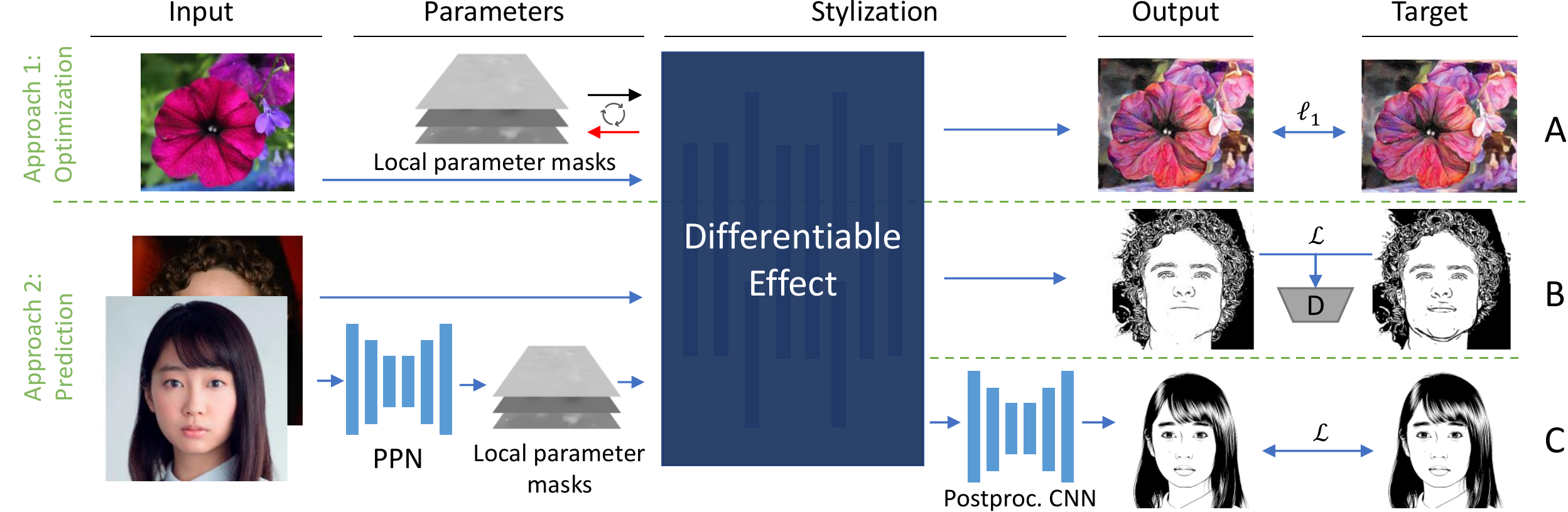}
\caption{\textbf{Overview of \WISE}. Differentiable effects can be adapted to example data, demonstrated for three use-cases: \textit{parametric style transfer} (\textbf{A}) optimizes parameter masks to match a hand-drawn or synthesized stylization target (\eg from \ac{NST} as in \Cref{fig:teaseer}) which enables style transfer results to remain editable and resolution-independent. \textit{Local parameter prediction} (\textbf{B}) trains \acsp{PPN} to predict parameter masks to adapt the effect to the content (\eg for facial structure enhancement as shown here). Combined with a postprocessing CNN, local parameter prediction can learn sophisticated \textit{image-to-image translation} tasks such as learning hand-drawn sketch-styles (\textbf{C}).
}
\label{fig:grad_sre}
\end{figure}

To counterbalance these limitations, this work aims to combine the strengths of \textit{heuristics-based} and \textit{learning-based} image stylization
by implementing algorithmic effects as \textit{differentiable} operations that can be trained to learn filter-based parameters aligning to certain reference styles. 
The goal is to enable (1) the creation of complex, example-based stylizations using lightweight algorithmic approaches that remain interpretable and can operate on very high image resolutions, and enable (2) the editing with artistic control on a fine-granular level according to design aspects.
To this end, we present \WISE, a \textit{whitebox} system for example-based image processing that can handle a multitude of stylization techniques in a common framework. 
Our system integrates existing algorithmic effects such as \ac{XDoG}-based stylization \cite{winnemoller2011xdog}, cartoon stylization \cite{winnemoller2006real}, watercolor effects \cite{bousseau2006interactive,wang2014towards}, and oilpaint effects \cite{semmo2016interactive},  by creating a library of differentiable image filters that match their shader kernel-based counterparts. We show that the majority of filters (\eg bilateral filtering) can be transformed into auto-differentiable formulations, while for the remaining filters, gradients can be approximated (\eg for color quantization).
Using our framework, effects can be adapted to reference styles using popular, deep network-based image-to-image translation losses. 
We train exemplary effects using both \ac{NST} and \ac{GAN}-based losses and show qualitatively and quantitatively that the results are comparable to state-of-the-art deep networks while retaining the advantages of filter-based stylization. 
To summarize, this paper contributes the following:
\begin{enumerate}
\item It provides an end-to-end framework for example-based image stylization using differentiable algorithmic filters. \Cref{fig:grad_sre} shows an overview of the system.
\item It demonstrates the applicability of style transfer-losses to adapt stylization effects to a reference style (\cref{fig:teaseer} and \Cref{fig:grad_sre} A). The results remain editable and resolution-independent. 
\item It shows that both global and local parametrizations of stylization effects can be optimized as well as predicted by a parameter prediction network (PPN). The latter can be trained on content-adaptive tasks (\Cref{fig:grad_sre} B).
\item It demonstrates that filters can be trained in combination with CNNs for improved generalization on image-to-image translation tasks. Combining the \ac{XDoG} effect with a simple post-processing \ac{CNN} (\Cref{fig:grad_sre} C) can achieve comparable results to state-of-the-art \ac{GAN}-based image stylization for hand-drawn sketch styles, but at much lower system complexity. 
\end{enumerate}

\section{Related Work}
\label{sec:RelatedWork}

\paragraph{Heuristics-based Stylization.}

In \ac{NPR}, image-based artistic rendering deals with emulating traditional artistic styles, using a pipeline of rendering stages \cite{kyprianidis2012state,rosin2012image,semmo2017neural}.
Commonly, edge detection and content abstraction are important parts of such pipelines.
The \ac{XDoG} filter \cite{winnemoller2011xdog,winnemoller2006real} is an extended version of the \ac{DoG} band-pass filter, and can be used to create smooth edge stylizations.
Furthermore, edge-aware smoothing filters such as bilateral filtering \cite{tomasi1998bilateral} or Kuwahara filtering \cite{kuwahara1976processing} can abstract image contents, and can be combined with image flow to adapt the results to local image structures \cite{kyprianidis2008image,kyprianidis2009image}. These techniques can be found in heuristics-based effects such as cartoon stylization~\cite{winnemoller2006real}, oil-paint abstraction \cite{semmo2016image} and image watercolorization \cite{bousseau2006interactive,wang2014towards},  each consisting of a series of rendering stages such as image blending, wobbling, pigment dispersion and wet-in-wet stylization.
For a comprehensive taxonomy of techniques, the interested reader is referred to the survey by Kyprianidis~\etal~\cite{kyprianidis2012state}.

These effects are typically parameterized globally, and can be further adjusted within pre-defined parameter ranges, or locally on a per-pixel level using parameter masks~\cite{semmo2016image}. 
In this work, we implement variants of the \ac{XDoG}, cartoon filtering, and watercolor pipeline in our framework using auto-differentiable formulations of each rendering stage. At runtime, users choose between one of these different effect pipelines; and results are generally achieved by optimizing the exact chain of filters as introduced in \cite{winnemoller2011xdog,winnemoller2006real,bousseau2006interactive,wang2014towards,semmo2016interactive}. 

\paragraph{Deep Learning-based Methods.}
With the advent of deep learning, \acp{CNN} for image generation and transformation have led to a range of impressive results. 

In \ac{NST}, first introduced by Gatys~\etal\cite{gatys2016image}, the stylistic characteristics of a reference image are transferred to a content image by matching deep feature statistics using an optimization process. 
Fast feed-forward networks have been trained to reproduce a single \cite{johnson2016perceptual} or even arbitrarily many styles \cite{huang2017arbitrary,gu2018arbitrary,park2019arbitrary}. Furthermore, there have been efforts to increase controllability of \acp{NST}, \eg by control over color \cite{GatysEBHS17}, sub-styles \cite{reimann2019locally}  or strokes \cite{Jing_2018_ECCV,reimann2022controlling}, however, the representations are not interpretable. Recently, style transfer has been formulated as a neurally-guided stroke rendering optimization approach \cite{kotovenko2021neuralstroke}, that retains interpretability, however, is slow to optimize. We show that our method can obtain comparable results to a state-of-the-art \ac{NST} \cite{kolkin2019style}, while retaining interactive editing control. 

\acp{GAN}, first introduced by Goodfellow~\etal\cite{goodfellow2014generative}, learn powerful generative networks that model the input distribution. They
have been widely used for conditional image generation tasks, with both paired \cite{isola2017image} and unpaired \cite{zhu2017unpaired} training data. In the stylization domain, it has found applications for collection style transfer \cite{chen2018gated,sanakoyeu2018styleaware}, cartoon generation \cite{chen2018cartoongan,su2020unpaired}, and sketch styles \cite{yi2019apdrawinggan}. However, domain-specific applications often require sophisticated losses and multiple networks to prevent artifacts \cite{chen2018cartoongan,YiXLLR20}. We show that our differentiable implementation of the \ac{XDoG} filter can be trained as a generator network in a \ac{GAN} framework and can produce comparable sketches to state-of-the-art (CNN-based) \acp{GAN}~\cite{yi2019apdrawinggan,YiXLLR20}.

\paragraph{Learnable Filters.}
While the previous end-to-end \acp{CNN} deliver impressive results, they are limited in their output resolution. 
A few recent methods have proposed training fast algorithmic filters to operate efficiently at high resolutions. Getreuer~\etal\cite{getreuer2018blade} introduce 
learnable 
approximations of algorithmic image filters, such as of the \ac{XDoG} filter. 
At run-time, a linear filter is selected per image pixel according to the local structure tensor; filters can be combined in pipelines for image enhancement \cite{garcia2020image}. 
Gharbi~\etal\cite{gharbi2017deep} train a \ac{CNN} to predict affine transformations for bilateral image enhancement, \eg to approximate edge-aware image filters or tone adjustments. 
The transform filters are predicted at a low resolution and then applied in full resolution to the image. 
\enquote{Exposure} framework \cite{hu2018exposure} combines learning linear image filters with reinforcement learning, where an actor-critic model decides which filters to include to achieve a desired photo enhancement effect. 

These methods have in common that they learn several simple, linear functions to approximate image processing operations \cite{getreuer2018blade,yan2016automatic,gharbi2017deep,hu2018exposure,moran2020deeplpf}.
Our framework consists of pipelines of differentiable filters as well. However, in contrast to previous work, we make a variety of heuristics-based stylization operators differentiable and learn to predict their parameterizations. 
Thereby, sophisticated stylization effects (\eg those found in stylization applications) can be ported and directly used in our framework.


\begin{figure}[t]
\includegraphics[width=\linewidth, trim={8cm 6.5cm 6cm 13.7cm},clip]{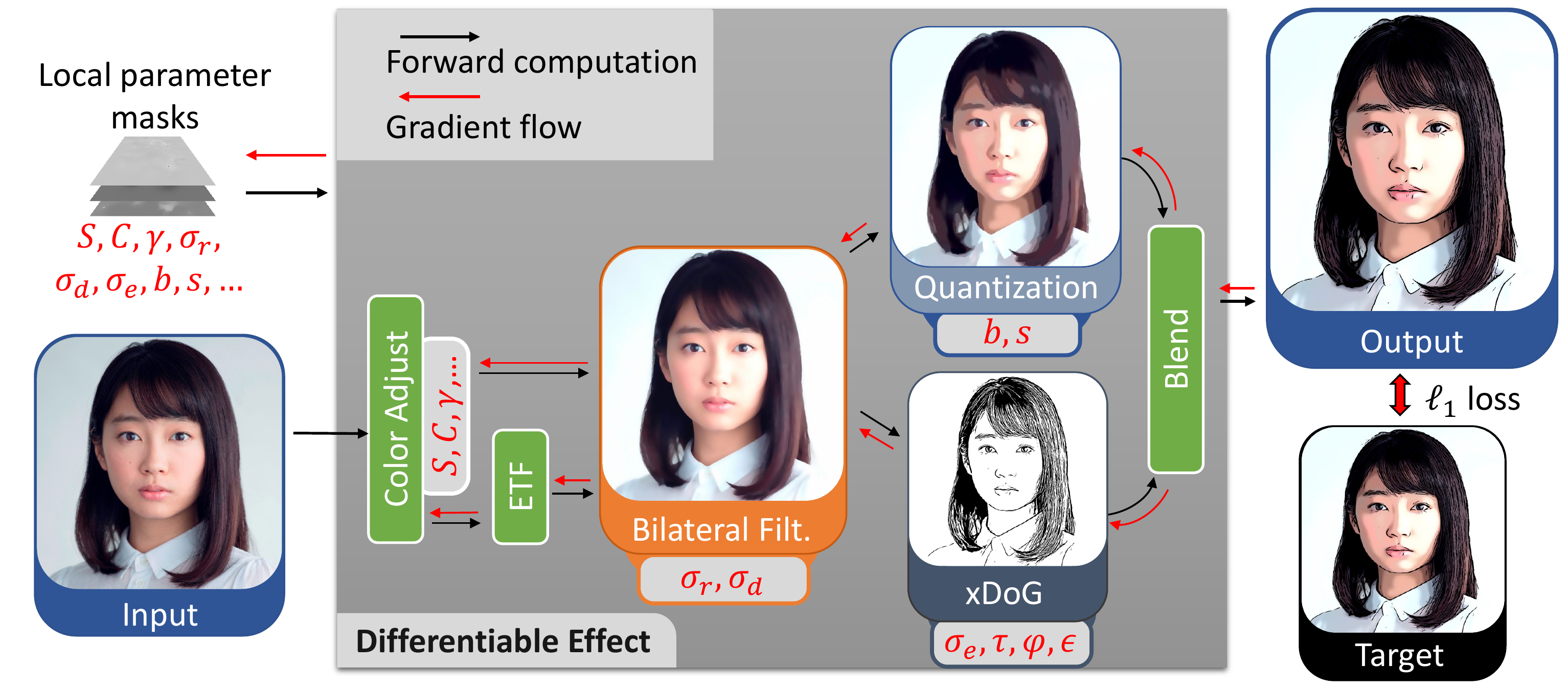}
\caption{\textbf{Exemplary differentiable effect pipeline.} Shown for the cartoon effect proposed by Winnemöller \etal\cite{winnemoller2006real}. Gradients are backpropagated from the loss to the filter parameter masks. In the effect, colors are first adjusted based on parameters such as saturation $S$, contrast $C$ or gamma $\gamma$, and then, using an \ac{ETF} \cite{kang2009flow}, orientation-aligned bilateral filtering \cite{kyprianidis2008image}, and \ac{XDoG} \cite{winnemoller2011xdog} are computed. Additionally, the image is quantized with respect to the number of bins $b$ and softness $s$.
}
\label{fig:toon_overview_pipeline}

\end{figure}

\section{Differentiable Image Filters}
\label{sec:DifferentiableFilters}

Heuristics-based stylization effects consist of pipelines of image filtering operations. To compute gradients for effect input parameters, all image operations within the pipeline are required to be differentiable with respect to their parameters and the image input.
Gradients throughout the pipeline can then be obtained by applying the chain rule.
\Cref{fig:toon_overview_pipeline} outlines the gradient flow from a loss function to the effect parameters by the cartoon pipeline example, gradients for parameters can be obtained both globally as well as locally using per-pixel parameter masks.

\setlength{\tabcolsep}{4pt}
\begin{table}[t]
\begin{center}
\caption{Differentiable filters by type and effect. Filters can be classified by their sampling approach, which is either point-based (PB) or in a fixed neighborhood (FN) or structure-adaptive neighborhood (SN). Some filters are non-differentiable and require numerical gradient (NG) approximations for training.}
\label{tab:filter_stages}

\begin{tabular}{lc|lll}
\hline\noalign{\smallskip}
Filtering operation                                  & Differentiable  & Car-      & Water-     & Oil-  \\
                                                     & Filter Type     & toon      & color      & paint  \\
\noalign{\smallskip}
\hline
\noalign{\smallskip} 
Anisotropic Kuwahara \cite{kyprianidis2009image}                 &     SN          &           & \ding{51}  &            \\   
Bilateral \cite{tomasi1998bilateral}                                            &     FN          &           & \ding{51}  & \ding{51} \\       
Bump Mapping / Phong Shading \cite{phong1975illumination}                                         &     FN       &           &            & \ding{51} \\  
Color Adjustment                                     &     PB          & \ding{51} & \ding{51}  & \ding{51} \\       
Color Quantization \cite{winnemoller2006real}        &     PB,NG       & \ding{51} &            &           \\   
Flow-based Gaussian Smoothing \cite{kyprianidis2008image}                               &     SN          & \ding{51} & \ding{51}  & \ding{51} \\       Gaps \cite{montesdeoca2017edge}                                                &     FN          &           & \ding{51}  &           \\ 
Joint Bilateral Upsampling \cite{kopf2007joint}      &     SN          &           &            & \ding{51} \\       
Flow-based Laplacian of Gaussian \cite{kyprianidis2011image}                                 &     SN          &           &            & \ding{51} \\      
Image Composition \cite{porter1984compositing}                                           &     PB           &           & \ding{51}  &           \\  
Orientation-aligned Bilateral \cite{kyprianidis2008image}   &     SN          & \ding{51} &            & \ding{51} \\         
Warping / Wobbling \cite{bousseau2006interactive}                                              &     FN          &           & \ding{51}  &           \\   
Wet-in-Wet \cite{wang2014towards}                                           &     SN          &           & \ding{51}  &           \\   
XDoG \cite{winnemoller2011xdog}                      &     SN          & \ding{51} &  \ding{51} & \ding{51}  \\          
\hline
\end{tabular}
\end{center}
\end{table}

In previous works, individual operations in such pipelines are typically implemented as shader kernels for fast GPU-based processing. To achieve an end-to-end gradient flow, we implement these operations in an auto-grad enabled framework, the implemented filtering stages are listed in \Cref{tab:filter_stages}.
Point-based and fixed-neighbourhood operations such as color space conversions, structure tensor computation~\cite{kyprianidis2008image}, or \ac{DoG}~\cite{winnemoller2011xdog} can generally be converted into differentiable filters by transforming any kernelized function into a sequence of its constituent auto-differentiable transformations. The exception to this are functions which are inherently not differentiable, such as color quantization, and which require the approximation of a numeric gradient. An example for numeric gradient approximation of color quantization is shown in the supplemental material.
Structure-adaptive neighborhood operations, such as the orientation-aligned bilateral filter and flow-based Gaussian smoothing filter, iteratively determine sampling locations based on the structure of the content (often oriented along a flow field). To preserve gradients and make use of the in-built fixed-neighborhood functions of auto-grad enabled frameworks, per-pixel iteration is transformed into a grid-sampling operation where neighborhood values are accumulated into a new dimension with size $D$ which represents the expected maximum kernel neighbourhood. A structure-adaptive filter transformation by example of the orientation-aligned bilateral filter is shown in the supplementary material. 

\paragraph{Implementation Aspects.}
We implement differentiable filters in PyTorch and create reference implementations of the same effects using OpenGL shaders. The learnability of each effect parameter is validated in a functional benchmark (shown in supplemental material) by optimizing the differentiable effect to match reference effects with randomized parameters. At inference time, OpenGL shaders can be interchangeably used with the differentiable effect, and can be efficiently executed on high-resolution images using parameters predicted on low-resolution images. At training time, memory usage of differentiable filters can be reduced by controlling their kernel-size (shown in supplemental material).


\section{Parameter Prediction} 
\label{sec:ParameterPrediction}

With the introduced differentiable filter pipelines, parameters can be optimized using image-based losses. To generalize to unseen data, we explore \acp{PPN} that are trained to predict global parameters or spatially varying (local) parameters.

\subsection{Parameter Prediction Networks} 
\label{sec:ParameterPredictionNetworks}

\paragraph{Global Parameter Prediction.}

We construct a \ac{PPN} that predicts the effect parameters of a stylized example image, given both the stylized and source image. Thereby, the network is trained to effectively reverse-engineer the stylization effect. During training, gradients are back-propagated through the effect, the parameters, and finally to the \ac{PPN}. 
Formally, let $I$ denote the input image, $T$ the target image, $O(\cdot)$ the differentiable effect,  $P_G(\cdot)$ the \ac{PPN} network. The loss for the global PPN is computed using an $\ell_1$ image space-based loss as:

\begin{equation}
\begin{split}
    \mathcal{L}_{\text{global}} = \| O(I,P_G(I))) - T\|_{1} 
\end{split}
\end{equation}

Our global \ac{PPN} architecture consists of a VGG backbone \cite{simonyan2014very} 
that extracts features of the input and stylized image and computes layer-wise Gram matrices to encode important style information \cite{gatys2016image}. The accumulated features are passed to a multi-head module to predict the final global parameters.
We found this network architecture to perform superior against other common architecture variants,  please refer to the supplementary material for an ablation study and details on the architecture. 

\paragraph{Local Parameter Prediction.}
While the global \ac{PPN} can predict settings of similar algorithmic effects, real-world, hand-drawn images often vary significantly based on the local content, which cannot be  modeled by a global  parameterization.
Therefore, we construct a \ac{PPN} to predict local \textit{parameter masks}. 
We use a U-Net architecture \cite{ronneberger2015u} for mask prediction at input resolution, where each output channel represents a parameter. 
Gradients are back-propagated to the \ac{PPN} through the differentiable effects for all parameter masks.

For training, a paired data \ac{GAN} approach is used, where a patch-based Pix2Pix discriminator \cite{isola2017image} matches the distribution of patches in the reference image and an additional weighted $\ell_1$ image space loss enforces a more strict pixel-wise similarity. 
Formally, let $D(\cdot)$  denote the discriminator, $\mathcal{L}_{\mathit{TV}}(\cdot)$ the total variation regularizer~\cite{johnson2016perceptual} to enforce smooth parameter masks, and $P_L(\cdot)$ the local \ac{PPN} which acts as the generator network. 
The final loss $\mathcal{L}$ for the \ac{PPN} generator is computed as:
\begin{equation}
\begin{split}
    \mathcal{L} = \mathbb{E}\big[& \alpha\log(1 - D(I,O(I,P_L(I))))  \\ & + \beta\| O(I,P_L(I)) - T\|_{1} + \gamma\mathcal{L}_{\mathit{TV}}(P_L(I)) \big]
\end{split}
\label{eq:objectivefunction}
\end{equation}

\subsection{PPN Experiments} 
\label{subsec:PPNExperiments}

We conduct several experiments to validate our approach for global and local parameter prediction.

\begin{table}[t]
\caption{\textbf{Global PPN loss functions.} The \acp{PPN} are trained with different loss functions on the \ac{NPR} benchmark~\cite{rosin2020nprportrait}. Networks trained in parameter space use reference parameters as the loss signal directly. We measure the \ac{SSIM} and \ac{PSNR}
}
\label{tbl:xdog_ppn_loss_functions}
\begin{center}
\begin{tabular}{|l|ccc|}
\hline
Loss domain & \ac{SSIM} & \ac{PSNR} & Parameter loss \\
\hline
Parameter space $\ell_1$/$\ell_2$  & 0.738/0.737 & 12.530/12.927 & \textbf{0.158}/0.162 \\
Image space  $\ell_1$/$\ell_2$ & \textbf{0.780}/0.764  & \textbf{13.875}/13.286  & 0.183/0.190 \\
\hline
\end{tabular}
\end{center}
\end{table}

\begin{figure}[tb]
\centering
\begin{subfigure}{0.155\linewidth}
\includegraphics[width=1\textwidth]{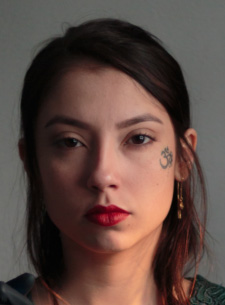}
\caption{Source}
\end{subfigure}
\begin{subfigure}{0.155\linewidth}
    \includegraphics[width=1\textwidth]{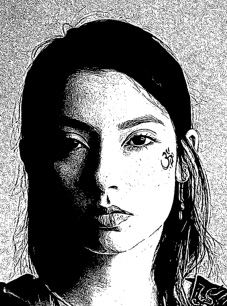}
    \caption{Predicted}
    \label{fig:global_predictions_ppn:pred_global1}
\end{subfigure}
\begin{subfigure}{0.155\linewidth}
    \includegraphics[width=1\textwidth]{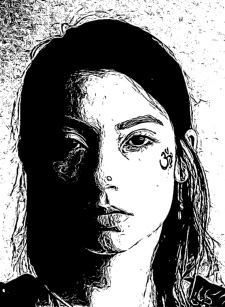}
    \caption{Reference}
    \label{fig:global_predictions_ppn:reference1}
\end{subfigure}
\hspace{2mm}
\begin{subfigure}{0.155\linewidth}
\includegraphics[width=1\textwidth]{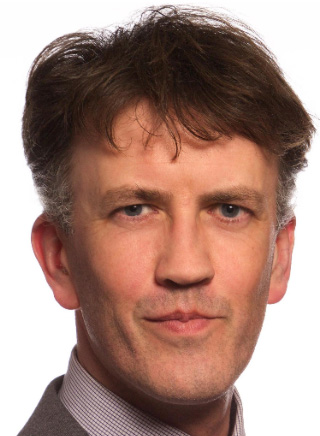}
\caption{Source}
\end{subfigure}
\begin{subfigure}{0.155\linewidth}
    \includegraphics[width=1\textwidth]{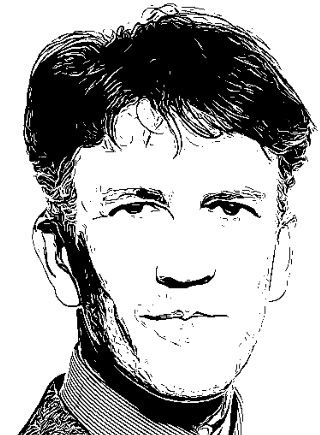}
    \caption{Predicted}
    \label{fig:global_predictions_ppn:pred_global2}
\end{subfigure}
\begin{subfigure}{0.155\linewidth}
    \includegraphics[width=1\textwidth]{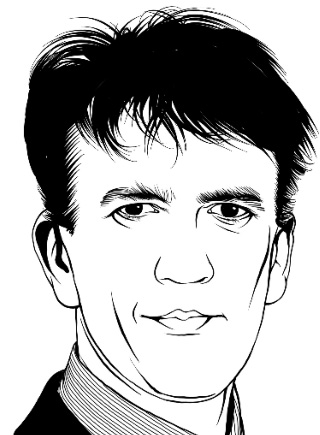}
    \caption{Reference}
    \label{fig:global_predictions_ppn:reference2}
\end{subfigure}
\caption{\textbf{Predictions using the global \ac{PPN} for \ac{XDoG}.} The stylized reference \protect\subref{fig:global_predictions_ppn:reference1} is synthetic (generated using the reference \ac{XDoG} implementation), while the reference \protect\subref{fig:global_predictions_ppn:reference2} is hand-drawn, taken from APDrawing~\cite{yi2019apdrawinggan}.
}
\label{fig:global_predictions_ppn}
\end{figure}

\paragraph{Global Parameter Prediction.}
We compare the loss function and loss space of global PPNs in (\Cref{tbl:xdog_ppn_loss_functions}). We  find that while directly predicting in parameter space (without obtaining gradients from the effect) yields closer parameter values, the highest visual accuracy is achieved using an $\ell_1$ image space-based loss. This validates the usefulness of the differentiable effect being part of the training pipeline.
Global \acp{PPN} can accurately match reference stylizations created by the same effect, as shown in \Cref{fig:global_predictions_ppn:pred_global1}. Furthermore, they can approximate similar hand-drawn styles, albeit with significant local deviations, as shown in \Cref{fig:global_predictions_ppn:pred_global2}.

\paragraph{Content-adaptive Effects.}

\begin{figure}[t]
\centering
\begin{subfigure}{0.49\linewidth}
  \includegraphics[width=\linewidth,trim={8cm 2.8cm 6cm 2.2cm},clip]{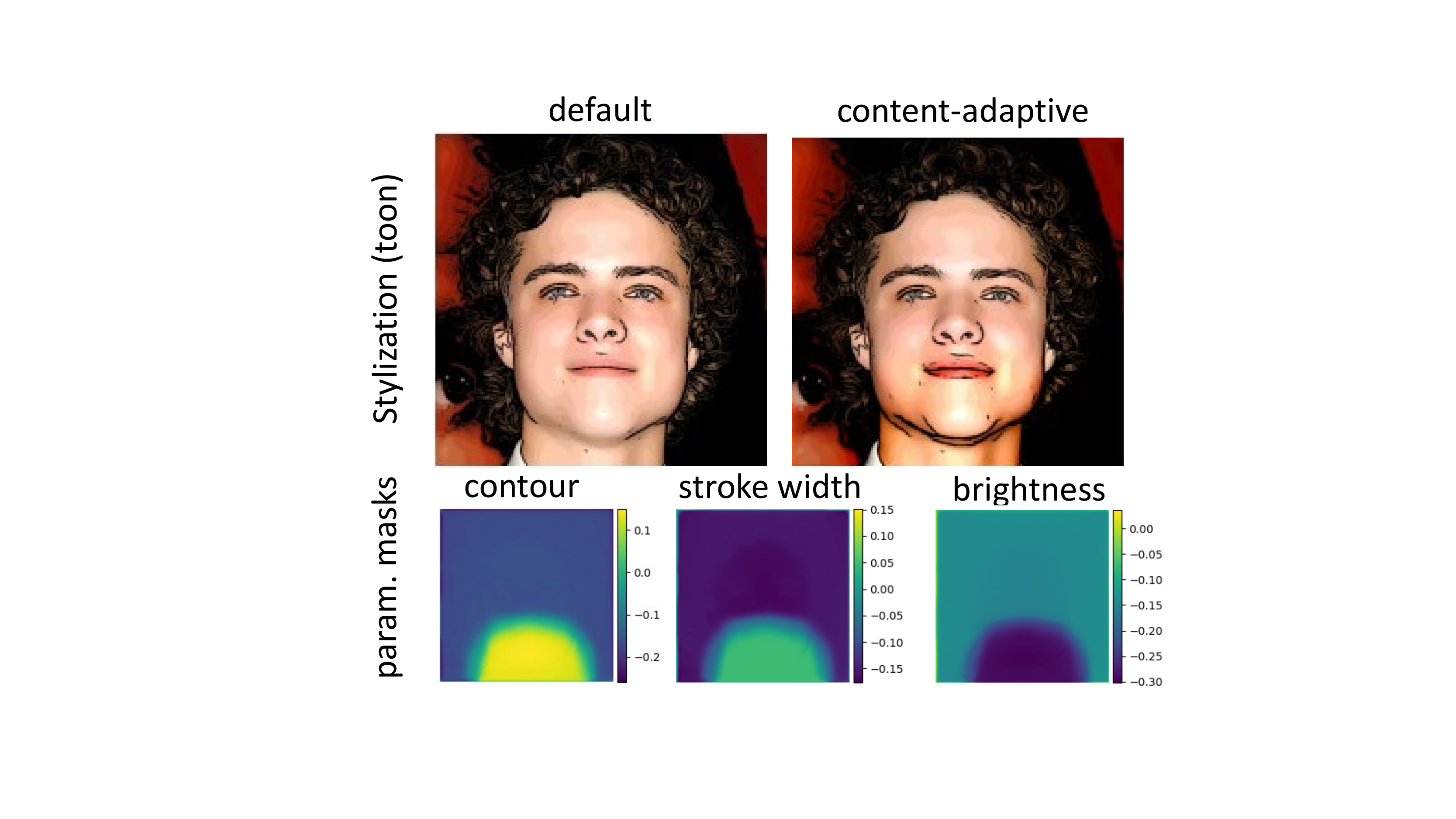}
    \caption{Chin contour enhancement}
    \label{fig:local_ppn_results:a}
\end{subfigure}
\begin{subfigure}{0.49\linewidth}
  \includegraphics[width=\linewidth,trim={8cm 2.8cm 6cm 2.2cm},clip]{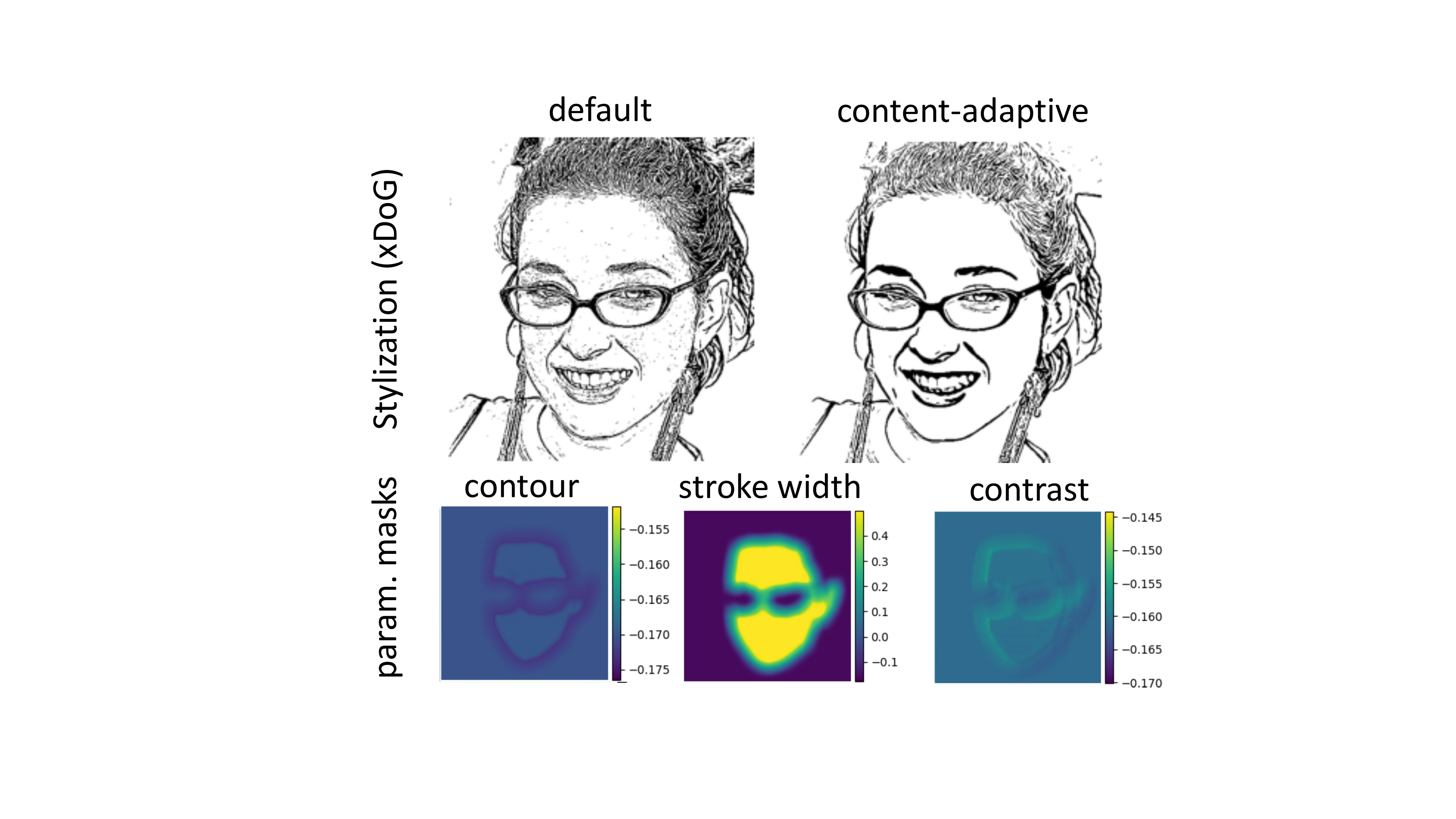}
    \caption{Selective detail reduction}
    \label{fig:local_ppn_results:b}
\end{subfigure}
\caption{\textbf{Local \ac{PPN} results.} Networks are trained on CelebAMask-HQ \cite{CelebAMask-HQ} to generate selective enhancements by predicting parameter masks. The default result is created by a global parameter configuration, while the shown predicted parameter masks  create the content-adaptive result. 
}
\label{fig:local_ppn_results}
\end{figure}

Using the previously described approach for local \ac{PPN} training, we demonstrate its applicability to several common problems that are often present in purely algorithmic image-stylization techniques. 
We consider three tasks to improve stylization quality: (1) highlighting facial features, for example by increasing contours at low-contrast edges such as the chin (\Cref{fig:local_ppn_results:a}), (2) selectively reducing details such as small wrinkles in the face  (\Cref{fig:local_ppn_results:b}), and (3) background removal (refer to supplemental material). 
We use the CelebAMask-HQ dataset \cite{CelebAMask-HQ} for training, which consists of 30,000 high-resolution face images and segmentation masks for all parts of the face. For the above tasks, we each create a synthetic training dataset by stylizing images using a reference effect and adjusting its parameters for certain parts of the face (obtained from dataset annotations) according to the task, \eg increasing the amount of contours in the chin area. %
In \Cref{fig:local_ppn_results} trained \ac{PPN} networks are evaluated by plotting the predicted local parameter masks together with the generated stylizations. 
It shows that the networks learn to predict parameter masks for the relevant regions accurately solely by observing pixels without additional supervision.

\begin{figure}[t]
\centering
\begin{subfigure}[b]{0.245\textwidth}%
    \includegraphics[width=\textwidth,height=0.113\textheight]{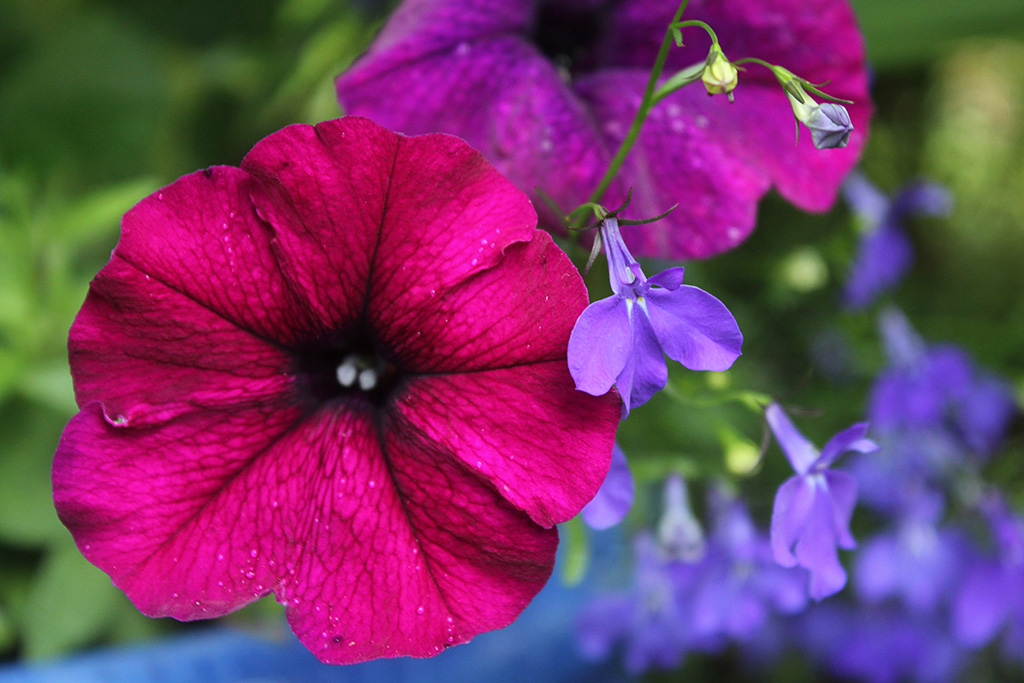}%
    \label{fig:flower:content}%
\end{subfigure}\hfill
\begin{subfigure}[b]{0.245\textwidth}
    \includegraphics[width=\textwidth,height=0.113\textheight]{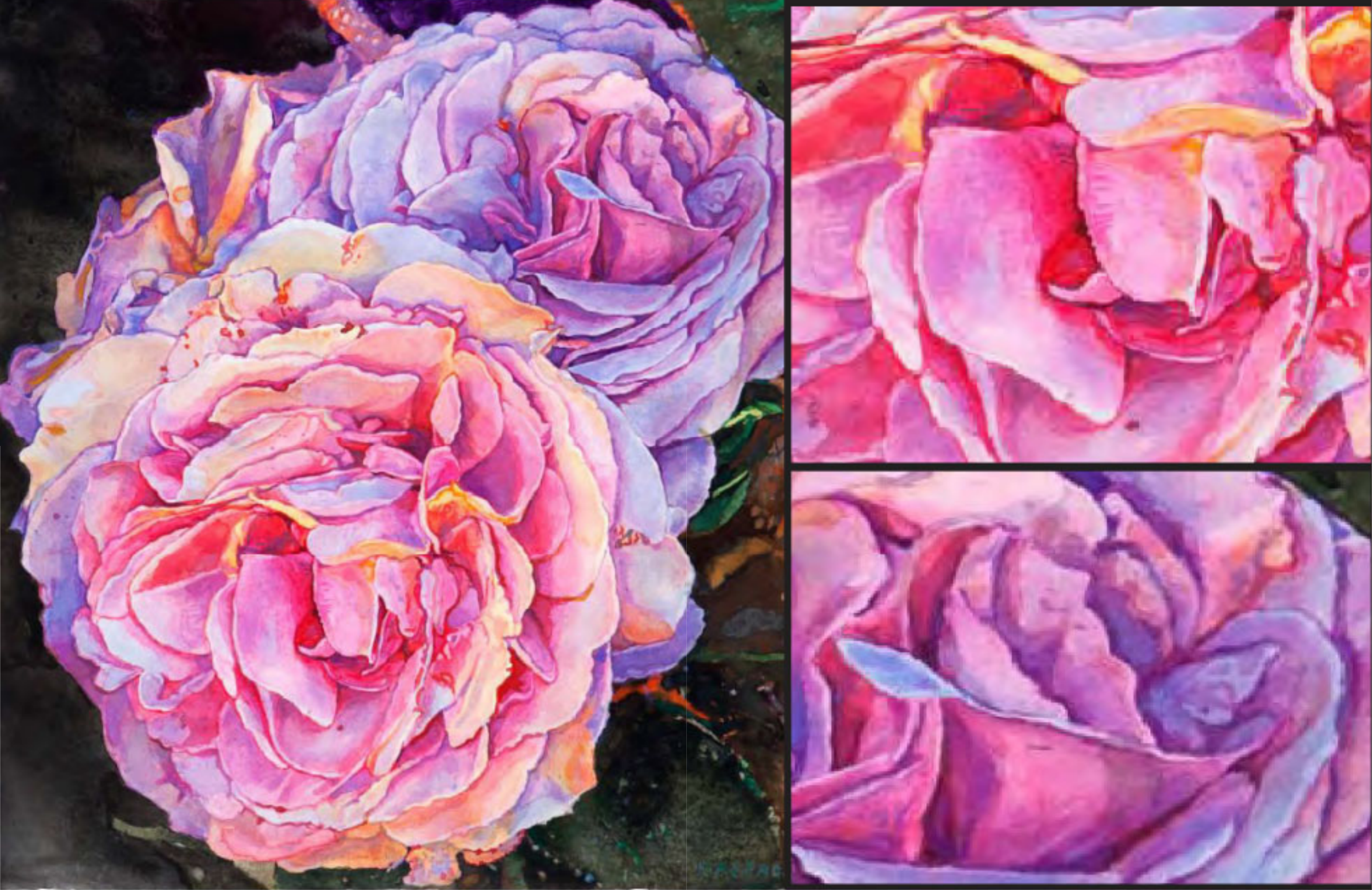}%
    \label{fig:flower:style_image}%
\end{subfigure}\hfill
\begin{subfigure}[b]{0.245\textwidth}%
    \includegraphics[width=\textwidth,height=0.113\textheight]{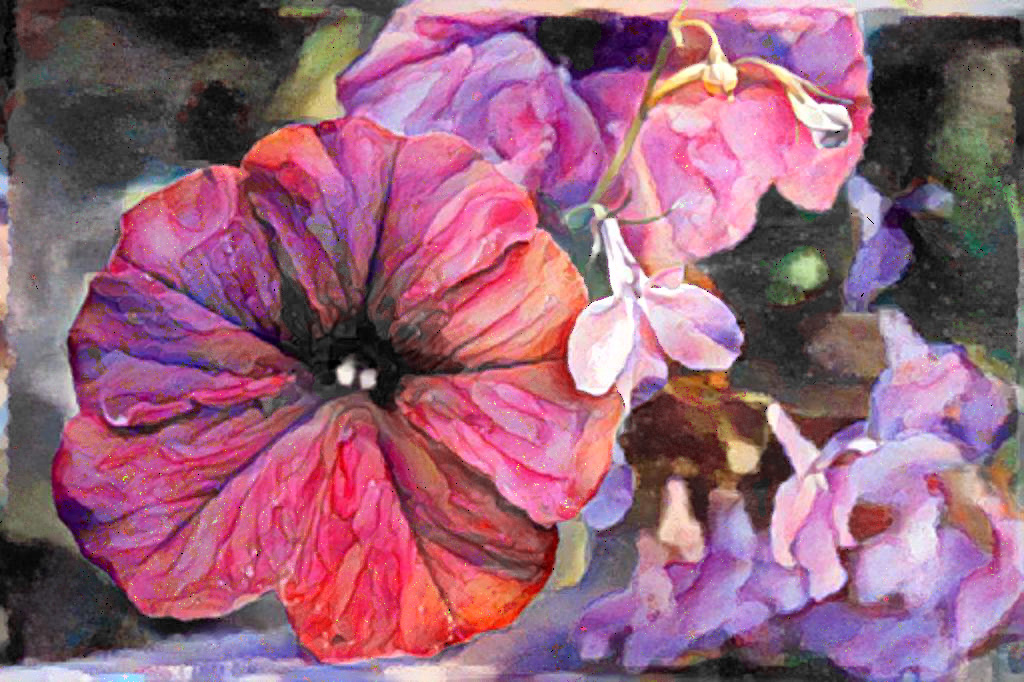}%
    \label{fig:flower:stylized_result}%
\end{subfigure}\hfill
\begin{subfigure}[b]{0.245\textwidth}%
    \includegraphics[width=\textwidth,height=0.113\textheight]{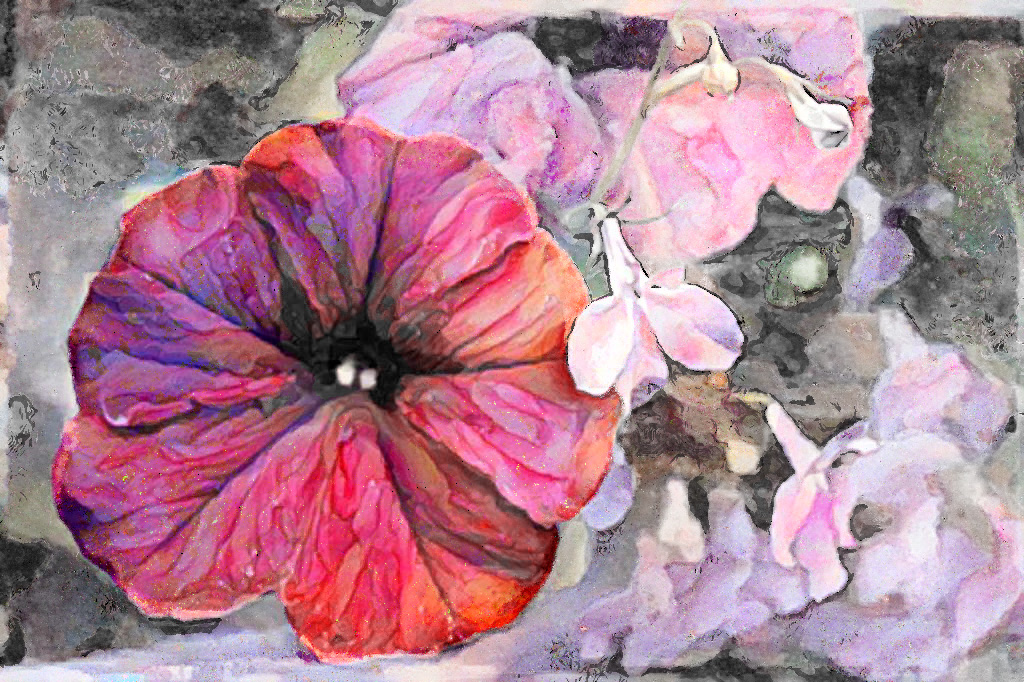}%
    \label{fig:flower:retouched_result}%
\end{subfigure}\\[0.5ex]

\begin{subfigure}[b]{0.245\textwidth}%
    \includegraphics[width=\textwidth,height=0.113\textheight]{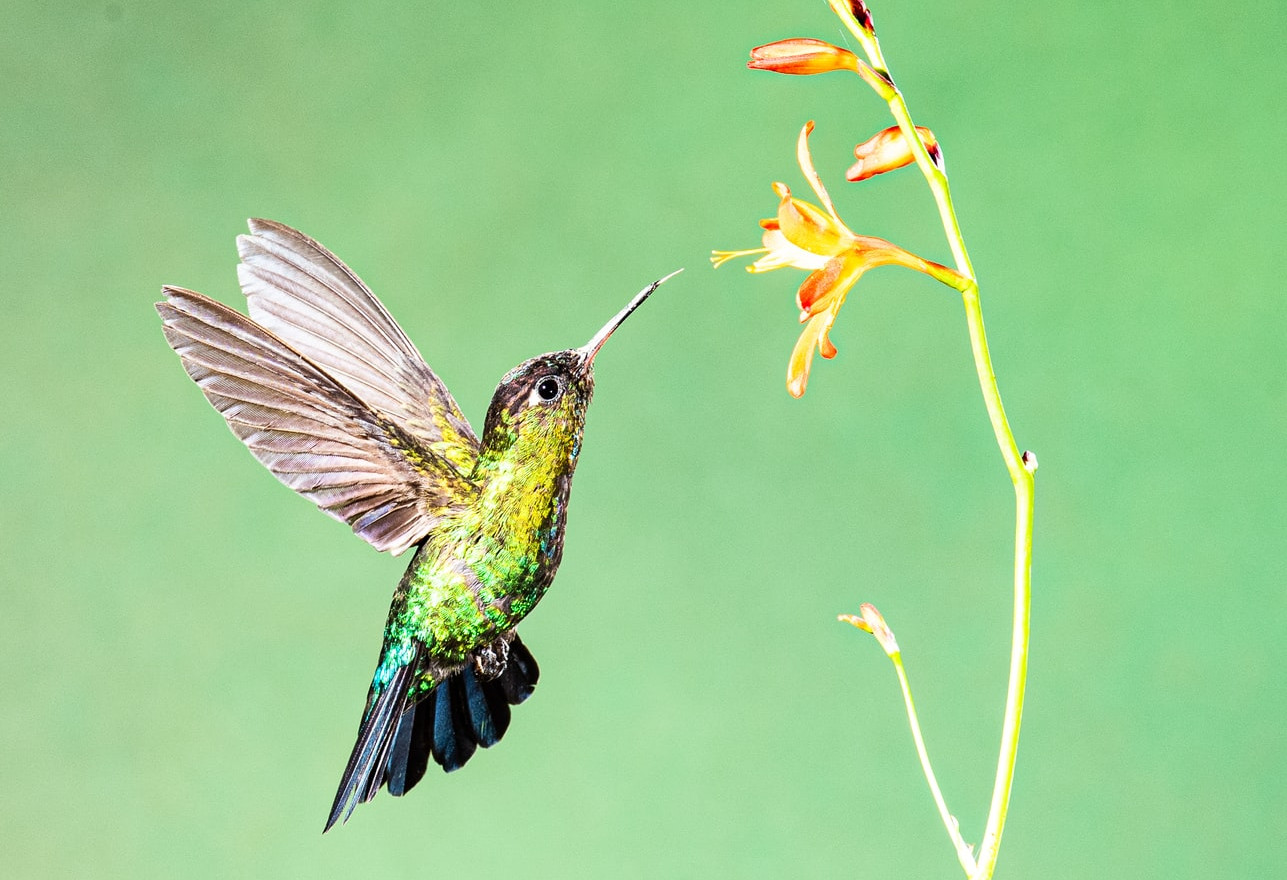}%
    \caption{Content image}%
    \label{fig:colibri:content}%
\end{subfigure}\hfill
\begin{subfigure}[b]{0.245\textwidth}%
    \includegraphics[trim={1.34in 0.11in 1.34in 0.11in},width=\textwidth,clip,height=0.113\textheight]{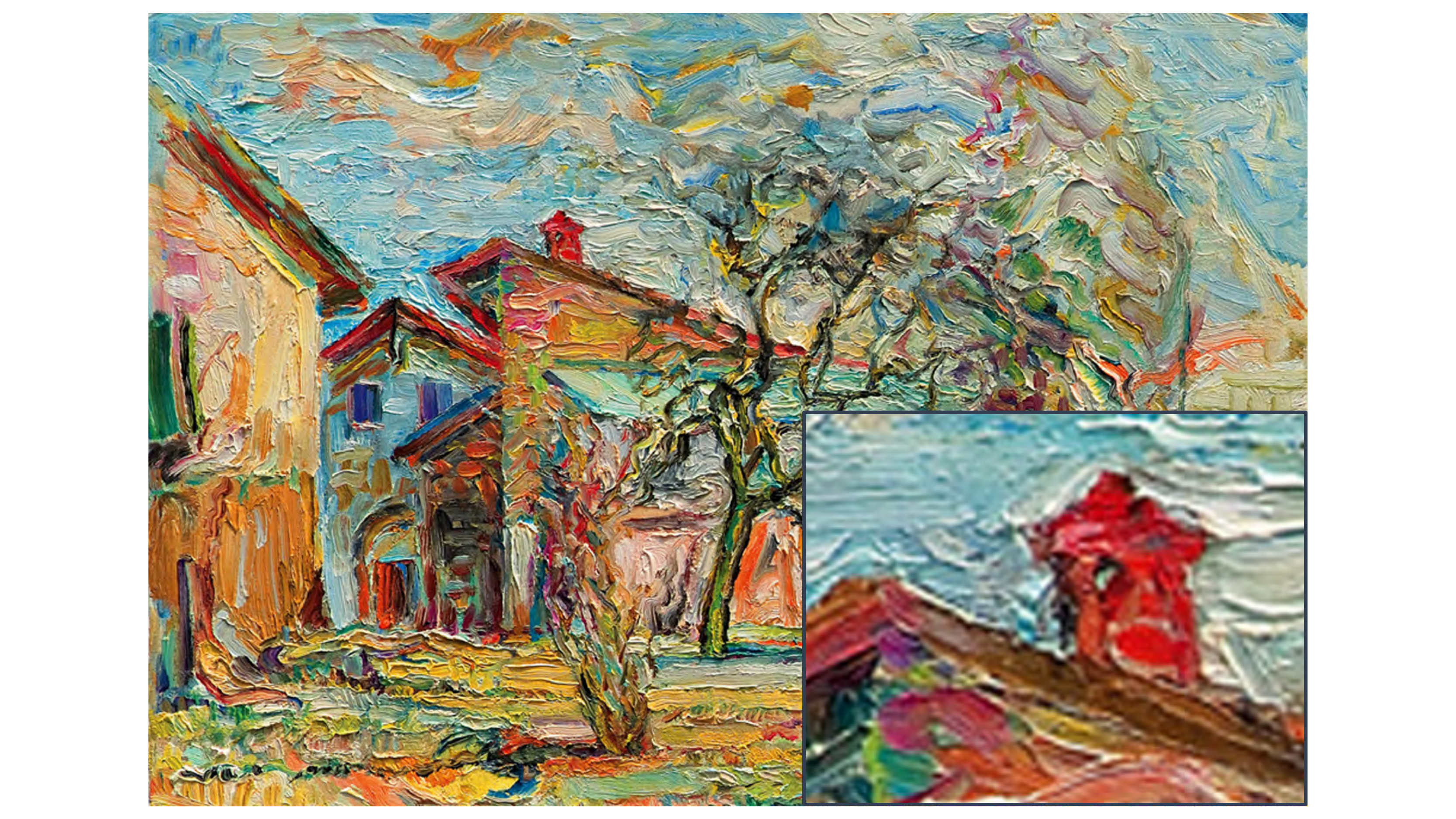}%
    \caption{Style image}%
    \label{fig:colibri:style}%
\end{subfigure}\hfill
\begin{subfigure}[b]{0.245\textwidth}%
    \includegraphics[trim={1.34in 0.11in 1.34in 0.11in},width=\textwidth,clip,height=0.113\textheight]{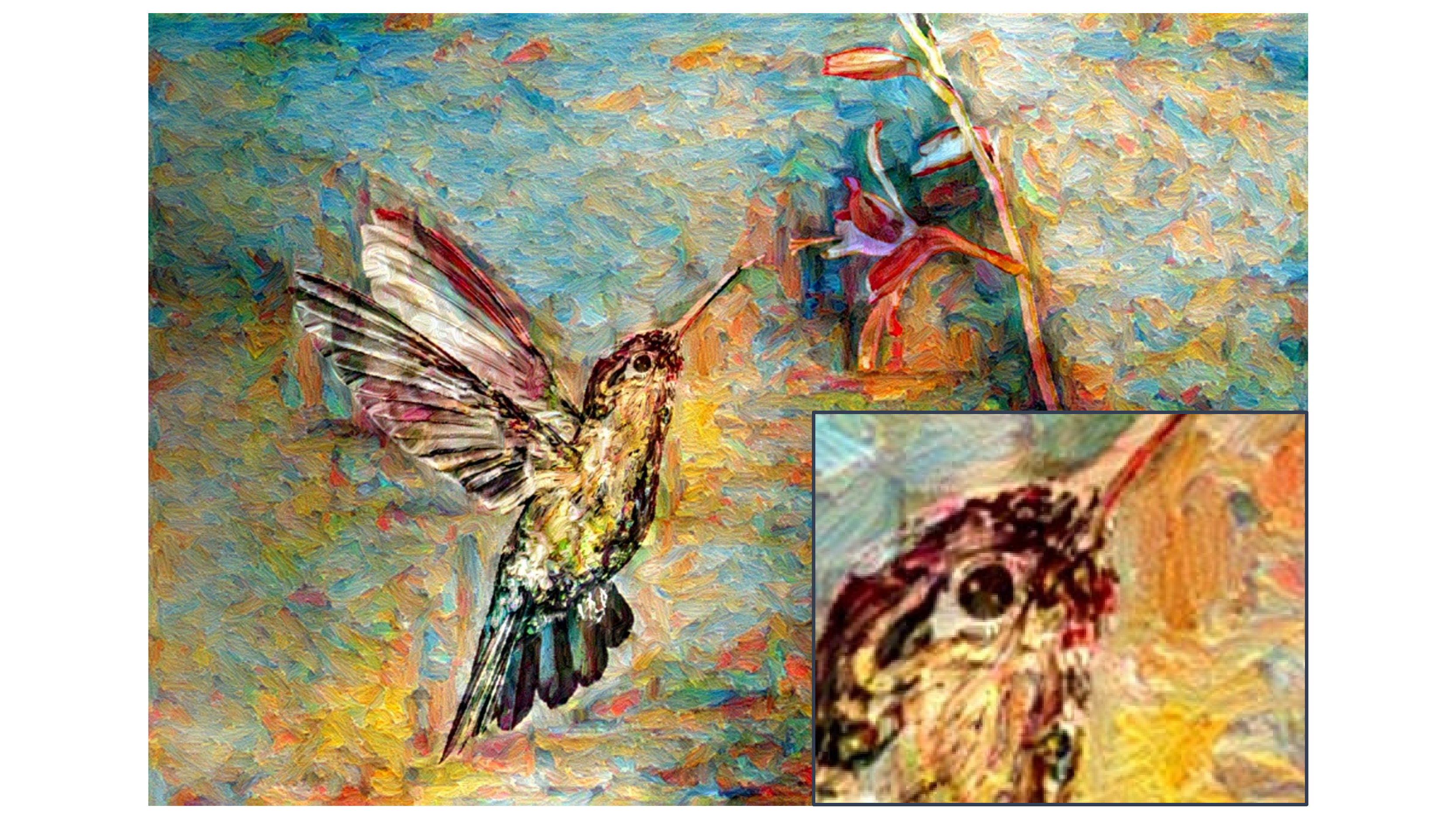}%
    \caption{Stylized result}%
    \label{fig:colibri:colibri_result}%
\end{subfigure}\hfill%
\begin{subfigure}[b]{0.245\textwidth}%
    \includegraphics[trim={1.34in 0.11in 1.34in 0.11in},width=\textwidth,clip,height=0.113\textheight]{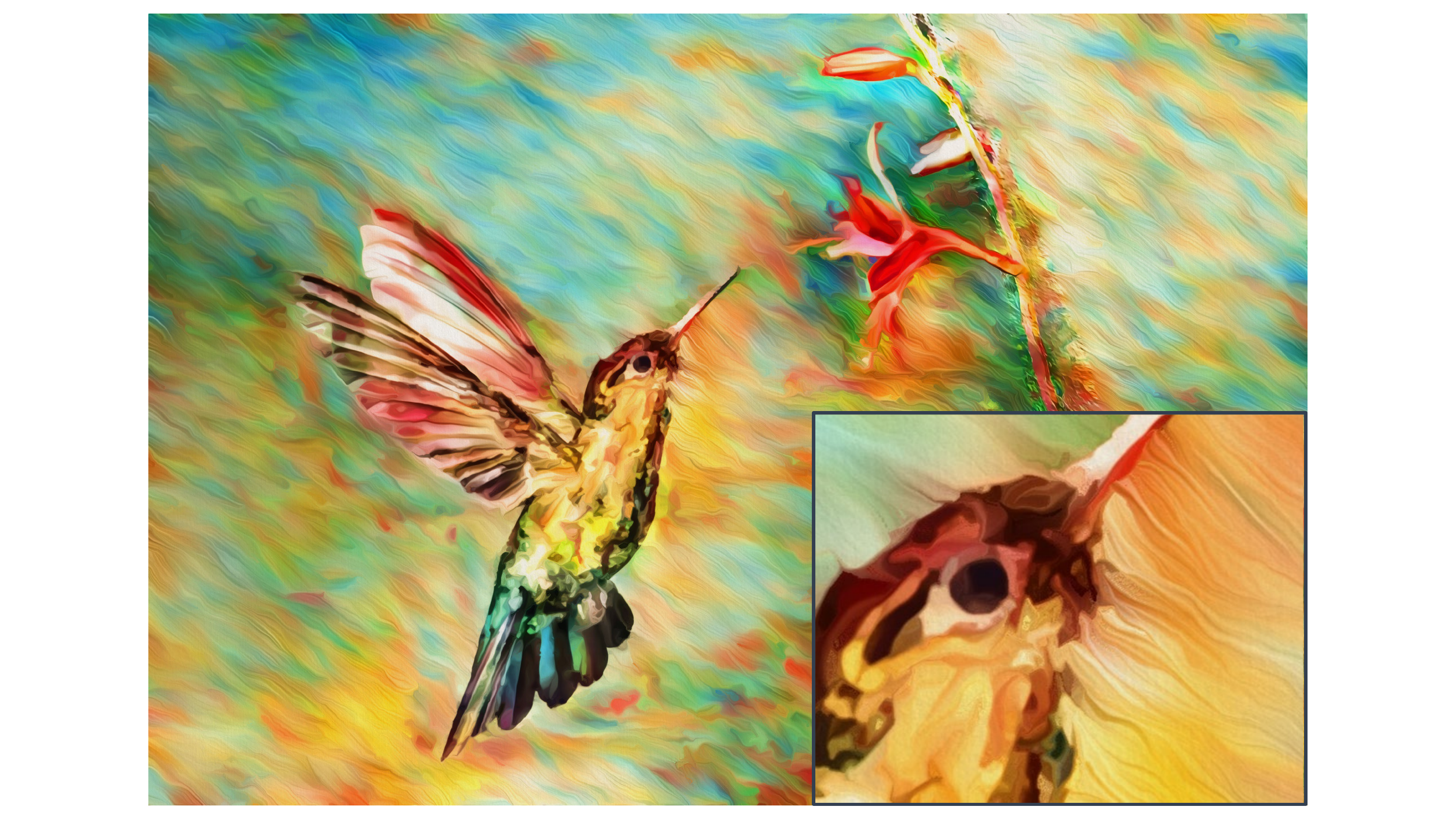}%
    \caption{Locally edited}%
    \label{fig:colibri:retouched_result}%
\end{subfigure}%
\caption{\textbf{Parametric style transfer.} Effect parameters are optimized (watercolor effect in top row, oilpaint effect in bottom row) to match the stylistic reference image \protect\subref{fig:colibri:style}. Users can then interactively edit the result \protect\subref{fig:colibri:colibri_result} by adjusting resulting parameter masks. In the top row, the saturation parameter mask is adjusted to highlight foreground objects \protect\subref{fig:colibri:retouched_result}. In the bottom row, the oilpaint-specific bump scale and flow-smoothing are adjusted in \protect\subref{fig:colibri:retouched_result} to make the background appear to be painted wet-in-wet with long brushstrokes.
}
\label{fig:flower_retouched}
\end{figure}

\begin{figure}[ht]
    \centering
        \begin{subfigure}[t]{0.32\textwidth}
                \includegraphics[width=\textwidth]{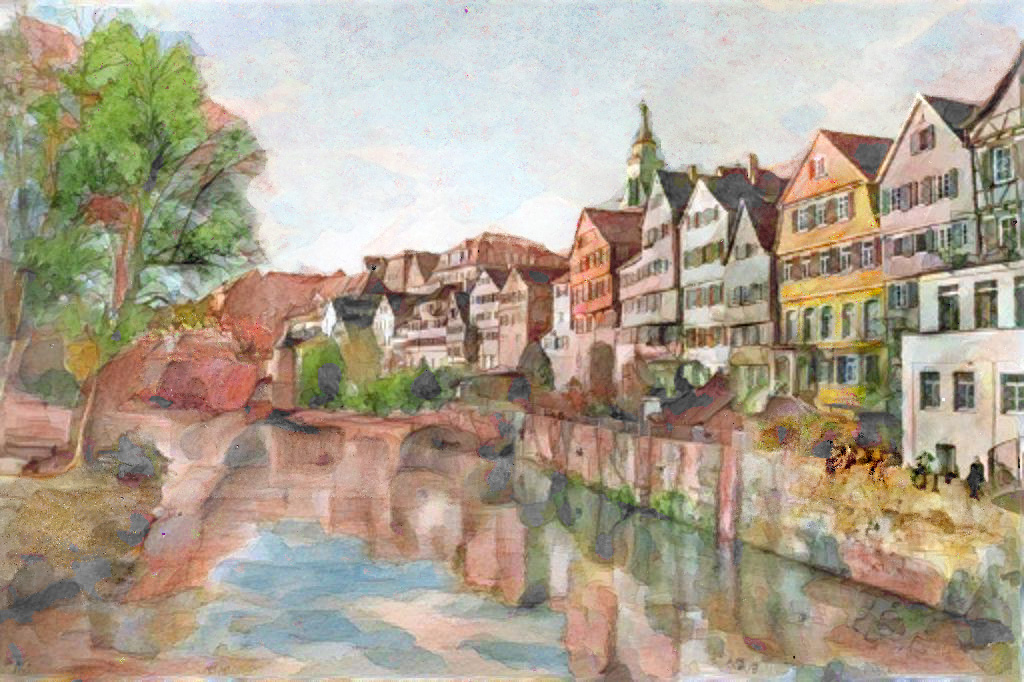}
          \caption{Watercolor $\ell_1$ optim }
          \label{fig:nstlosses:l1optim}
        \end{subfigure}
        \begin{subfigure}[t]{0.32\textwidth}
                \includegraphics[width=\textwidth]{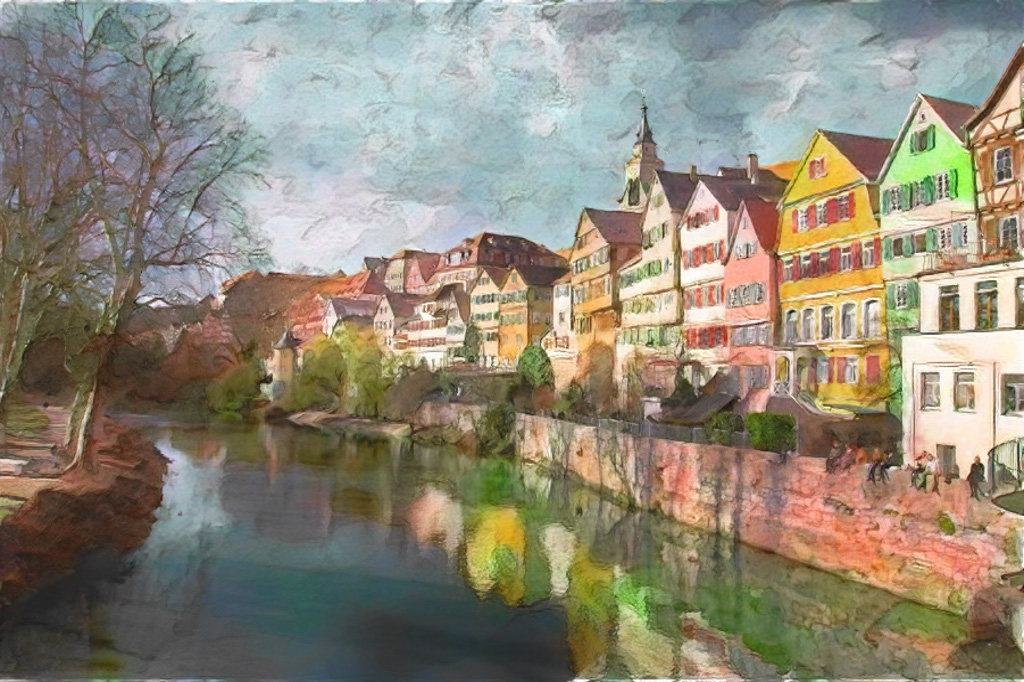}
          \caption{Oilpaint $\mathcal{L}_C + \mathcal{L}_S$}
          \label{fig:nstlosses:oillslc}
        \end{subfigure}
        \begin{subfigure}[t]{0.32\textwidth}
                \includegraphics[width=\textwidth]{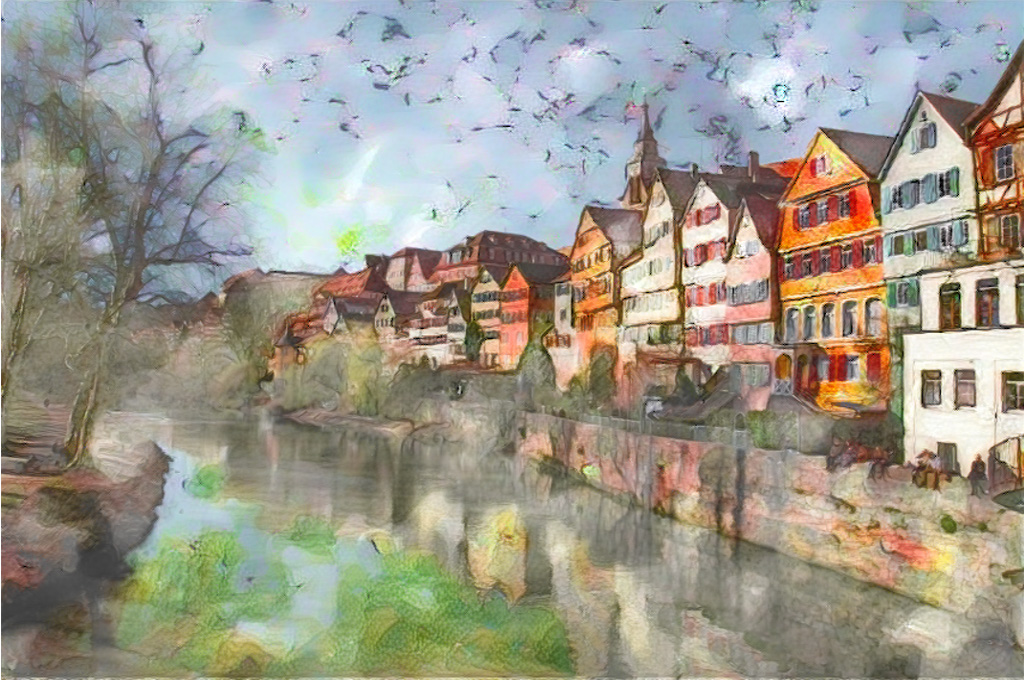}
          \caption{Watercolor $\mathcal{L}_C + \mathcal{L}_S$}
          \label{fig:nstlosses:watercolorlslc}
        \end{subfigure}
        \caption{Optimization of parametric style transfer with different losses. In a), the effect is optimized by matching the result of STROTSS style transfer \cite{kolkin2019style} using an $\ell_1$ loss. In b) and c), the effect is directly optimized using neural style transfer \cite{gatys2016image} losses $\mathcal{L}_C + \mathcal{L}_S$. The content/style image are the same as in \Cref{fig:teaseer}.}
        \label{fig:nstlosses}
    \end{figure}

\section{Applications}
\label{sec:applications}

Using our framework for global and local parameter prediction for differentiable algorithmic pipelines, example-based stylization with closely related references is made possible. To adapt to real-world, more diverse example data, our framework can be integrated with existing stylization paradigms. In the following, we demonstrate our approach for the task of (statistics-based) style transfer reconstruction and \ac{GAN}-based image-to-image translation based on the APDrawing dataset \cite{yi2019apdrawinggan}.

\subsection{Style Transfer}
We investigate the combination of iterative style transfer and algorithmic effects.
We use \ac{STROTSS} by Kolkin \etal\cite{kolkin2019style} to create stylized references for our effect.
We subsequently try to recreate the style transfer result with our algorithmic effects by optimizing parameter masks (\Cref{fig:flower_retouched}). 
For this, a $\ell_1$ loss in image space is again used to match effect output and reference image.  

\begin{table}[t]
    \caption{Comparison of our method to \ac{STROTSS} \cite{kolkin2019style} for style loss $\mathcal{L}_S$, content loss $\mathcal{L}_C$, and the $\ell_1$ difference between respective results. We test against in-domain style images and against a set of common (arbitrary domain) NST styles. In each case, results are averaged over 10 styles and NPRB \cite{rosin2020nprportrait} as content\protect\footnotemark}
    \label{tbl:strotss_all_pipelines}
    \begin{center}
    \begin{tabular}{|ll |cccccc|}
    \hline
    & & \ac{XDoG} & Cartoon & \multicolumn{2}{c}{Watercolor}     & \multicolumn{2}{c|}{Oilpaint} \\
    \multicolumn{2}{|r|}{style domain:} &  bw-drawing   &  cartoon       & watercolor  & common & oilpaint & common \\
    \hline
    \multirow{2}{*}{$\mathcal{L}_S$} & \acsu{STROTSS} & 0.340 & 0.289 & 0.351 & 0.39 & 0.209 & 0.39  \\
    & Our results & 0.246 & 0.406 & 0.359  & 0.42  & 0.384  & 0.52 \\
    \hline
    \multirow{2}{*}{\(\mathcal{L}_C\)} & \acsu{STROTSS} & 0.099 & 0.094 & 0.081 & 0.036 & 0.037 & 0.036 \\
    & Our results & 0.172 & 0.148 & 0.092 & 0.034 & 0.023 & 0.033\\
    \hline
    \multicolumn{2}{|r|}{$\ell_1$ Difference} & 0.188 & 0.136 & 0.007  & 0.021 & 0.036 & 0.039\\
    \hline
    \end{tabular}
    \end{center}
    \end{table}

\paragraph{Optimization.}
We run the style transfer algorithm~\cite{kolkin2019style} for 200 steps to create a stylized reference with a resolution of 1024 $\times$ 1024 pixels. The local parameter masks are then optimized using 1000 iterations of Adam~\cite{kingma2014adam} and a learning rate of 0.1. The learning rate is decreased by a factor of 0.98 every 5 iterations starting from iteration 50. To avoid the generation of artifacts in parameter masks, we smooth all masks at increasing iterations (10, 25, 50, 100, 250, 500) using a Gaussian filter. Alternatively to image-space matching, parameters can also be directly optimized using NST~\cite{gatys2016image} losses $\mathcal{L}_S+\mathcal{L}_C$, however this often fails to transfer more complex stylistic elements \cref{fig:nstlosses}. 
As optimizing effect parameters to affect the output is harder than direct pixel optimization (as done in NST), stricter supervision (i.e., pixel-wise losses) is necessary to achieve similar outputs. Directly using $\mathcal{L}_S+\mathcal{L}_C$ could however be appropriate for photographic style transfer.

\footnotetext{Exemplary style images and results in suppl. material.}
    
\paragraph{Results.}
As \Cref{tbl:strotss_all_pipelines} shows, parametric style transfer works better for effects that have a high expressivity and are closer to hand-drawn styles, such as the watercolor or oilpaint, compared to more restricted effects such as \ac{XDoG} or cartoon. 
After the generation of local parameter masks, the parameters can be refined by the user as shown in \Cref{fig:colibri:retouched_result}. By optimizing parameters, we obtain an interpretable \enquote{whitebox} representation of a style that, in contrast to current pixel-optimizing \acp{NST} \cite{gatys2016image,johnson2016perceptual,kolkin2019style}, retains controllability according to artistic design aspects. Furthermore, our method is resolution independent, \ie parameter masks can be optimized at lower resolutions and then scaled up to high resolutions for editing. In \Cref{fig:colibri:retouched_result} (bottom row) the effect is applied at 4096 $\times$ 4096 pixels, while current style transfers are mostly memory-limited to much lower resolutions.
We further compare matching performance of in-domain styles with a set of common NST styles and observe that the $\ell_1$ difference (\Cref{tbl:strotss_all_pipelines}) is only marginally higher for the latter. Thus, highly parameterized effects such as watercolor or oilpaint can emulate any out-of-domain style by per-pixel optimization of parameter combinations reasonably well. However, as this creates highly fragmented parameter masks, a tradeoff between generalizability and interpretability of masks can be made using a weighted total variation-loss.

\subsection{GAN-based image-to-image translation with PPNs}\label{subsec:CombiningEffects}

For learning a style distribution, \ie the characteristics of an artistic style over a larger collection of artworks, GAN-based approaches have achieved impressive results \cite{chen2018cartoongan,wang2020learning,yi2019apdrawinggan}. We investigate training \acp{PPN} in a GAN-setting for image-to-image translation. 
While the global and local \acp{PPN} discussed in \Cref{sec:ParameterPrediction} can match in-domain styles very well (\Cref{subsec:PPNExperiments}), they cannot produce local image structures that are not synthesizable by their constituent image filters. Artistic reference styles, however, often contain stylistic elements that have not been modeled in the heuristics-based filter - this holds especially true for more simple effects such as xDoG. 
For reference styles that are stylistically close to such effects (\eg xDoG), such as line-drawings, we hypothesize that combining our filter pipeline with a lightweight \ac{CNN}-based post-processing operation and learning them end-to-end can close the domain gap while retaining the positive properties of the filter parametrization and beeing computationally efficient.


\paragraph{Dataset.}
For our experiment, we select the APDrawing~\cite{yi2019apdrawinggan} dataset which consists of closely matching photos and their hand-drawn stylistic counterparts. Its hand-drawn images are reasonably similar to the \ac{XDoG} results, while still containing many stylistic abstractions that cannot be emulated solely by the \ac{XDoG}. 
We hypothesize, that re-creating such an effect entails both edge detection and content abstraction, which could be performed by our differential \ac{XDoG} pipeline combined with a separate convolution network for content abstraction. 

APDrawing contains a set of 140 portrait photographs along with paired drawings of these portraits. 
The \ac{GAN}-based local \ac{PPN} approach introduced in \Cref{sec:ParameterPrediction} is used to train on this paired dataset, where solely the generator is extended using a \ac{CNN} $N(\cdot)$ to post-process the \ac{XDoG} output, \ie following \Cref{eq:objectivefunction} our generator now combines \ac{PPN}, effect and \ac{CNN}: $N(O(I, P_L(I)))$.

\begin{figure}[t]
    \centering
    \small{
    \begin{subfigure}[b]{0.20\linewidth}
        \includegraphics[width=\textwidth, trim={0cm 0.3cm 0cm 0.3cm},clip]{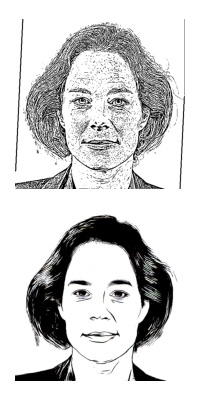}
    \end{subfigure}
    \begin{subfigure}[b]{0.20\linewidth}
        \includegraphics[width=\textwidth, trim={0cm 0.3cm 0cm 0.3cm},clip]{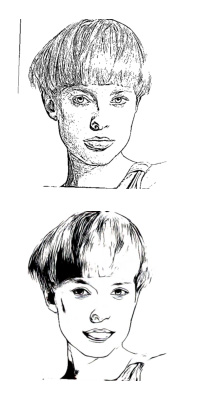}
    \end{subfigure}
    \begin{subfigure}[b]{0.20\linewidth}
        \includegraphics[width=\textwidth, trim={0cm 0.3cm 0cm 0.3cm},clip]{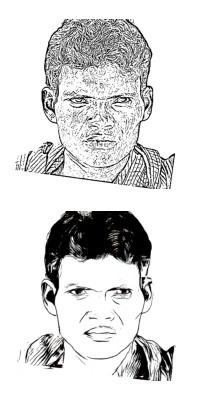}
    \end{subfigure}
    \begin{subfigure}[b]{0.20\linewidth}
        \includegraphics[width=\textwidth, trim={0cm 0.3cm 0cm 0.3cm},clip]{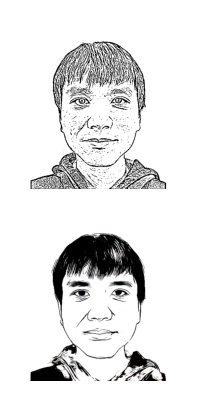}
    \end{subfigure}
    \vspace{\capmargin}
    \caption{\textbf{Learned task seperation}. Results on APDrawing~\cite{yi2019apdrawinggan} after XDoG (top) and then after convolution (bottom) are shown.
    }
    \label{fig:apdrawing_2stages}
    }
\end{figure}

\begin{figure}[t]
\captionsetup[subfigure]{justification=centering}
\centering
\begin{subfigure}[b]{0.16\textwidth}
    \includegraphics[width=\textwidth]{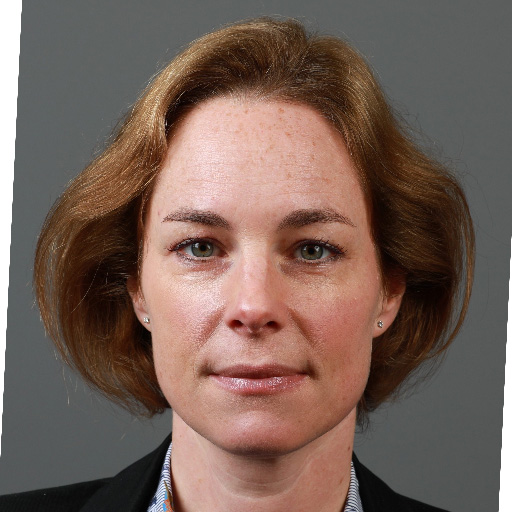}
\end{subfigure}
\begin{subfigure}[b]{0.16\textwidth}
    \includegraphics[width=\textwidth]{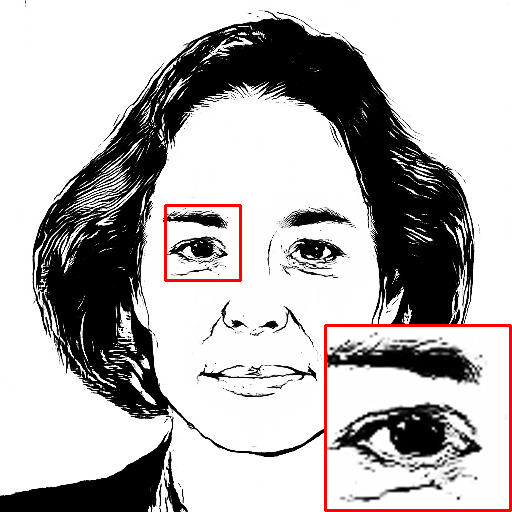}
\end{subfigure}
\begin{subfigure}[b]{0.16\textwidth}
    \includegraphics[width=\textwidth]{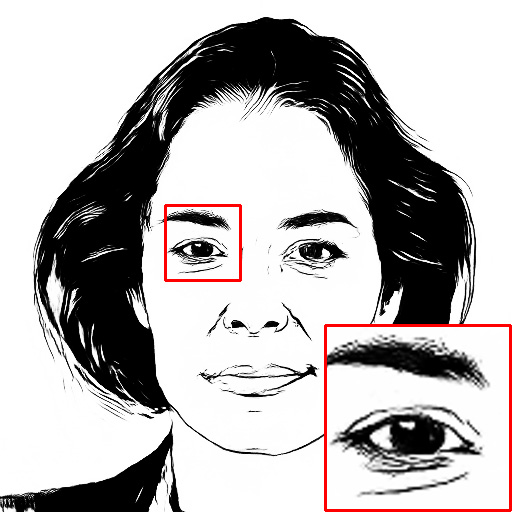}
\end{subfigure}
\begin{subfigure}[b]{0.16\textwidth}
    \includegraphics[width=\textwidth]{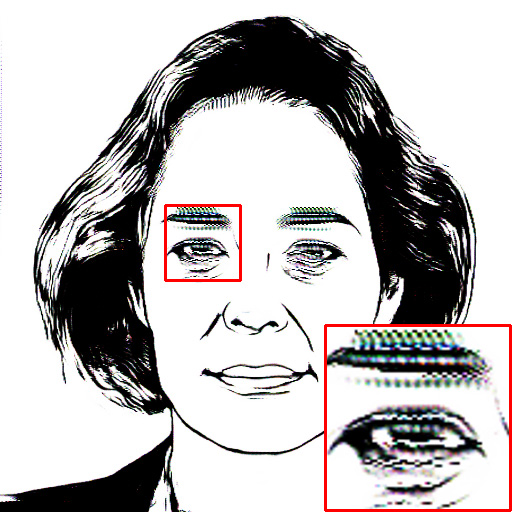}
\end{subfigure}
\begin{subfigure}[b]{0.16\textwidth}
    \includegraphics[width=\textwidth]{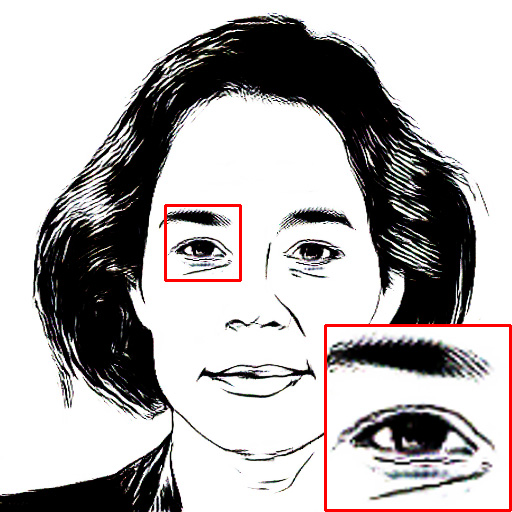}
\end{subfigure}
\begin{subfigure}[b]{0.16\textwidth}
    \includegraphics[width=\textwidth]{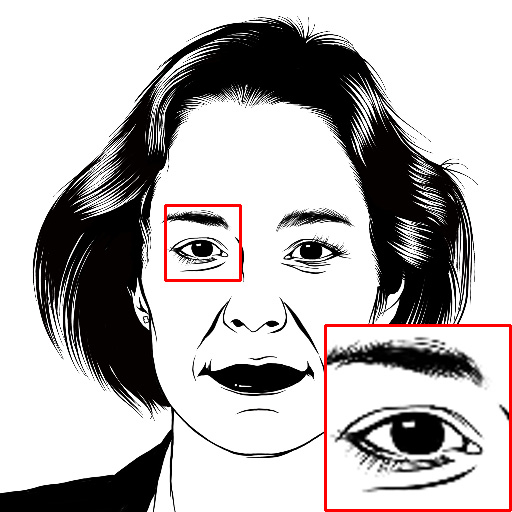}
\end{subfigure}

\begin{subfigure}[b]{0.16\textwidth}
    \includegraphics[width=\textwidth]{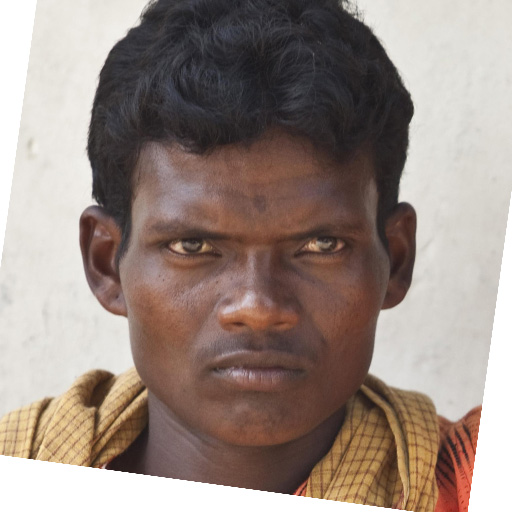}
    \caption{Input\newline}
    \label{fig:apdrawing_comparison:a}
\end{subfigure}
\begin{subfigure}[b]{0.16\textwidth}
    \includegraphics[width=\textwidth]{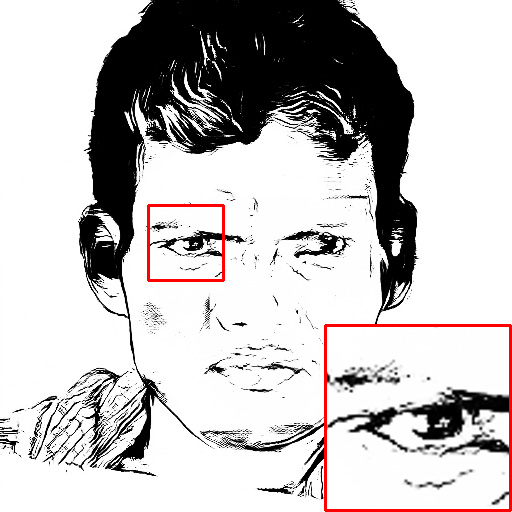}
    \caption{APDraw. -\ac{GAN}~\cite{yi2019apdrawinggan}}
    \label{fig:apdrawing_comparison:c}        
\end{subfigure}
\begin{subfigure}[b]{0.16\textwidth}
    \includegraphics[width=\textwidth]{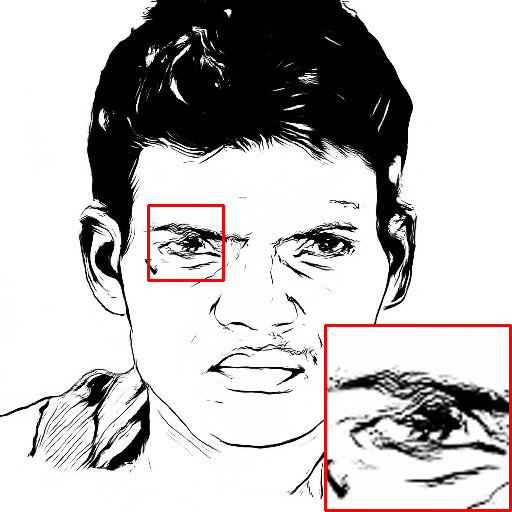}
    \caption{APDraw-ing++~\cite{YiXLLR20}}
    \label{fig:apdrawing_comparison:ap++}        
\end{subfigure}
\begin{subfigure}[b]{0.16\textwidth}
    \includegraphics[width=\textwidth]{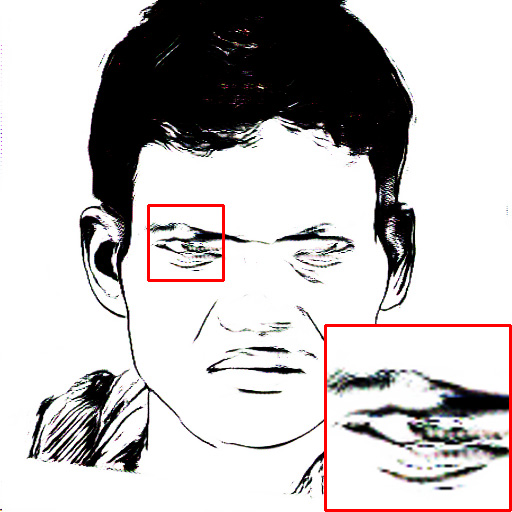}
    \caption{Ours - \ac{CNN}-only}
    \label{fig:apdrawing_comparison:cnnonly}    
\end{subfigure}
\begin{subfigure}[b]{0.16\textwidth}
    \includegraphics[width=\textwidth]{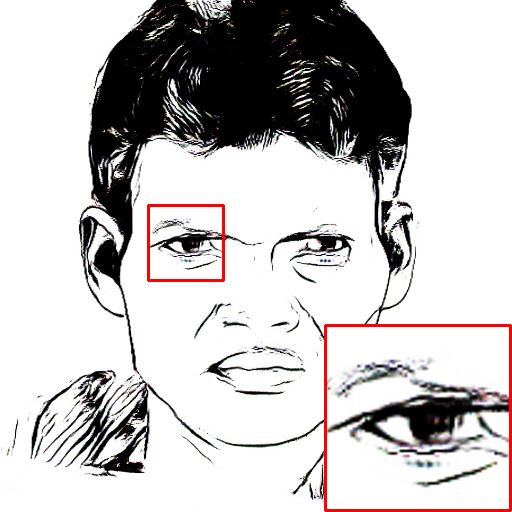}
    \caption{Ours - \ac{XDoG}+\ac{CNN}}
    \label{fig:apdrawing_comparison:ours}        
\end{subfigure}
\begin{subfigure}[b]{0.16\textwidth}
    \includegraphics[width=\textwidth]{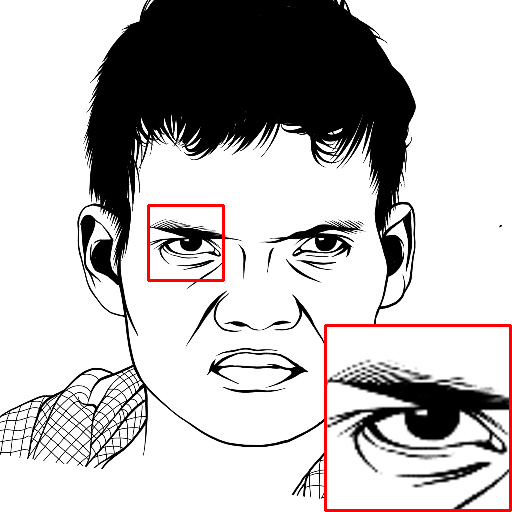}
    \caption{Ground truth}
    \label{fig:apdrawing_comparison:e}        
\end{subfigure}
\caption{\textbf{Results on APDrawing.} While APDrawing \ac{GAN} \protect\subref{fig:apdrawing_comparison:c} and APDrawing++ \ac{GAN} \protect\subref{fig:apdrawing_comparison:ap++} can produce inconsistent lines, our proposed method \protect\subref{fig:apdrawing_comparison:ours} generally produces flow-consistent lines. The differentiable filter in our approach is important for consistent quality, as solely using an image translation \ac{CNN} \cite{johnson2016perceptual} often produces local dithering artefacts (upper row) and patchy features (lower row)  \protect\subref{fig:apdrawing_comparison:cnnonly}.
}
\label{fig:apdrawing_comparison}

\end{figure}

\paragraph{Architecture.}

We investigate the efficacy of each component in our proposed approach for APDrawing in \Cref{tbl:ap_drawing_ablation_study}. Following Yi~\etal\cite{YiXLLR20}, we measure the \ac{FID} score~\cite{heusel2017gans} and \ac{LPIPS} \cite{zhang2018perceptual} to the test set. We train for 200 epochs and otherwise use the same hyperparameters as Pix2Pix~\cite{isola2017image}. 
We observed that using the ResNet-based architecture for image translation introduced by Johnson~\etal\cite{johnson2016perceptual} works best for the post-processing CNN. Furthermore integrating the \ac{XDoG} in the pipeline improves the results \vs a convolutional-only pipeline. Note that this combination of algorithmic effects and CNNs in a training pipeline is only made possible by our introduced approach for end-to-end differentiable filter pipelines and PPNs.
Omitting the \ac{PPN} and using fixed parameters for \ac{XDoG} significantly degrades the results, which validates the integrated training of filter and CNN.  \Cref{fig:apdrawing_2stages} visualizes the results after the \ac{XDoG} stage and after post-processing using a \ac{CNN}; the learned separation of edge detection and abstraction is apparent.
Further, we observe that training with \ac{XDoG} as a postprocessing instead of as a preprocessing step does not converge. 
All architecture choices are extensively evaluated in an ablation study, please refer to the supplemental material.

\paragraph{Results.}
While the \ac{CNN} alone (without \ac{XDoG}) already achieves good \ac{FID} and \ac{LPIPS} scores, we show in \Cref{fig:apdrawing_comparison} that it creates major artifacts especially around eyebrows and eyes, which are not detected by those metrics. 

\begin{wraptable}{r}{6.5cm}
    \caption{Our results on APDrawing~\cite{yi2019apdrawinggan}
    }\label{tbl:ap_drawing_ablation_study}
    \begin{center}
    \begin{threeparttable}
    \begin{tabular}{| l l l c c |}
    \hline
    \scriptsize{\ac{PPN}} & \scriptsize{\ac{XDoG}}  & CNN & \ac{FID} & \ac{LPIPS}\\
    \hline
    \ding{55} & \ding{55} & U-Net  & 71.26 & 0.322\\
    \ding{55} & \ding{55} & ResNet & 62.44 & \textbf{0.275}\\
    \hline
    
    \ding{55}$\tnote{1} $ & \ding{51} & U-Net & 75.40 & 0.329\\
    \ding{55}$\tnote{1} $ & \ding{51} & ResNet & 71.56 & 0.305\\
    
    \hline
    
    \ding{51} & \ding{51} & U-Net & 89.93 & 0.366\\
    \ding{51} & \ding{51} & ResNet & \textbf{60.55} & 0.285\\
    
    \hline
    \hline
    
    \multicolumn{3}{|c}{APDrawing \ac{GAN}} & 62.14\tnote{2} & 0.291\tnote{2} \\ 
    \multicolumn{3}{|c}{APDrawing++} & \textbf{54.40} & \textbf{0.258}\tnote{2} \\
    \multicolumn{3}{|c}{Train \vs Test} & 49.72 & -\\
    \hline
    \end{tabular}
           \begin{tablenotes} 
          \footnotesize
          \item[1] a fixed parameter preset is used  
          \item[2] results obtained from \cite{yi2019apdrawinggan}\cite{YiXLLR20} 
        \end{tablenotes}
      \end{threeparttable}
    \end{center} \vspace{-1.0cm}
    \end{wraptable}

Compared to the APDrawing \ac{GAN} approach by Yi~\etal\cite{yi2019apdrawinggan}, our model improves the \ac{FID} score (\Cref{tbl:ap_drawing_ablation_study}). 
The state-of-the-art APDrawing++ \cite{YiXLLR20} improves on these metrics and quantitatively performs better than our model,  however qualitatively it can suffer from artifacts in small structures such as the eyes  (\Cref{fig:apdrawing_comparison:ap++}) whereas our approach leads to more consistent lines.
We note that their approach consists of a sophisticated combination of several losses and task-specific discriminators that require facial landmarks to train multiple local generator networks for facial features such as eyes, nose, and mouth separately.
This limits their generalizability to other datasets, while our approach, on the other hand, represents a general setup for image-to-image translation consisting of a globally trained \ac{CNN} and a simple effect, making it applicable to any paired training data without further annotation requirements.


\section{Discussion}
\label{sec:Discussion}

\paragraph{Applicability.}%
In the previous sections, we have demonstrated the applicability of differentiable filters to several example-based stylization tasks using four established heuristics-based filter pipelines. 
Their constituent image filters (\Cref{tab:filter_stages}) form a common basis of many image-based artistic rendering approaches \cite{kyprianidis2012state}. We expect that other filtering-based effects, such as pencil-hatching \cite{Lu2012hatching}, or stippling \cite{son2011structure}, can be transferred to our framework with relative ease due to their pipeline-based, GPU-optimized formulations. Stroke-based rendering approaches, on the other hand, are typically optimized globally \cite{nakano2019neural} or locally \cite{hertzmann1998painterly}, and are thus challenging to transform into differentiable formulations.
However, a recent approach by Liu \etal\cite{liu2021paint} has shown that strokes can be predicted in a single feedforward pass of a \ac{CNN}, which could be regarded as a complementary approach.
 
\paragraph{Limitations.}%
Our \ac{PPN}-based approaches make use of a paired data training regime. While paired data can be synthetically generated for content-adaptive effects aiming at solving filter-specific problems, datasets with paired real-world paintings are subject to limited availability. As our training approach follows Pix2Pix GAN \cite{isola2017image}, future work extending the method to train with unpaired training losses, such as cycle-consistency losses \cite{zhu2017unpaired}, could alleviate this limitation.
An inherent limitation of predicting parameters in comparison to directly predicting pixels (as with convolutional \acp{GAN}), remains the constraint of only being able to produce styles that lie in the manifold of achievable effects of the underlying image filters. While this can be mitigated using a post-processing \ac{CNN}, this represents a trade-off with respect to interpretability and range of low-level control (we examine this aspect in the supplemental material). On the other hand, our parametric style transfer is able to match arbitrary styles when optimizing highly parameterized effects such as watercolor. Training a PPN with such an effect on a large dataset, \eg using unpaired training, could similarly already have sufficient representation capability without postprocessing CNNs.


\section{Conclusions}
\label{sec:Conclusions}

In this work, we propose the combination of algorithmic stylization effects and example-based learning by implementing heuristics-based stylization effects as differentiable operations and learning their parametrizations.  
The results show that both optimization of parameters, \eg to achieve style transfers, and their global and local prediction, \eg for content-adaptive effects, are viable approaches for example-based algorithmic stylizations. Our experiments demonstrate that our approach is especially suitable for applications that require fast adaptation to new styles while retaining full artistic control and low computation times for high image resolutions. 
Furthermore, stylizations beyond the filters' abstraction capabilities are achieved by adding convolutional post-processing. This approach can generate results on-par with state-of-the-art \ac{CNN}-based methods. 
For future research, learning the composition of filters as building blocks of a generic algorithmic effect pipeline would allow for seamless integration of user control and example-based stylization without the limitation to a specific stylization technique.

\paragraph{Acknowledgments.}
We thank Sumit Shekhar and Sebastian Pasewaldt for comments and discussions. This work was partially funded by the BMBF through grants 01IS18092 (\enquote{mdViPro}) and 01IS19006 (\enquote{KI-LAB-ITSE}).

\FloatBarrier
{\small
\bibliographystyle{ieee_fullname}
\bibliography{main}
}

\begin{acronym}
\acro{ANE}{Apple Neural Engine}
\acro{API}{Application Programming Interface}
\acro{CG}{Computer Graphics}
\acro{CNN}{Convolutional Neural Network}
\acro{CPU}{Central Processing Unit}
\acro{CV}{Computer Vision}
\acro{DL}{Deep Learning}
\acro{DNN}{Deep Neural Network}
\acro{EPE}{Endpoint Error}
\acro{ETF}{Edge Tangent Flow}
\acro{FPS}{Frames per second}
\acro{GPU}{Graphics Processing Unit}
\acro{GAN}{Generative Adversarial Network}
\acro{K}{Kilo, thousand}
\acro{M}{Million}
\acro{MB}{Megabyte}
\acro{NST}{Neural Style Transfer}
\acro{NPR}{Non-photorealistic Rendering}
\acro{NPU}{Neural Processing Unit}
\acro{ReLU}{Rectified Linear Unit}
\acro{TFLOPS}{$10^{12}$ Floating Point Operations Per Second}
\acro{PPN}{Parameter Prediction Network}
\acro{DoG}{Difference-of-Gaussians}
\acro{XDoG}{eXtended difference-of-Gaussians}
\acro{SSIM}{Structural Similarity Index Measure}
\acro{PSNR}{Peak Signal-to-Noise Ratio}
\acro{FID}{Fréchet Inception Distance}
\acro{LPIPS}{Learned Perceptual Image Patch Similarity}
\acro{BLADE}{Best Linear Adaptive Enhancement}
\acro{STROTSS}{Style Transfer by Relaxed Optimal Transport and Self-Similarity}
\end{acronym}


\newpage
\phantomsection
    \refstepcounter{chapter}%
    \stepcounter{section}%
    \setcounter{section}{0}%
    \setcounter{subsection}{0}%
    \setcounter{figure}{0}
    \setcounter{table}{0}
    \setcounter{equation}{0}
    \setcounter{footnote}{0}%
    \begin{center}%
        \let\newline\\
        {\Large \bfseries\boldmath
            \pretolerance=10000
            Supplemental Material \par}\vskip .8cm
    \end{center}%

Due to space restrictions, some details had to be omitted from the main paper; we present those details here. In \Cref{sup:sec:differentiableFilters} we describe filter modifications to obtain gradients using auto-grad enabled frameworks on the example of the seperated orientation-aligned bilateral filter and color quantization, and elaborate on filter pipelines concerning  their structure, learnability and memory consumption. In \Cref{sup:sec:PPN}, we elaborate on training global \acp{PPN} by presenting three \ac{PPN} architectures and evaluating their performance in an ablation study.  In \Cref{sup:sec:STstyletransfer} we give visual examples for the style transfer capability of different effects. In \Cref{sup:sec:apdrawing_gan}, we provide additional details and results for the architecture ablation study on image-to-image translation tasks using the APDrawing dataset\cite{yi2019apdrawinggan} as well as editability of such effects. Finally, in \Cref{sup:sec:AdditionalResults}, additional results for parametric style transfer, image-to-image translation, and effect variants are shown. Further, our  supplemental video\footnote{\href{https://youtu.be/wIndN7cr0PE}{https://youtu.be/wIndN7cr0PE}} demonstrates parametric style transfer optimization and interactive editing for the images shown in the main paper.

\section{Differentiable Filters}
\label{sup:sec:differentiableFilters}
In the main paper we note that while most filters are straight-forward to implement in auto-grad enabled frameworks, for some several filters re-formulations or approximations are needed.
 Specifically, for structure-adaptive neighborhood operations, a re-formulation to retain sub-gradients in adaptive kernel neighbourhoods must be employed, and for non-differentiable operations a numeric gradient approximation is needed. In the following, we show an example for each.

\subsection{Bilateral Filter} 

Commonly used techniques in image filtering pipelines are bilateral filters. The bilateral filter \cite{tomasi1998bilateral} using Gaussians $G$ of size $\sigma_d$ (distance kernel) and $\sigma_r$ (range kernel) for an image \(I\in\mathbb{R}^{C\times W\times H}\) is defined for a pixel coordinate $x$ by  
\begin{equation}
    \hat{I}(x) = \frac{1}{W} \sum_{y\in\Omega(x)}I(y)G_{\sigma_{d}}\left(\lVert y-x\rVert\right)G_{\sigma_{r}}\left(\lVert I(y)-I(x)\rVert\right)
\end{equation}

where $\Omega(x)$ denotes the window centered in $x$ and $y\in \Omega$ represent pixel coordinates of the kernel neighbourhood. $W$ represents the weight normalization term, which we omit in the following for the sake of brevity. Computing the bilateral filter for large images and at large kernel neighbourhoods is, however, computationally expensive. Several approaches have been proposed to approximate the filter by separation into multiple passes for improved efficiency.

For image abstraction and stylization, the \textit{orientation-aligned separated bilateral filter}\cite{kyprianidis2008image,kang2009flow} has shown a good trade-off between quality and performance.
The filter is guided in two passes along the gradient and tangent direction of an edge tangent field. The tangent field of the input image is obtained by an eigenanalysis of the smoothed structure tensor \cite{brox2006adaptive}; for more details the reader is kindly referred to the work of Kyprianidis and D{\"o}llner \cite{kyprianidis2008image}.
The tangent field computation consists of point-based and fixed-neighbourhood kernels only, and thus is straightforward to implement using common auto-differentiable functions. However, the following separated bilateral filter is iterative and structure adaptive (\ie the size of the kernel neighbourhood depends on the content), and thus cannot be ported to the fixed-neighbourhood functions (\eg convolutions) of auto-grad enabled frameworks in a straightforward way. 

Specifically, in the work of Kyprianidis and D{\"o}llner \cite{kyprianidis2008image}, the filter response at a sampling point $x$ of a pass of the separated orientation-aligned bilateral filter in direction $t$ (gradient or tangent direction) is defined as:
\begin{equation}
    \hat{I}(x) = I(x) + \sum_{y\in\Omega_t(x)}I(y)G_{\sigma_{d}}\left(\lVert y-x\rVert\right)G_{\sigma_{r}}\left(\lVert I(y)-I(x)\rVert\right)
    \label{sup:eq:orientationaligned}
\end{equation}
 where \(\Omega_t(x) = \{x+kt,|\,k\in[-N,N] \wedge k\in \mathbb{Z}\}\) represents sampling positions along the direction $t$ defined in $x$, and $N$ denotes the cut-off kernel radius. $N$ is typically based on $\sigma_{d}$, \eg set to $\lfloor\sfrac{2\sigma_{d}}{\Vert t\Vert}\rfloor$, and the kernel size locally depends on the magnitude of the direction vector $t$. 

It would be possible to implement a custom kernel with associated backward pass for the above to manually compute (sub-) gradients for both input and parameters, and similarly repeat the procedure for every other structure-adaptive filter in \Cref{tab:filter_stages} of the main paper. However, as our goal is to minimize the effort of porting existing (shader-based) filters to our framework while maximizing reusability and portability, we propose to implement the filters using common auto-differentiable components.
To this end, we reformulate \Cref{sup:eq:orientationaligned} into a grid sampling-based operation.  Specifically, we use spatial coordinate grids \(C \in\mathbb{R}^{2 \times W\times H }\)  and \(T\in\mathbb{R}^{ W\times H \times 2D}\) that represent the mapping of output pixel locations to input pixels for grid sampling. Grid $C$ is the identity mapping, \ie contains coordinates $C_{wh} = (w,h)$. Grid $T$ contains the sampling offsets in dimension $D$: \(T_{wh} =\{kt_{wh},|\,k\in[-D,D] \wedge k\in \mathbb{Z}\}\). Then the neighbourhood sampling values \(V\in\mathbb{R}^{C\times W\times H \times 2D}\) for the entire image $I$ are computed as:
\begin{equation}
   V = \mathcal{I}(C+T)\delta\left[G_{\sigma_{d}}\left(\lVert T\rVert\right)\right]  G_{\sigma_{r}}\left(\lVert \mathcal{I}(C+T)-I\rVert\right)
    \label{eq:bilateral_unfolded_equation}
\end{equation}
where in slight abuse of notation, $\mathcal{I}(C+T)$ denotes bilinear grid sampling \cite{jaderberg2015spatial} of $I$ over the coordinate grid of unfolded kernels. It is thus equivalent to retrieving $I(\Omega_t(x))$ for every pixel $x$ with fixed number of samples.
To clip values outside of the adaptive kernel neighborhood, $\delta$ is computed as: 
\begin{equation}
    \delta_{whd}(v) = 
    \begin{cases}
    v & \text{if } \lVert T_{whd} \rVert \leq N_{wh}\\
    0,              & \text{otherwise}
    \end{cases}
\end{equation}
Note that multiplications between tensors in \Cref{eq:bilateral_unfolded_equation} are performed elementwise. Filter responses in dimension $D$ are then folded (summed) to obtain the response over the full kernel neighborhood (\ie the filtered image $\hat{I}$):

\begin{equation}
     \hat{I} = I + \sum_{d=1}^D V_d
    \label{eq:bilateral_folded_equation}
\end{equation}
 
 The size of dimension $D$ can be varied for performance-quality trade-offs (as shown in Fig.4 in the main paper).

\subsection{Color Quantization} 
Color quantization is often applied to achieve a flat, cartoon-like impression (\Cref{fig:toon_overview_pipeline}, main paper). 
It is usually defined using the floor function: $y = \sfrac{\lfloor{xb} \rfloor + 0.5}{b}$~\cite{winnemoller2006real},
where $x$ denotes the color value before and $y$ the color value after quantization, and $b$ (number of quantization bins) is the parameter that should receive gradients. 
Due to the non-differentiable nature of the floor function, we introduce a differentiable approximation. Common differentiable proxies like the straight-through estimator\cite{bengio2013estimating} work well for approximating the gradient of quantization functions with fixed numbers of quantization bins. However, for color quantization we want to differentiate with respect to the strength of the quantization or the number of bins, which is not possible using common differentiable proxies.
As reasoning about the number of quantization bins requires information about the complete image, a global optimization is formulated instead:
\begin{equation}
f(r)=\sum_i \sign\bigg(\bm{y}_i-\frac{\lfloor r \bm{x}_i \rfloor + 0.5}{r}\bigg)\sign(\bm{g}_i)
\label{eq:objective_function}
\end{equation}
where $\bm{x}$ denotes a vector containing the input pixel values, $\bm{y}$ the  outputs of $y(\bm{x})$, and $\bm{g}$ the gradient vector for $\bm{y}$. 

Intuitively, $f$ sums values for each pixel, which are positive if the pixel's gradient points in the direction of $r$ and negative otherwise. The maximum number of per-pixel gradients should point in the direction of $r$.

We choose $\hat{b} = \argmax_r f(r)$ to derive the optimal value for the parameter $b$, which minimizes the learning target. This value is obtained in a single optimization step.
To obtain a continuous gradient, the difference $b-\hat{b}$ is scaled by the gradient magnitude $|g_i|$ of the loss function computed with respect to all pixels $y_i\in\bm{y}$, such that gradients for $b$ decrease, once the optimal value is approached:
\begin{equation}
\frac{\partial y}{\partial b}\coloneqq \sum_i{|g_i|(b-\hat{b})}
\end{equation}

\begin{figure}
    \centering
    \includegraphics[width=\linewidth]{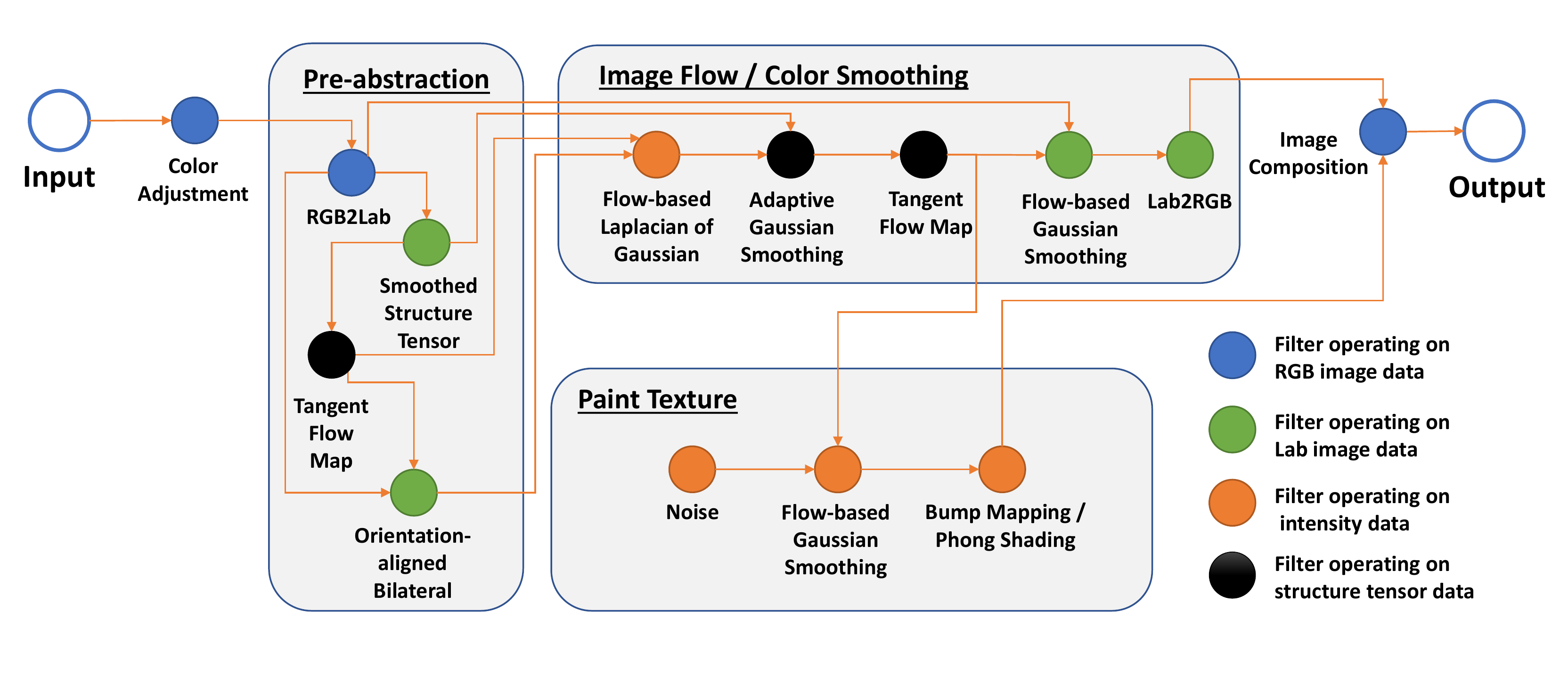}
    \caption{Oilpaint filter pipeline adapted from Semmo~\etal\cite{semmo2016image}. }
    \label{fig:oilpaint_pipeline_wise}
\end{figure}

\subsection{Differentiable filter pipeline - Oilpaint}

As noted in the main paper, we implement differentiable filter pipelines analogous to their originally published versions and show an example for the toon pipeline in \Cref{fig:toon_overview_pipeline} of the main paper. In  \Cref{fig:oilpaint_pipeline_wise} we further show the filter pipeline of the oilpaint effect \cite{semmo2016image}. In contrast to Semmo~\etal we do not use color palette extraction and color quantization and instead use a color adjustment layer.

\begin{table}[h]
\caption{Parameter optimization to compare learned and target stylizations. 
Scores are computed on \ac{NPR} benchmark \cite{rosin2020nprportrait} for both high-quality (level I) and low-quality (level III) portraits.}
\label{tbl:benchmark_full_p}
\begin{center}
\begin{tabular}{| l | c c | c c |}
\hline
& \multicolumn{2}{c|}{NPR level I} & \multicolumn{2}{c|}{NPR level III} \\
\cline{2-5}
Loss & SSIM & PSNR & SSIM & PSNR \\
\hline
Cartoon $\ell_2$ & 0.937 & 28.144 & 0.958 & 34.124  \\
Cartoon $\ell_1$ & 0.931 & 26.939 & 0.947 & 30.429  \\
Watercolor $\ell_2$ & 0.922 & 30.634 & 0.897 & 32.342 \\
Watercolor $\ell_1$ & 0.883 & 31.032 & 0.895 & 29.048 \\
\hline
\end{tabular}
\end{center}
\end{table}

\subsection{Functional Benchmark}
To verify that all parameters of an effect pipeline can be learned, we set up a functional benchmark to evaluate how well effects can be adapted to an example stylization in the same domain.
We developed an OpenGL-based \enquote{ground truth} implementation to generate reference stylizations, where parameter values are randomized during the benchmark.  
Using an image-based loss, gradients are then backpropagated to optimize the parameters of the differentiable effect using gradient descent. 
We measure the \ac{SSIM} and \ac{PSNR}. 
 \Cref{tbl:benchmark_full_p} shows that both the cartoon and watercolor effect achieve very high similarity to their reference using both $\ell_1$ and $\ell_2$ losses. The photographic quality level of the input images does not seem to play a role in the style-matching capability. 
As the \ac{XDoG} filter is contained in both cartoon and watercolor pipelines, it is not evaluated separately.

\subsection{Implementation Aspects - Memory}
\begin{figure}[t]
\centering
\begin{subfigure}[t]{0.3\linewidth}
    
     \begin{tikzpicture}
        \begin{scope}[
        node distance = 11mm,
            inner sep = 0pt,spy using outlines={rectangle, red, magnification=4, every spy on node/.append style={thick} }
          ]
    \node (n0)  { \includegraphics[width=\textwidth,trim={0cm 10cm 0cm 4cm},clip]{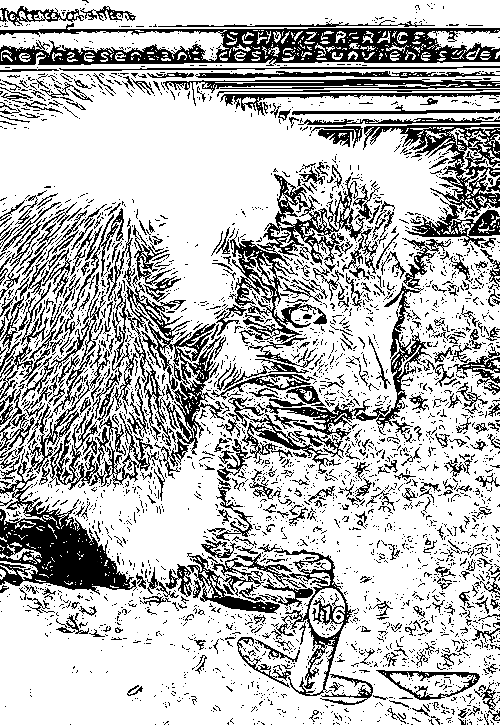}};
    \spy [blue,size=1.3cm] on (0.3,-0.25) in node[below right=of n0.north west] at (-2.4,0.985);
    \end{scope}
    \draw[dashed,blue,line width=0.8pt] (tikzspyonnode.south east) -- (tikzspyinnode.south east);
    \draw[dashed,blue,line width=0.8pt] (tikzspyonnode.north west) -- (tikzspyinnode.north west);
    \end{tikzpicture}

    \caption{\SI{0.98}{\gibi\byte}, $\SI{0.12}{\second}$}
    \label{fig:dim_kernsize_parameter:a}
\end{subfigure}    
\begin{subfigure}[t]{0.3\linewidth}
    \begin{tikzpicture}
        \begin{scope}[
        node distance = 11mm,
            inner sep = 0pt,spy using outlines={rectangle, red, magnification=4, every spy on node/.append style={thick} }
          ]
    \node (n0)  { \includegraphics[width=\textwidth,trim={0cm 10cm 0cm 4cm},clip]{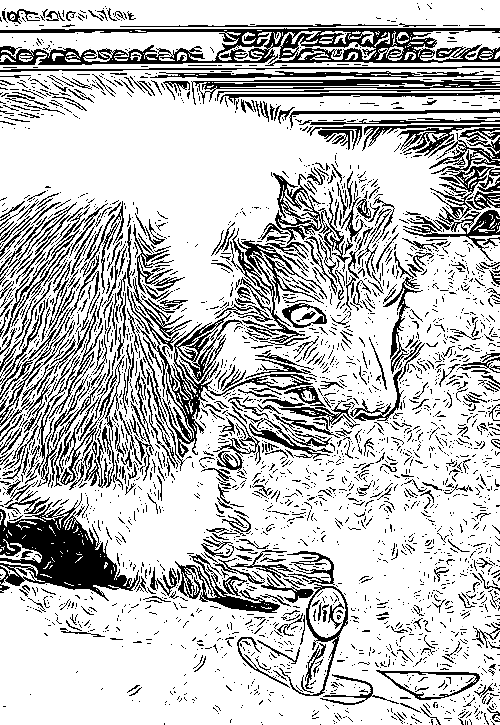}};
    \spy [blue,size=1.3cm] on (0.3,-0.25) in node[below right=of n0.north west] at (-2.4,0.985);
    \end{scope}
    \draw[dashed,blue,line width=0.8pt] (tikzspyonnode.south east) -- (tikzspyinnode.south east);
    \draw[dashed,blue,line width=0.8pt] (tikzspyonnode.north west) -- (tikzspyinnode.north west);
    \end{tikzpicture}
    \caption{\SI{1.51}{\gibi\byte}, $\SI{0.17}{\second}$}
    \label{fig:dim_kernsize_parameter:b}
\end{subfigure}  
\begin{subfigure}[t]{0.3\linewidth}
     \begin{tikzpicture}
        \begin{scope}[
        node distance = 11mm,
            inner sep = 0pt,spy using outlines={rectangle, red, magnification=4, every spy on node/.append style={thick} }
          ]

    \node (n0)  { \includegraphics[width=\textwidth,trim={0cm 10cm 0cm 4cm},clip]{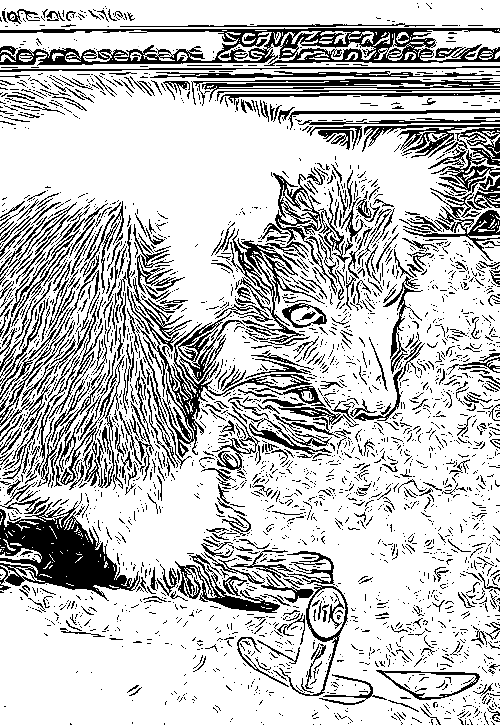}};
    \spy [blue,size=1.3cm] on (0.3,-0.25) in node[below right=of n0.north west] at (-2.4,0.985);
    \end{scope}
    \draw[dashed,blue,line width=0.8pt] (tikzspyonnode.south east) -- (tikzspyinnode.south east);
    \draw[dashed,blue,line width=0.8pt] (tikzspyonnode.north west) -- (tikzspyinnode.north west);
    \end{tikzpicture}
    \caption{OpenGL reference}
    \label{fig:dim_kernsize_parameter:c}
\end{subfigure} 

\caption{GPU VRAM and run-time in seconds for varying kernel sizes $D=1$ \protect\subref{fig:dim_kernsize_parameter:a} and $D=5$ \protect\subref{fig:dim_kernsize_parameter:b} for XDoG. 
(Image resized to 1MP from \cite{Brummer_2019_CVPR_Workshops}). 
}
\label{fig:dim_kernsize_parameter}
\end{figure}

During training, limiting the memory consumption of individual filters is important, as they operate in the full input resolution. Generally, filters that accumulate sampling of multiple different locations in a single tensor have the highest memory usage. For our filters, wet-in-wet stylization, Kuwahara, and bilateral filtering are the most expensive with \SIrange{3}{5}{\giga\byte} peak memory usage for \SI{1}{\mega\pixel} input images. The trade-off between the quality of the generated results and computing resources can be controlled by the kernel size parameter $D$ as shown in \Cref{fig:dim_kernsize_parameter}. To further reduce memory consumption during training, we use gradient checkpointing to recompute intermediate gradients on the fly.  Thereby, activations are stored only at the end of each filter, which is usually a single RGB image. Intermediate activations are re-computed on the fly during back-propagation. The peak memory usage thus then depends on the single most memory-intensive filter. For example, the non-checkpointed XDoG uses 4GB of GPU memory at peak for 1MP resolution input, while the checkpointed version uses 2GB. The loss in speed is small and can be mitigated through the larger batch size that can be processed with the available memory.

\FloatBarrier
\section{Global \acs{PPN}}\label{sup:sec:PPN}

This section provides details on the global \ac{PPN}, as well as an ablation study for architecture variants and losses. 

\subsection{Architecture}

\begin{figure*}[h]
\centering
\begin{subfigure}{0.7\textwidth}
    \includegraphics[width=\textwidth]{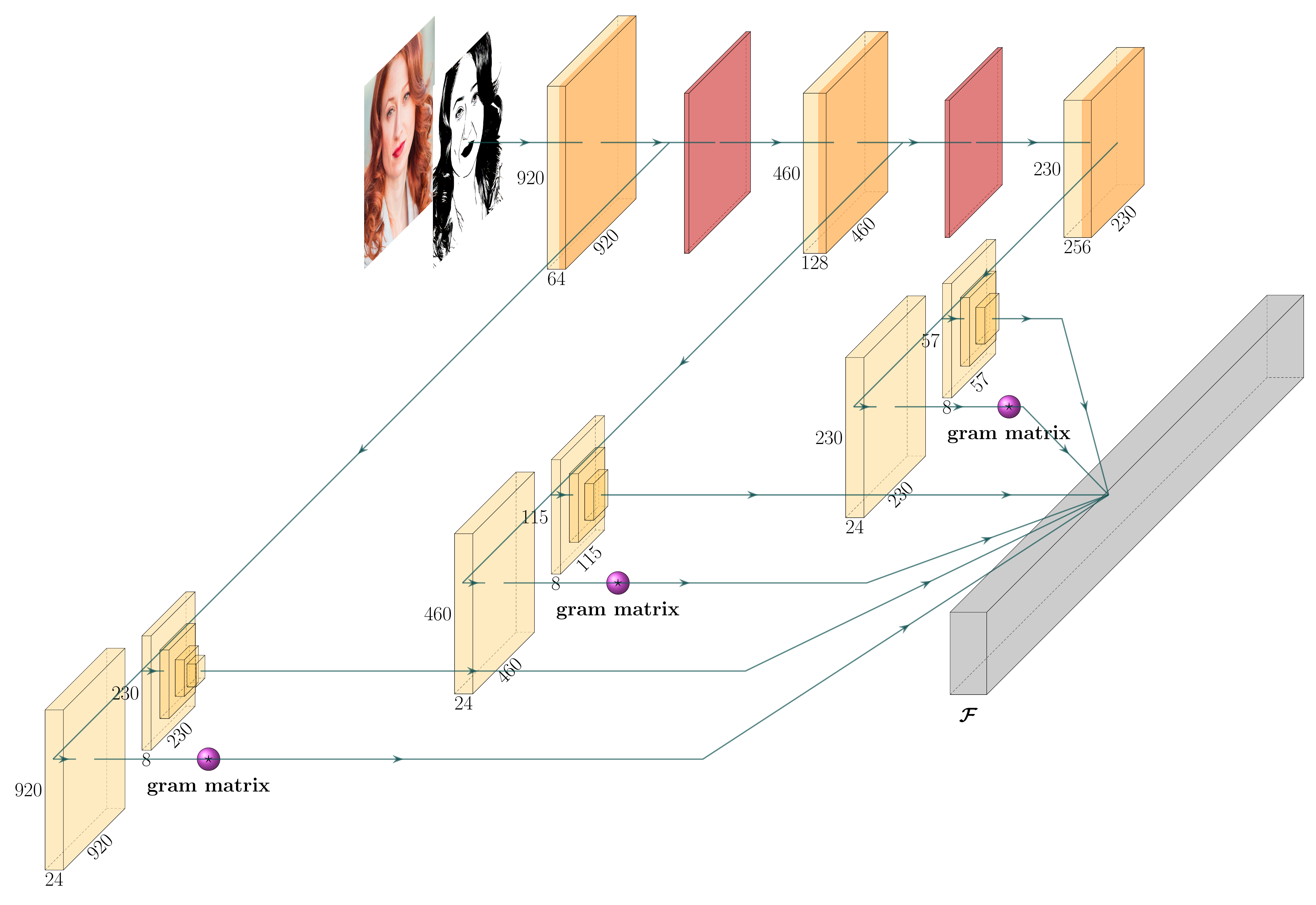}
    \caption{Multi-feature extractor}
    \label{sup:fig:multi_feature}
\end{subfigure}
\\
\begin{subfigure}{0.7\textwidth}
    \centering
    \begin{minipage}{.1cm}
    \vfill
    \end{minipage}
    \includegraphics[trim={1cm 0cm 0 0},clip,width=1\textwidth]{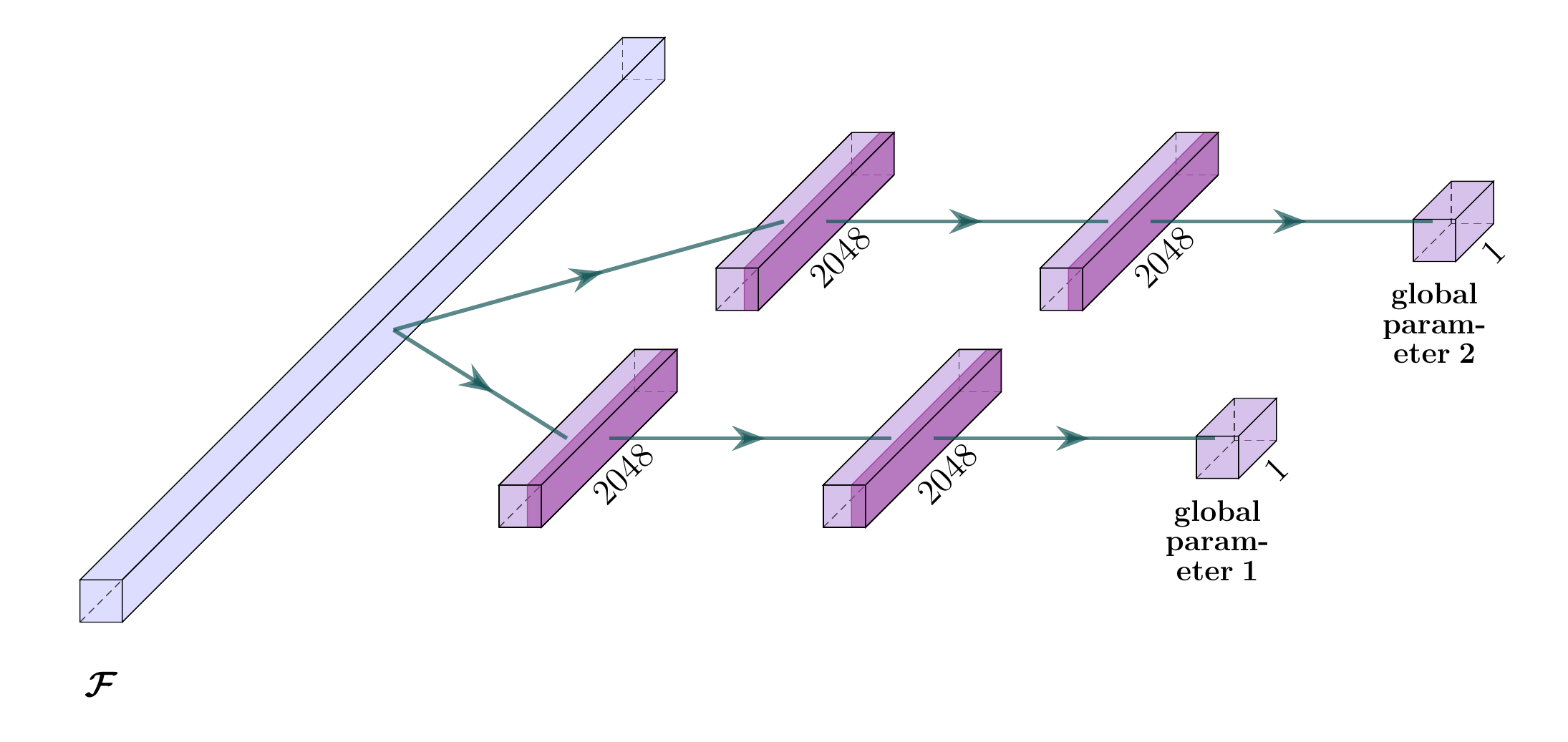} 
    \caption{Multi-head module}
    \label{sup:fig:multi_head}
\end{subfigure}
\caption{\textbf{Multi-feature extractor \protect\subref{sup:fig:multi_feature}:} Features are extracted from all convolution stages of a pre-trained VGG-11 network. We only show the first 3 convolution stages of the VGG-11 here for brevity. The features are passed through small stacks of strided convolutions with stride 4 and kernel size $4\times4$ pixels to generate compact representations. In parallel, the same features are passed through a single $1\times1$ pixel convolution and a subsequent operation to calculate the Gram matrices. All compact features $\mathcal{F}$ are concatenated. Convolution operations are visualized in yellow, ReLU layers in orange, and pooling layers in red. \textbf{Multi-head module \protect\subref{sup:fig:multi_head}:} After the extraction of all features, we continue with multiple streams of linear layers. A separate stream of 3 linear layers is created for each global parameter: The first two layers process the features and have ReLU activations. The last layer outputs the final prediction as a single number. Fully connected layers are visualized in light violet and ReLU layers in dark violet. 
} 
\label{sup:fig:multifeature_multihead}
\end{figure*}

The network architecture receives two images: an input image and a stylized image, which has been generated with unknown parameters. We develop and benchmark three network architectures for parameter prediction, as follows:

\begin{enumerate}
\item \textbf{SimpleNet}: 
A simple convolution network with three Conv-ReLU layers (channel count: 16,32,64) followed by MaxPooling layers. All features are concatenated at the end and transformed with an additional linear layer to yield the global parameters. Input and stylized image are concatenated along the channel dimension and subsequently passed through the network. 
\item \textbf{ResNet + Multi-head}: 
A ResNet50 architecture \cite{he2016deep} without BatchNormalization layers is used to extract features. 
We initialize the ResNet with pre-trained weights on the ImageNet dataset \cite{deng2009imagenet}. 
Again, input and stylized image are concatenated along the channel dimension before passing them through the network. Features $\mathcal{F}$ are extracted after the last convolution layer and passed to a custom multi-head module depicted in \Cref{sup:fig:multi_head}. 
The streams of linear layers process the features for each global parameter independently.
\item \textbf{Multi-feature + Multi-head}: 
A VGG backbone without BatchNormalization and pre-trained weights on ImageNet is used. 
Input and stylized image are passed separately through the feature extractor in two passes and all compact features are accumulated afterward. 
We extract features and compute Gram matrices for every convolution step as depicted in \Cref{sup:fig:multi_feature}. 
As the Gram matrices contain essential information about style, we hypothesize that these features help understand the input stylizations.
The accumulated features $\mathcal{F}$ from both images are passed to the multi-head module of \Cref{sup:fig:multi_head} to predict the final global parameters.
\end{enumerate}

The design of the \ac{PPN} architectures is motivated by two assumptions: 
(1) using a multi-head module for independent processing of each parameter should improve the accuracy as parameters for algorithmic effects model 
independent aspects of the stylization process and thus benefit from learning parameter-specific representations; 
(2) extraction of features at multiple scales (multi-feature) should be beneficial for the \ac{PPN}'s performance, as parameter changes can affect larger parts of the image as well as small details. 
As Huang~\etal\cite{huang2017arbitrary} find that normalization layers also perform normalization of style, we suspect that leaving out BatchNormalization in all architectures would yield better results

\FloatBarrier
\subsection{Ablation Study}
We conduct an ablation study for the previously discussed \ac{PPN} architectures and loss functions. 
We use the \ac{XDoG} filter effect \cite{winnemoller2011xdog} and compute the \ac{SSIM} and \ac{PSNR} metric in image space between the predicted and reference stylizations.
We also indicate the mean absolute difference ($\ell_1$) between the predicted and ground truth global parameters. 

\paragraph{Loss Functions.}
For our ablation study, we compare parameter space- and image space losses. 
Parameter space losses are computed from the output of the \ac{PPN} directly (\ie in parameter space), thus the differentiable effect is not used during training. 
For image space losses, the predicted parameters are used to parameterize the differentiable effect, its output is then compared to the target output image using a pixel-wise similarity metric and gradients are back-propagated through the effect back to the \ac{PPN}. 

\paragraph{Network Training.} 
For training, we use a random extract of 7000 images from the FFHQ-dataset of portrait images \cite{karras2019style} at a resolution of $1024 \times 1024$ pixels. For data augmentation, images are randomly cropped to resolution $920 \times 920$ pixels. 
We use Adam \cite{kingma2014adam} with a learning rate of $10^{-5}$ and train for 25 epochs, using a virtual batch size of 64. 
For testing, we use all 60 portrait images of the \ac{NPR} benchmark portrait \cite{rosin2020nprportrait} and randomly generated global parameters. 
The final parameter predictions are passed through a $\tanh$ activation function for all networks and multiplied by 0.5 to yield parameters in the range $[-0.5, 0.5]$.

\paragraph{Results.} 
Results are summarized in \Cref{sup:tbl:xdog_ppn_ablation}. 
Interestingly, the simple network architecture performs better than or similar to the ResNet in most experiments. 
We argue that for the ResNet, too much spatial information gets lost, as features are derived only after all downsampling and convolution operations. 
The multi-feature architecture effectively recovers the ability to derive representative features  at all scales, as shown by Zhang~\etal\cite{zhang2018perceptual}, and outperforms the other two options.
We observe that computing the loss in image space instead of parameter space generally yields better results in terms of \ac{SSIM} and \ac{PSNR}, even though the ground-truth parameters are not directly available to the network in this case. 
As expected, computing the loss in parameter space still leads to a closer approximation of the ground truth parameters as measured by the $\ell_1$ distance between ground truth and predicted parameters. 
If gradients are used as a learning signal, the model learns to use parameters more effectively to achieve better results, sometimes deviating more from the ground truth parameters. 
Some parameter changes within the \ac{XDoG} effect (\eg changing either contour or blackness) can similarly impact the final outcome. 
We reason that the gradients, which are derived from our differentiable effects, enable the model to build a better understanding of how the various parameters influence the final outcome.

\begin{table}
\begin{center}
\caption{Global parameter prediction: ablation study for \ac{XDoG} on FFHQ\cite{karras2019style}. For network variants SimpleNet (SN), ResNet + Multi-head (R+M), Multi-feature + Multi-head (M+M), using image space-based losses $I[{\ell_{1/2}}]$ and parameter space-based losses $P[\ell_{1/2}]$. }
\label{sup:tbl:xdog_ppn_ablation}
\begin{tabular}{|l|l|c|c|c|}
\hline
Network & Loss & \ac{SSIM} & \ac{PSNR} &  $P[\ell_1]$ \\ \cline{3-5}
 & &  \multicolumn{3}{c|}{-- across all NPR levels --} \\
\hline
M+M & $I[{\ell_2}]$ & 0.764 & 13.286 & 0.190\\
SN & $I[{\ell_2}]$ & 0.733 & 12.523 & 0.218 \\
R+M & $I[{\ell_2}]$ & 0.726 & 12.234 & 0.235\\
\hline

M+M & $P[\ell_2]$ & 0.738 & 12.530 & \textbf{0.158} \\
SN & $P[\ell_2]$ & 0.686 & 11.277 & 0.197\\
R+M & $P[\ell_2]$ & 0.677 & 10.874 & 0.205 \\
\hline

M+M & $I[{\ell_1}]$ & \textbf{0.780} & \textbf{13.875} & 0.183 \\
SN & $I[{\ell_1}]$ & 0.728 &  12.407 & 0.227\\
R+M & $I[{\ell_1}]$ & 0.721 & 12.038 & 0.236\\
\hline

M+M & $P[\ell_1]$ &  0.737 & 12.927 & 0.162\\
SN & $P[\ell_1]$ & 0.697 & 11.805 & 0.200 \\
R+M & $P[\ell_1]$ &  0.692 & 11.408 & 0.200 \\
\hline
\end{tabular}
\end{center}
\end{table}

\FloatBarrier

\section{Style Transfer Capability}\label{sup:sec:STstyletransfer}

\begin{figure}[h]
\centering
    \begin{subfigure}[b]{0.24\textwidth}
            \includegraphics[width=\textwidth]{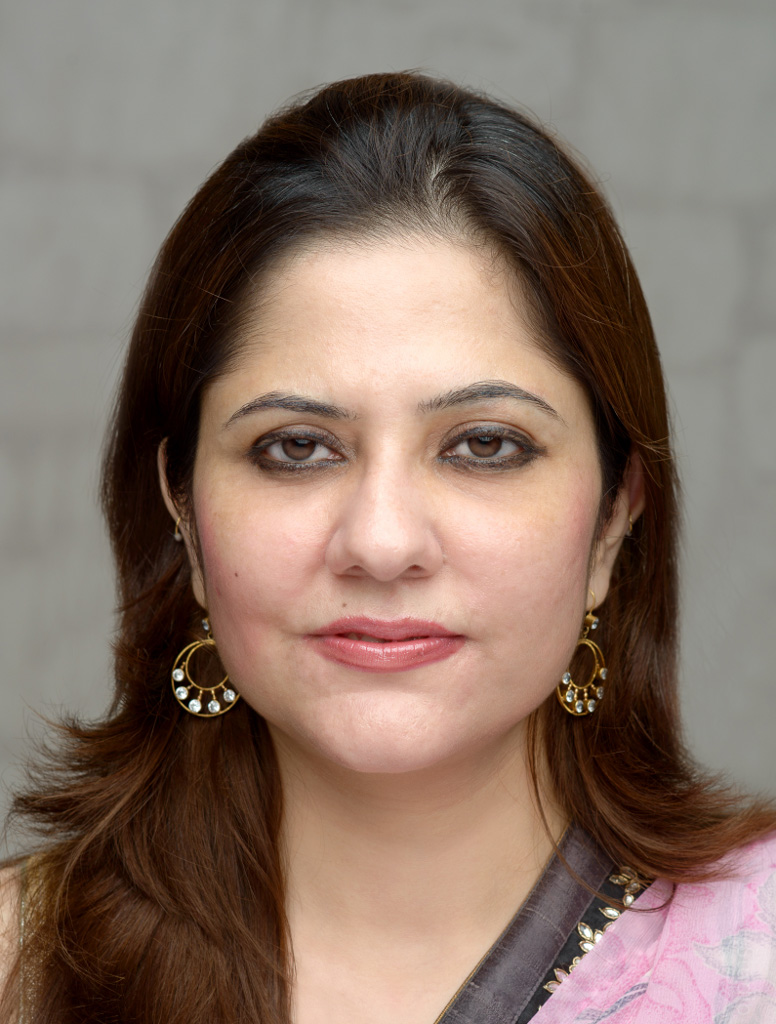}
      \caption{Content image}
      \label{fig:st_effects:content1}
    \end{subfigure}
    \begin{subfigure}[b]{0.24\textwidth}
            \includegraphics[width=\textwidth]{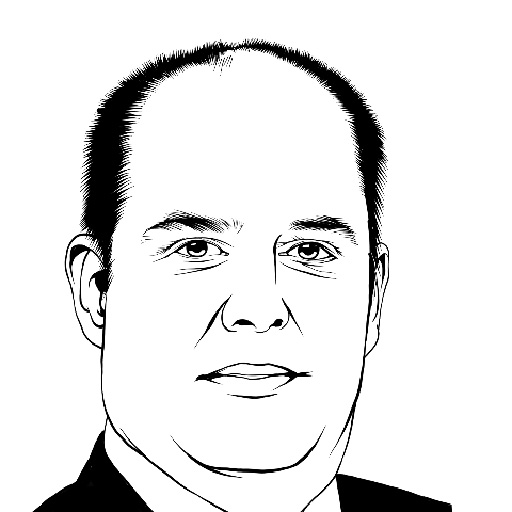}
      \caption{Style image}
      \label{fig:st_effects:style1}
    \end{subfigure}
    \begin{subfigure}[b]{0.24\textwidth}
            \includegraphics[width=\textwidth]{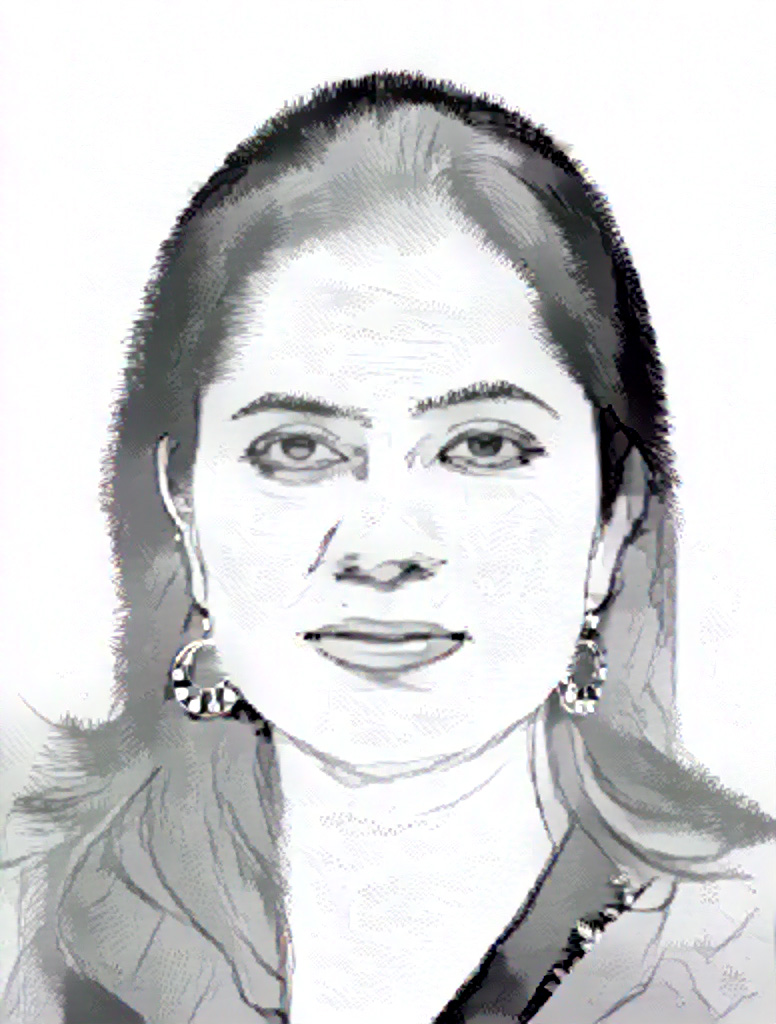}
          \caption{STROTSS \cite{kolkin2019style}}
    \end{subfigure}
    \begin{subfigure}[b]{0.24\textwidth}
            \includegraphics[width=\textwidth]{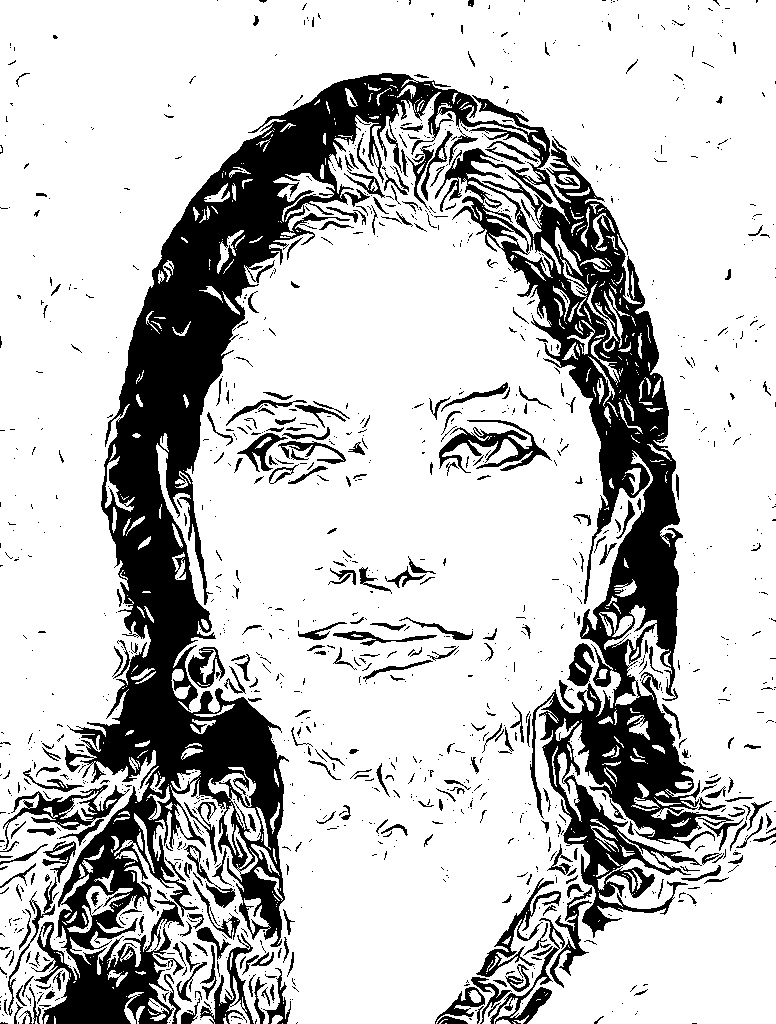}
      \caption{XDoG optim}
    \end{subfigure}

    \begin{subfigure}[b]{0.24\textwidth}
            \includegraphics[width=\textwidth]{supplemental/st_compare/1}
      \caption{Content image}
    \end{subfigure}
    \begin{subfigure}[b]{0.24\textwidth}
            \includegraphics[width=\textwidth]{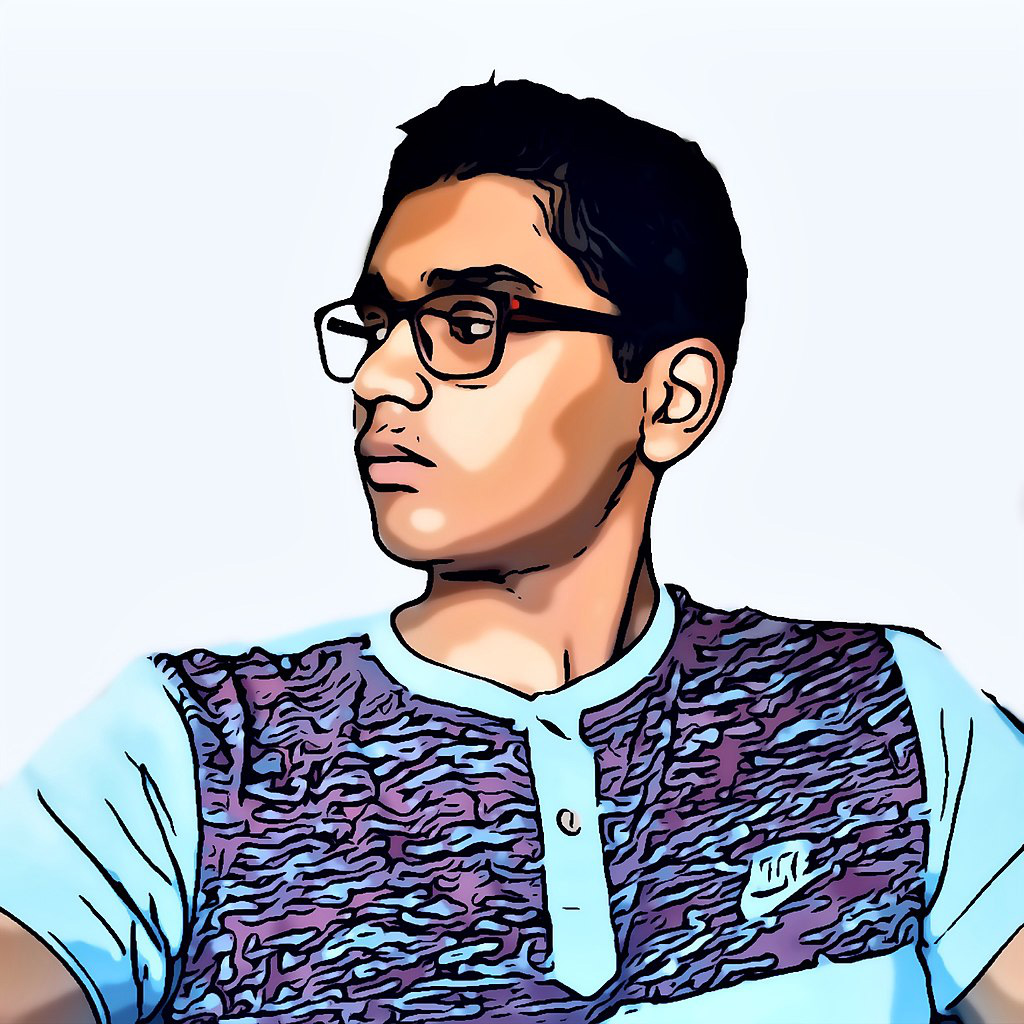}
      \caption{Style image}
      \label{fig:st_effects:style2}
    \end{subfigure}
    \begin{subfigure}[b]{0.24\textwidth}
            \includegraphics[width=\textwidth]{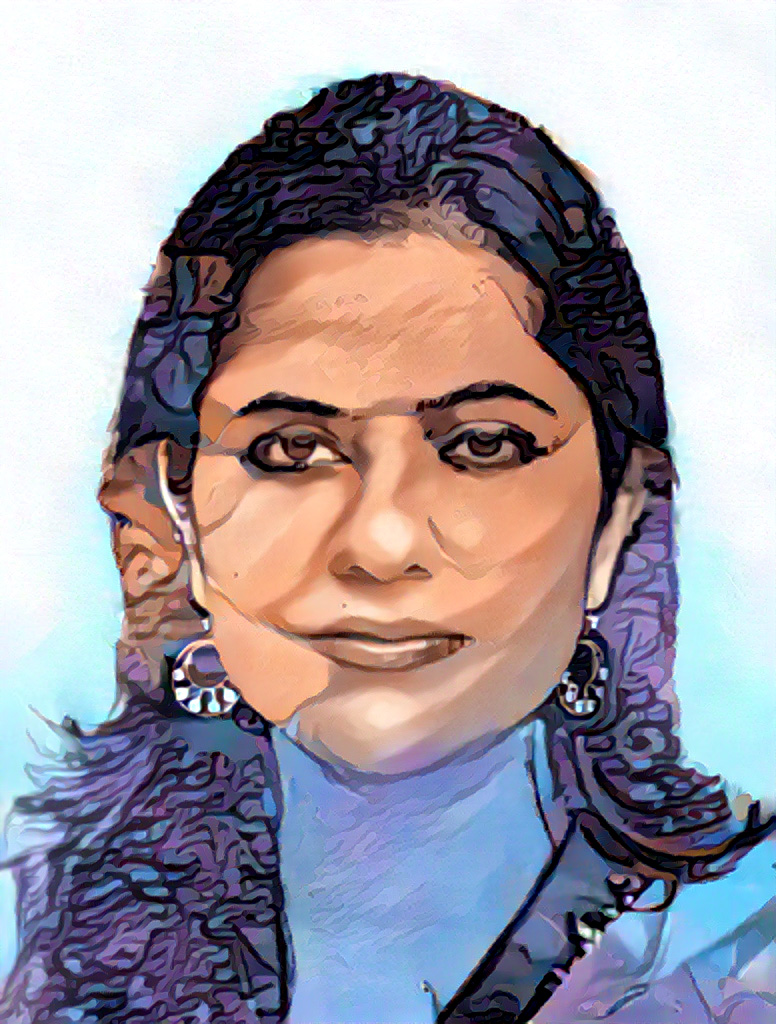}
          \caption{STROTSS \cite{kolkin2019style}}
    \end{subfigure}
    \begin{subfigure}[b]{0.24\textwidth}
            \includegraphics[width=\textwidth]{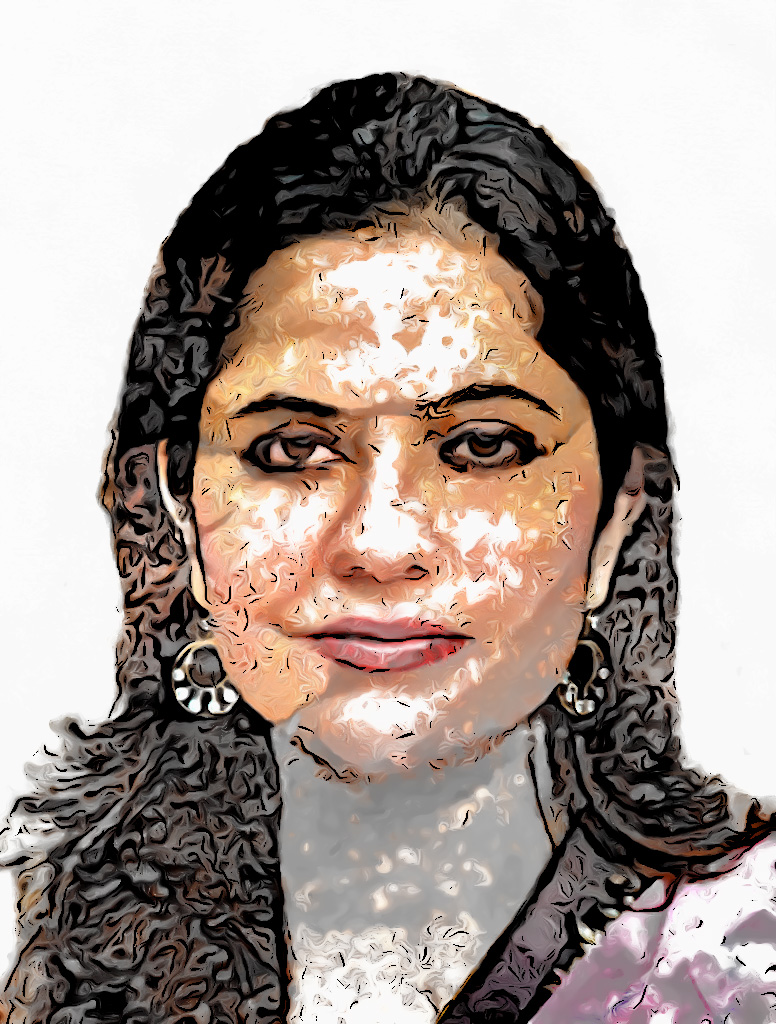}
      \caption{Cartoon optim}
    \end{subfigure}

    \begin{subfigure}[b]{0.24\textwidth}
            \includegraphics[width=\textwidth]{supplemental/st_compare/1}
      \caption{Content image}
    \end{subfigure}
    \begin{subfigure}[b]{0.24\textwidth}
            \includegraphics[width=\textwidth]{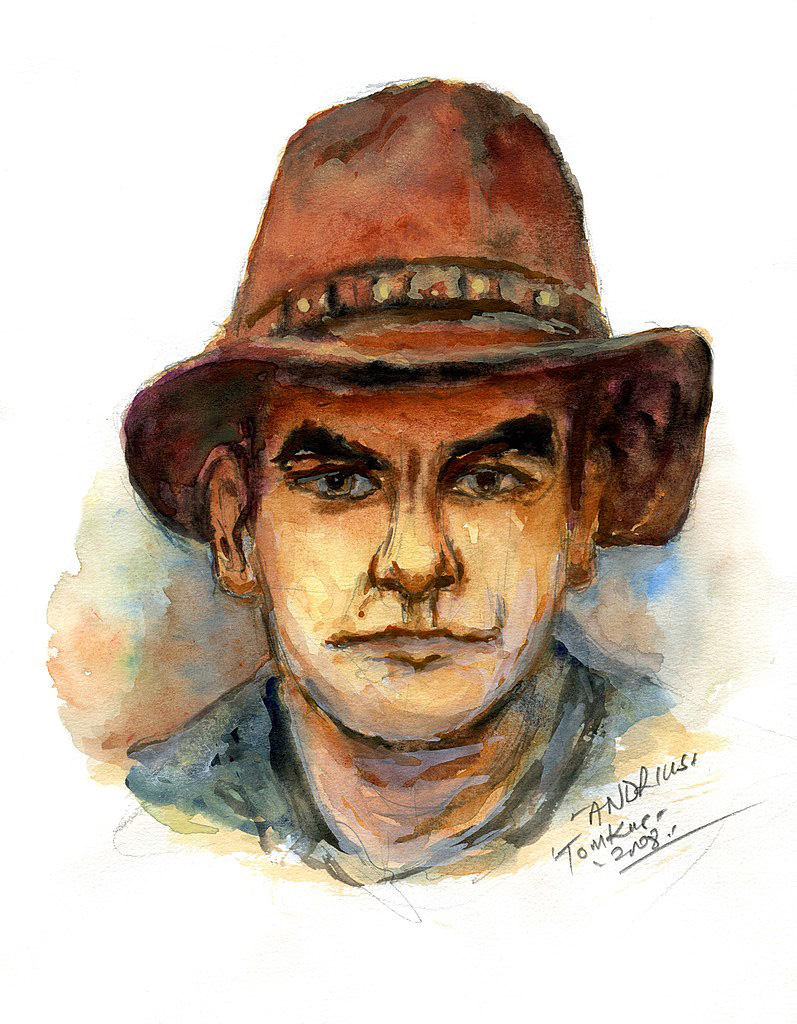}
      \caption{Style image}
      \label{fig:st_effects:style3}
    \end{subfigure}
    \begin{subfigure}[b]{0.24\textwidth}
            \includegraphics[width=\textwidth]{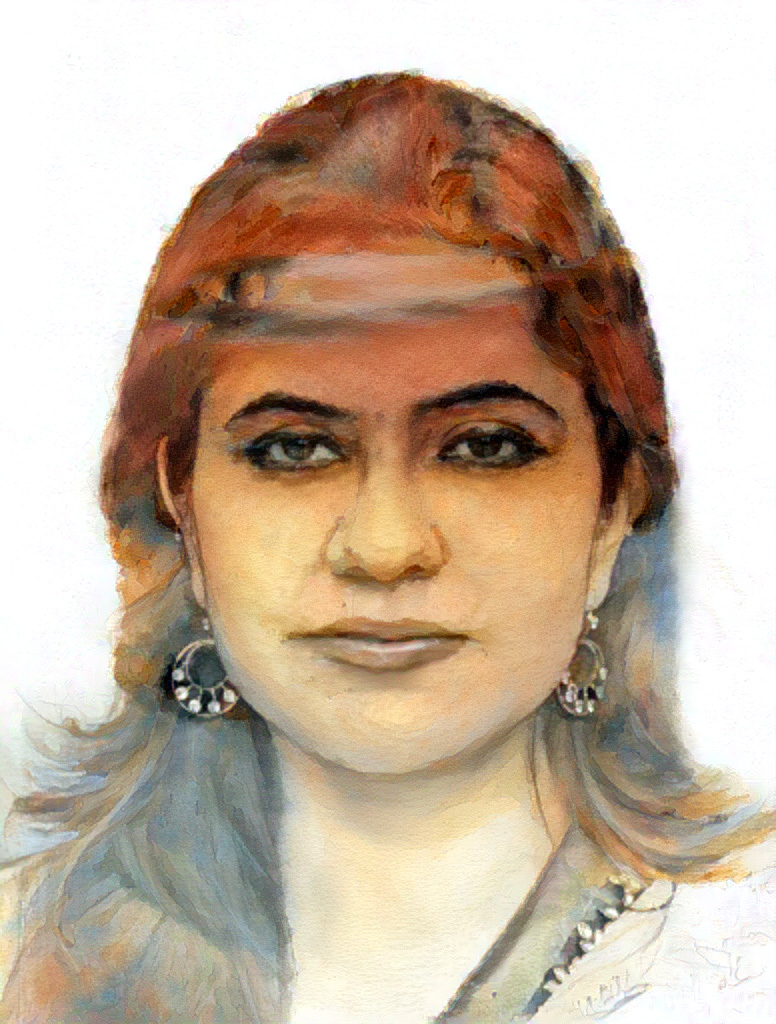}
          \caption{STROTSS \cite{kolkin2019style}}
          \label{fig:st_effects:reference3}
    \end{subfigure}
    \begin{subfigure}[b]{0.24\textwidth}
            \includegraphics[width=\textwidth]{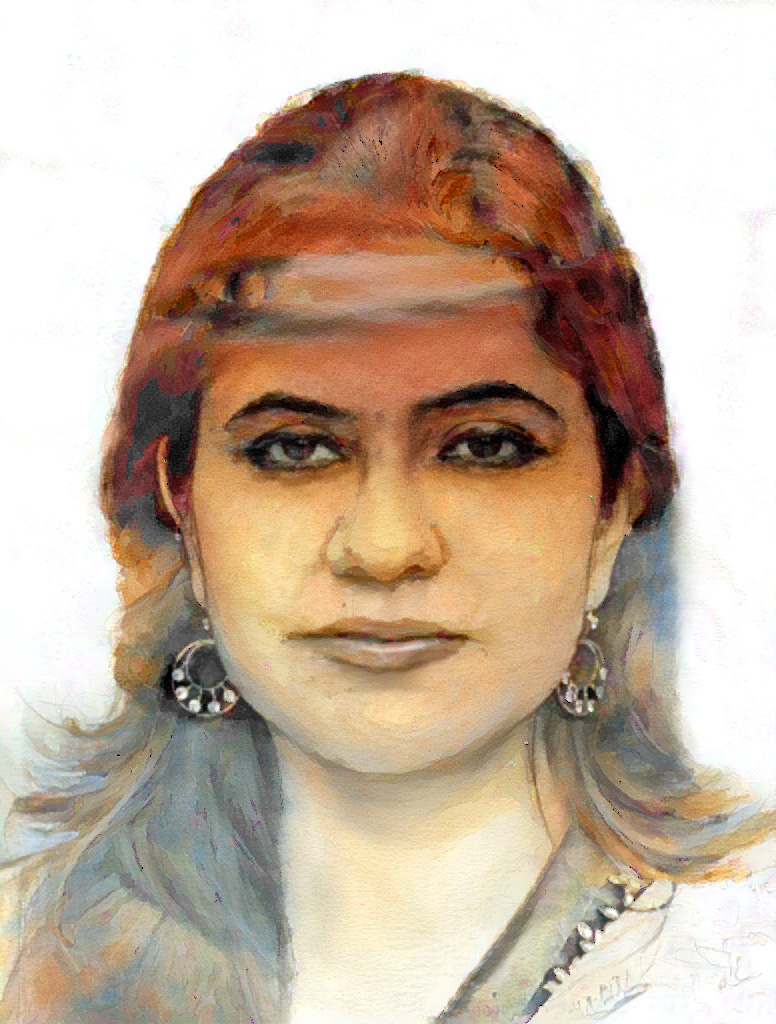}
      \caption{\footnotesize{Watercolor optim}}
      \label{fig:st_effects:optresult3}

    \end{subfigure}
        \begin{subfigure}[b]{0.24\textwidth}
            \includegraphics[width=\textwidth]{supplemental/st_compare/1}
      \caption{Content image}
    \end{subfigure}
    \begin{subfigure}[b]{0.24\textwidth}
            \includegraphics[width=\textwidth]{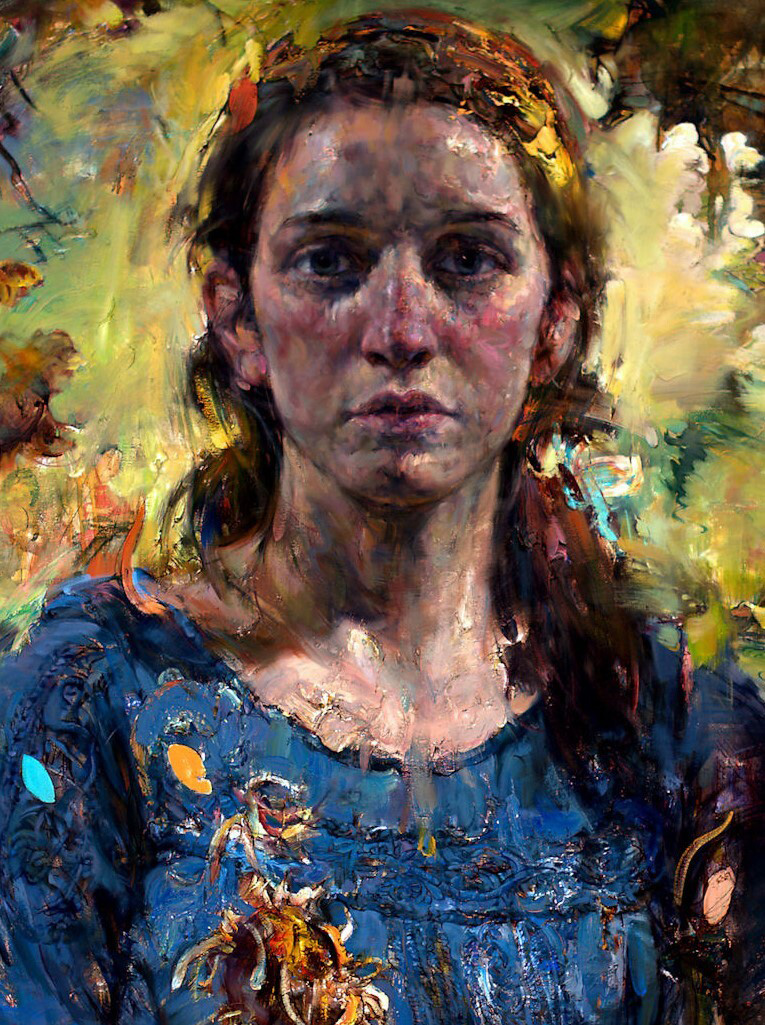}
      \caption{Style image}
      \label{fig:st_effects:style4}
    \end{subfigure}
    \begin{subfigure}[b]{0.24\textwidth}
            \includegraphics[width=\textwidth]{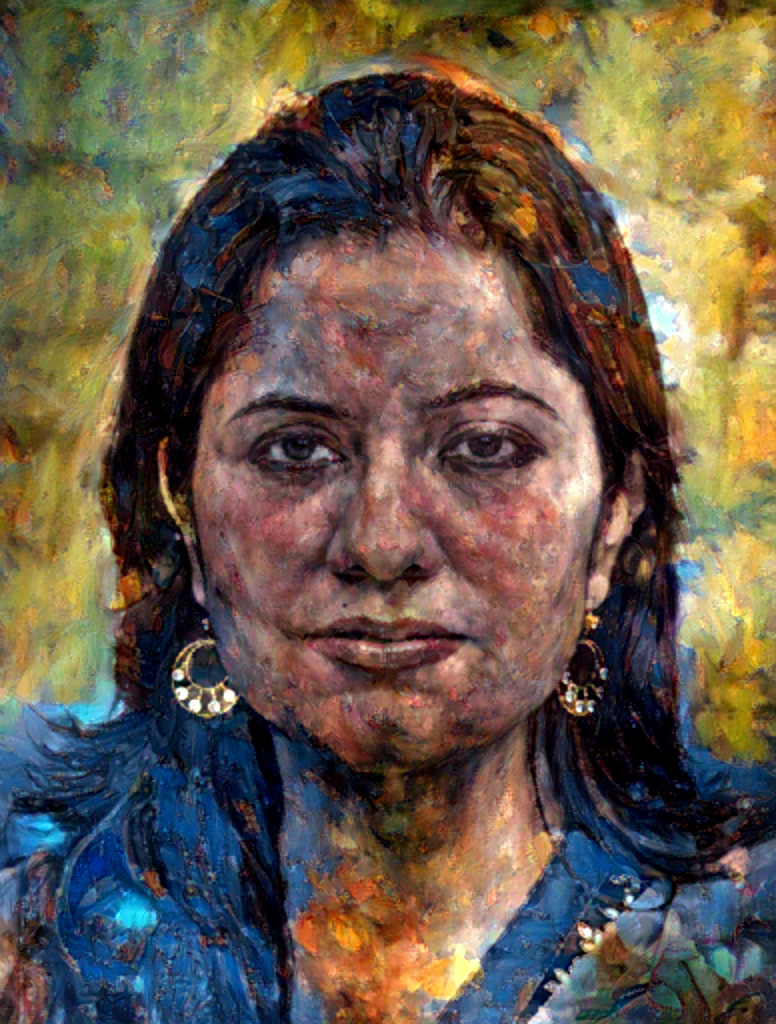}
          \caption{STROTSS \cite{kolkin2019style}}
          \label{fig:st_effects:reference4}
    \end{subfigure}
    \begin{subfigure}[b]{0.24\textwidth}
            \includegraphics[width=\textwidth]{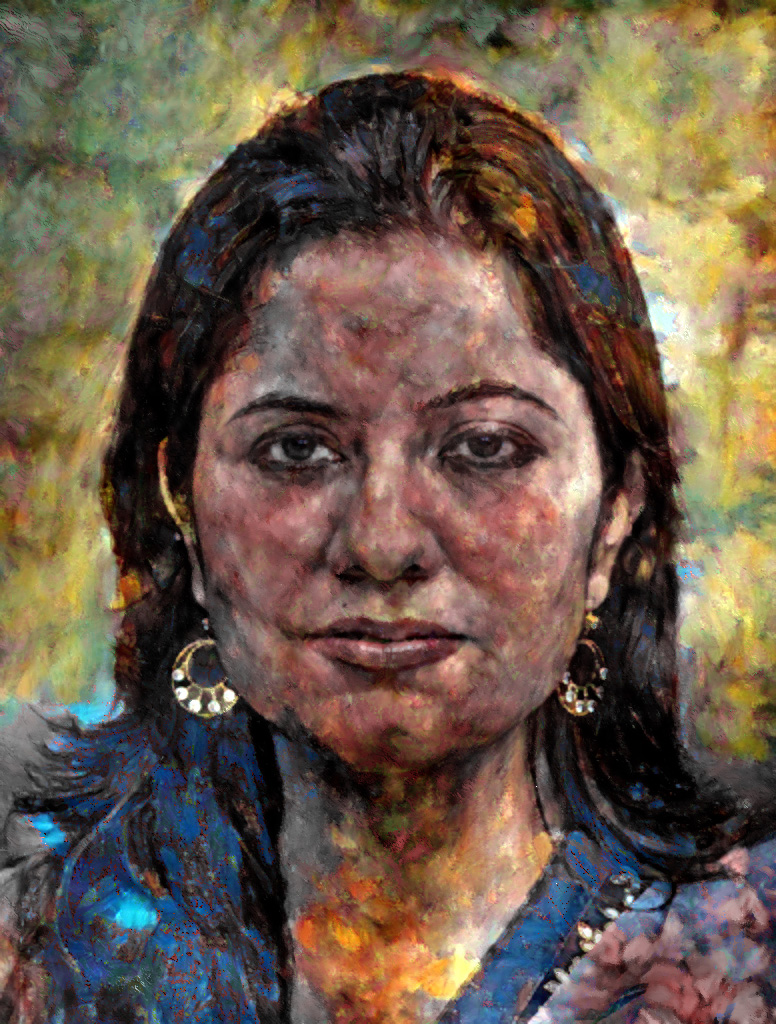}
      \caption{Oilpaint optim}
      \label{fig:st_effects:optresult4}
    \end{subfigure}
    \caption{Local parameter optimization using different algorithmic effects}
    \label{fig:strotss_all_pipelines}
\end{figure}

\newcommand{\specFigSize}{0.22\textwidth}
\newcommand{\halfFigSize}{0.155\textwidth}

\begin{figure}[tb]
\tikzset{every node/.style={inner sep=0pt,outer sep=0pt}}
\settoheight{\tempdima}{\includegraphics[width=\specFigSize]{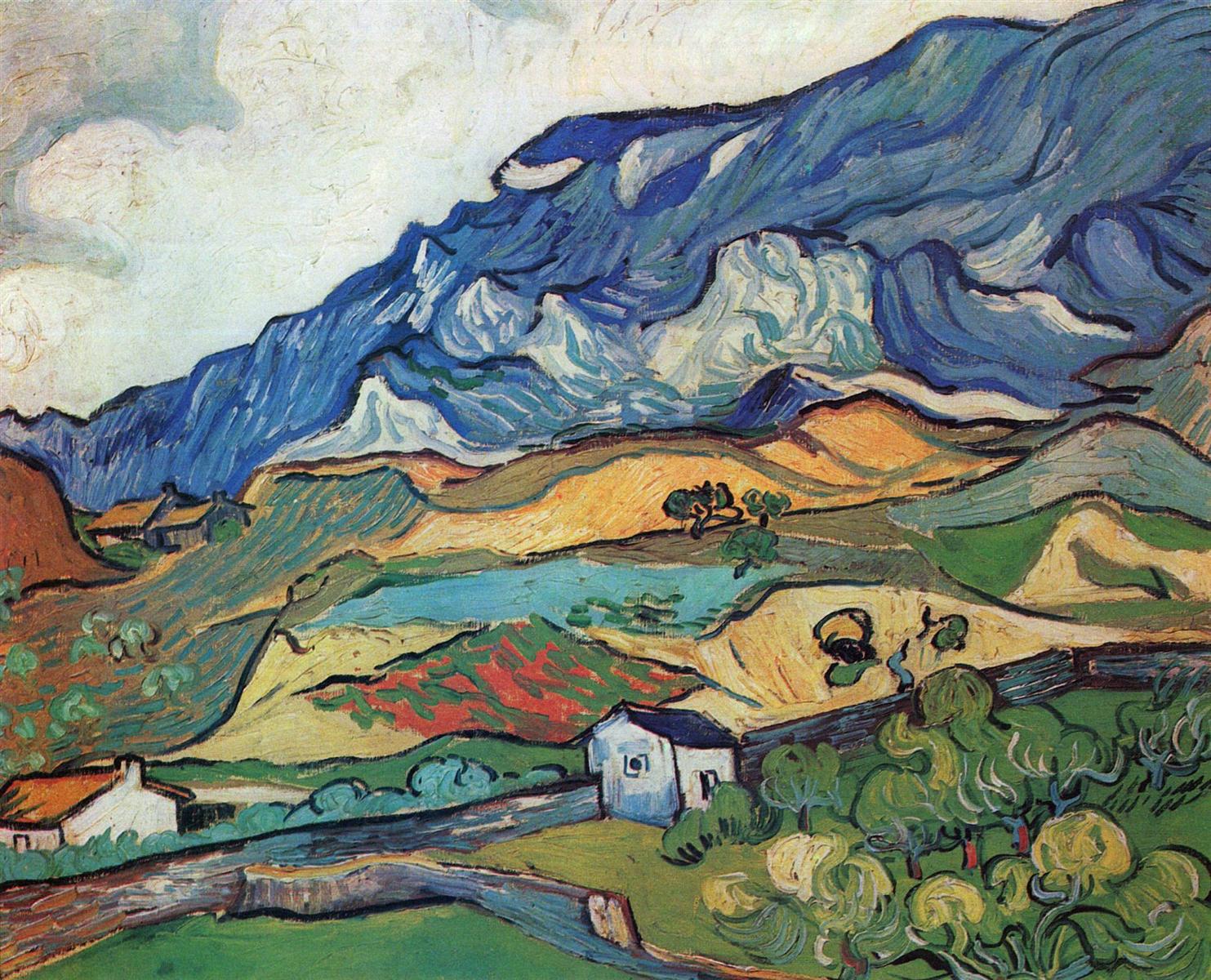}}%
	\centering\begin{tabular}{cc@{}c@{}c@{}c@{}}
    Style & STROTTS\cite{kolkin2019style} & Watercolor & Oilpaint \\
		
\begin{tikzpicture}
    \node{\includegraphics[width=\specFigSize]{supplemental/st_common_grid/styles/les_apilles.jpg}};%
\end{tikzpicture}&
\begin{tikzpicture}
    \node{\includegraphics[clip,width=\specFigSize]{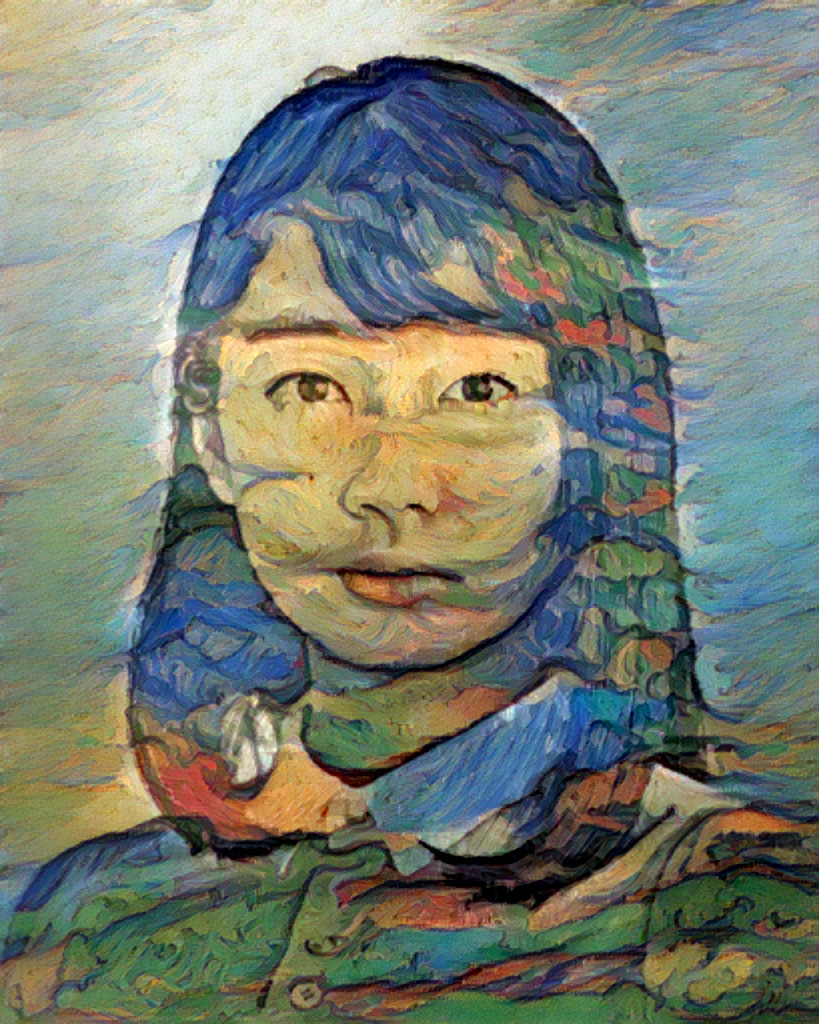}};%
\end{tikzpicture}&
\begin{tikzpicture}
    \node{\includegraphics[width=\specFigSize]{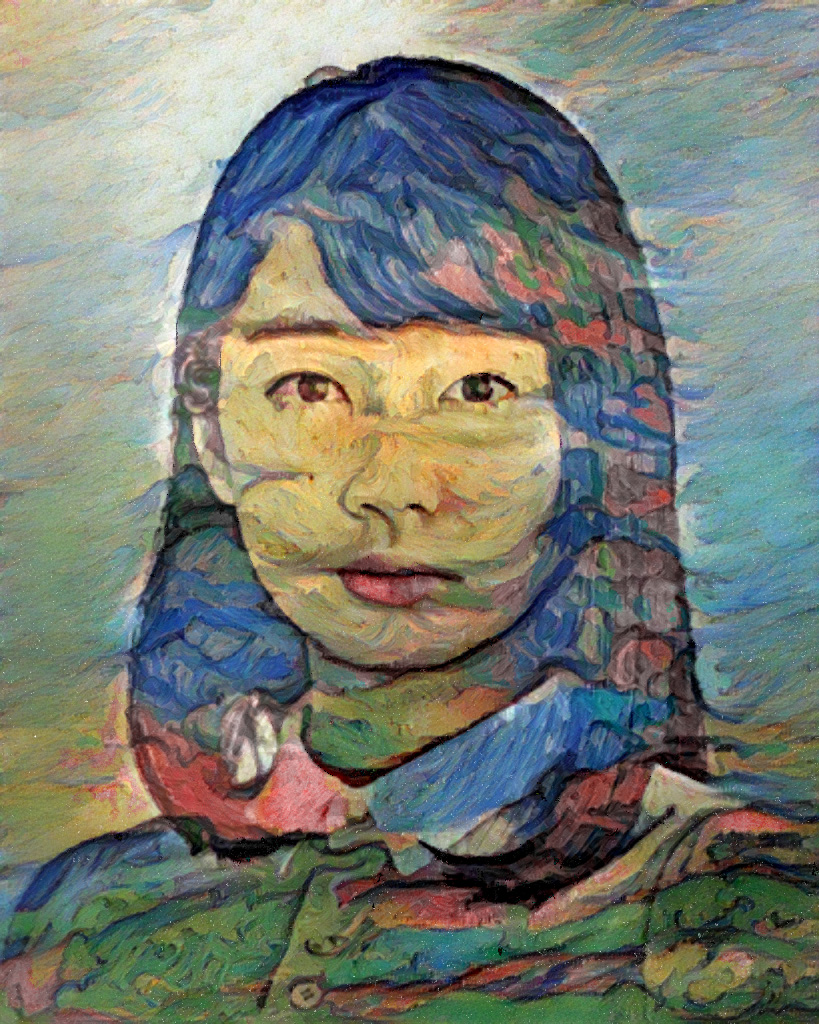}};%
\end{tikzpicture}&
\begin{tikzpicture}
    \node{\includegraphics[width=\specFigSize]{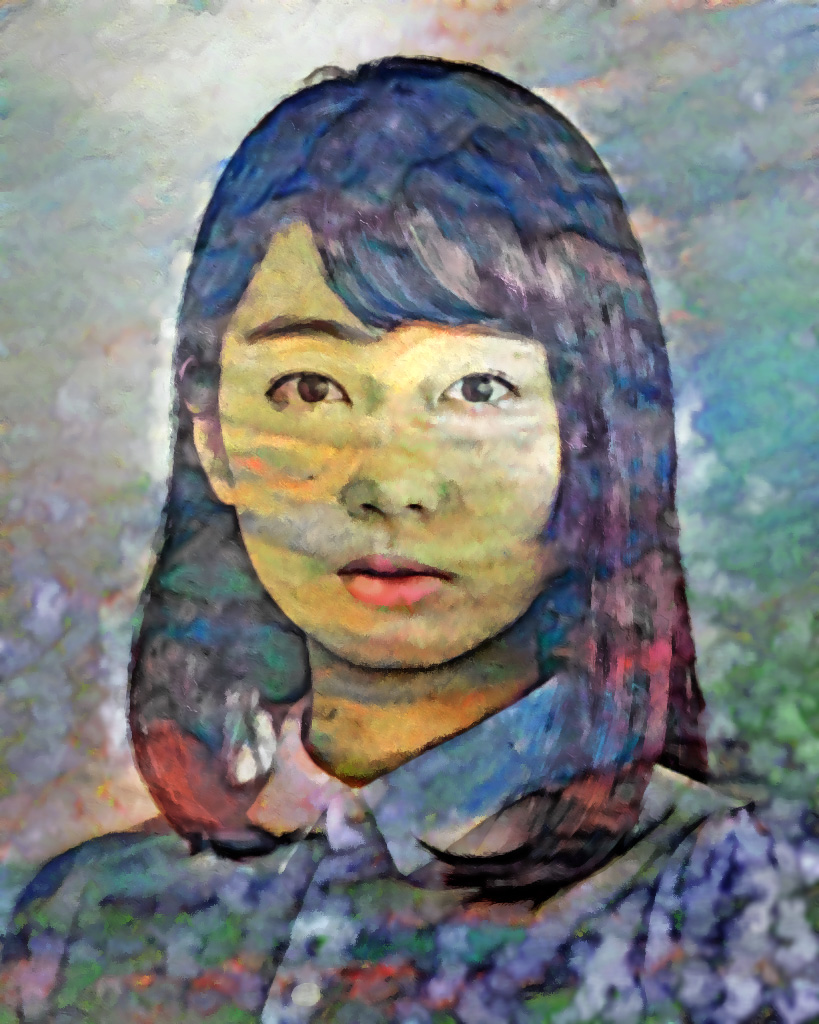}};%
\end{tikzpicture}\\

\begin{tikzpicture}
    \node{\includegraphics[width=\specFigSize]{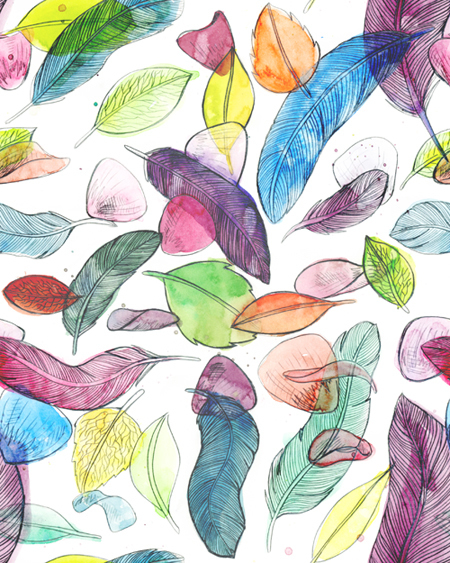}};%
\end{tikzpicture}&
\begin{tikzpicture}
    \node{\includegraphics[clip,width=\specFigSize]{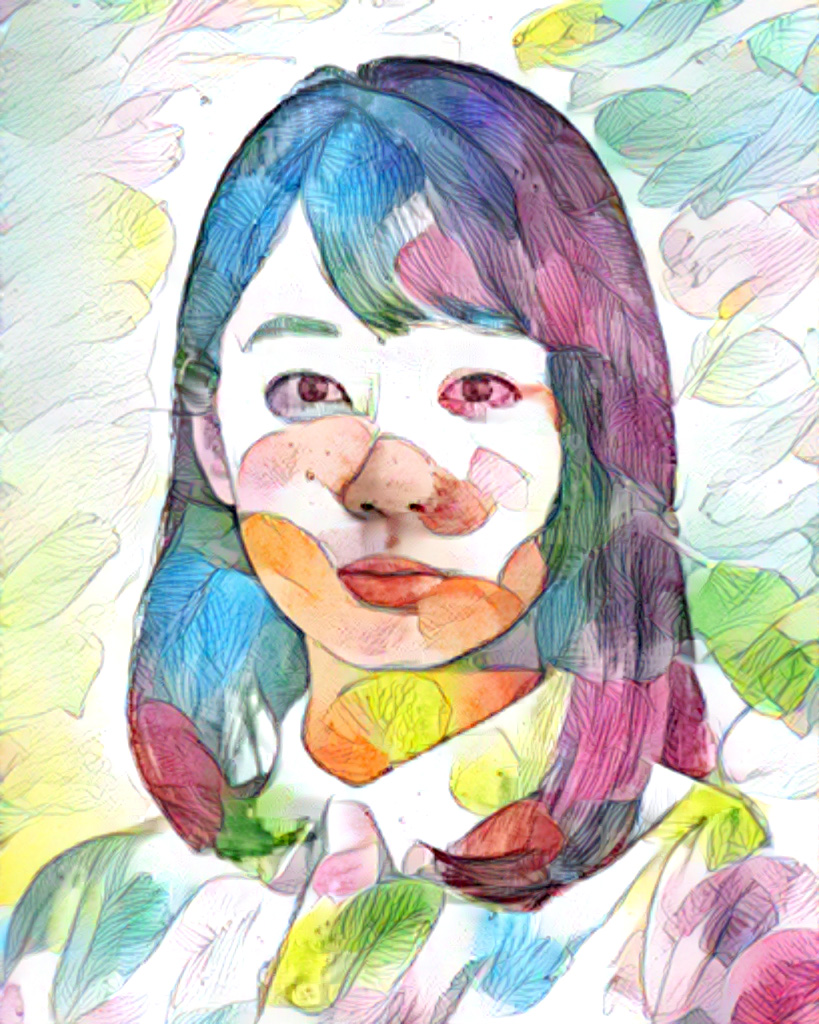}};%
\end{tikzpicture}&
\begin{tikzpicture}
    \node{\includegraphics[width=\specFigSize]{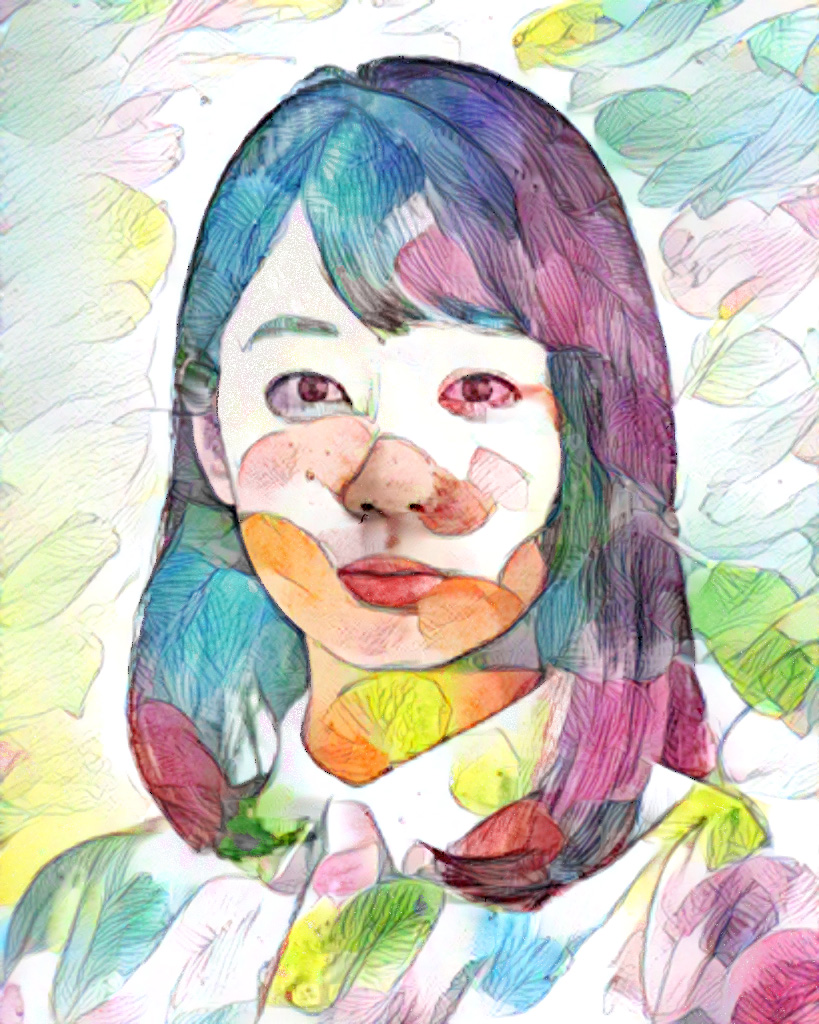}};%
\end{tikzpicture}&
\begin{tikzpicture}
    \node{\includegraphics[width=\specFigSize]{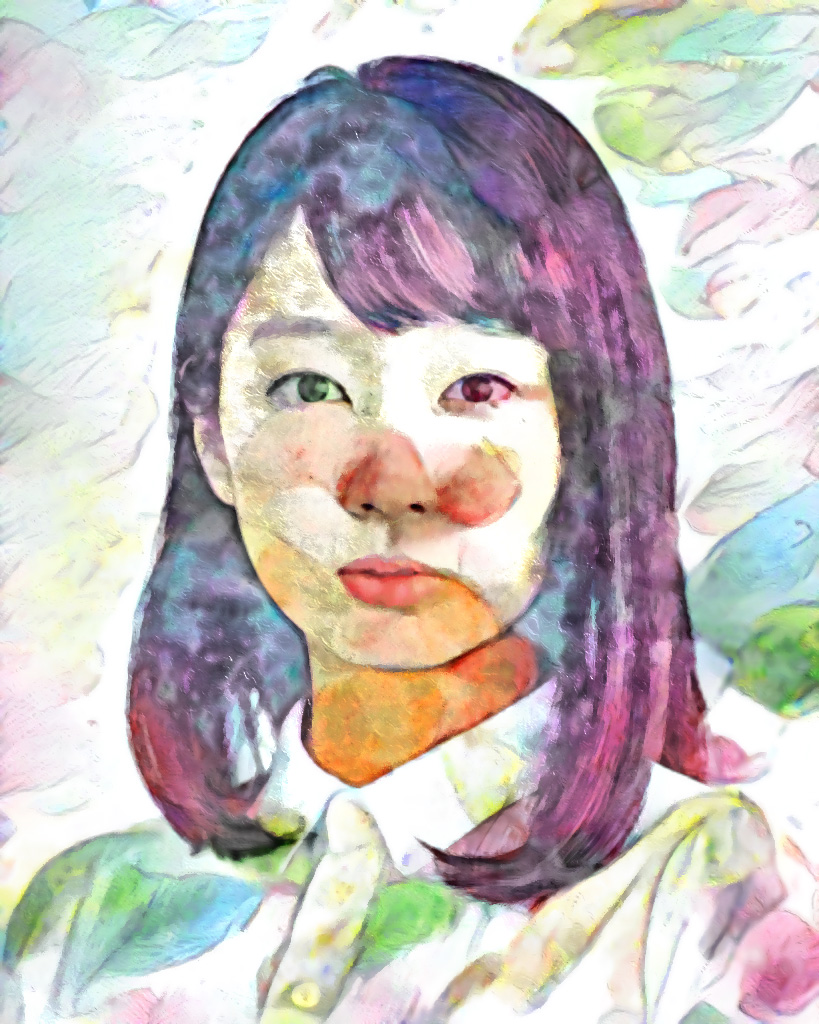}};%
\end{tikzpicture}\\

\begin{tikzpicture}
    \node{\includegraphics[width=\specFigSize]{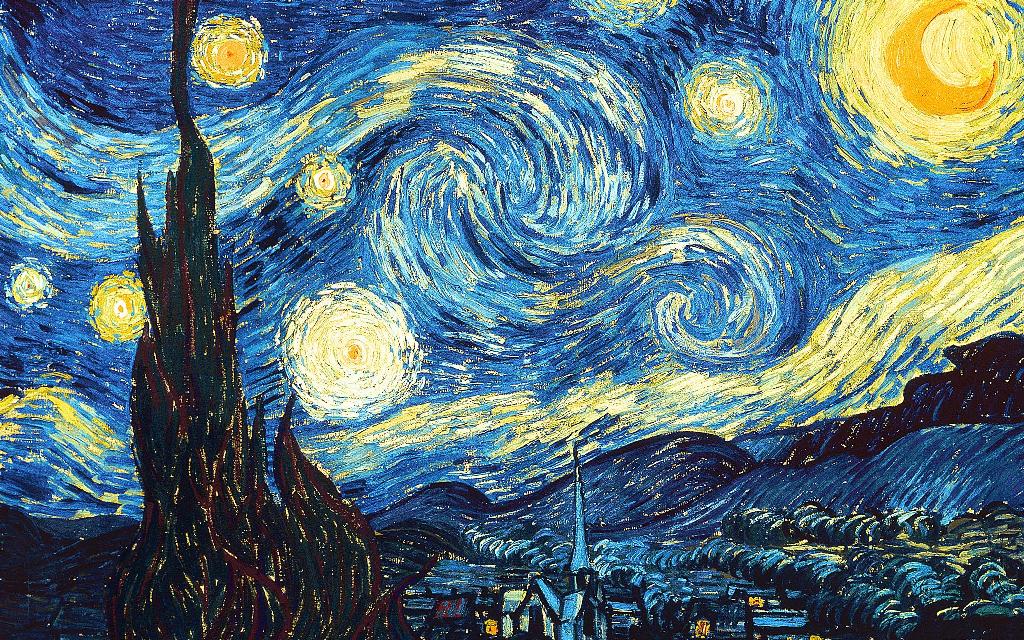}};%
\end{tikzpicture}&
\begin{tikzpicture}
    \node{\includegraphics[clip,width=\specFigSize]{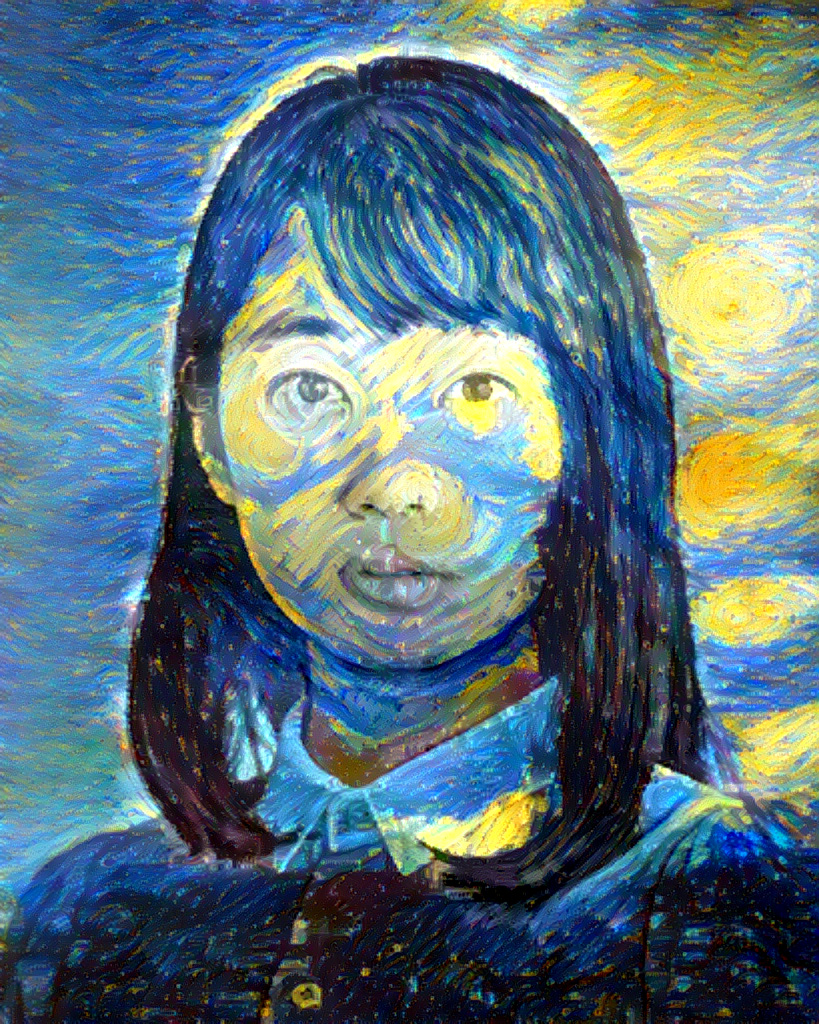}};%
\end{tikzpicture}&
\begin{tikzpicture}
    \node{\includegraphics[width=\specFigSize]{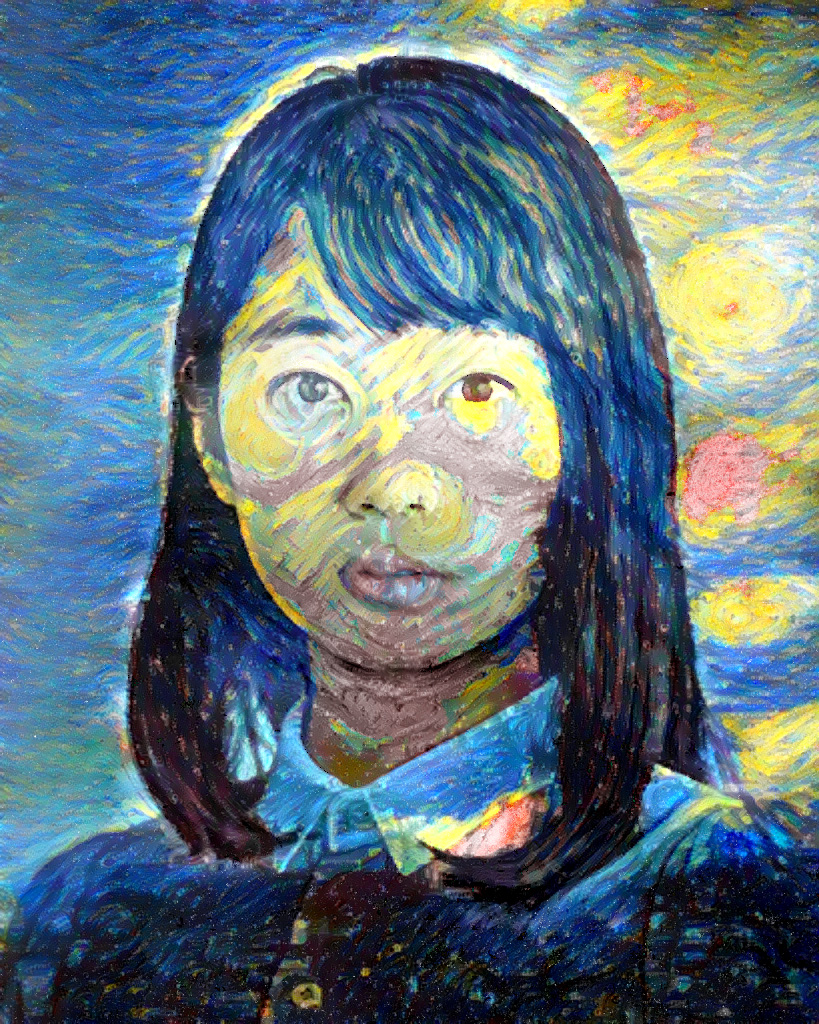}};%
\end{tikzpicture}&
\begin{tikzpicture}
    \node{\includegraphics[width=\specFigSize]{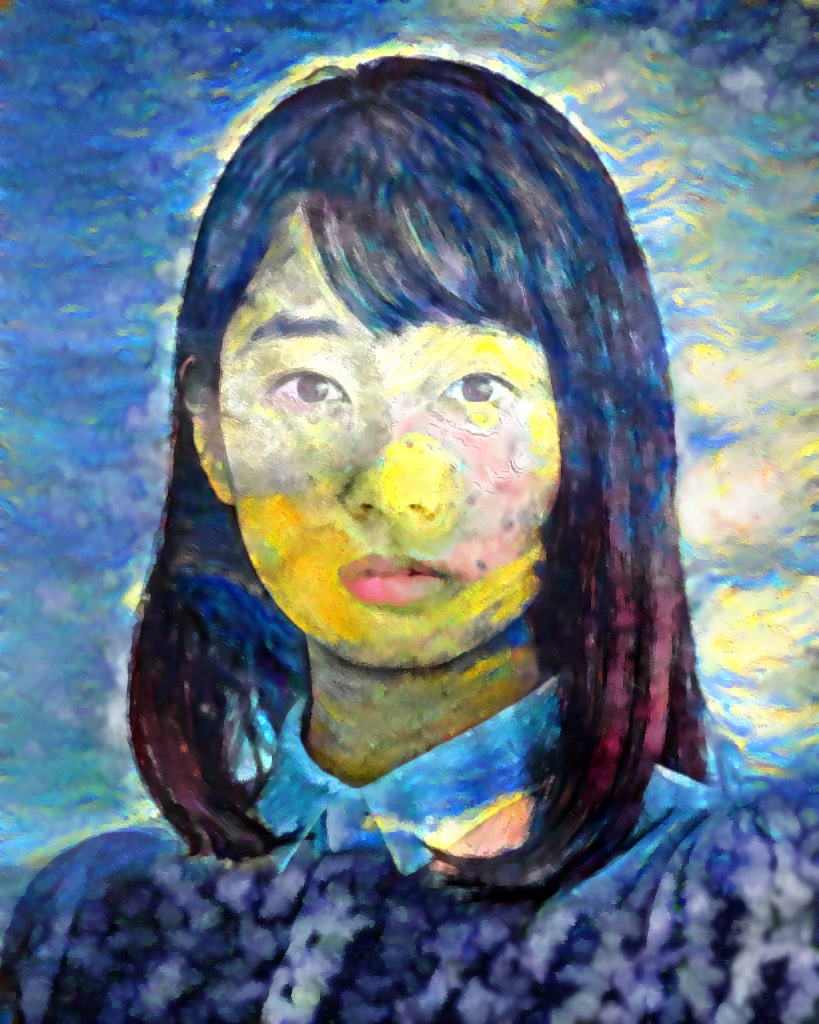}};%
\end{tikzpicture}\\

\begin{tikzpicture}
    \node{\includegraphics[width=\specFigSize]{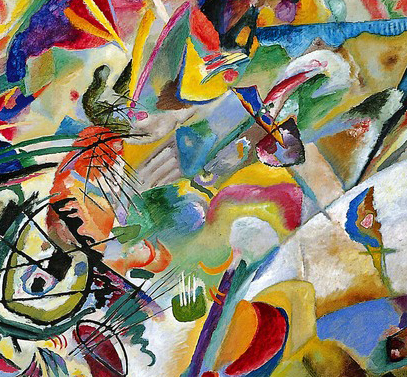}};%
\end{tikzpicture}&
\begin{tikzpicture}
    \node{\includegraphics[clip,width=\specFigSize]{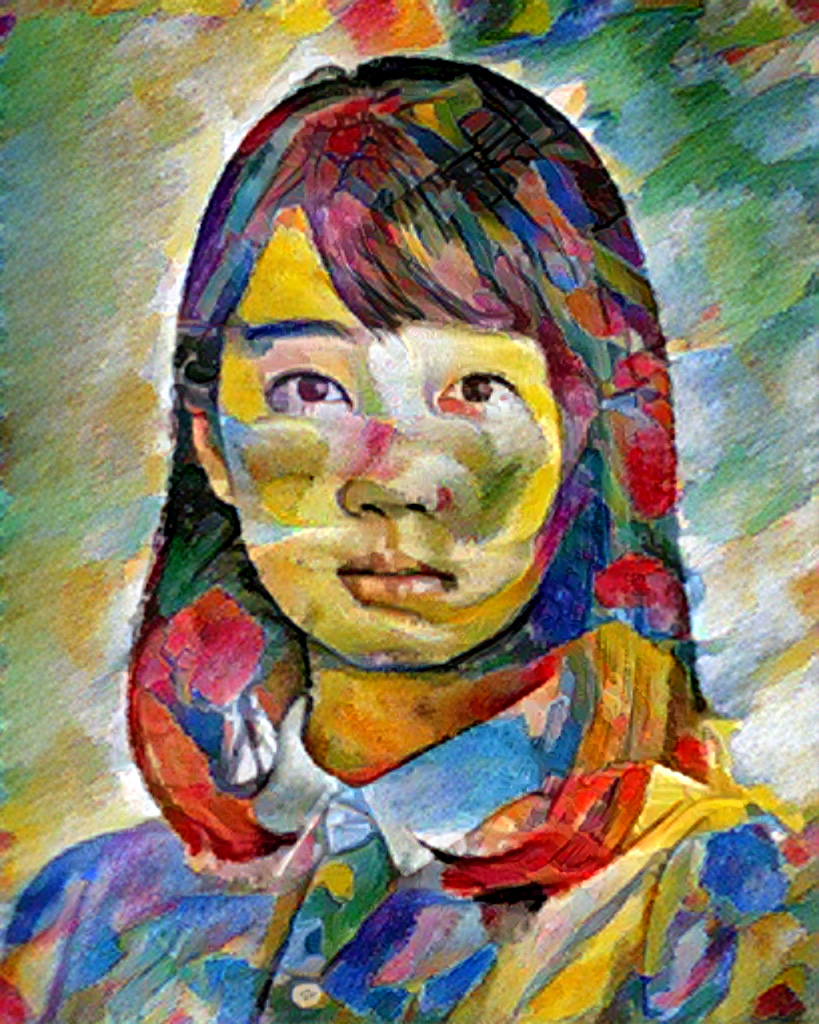}};%
\end{tikzpicture}&
\begin{tikzpicture}
    \node{\includegraphics[width=\specFigSize]{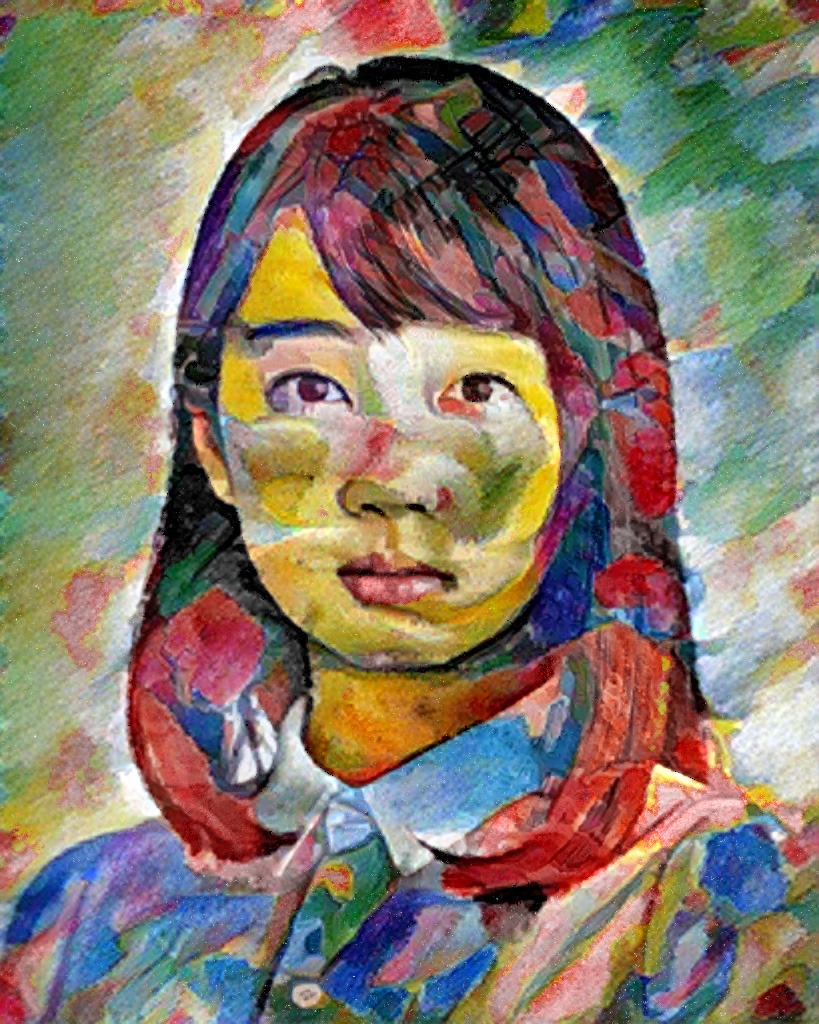}};%
\end{tikzpicture}&
\begin{tikzpicture}
    \node{\includegraphics[width=\specFigSize]{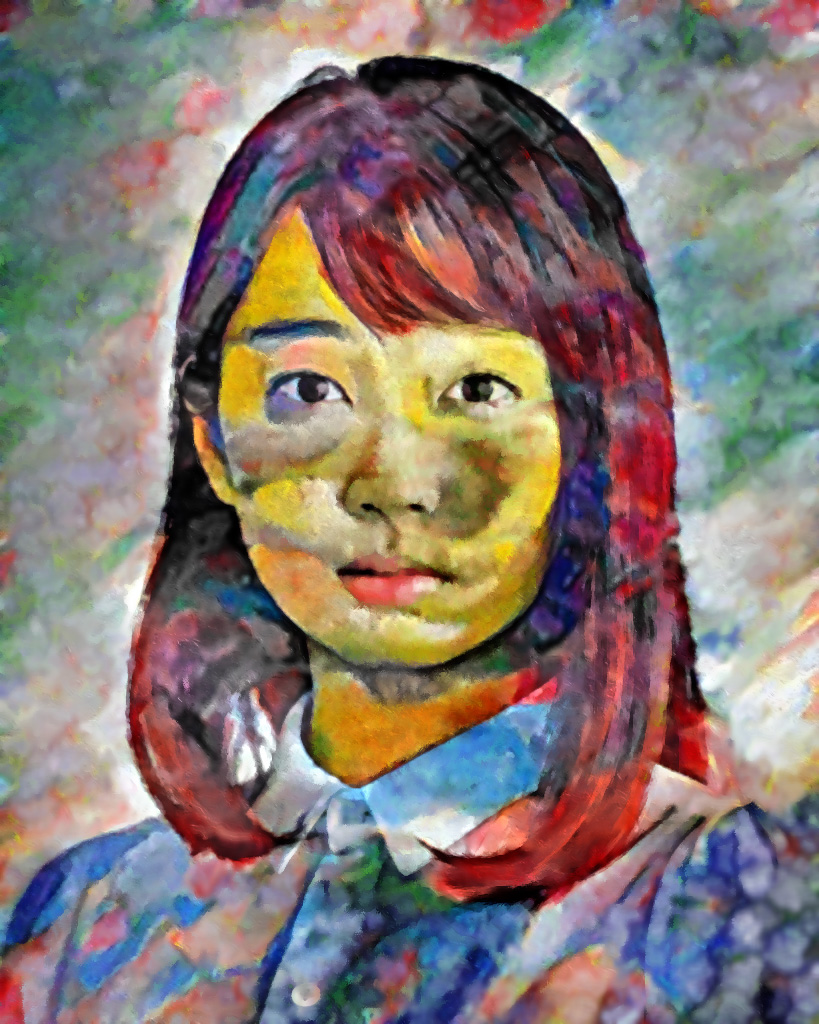}};%
\end{tikzpicture}\\

\begin{tikzpicture}
    \node{\includegraphics[width=\specFigSize]{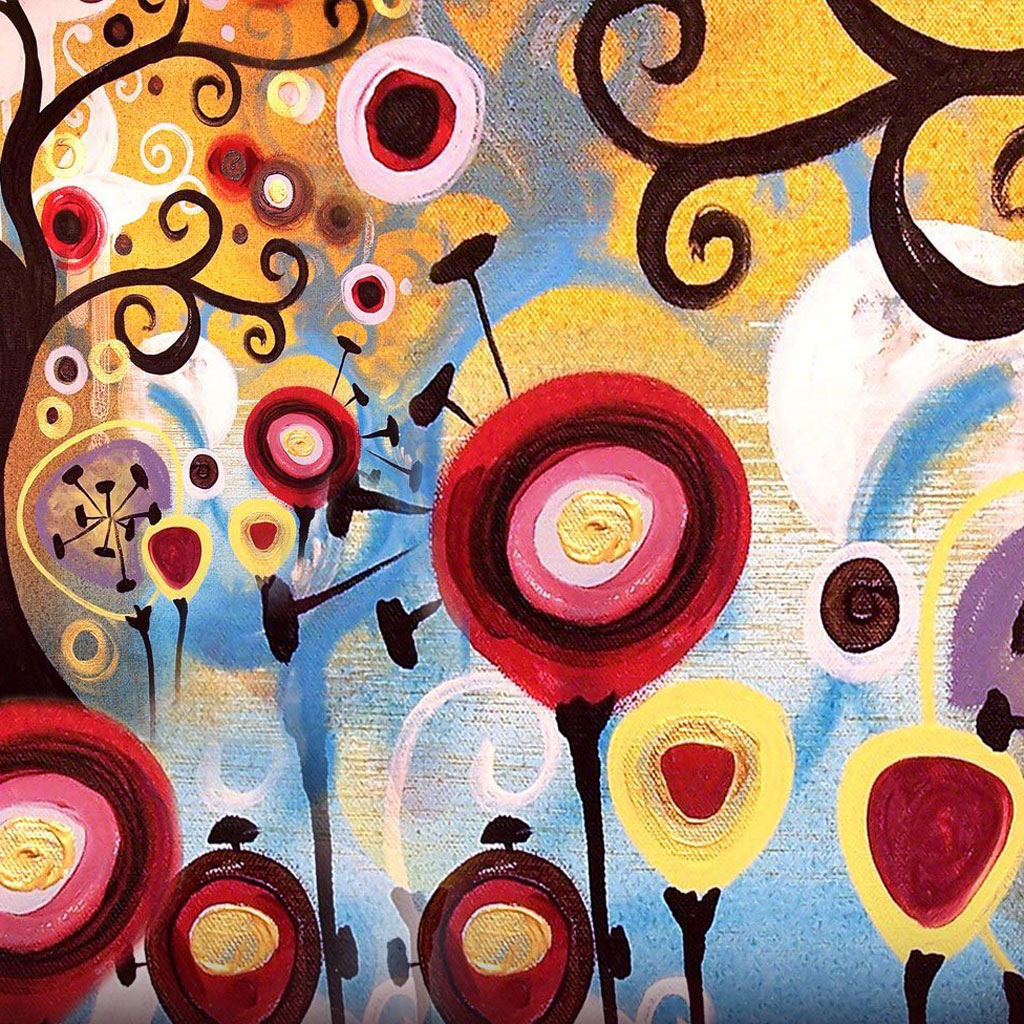}};%
\end{tikzpicture}&
\begin{tikzpicture}
    \node{\includegraphics[clip,width=\specFigSize]{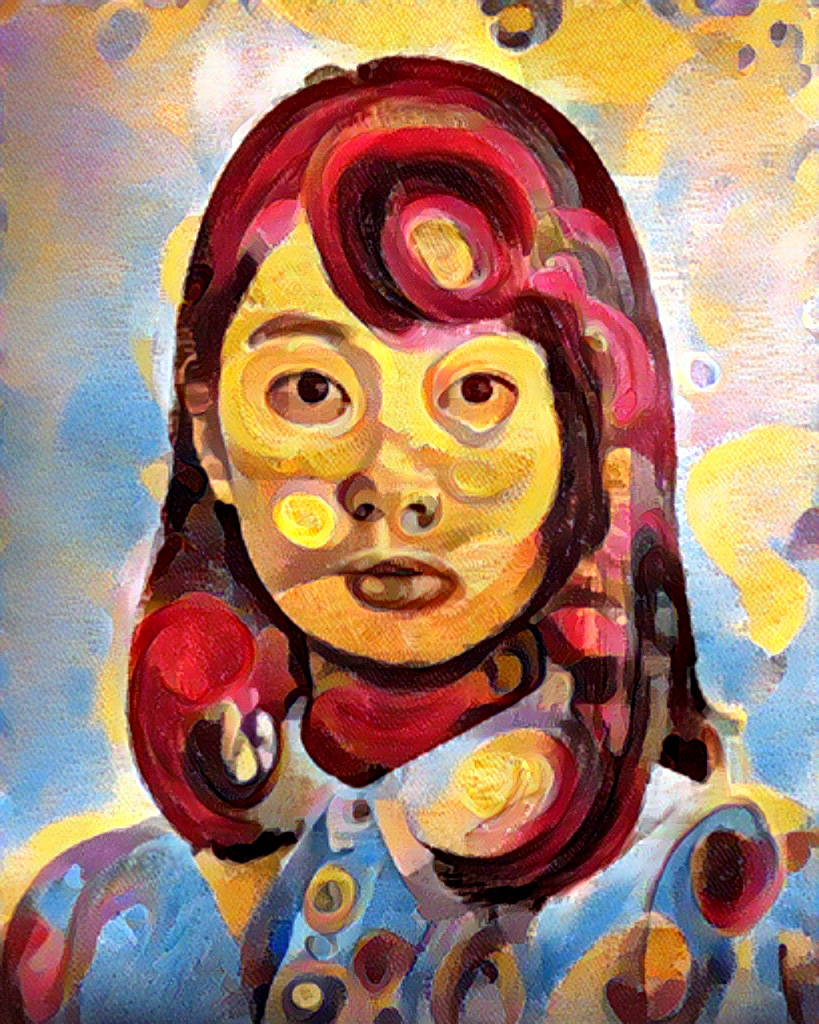}};%
\end{tikzpicture}&
\begin{tikzpicture}
    \node{\includegraphics[width=\specFigSize]{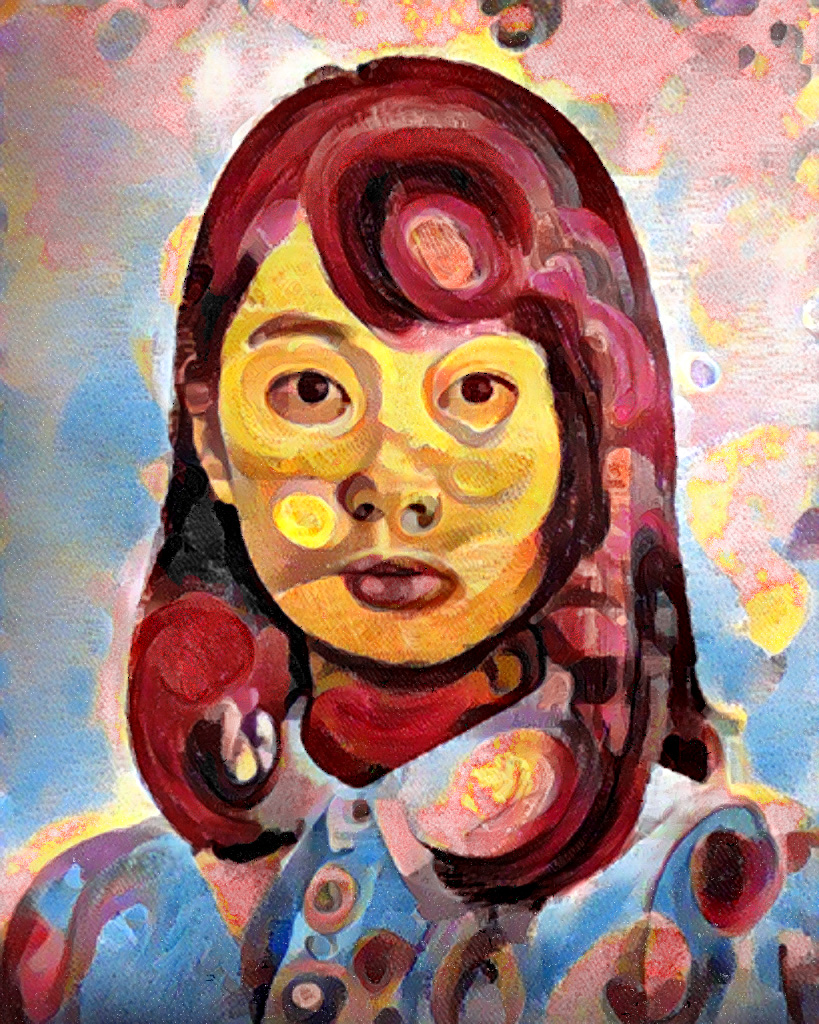}};%
\end{tikzpicture}&
\begin{tikzpicture}
    \node{\includegraphics[width=\specFigSize]{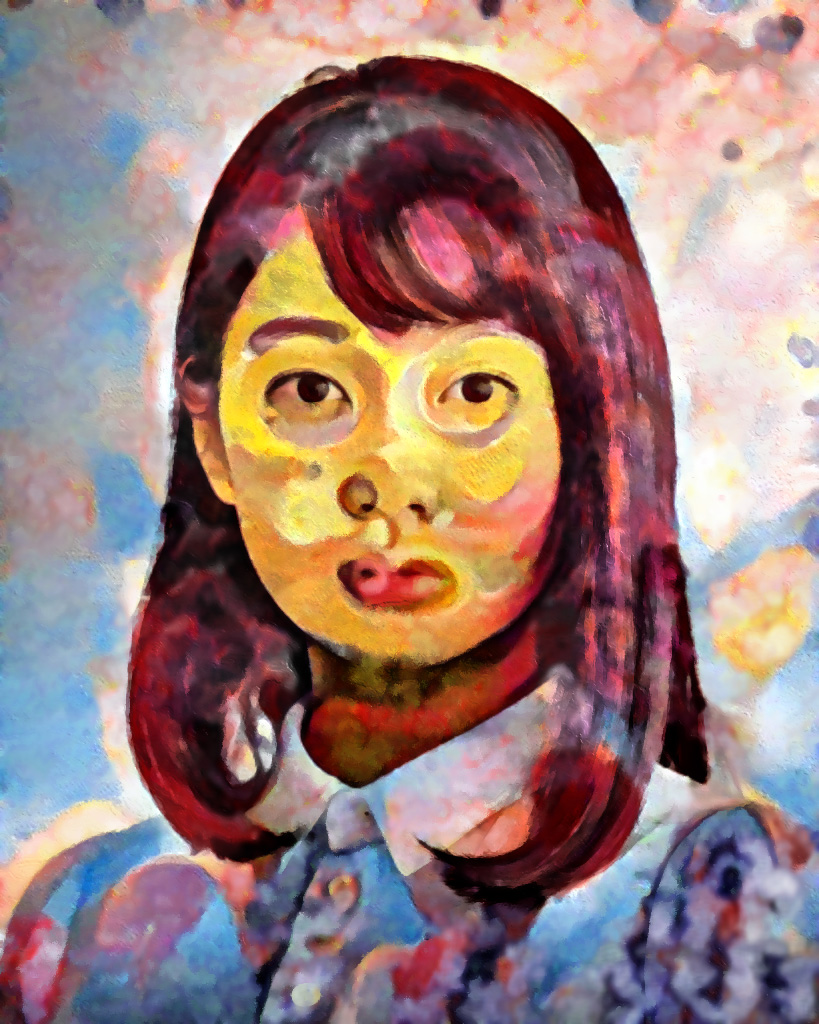}};%
\end{tikzpicture}\\

	\end{tabular}
	\caption{Style transfer capability on common NST styles.}
	\label{fig:commonnst1}
\end{figure}

\begin{figure}[tb]
\tikzset{every node/.style={inner sep=0pt,outer sep=0pt}}
	\centering\begin{tabular}{ccccc}
    Style & STROTTS\cite{kolkin2019style} & Watercolor & Oilpaint \\
		
\begin{tikzpicture}
    \node{\includegraphics[width=\specFigSize]{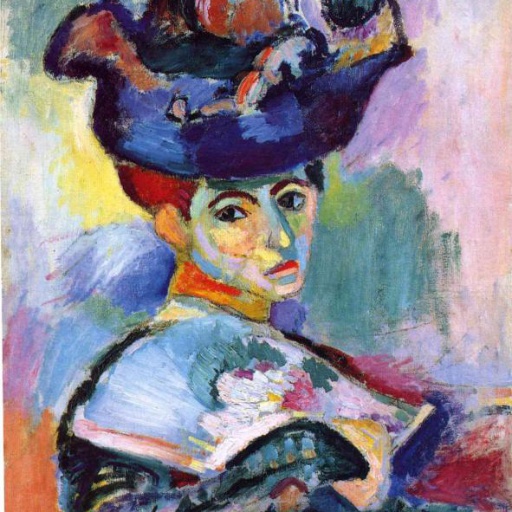}};%
\end{tikzpicture}&
\begin{tikzpicture}
    \node{\includegraphics[clip,width=\specFigSize]{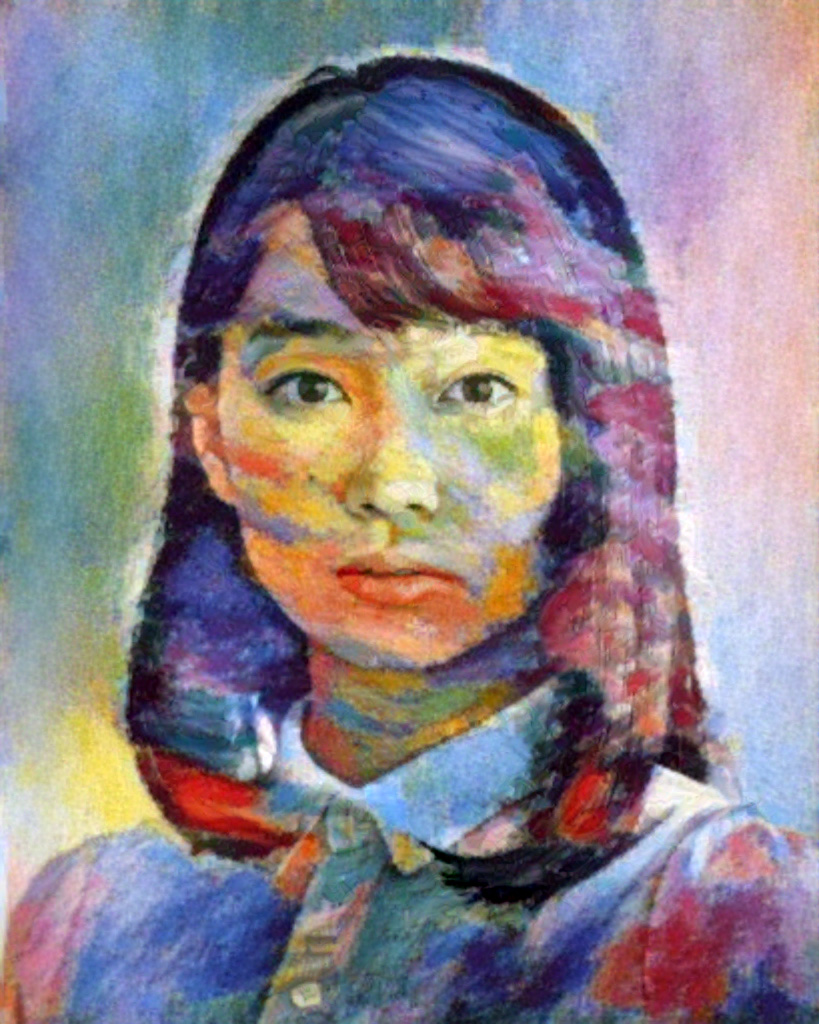}};%
\end{tikzpicture}&
\begin{tikzpicture}
    \node{\includegraphics[width=\specFigSize]{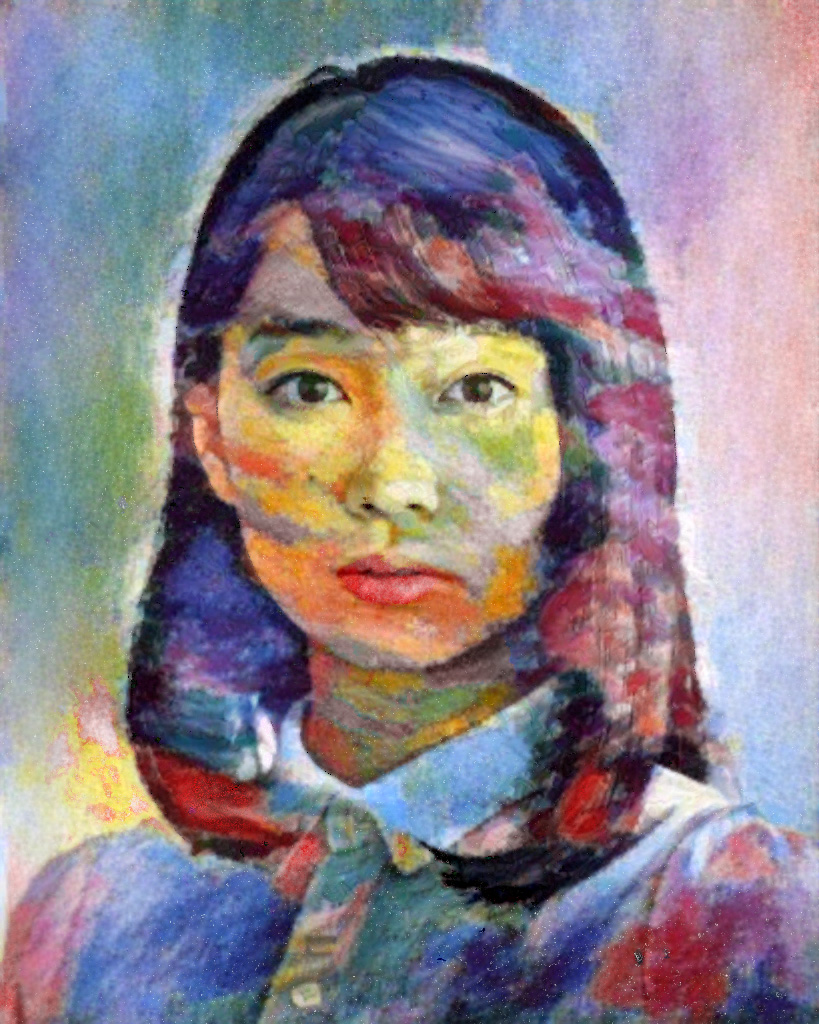}};%
\end{tikzpicture}&
\begin{tikzpicture}
    \node{\includegraphics[width=\specFigSize]{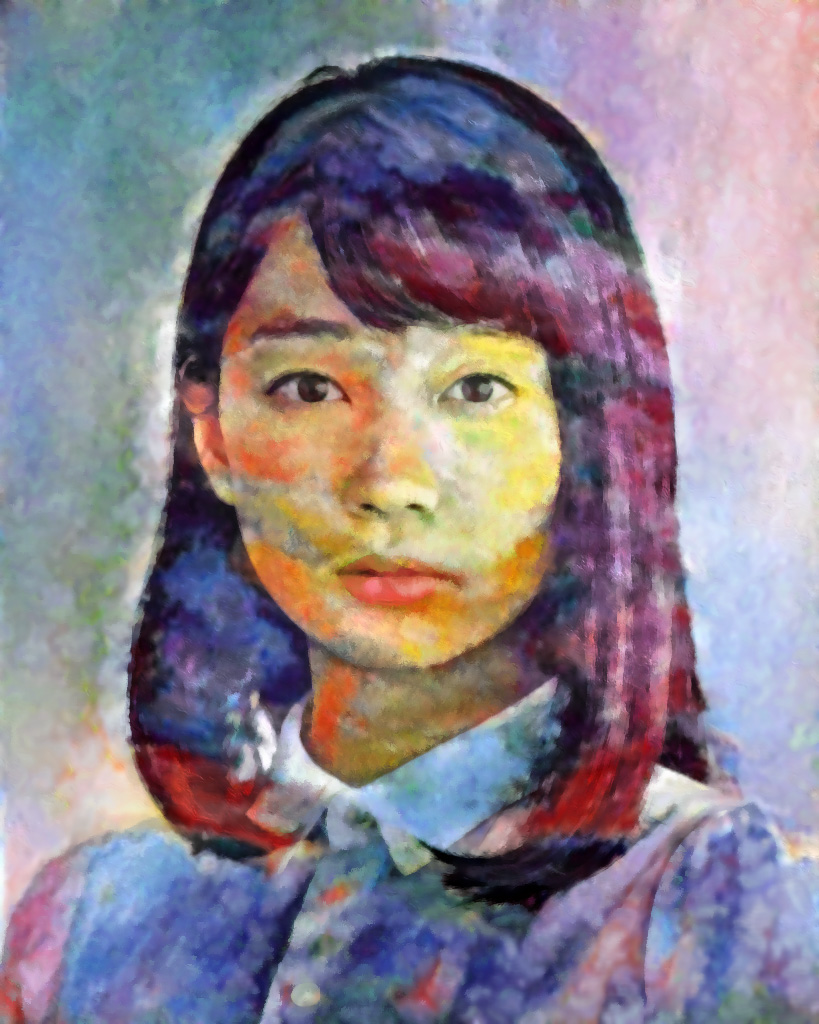}};%
\end{tikzpicture}\\

\begin{tikzpicture}
    \node{\includegraphics[width=\specFigSize]{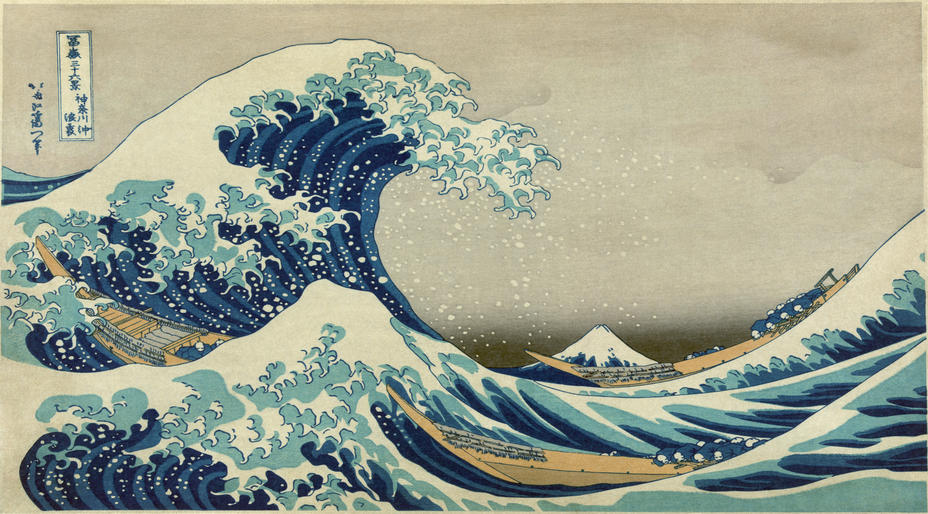}};%
\end{tikzpicture}&
\begin{tikzpicture}
    \node{\includegraphics[clip,width=\specFigSize]{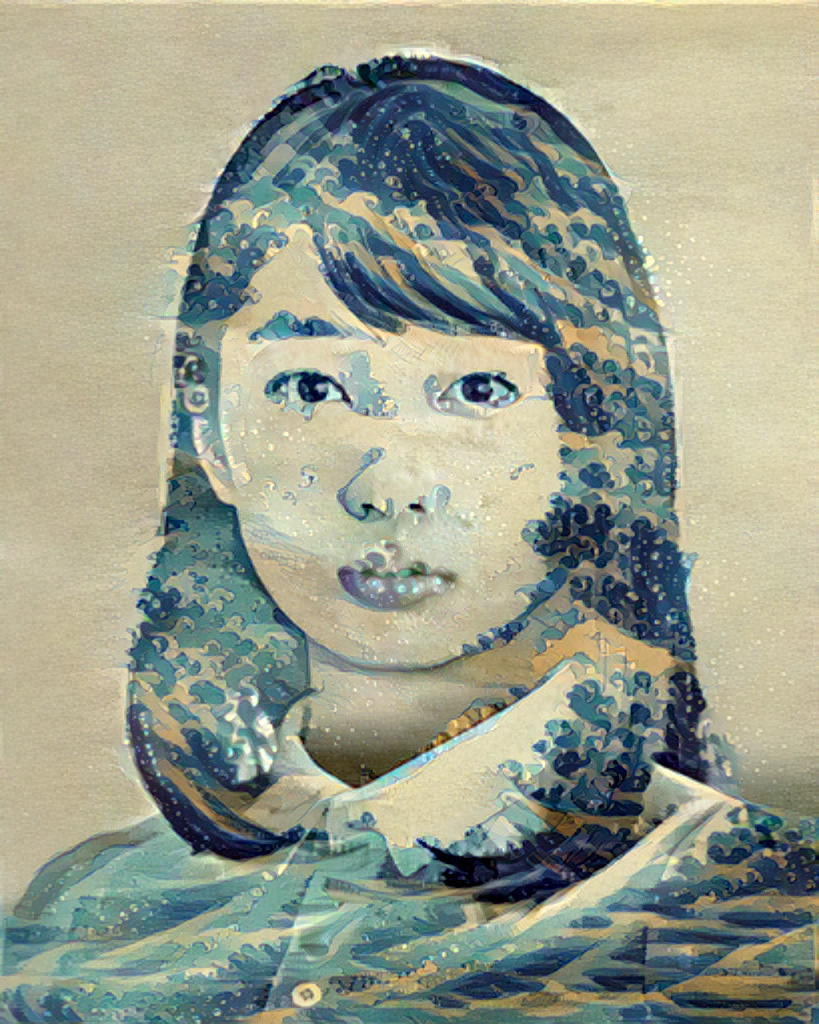}};%
\end{tikzpicture}&
\begin{tikzpicture}
    \node{\includegraphics[width=\specFigSize]{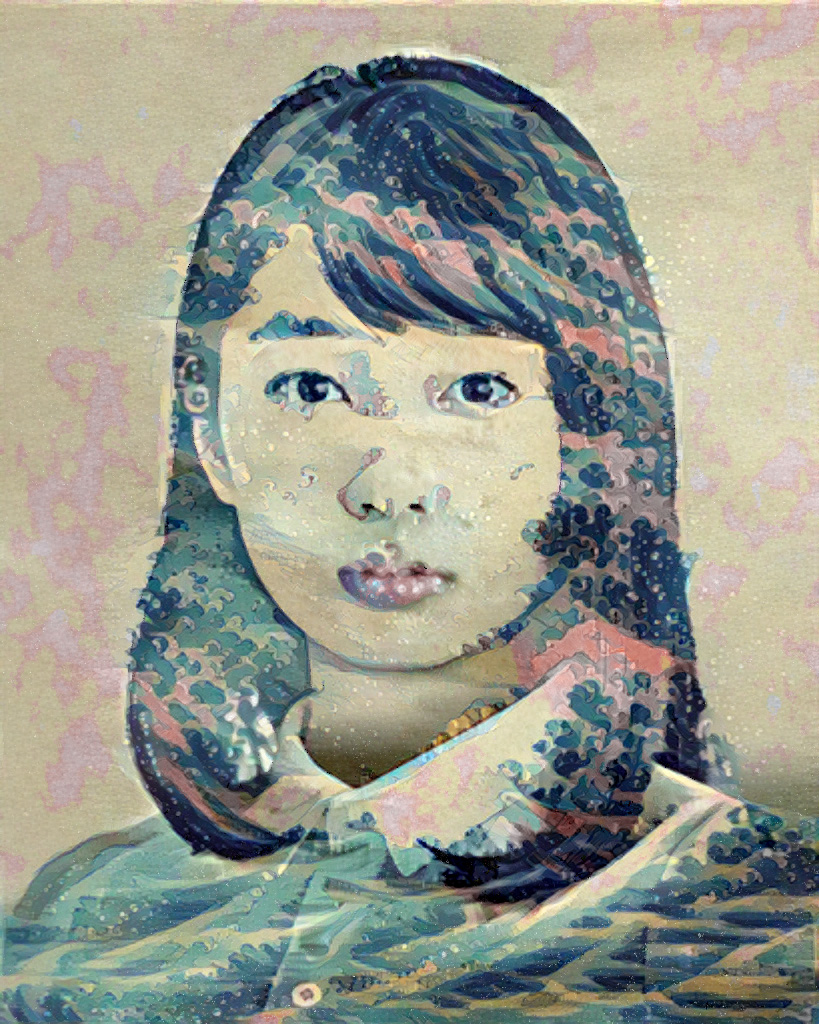}};%
\end{tikzpicture}&
\begin{tikzpicture}
    \node{\includegraphics[width=\specFigSize]{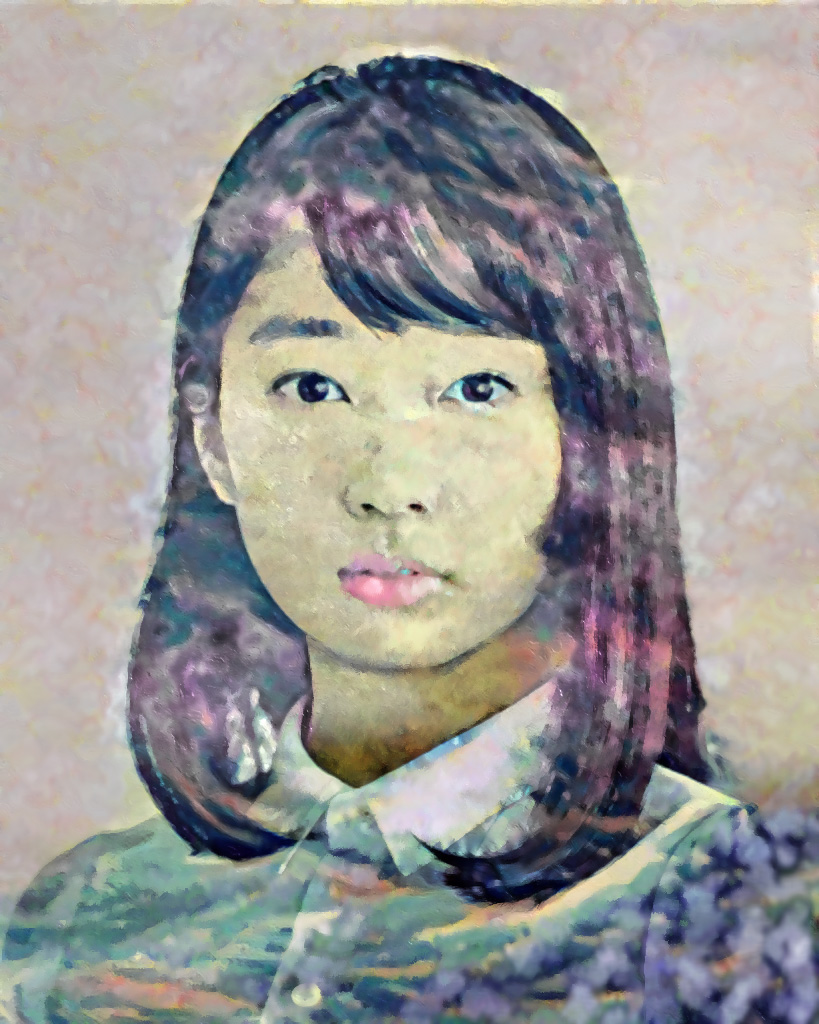}};%
\end{tikzpicture}\\

\begin{tikzpicture}
    \node{\includegraphics[width=\specFigSize]{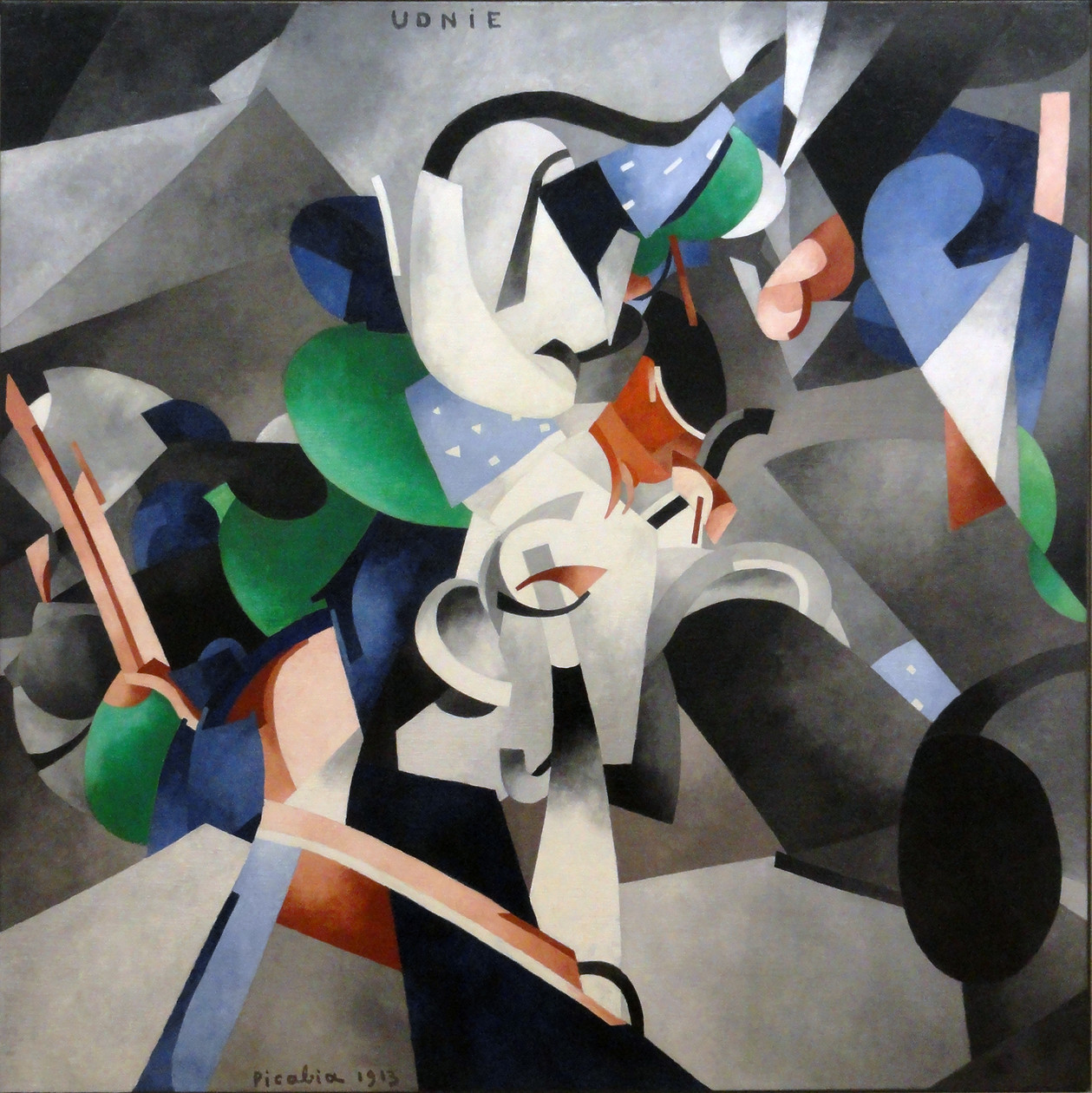}};%
\end{tikzpicture}&
\begin{tikzpicture}
    \node{\includegraphics[clip,width=\specFigSize]{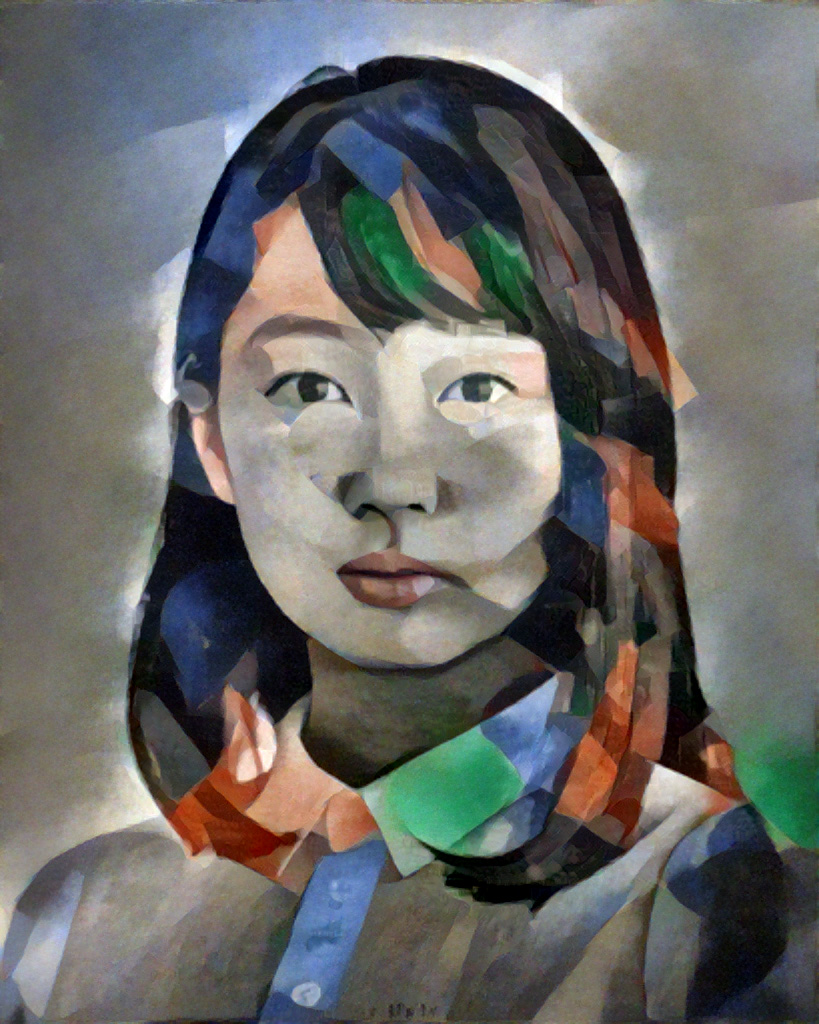}};%
\end{tikzpicture}&
\begin{tikzpicture}
    \node{\includegraphics[width=\specFigSize]{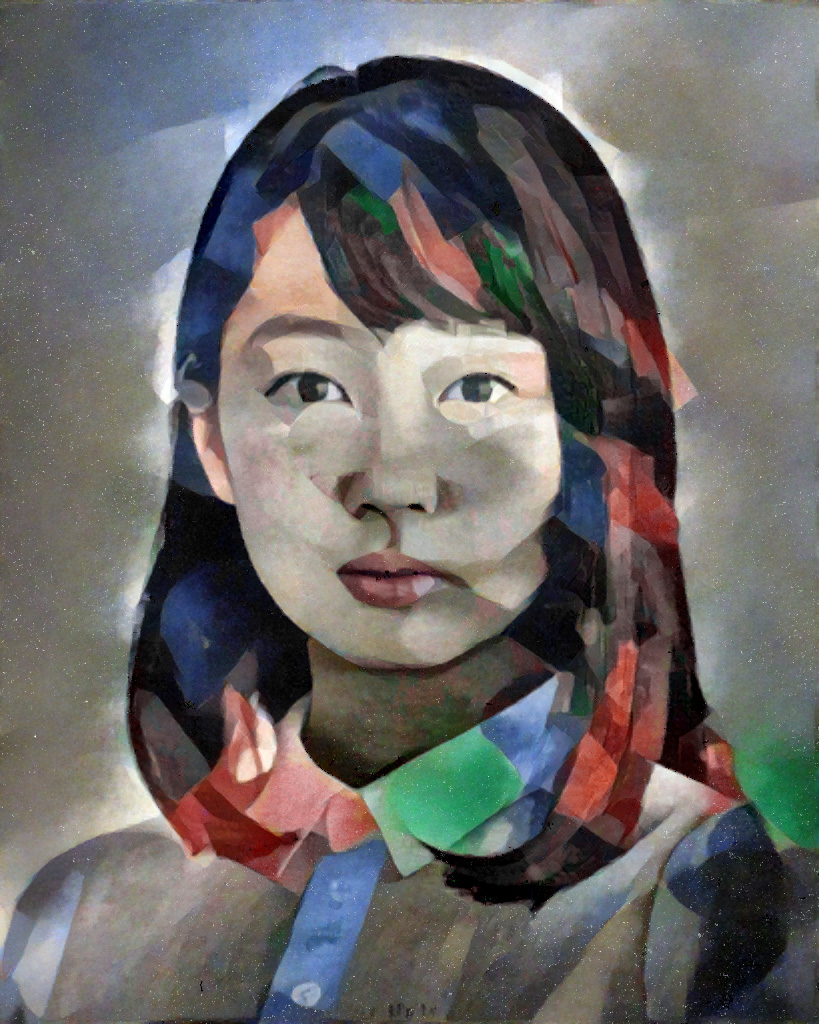}};%
\end{tikzpicture}&
\begin{tikzpicture}
    \node{\includegraphics[width=\specFigSize]{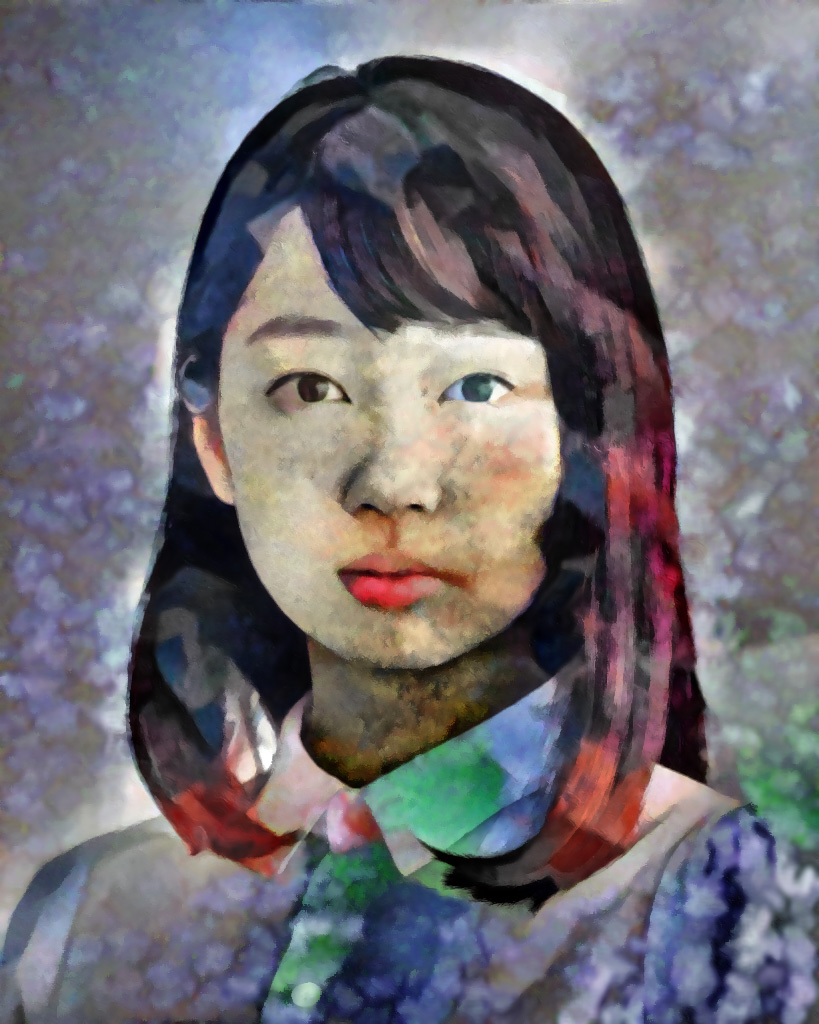}};%
\end{tikzpicture}\\

\begin{tikzpicture}
    \node{\includegraphics[width=\specFigSize]{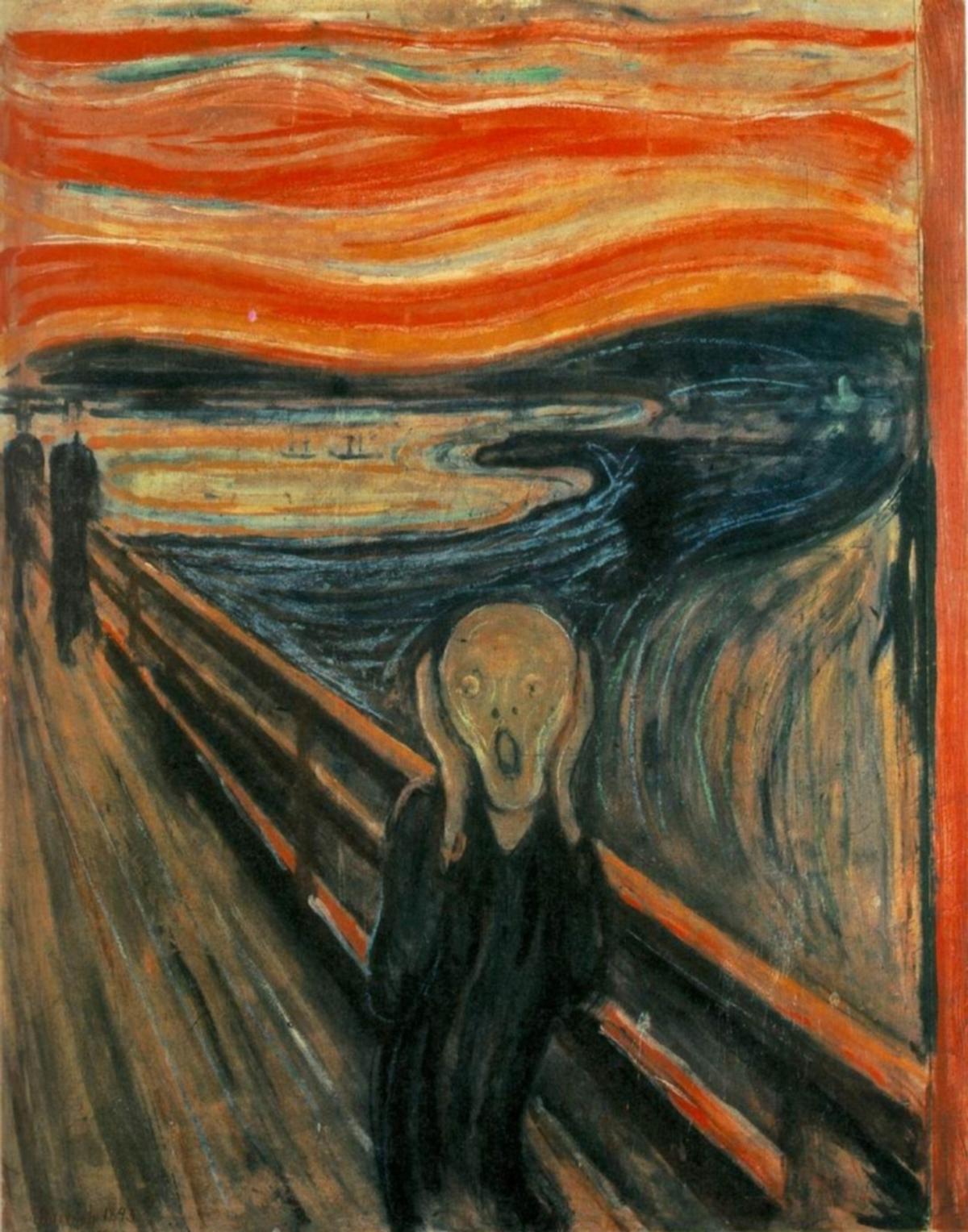}};%
\end{tikzpicture}&
\begin{tikzpicture}
    \node{\includegraphics[clip,width=\specFigSize]{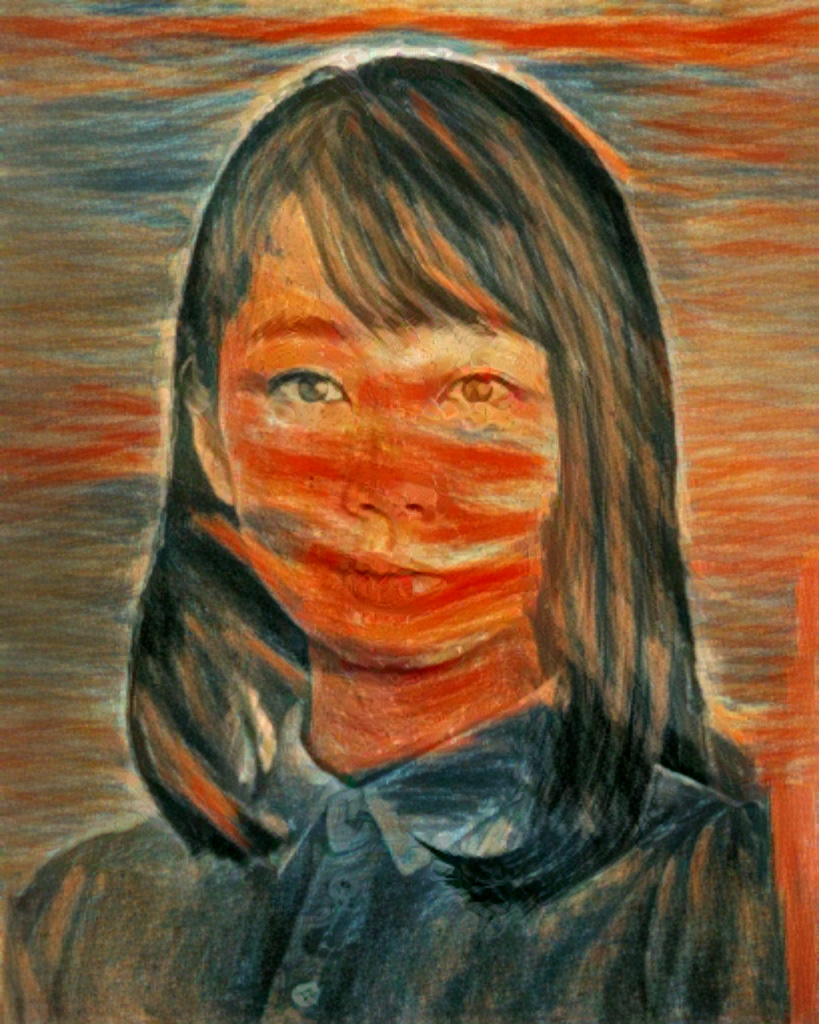}};%
\end{tikzpicture}&
\begin{tikzpicture}
    \node{\includegraphics[width=\specFigSize]{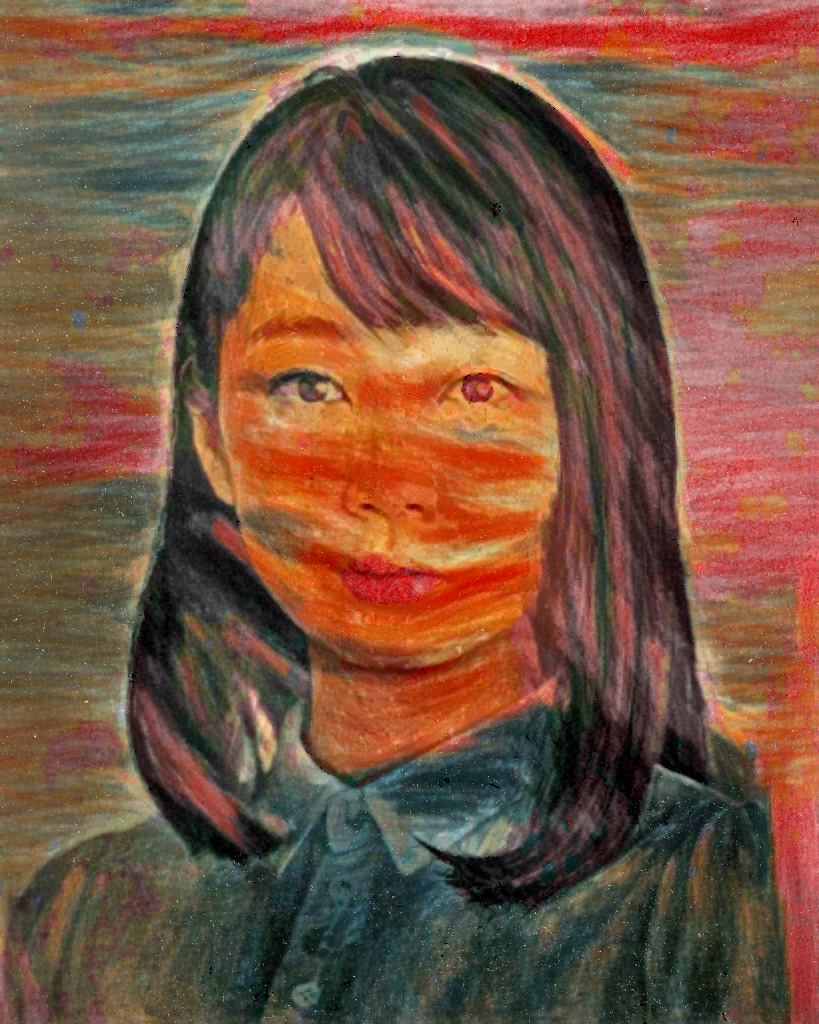}};%
\end{tikzpicture}&
\begin{tikzpicture}
    \node{\includegraphics[width=\specFigSize]{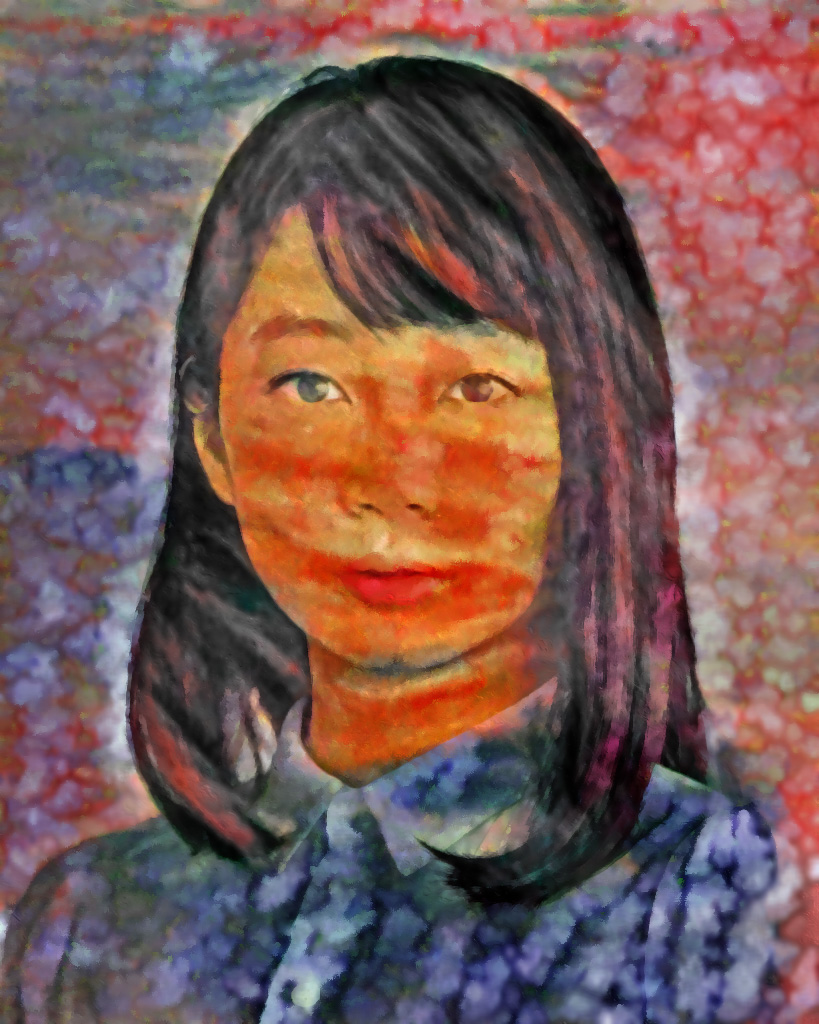}};%
\end{tikzpicture}\\

\begin{tikzpicture}
    \node{\includegraphics[width=\specFigSize]{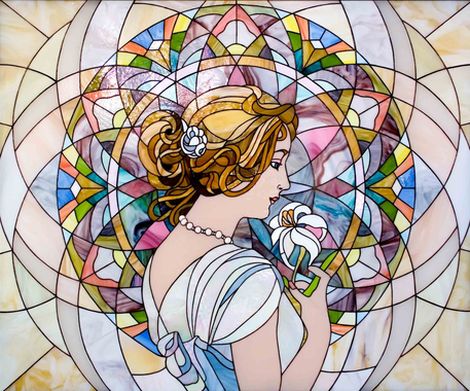}};%
\end{tikzpicture}&
\begin{tikzpicture}
    \node{\includegraphics[clip,width=\specFigSize]{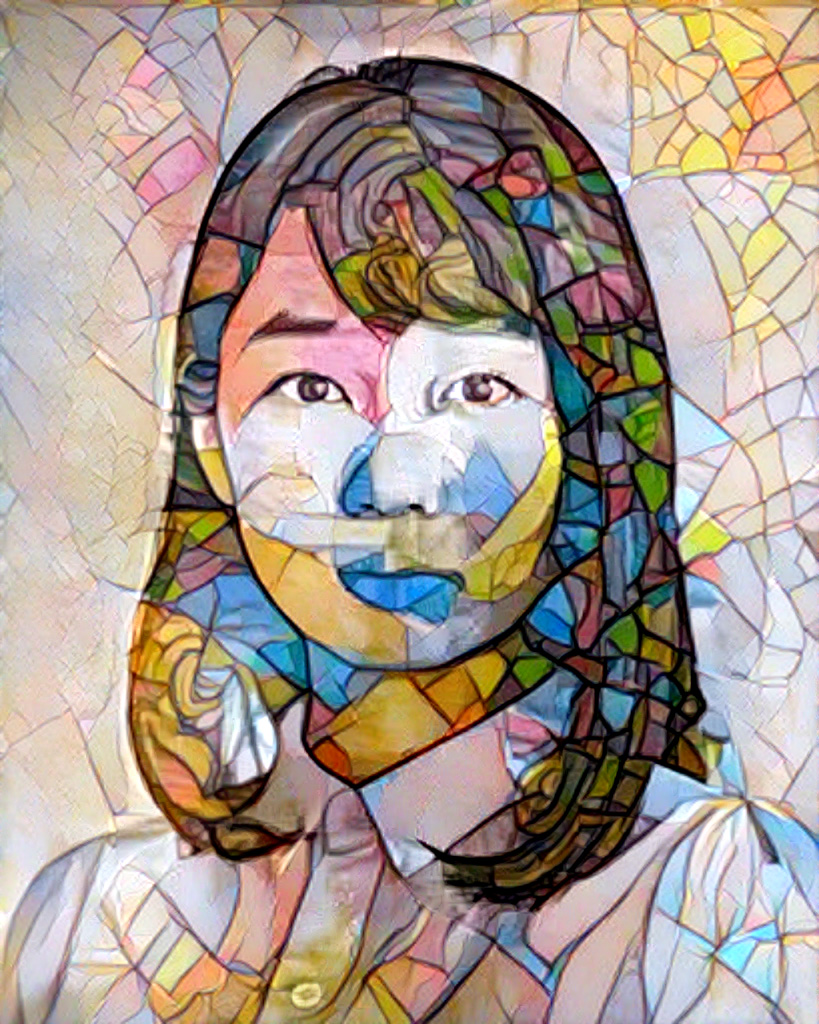}};%
\end{tikzpicture}&
\begin{tikzpicture}
    \node{\includegraphics[width=\specFigSize]{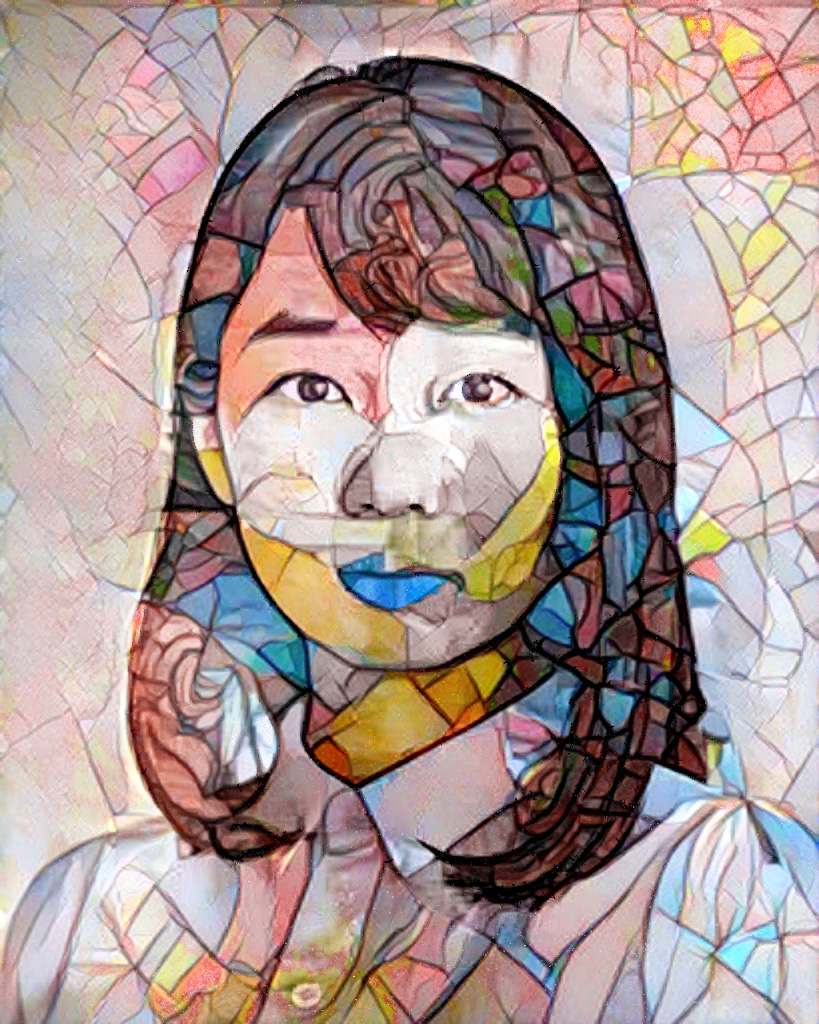}};%
\end{tikzpicture}&
\begin{tikzpicture}
    \node{\includegraphics[width=\specFigSize]{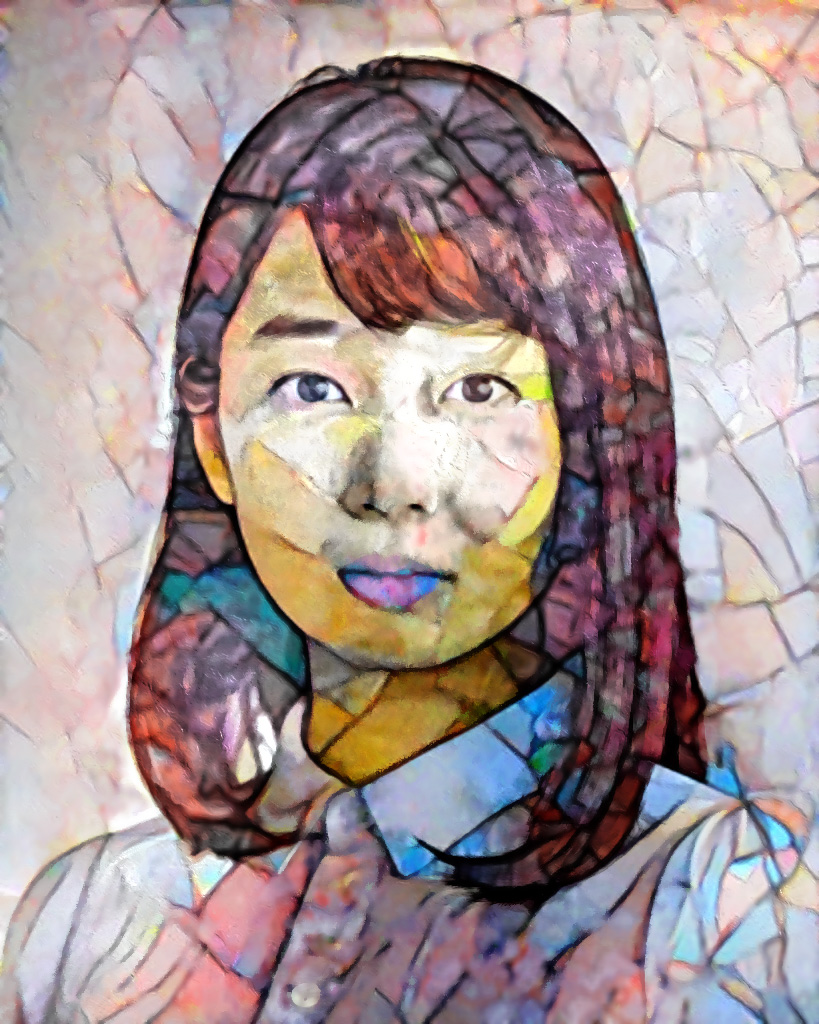}};%
\end{tikzpicture}\\

	\end{tabular}
	\caption{Style transfer capability on common NST styles.}
		\label{fig:commonnst2}
\end{figure}

In \Cref{tbl:strotss_all_pipelines}  of the main paper we compared the potential of our differentiable effects to be optimized in a (general) style transfer framework. In \Cref{fig:strotss_all_pipelines} we show example images from the in-domain optimization for the implemented effects. For this, we used content images from the \ac{NPR} benchmark dataset\cite{rosin2020nprportrait} and several style images from the same artistic domain for each effect (one example per effect is shown \Cref{fig:strotss_all_pipelines} \protect\subref{fig:st_effects:style1},\protect\subref{fig:st_effects:style2},\protect\subref{fig:st_effects:style3},\protect\subref{fig:st_effects:style4}). 

The reference for optimization is created by STROTSS \cite{kolkin2019style} style transfer. Note that it does not necessarily produce references that would be considered strictly in-domain  with the style image (\eg cartoon or line drawing domains) as the style transfer method has no concept of the actual drawing techniques used. The created reference (\eg \Cref{fig:st_effects:reference3}) is then matched as closely as possible during local parameter optimization. As is visible, only the Watercolor and Oilpaint effects are able to produce closely matching results \Cref{fig:st_effects:optresult3}. The Cartoon and \ac{XDoG} effects can, on the other hand, only be optimized with references that are closer to their range of achievable effects and are not suited for the general style transfer case. 

In 	\Cref{fig:commonnst1}  and \Cref{fig:commonnst2} we further show the general style transfer capability of oilpaint and watercolor on the set of common NST styles measured in Tab. 2. It is visible that oilpaint achieves a decent matching on most styles, but fails to create structure in some regions, while watercolor achieves an almost perfect matching of all references.

\clearpage
\FloatBarrier
\section{GAN-based image-to-image translation with PPNs}
\label{sup:sec:apdrawing_gan}

Here, we provide additional details to GAN-based image-to-image translation with PPNs. In an ablation study (\Cref{sup:sec:apdrawing_gan_ablation}) we elaborate on the architecture evaluation section on combining algorithmic effects and \acp{CNN} in the main paper (\Cref{subsec:CombiningEffects}). We then provide additional comparisons to state of the art method, show predicted parameter masks and evaluate the methods editability in \Cref{sup:sec:apdrawing_gan_editability}.

\subsection{Ablation study for APDrawing Stylization}\label{sup:sec:apdrawing_gan_ablation}

\begin{table*}[hb]\vspace{\figmargin}\vspace{\figmargin}
\begin{center}
\caption{Ablation study for our model on the APDrawing \cite{yi2019apdrawinggan} dataset. The \ac{FID} score between the train and the test set can be used as a baseline for all results.}
\label{sup:tbl:ap_drawing_ablation_study}
\vspace{0.1cm}
\begin{tabular}{|c|l|c|c|c|c|}
\hline
\ac{XDoG} & \ac{PPN} weights & Postprocessing  &  Dropout & \ac{FID} score & Key in \\
          &                  & Architecture   &          &                & \Cref{sup:fig:apdrawing_comparison_extra}\\
\hline
\ding{55} & - & U-Net & \ding{51} & 75.27 & - \\
\ding{55} & - & U-Net & \ding{55} & 71.26 & - \\
\ding{55} & - & ResNet & \ding{51} & 70.00 & -\\
\ding{55} & - & ResNet & \ding{55} & 62.44 & (a) \\
\hline
\ding{51} & Random & U-Net & \ding{51} & 86.73 & - \\
\ding{51} & Random & U-Net & \ding{55} & 89.93 & - \\
\ding{51} & Random & ResNet  & \ding{51} & 71.92 & -\\
\ding{51} & Random & ResNet  & \ding{55} & \textbf{60.55} & (c) \\
\hline
\ding{51} & ImageNet & U-Net & \ding{51} & 100.41 & -\\
\ding{51} & ImageNet & U-Net & \ding{55} & 79.56  & -\\
\ding{51} & ImageNet & ResNet  & \ding{51} & 76.25 & -\\
\ding{51} & ImageNet & ResNet  & \ding{55} & 64.81 & -\\
\hline
\ding{51} & - & U-Net & \ding{51} & 71.64 & -\\
\ding{51} & - & U-Net & \ding{55} & 75.40 & -\\
\ding{51} & - & ResNet  & \ding{51} & 71.56 & -\\
\ding{51} & - & ResNet  & \ding{55} & 73.77 & -\\
\hline
\hline
\multicolumn{4}{|c}{APDrawing \acs{GAN} \cite{yi2019apdrawinggan}} & 62.14 & (b) \\
\multicolumn{4}{|c}{Train vs. Test} & 49.72 & -\\
\hline
\end{tabular}
\vspace{\figmargin}\vspace{\figmargin}
\end{center}
\end{table*}

The results are summarized in \Cref{sup:tbl:ap_drawing_ablation_study}. 
For the ResNet-50 backbone of the \ac{PPN}, we compare initialization with random weights and weights pre-trained on ImageNet. 
For the convolutional postprocessing network in our generator, we compare the ResNet-based network architecture of Johnson~\etal\cite{johnson2016perceptual} and a classical U-Net\cite{ronneberger2015u} without residual blocks following Isola \etal\cite{isola2017image}\footnote{We adapt the implementation of \url{https://github.com/junyanz/pytorch-CycleGAN-and-pix2pix}}. 
We also investigate the effect of dropout. As it decreases the \ac{FID} compared to architectures trained without dropout, the model clearly benefits from having the full learning capacity at its disposal during training. Also, pre-trained weights on ImageNet do not increase the ability of the model to adapt to data successfully.

We also test whether parameter prediction and thus differentiable effects are necessary by comparing our method to the same approach using a fixed parameter preset, which is optimized for portrait data specifically. 
Our \ac{PPN} network can dynamically adapt parameters locally to input images. 
As the results using our \ac{PPN} and those with a fixed preset differ by a large margin, we conclude that parameter prediction and thus differentiable effects play an important role in the success of our method. We do not include results using only the differentiable \ac{XDoG} effect in conjunction with a \ac{PPN}. All experiments without a separate convolution network failed with mode collapse of the generator, \ie the \ac{PPN} started to predict the same parameters regardless of the input. 
We argue that this behavior is related to the missing ability of the \ac{XDoG} effect to model the artist's style accurately.

\subsection{Additional results and editability}\label{sup:sec:apdrawing_gan_editability}

\begin{figure*}[!hb]
    \centering
    \small{
    \begin{subfigure}[b]{0.24\textwidth}
        \includegraphics[width=\textwidth]{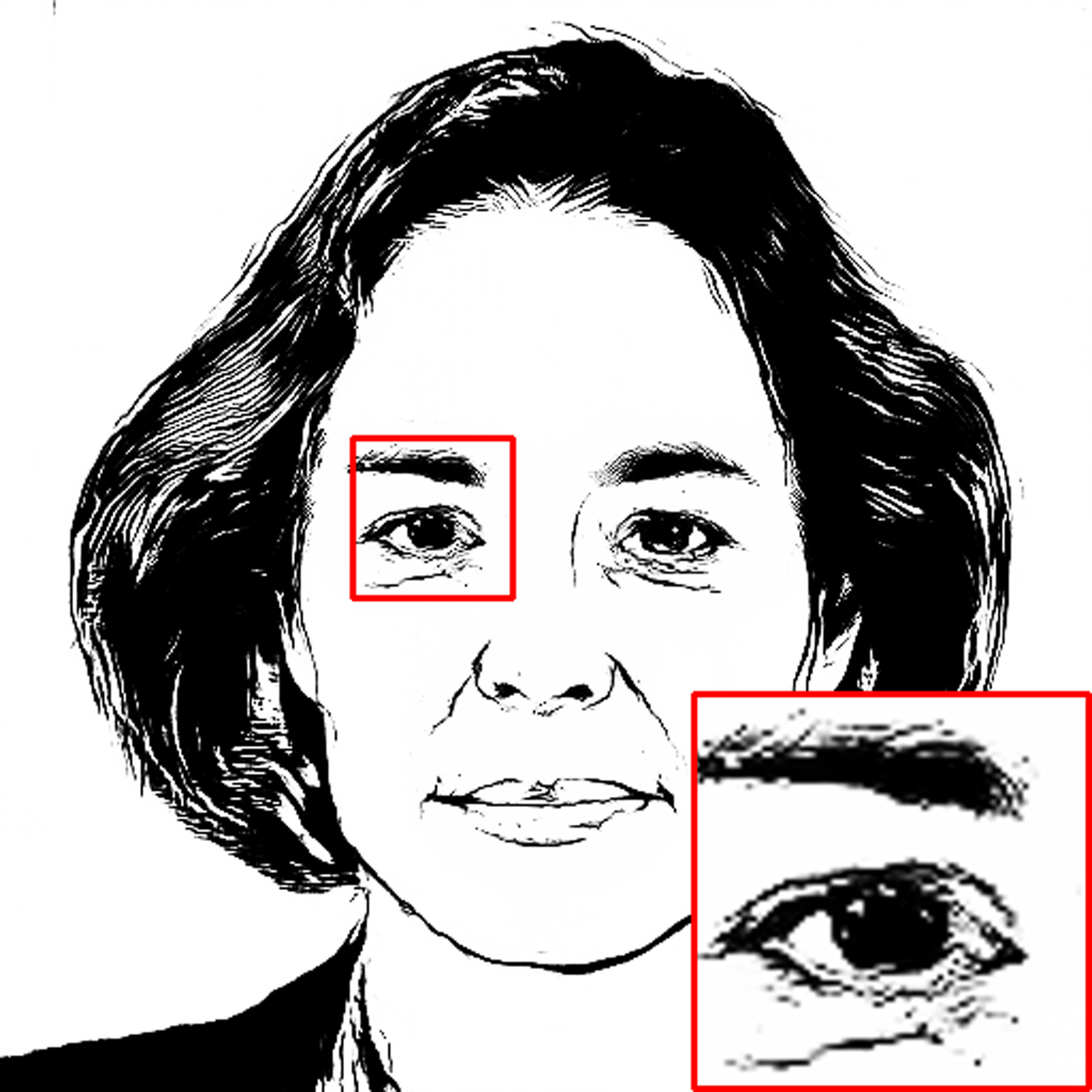}
    \end{subfigure}
    \begin{subfigure}[b]{0.24\textwidth}
        \includegraphics[width=\textwidth]{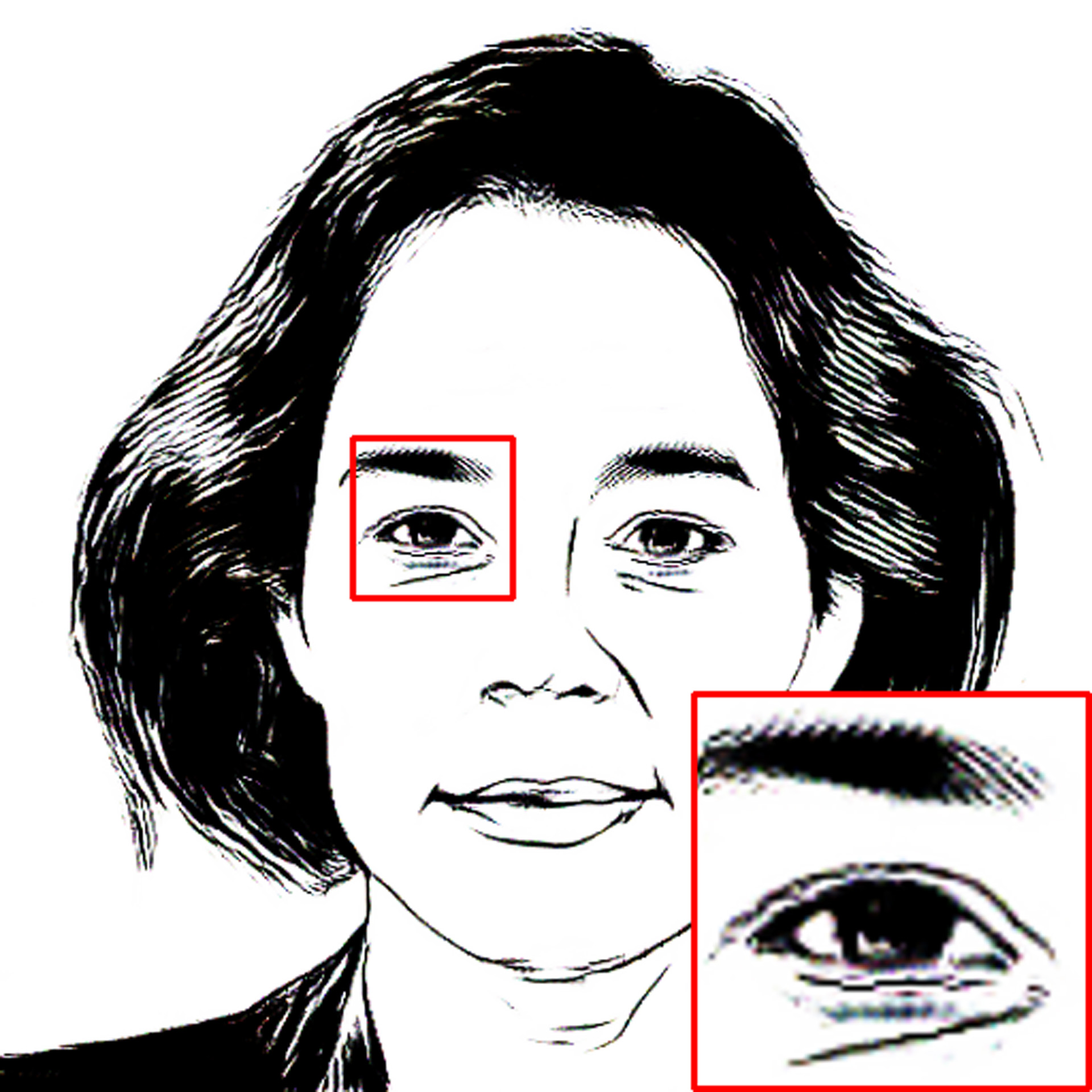}
    \end{subfigure}
    \begin{subfigure}[b]{0.24\textwidth}
        \includegraphics[width=\textwidth]{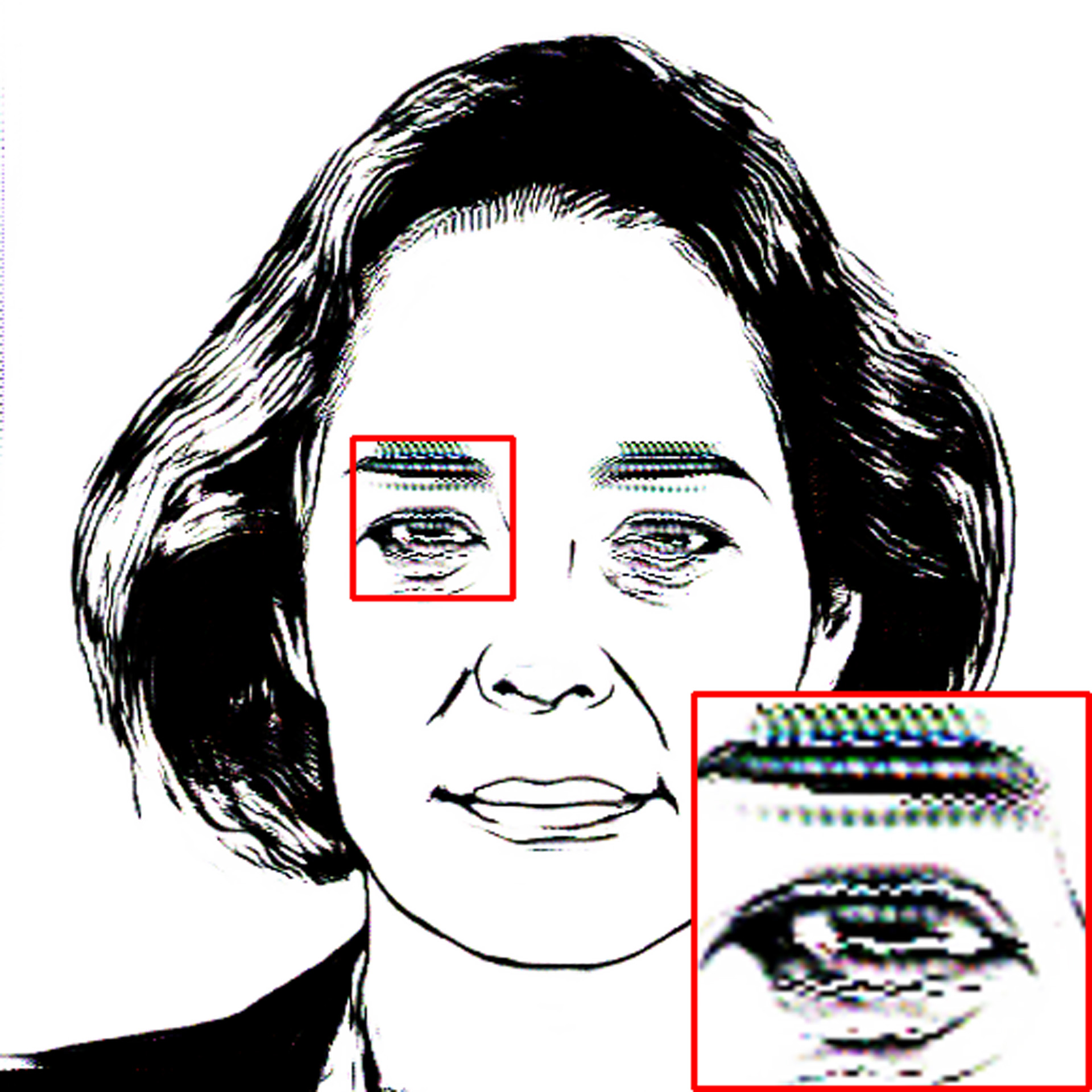}
    \end{subfigure}
    \begin{subfigure}[b]{0.24\textwidth}
        \includegraphics[width=\textwidth]{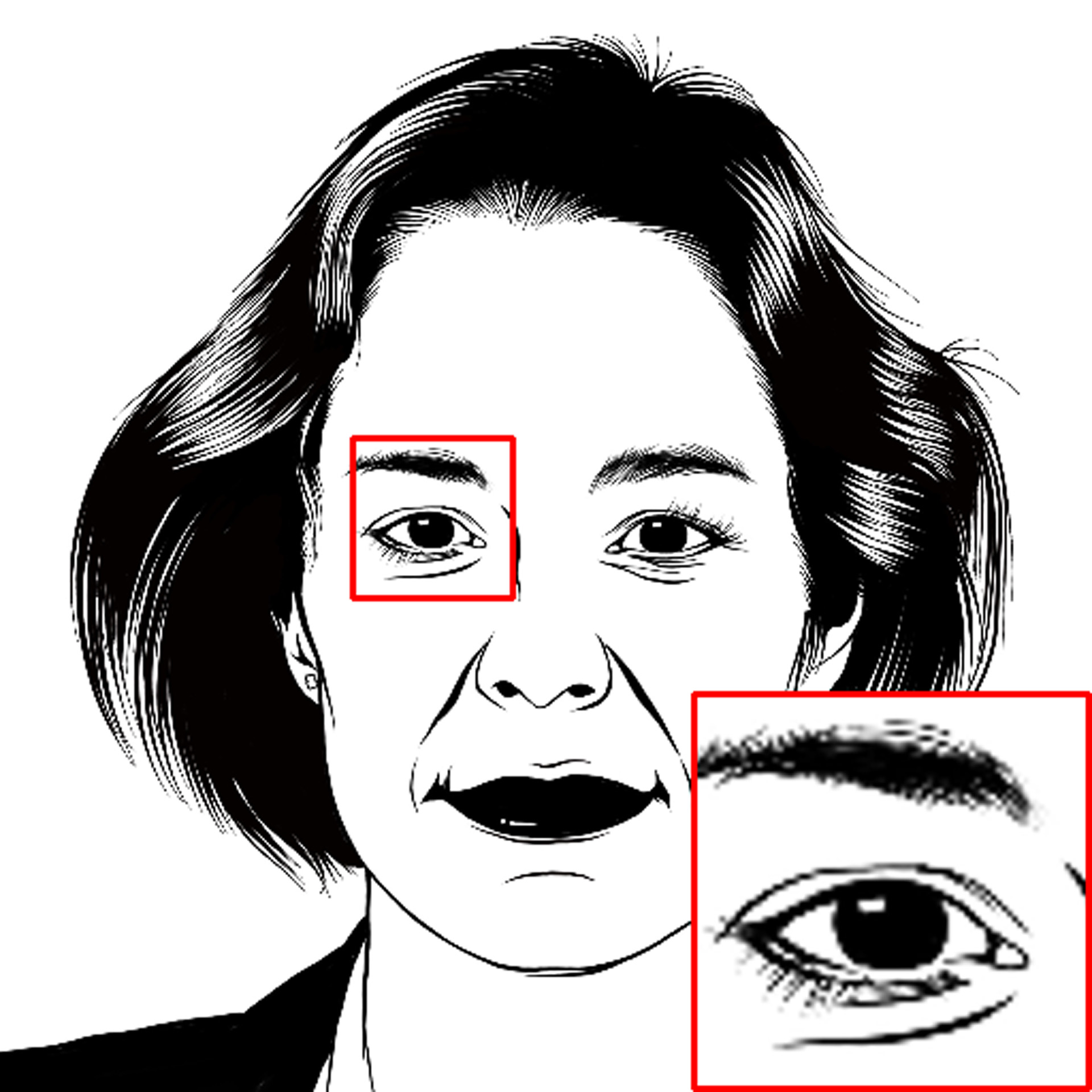}
    \end{subfigure}
    
    \begin{subfigure}[b]{0.24\textwidth}
        \includegraphics[width=\textwidth]{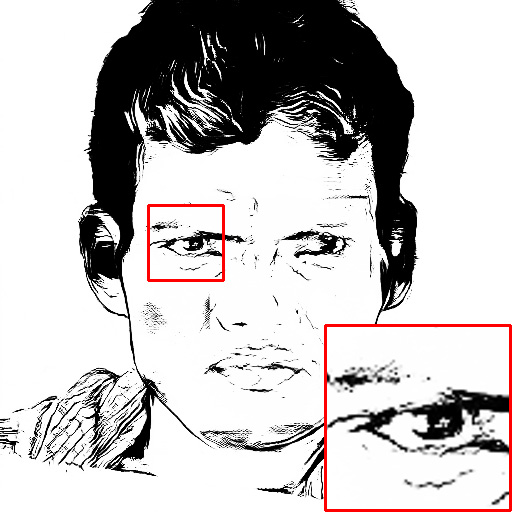}
    \end{subfigure}
    \begin{subfigure}[b]{0.24\textwidth}
        \includegraphics[width=\textwidth]{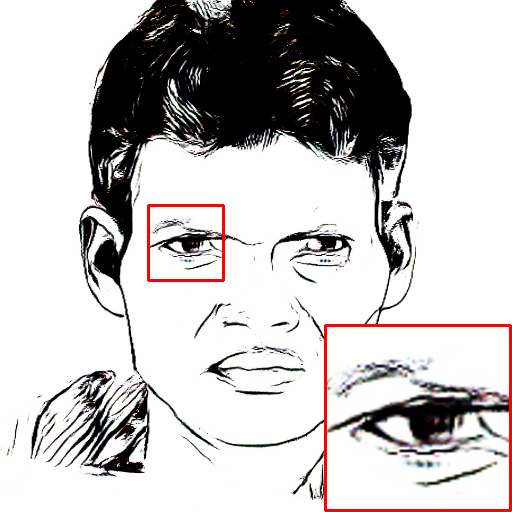}
    \end{subfigure}
    \begin{subfigure}[b]{0.24\textwidth}
        \includegraphics[width=\textwidth]{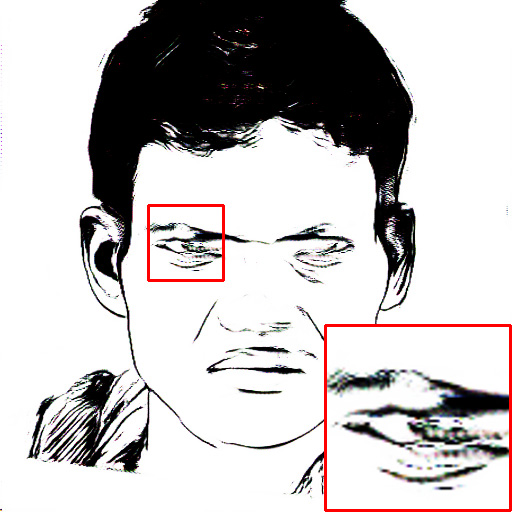}
    \end{subfigure}
    \begin{subfigure}[b]{0.24\textwidth}
        \includegraphics[width=\textwidth]{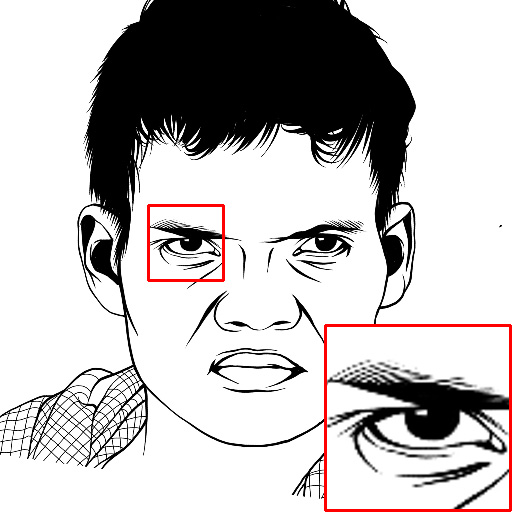}
    \end{subfigure}

    \begin{subfigure}[b]{0.24\textwidth}
        \includegraphics[width=\textwidth]{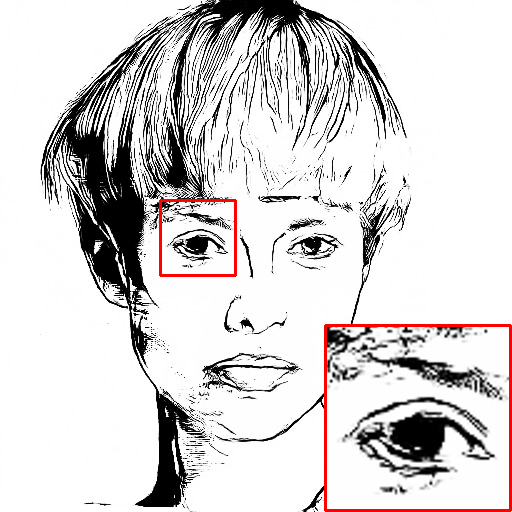}
    \end{subfigure}
    \begin{subfigure}[b]{0.24\textwidth}
        \includegraphics[width=\textwidth]{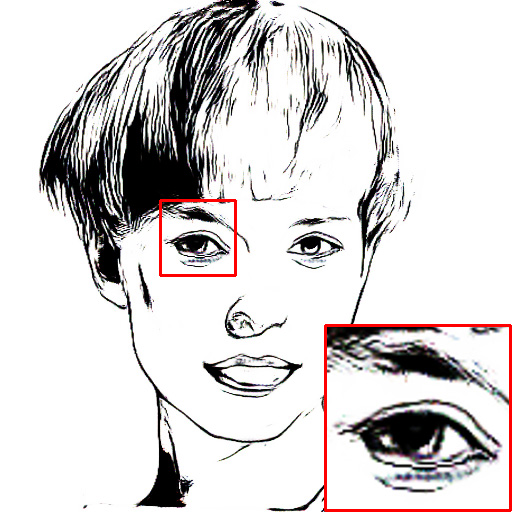}
    \end{subfigure}
    \begin{subfigure}[b]{0.24\textwidth}
        \includegraphics[width=\textwidth]{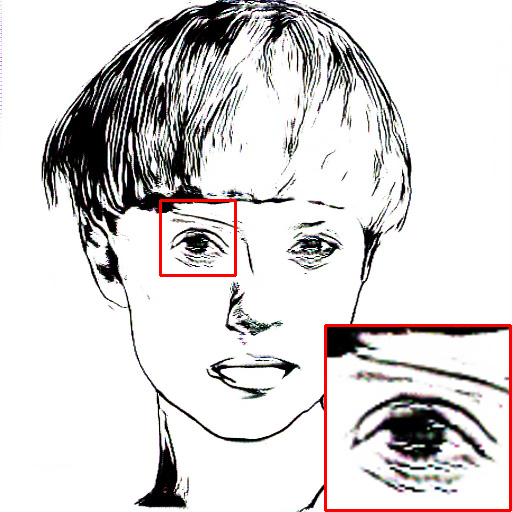}
    \end{subfigure}
    \begin{subfigure}[b]{0.24\textwidth}
        \includegraphics[width=\textwidth]{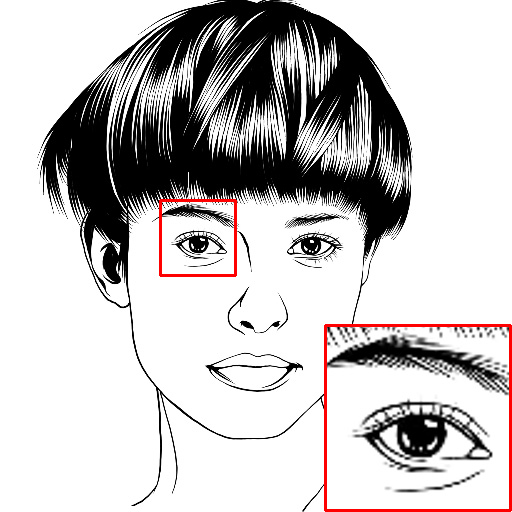}
    \end{subfigure}
    
    \begin{subfigure}[b]{0.24\textwidth}
        \includegraphics[width=\textwidth]{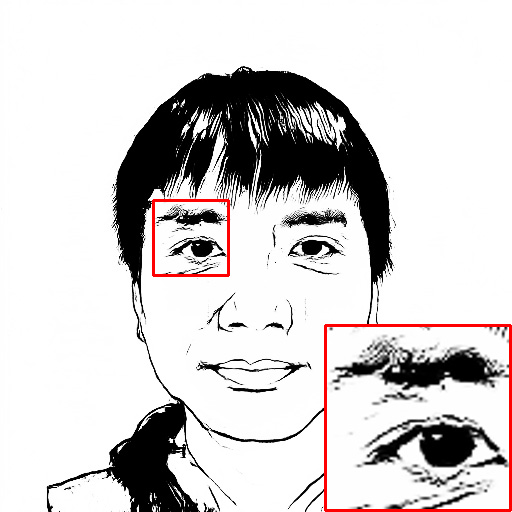}
        \caption{Yi \etal\cite{yi2019apdrawinggan}}
    \end{subfigure}
    \begin{subfigure}[b]{0.24\textwidth}
        \includegraphics[width=\textwidth]{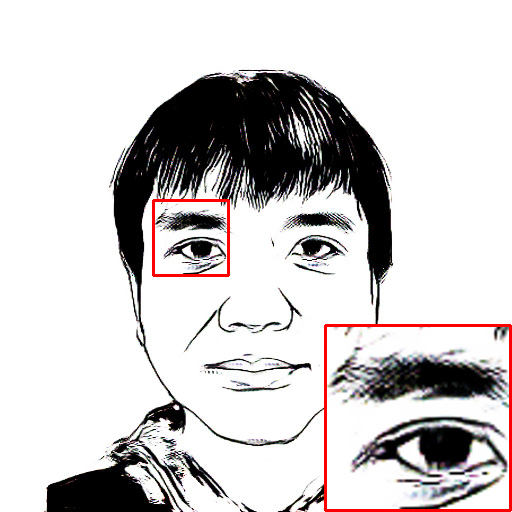}
        \caption{Our method}
    \end{subfigure}
    \begin{subfigure}[b]{0.24\textwidth}
        \includegraphics[width=\textwidth]{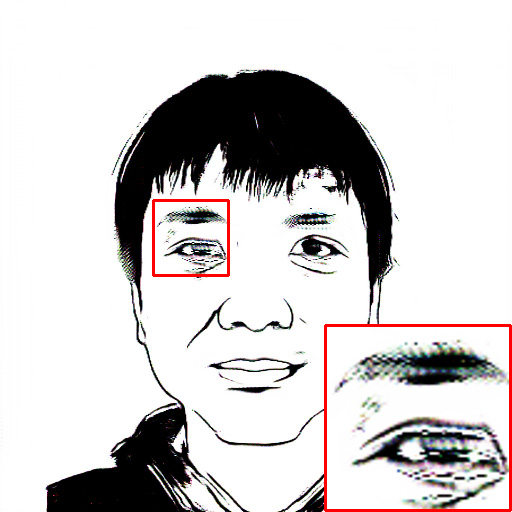}
        \caption{CNN-only}
    \end{subfigure}
    \begin{subfigure}[b]{0.24\textwidth}
        \includegraphics[width=\textwidth]{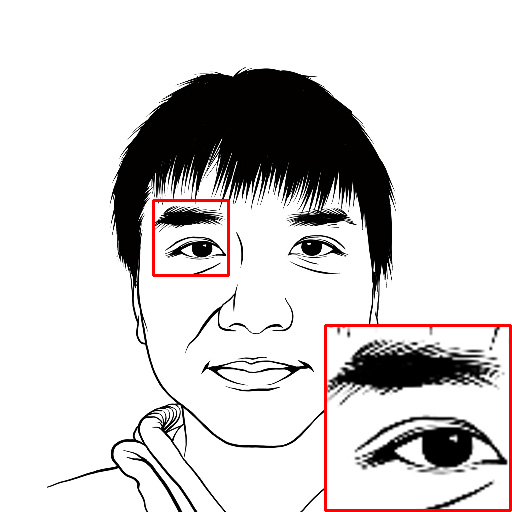}
        \caption{Ground truth}
    \end{subfigure}
    
    \caption{Results on the APDrawing \cite{yi2019apdrawinggan} dataset, our method uses \ac{XDoG} and a postprocessing \ac{CNN}. Our method, in contrast to APDrawing \acp{GAN}\cite{yi2019apdrawinggan,YiXLLR20}, often produces more consistent lines \eg for the eyes, and does not need any known facial landmarks at both training and inference time. 
    \label{sup:fig:apdrawing_comparison_extra}
    }
    }
    \end{figure*}

\begin{figure*}
    \centering
      \begin{tabular}[c]{c}
        \begin{subfigure}[c]{0.6\textwidth} 
            \includegraphics[width=\textwidth, trim={0.65cm 0 23cm 0},clip]{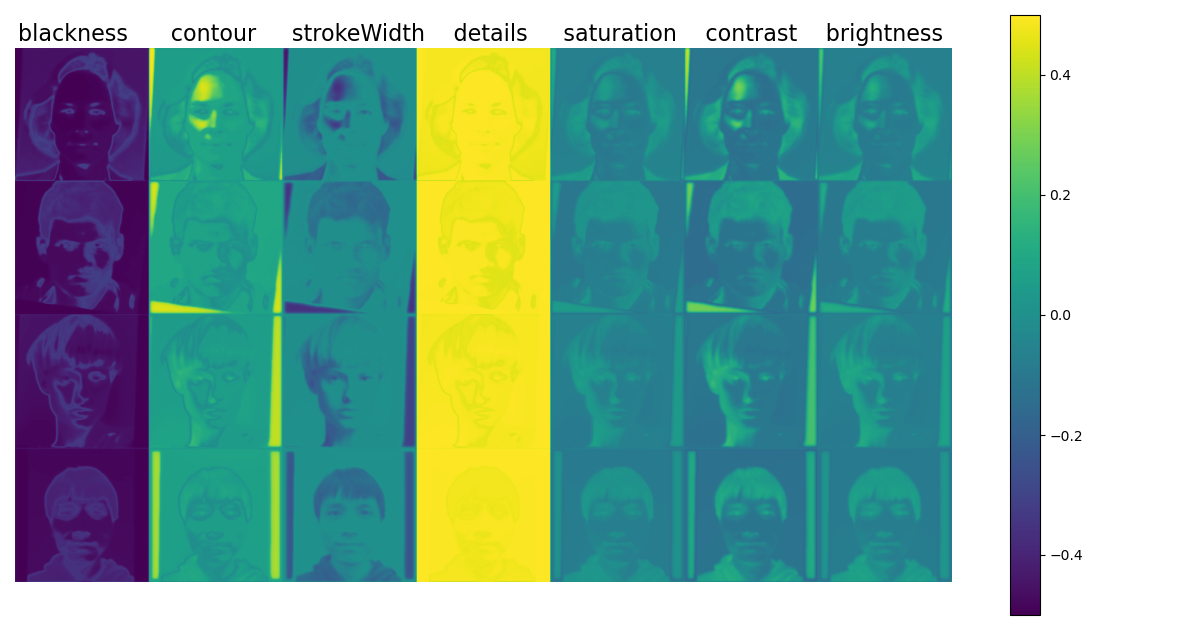}
        \end{subfigure}\\
        \begin{subfigure}[c]{0.6\textwidth} 
            \includegraphics[width=\textwidth, trim={19.5cm 0 4cm 0},clip]{supplemental/parameter_plot}
        \end{subfigure}
      \end{tabular}   
        \caption{In addition to the generated outputs in figure \ref{sup:fig:apdrawing_comparison_extra}, we show the parameter masks that have been predicted for the first stage of our method (XDoG). See \Cref{fig:apdrawing_2stages} of the main paper for a plot of the intermediate results after processing with XDoG.}
        \label{sup:fig:apdrawing_parameter_plot}
    \end{figure*}

\paragraph{Results.} \Cref{sup:fig:apdrawing_comparison_extra} shows additional comparisons of our model trained on APDrawing~\cite{yi2019apdrawinggan} against state of the art methods. We plot the  parameter masks of our results in the latter figure in \Cref{sup:fig:apdrawing_parameter_plot}.

\begin{figure}
        \centering
        \includegraphics[width=\textwidth]{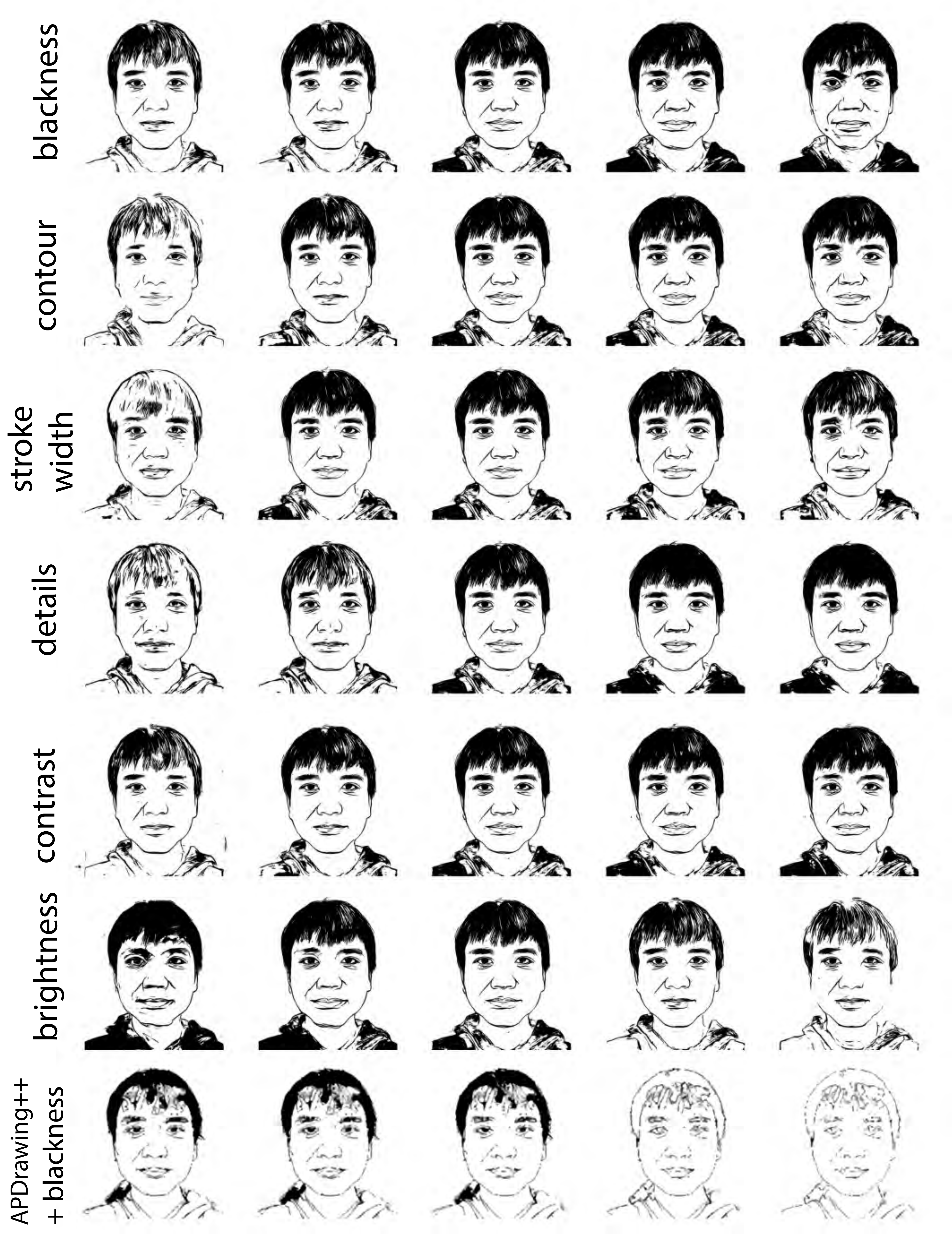}
        \caption{Adjusting parameters of the \ac{XDoG} effect after prediction using our \ac{XDoG}+CNN method trained on APDrawing \cite{yi2019apdrawinggan}. Each row corresponds to global changes of one parameter from lowest to highest value in the visual range, where unedited results (i.e., the predicted parameters) are shown in the middle column.}
        \label{fig:apdrawing_adapt_parameters}
    \end{figure}

\paragraph{Editability.}
In \Cref{fig:apdrawing_adapt_parameters} we show that while the postprocessing CNN reduces the stylistic variance that can be achieved with  \ac{XDoG}, results main editable vs. results of APDrawingGAN \cite{yi2019apdrawinggan} and also remain stylistically close to the reference. Combining pretrained GANs such as APDrawingGAN++ \cite{YiXLLR20} with a \ac{XDoG} effect for postprocessing, on the other hand, does not yield visually appealing results and results deviate from the style of the reference, as shown in the last row, thus validating our integrated training approach.

\FloatBarrier
\section{Additional Results}
\label{sup:sec:AdditionalResults}
This section provides additional results of our method, \ie for content-adaptive effects \Cref{sup:sec:AdditionalResults:contentadaptive}, and parametric style transfer \Cref{sup:sec:AdditionalResults:styletransfer}. 
Further results on APDrawing depicted in \Cref{sup:fig:apdrawing_npr} demostrate that our model generalizes to portraits from other datasets, such as the \ac{NPR} benchmark~\cite{rosin2020nprportrait}. 
Finally, we show baseline results of our implemented algorithmic differentiable effects with different parameter variants in \Cref{sup:sec:AdditionalResults:effectpresets}.

\subsection{Content-adaptive Effects}\label{sup:sec:AdditionalResults:contentadaptive}

\begin{figure}[!h]
    \centering
    \includegraphics[trim={0cm 0.4cm 0cm 0.4cm},width=1.0\textwidth,clip]{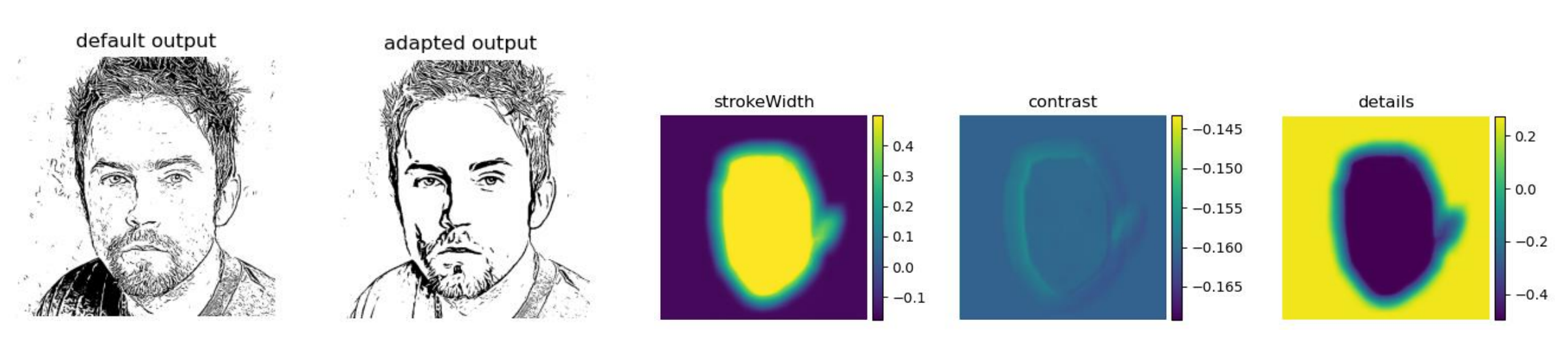}
    \caption{Facial feature enhancement and predicted local parameter maps.}
    \label{fig:facialenhancement}
\end{figure}

\begin{figure}[!h]
    \centering
    \includegraphics[width=0.75\textwidth]{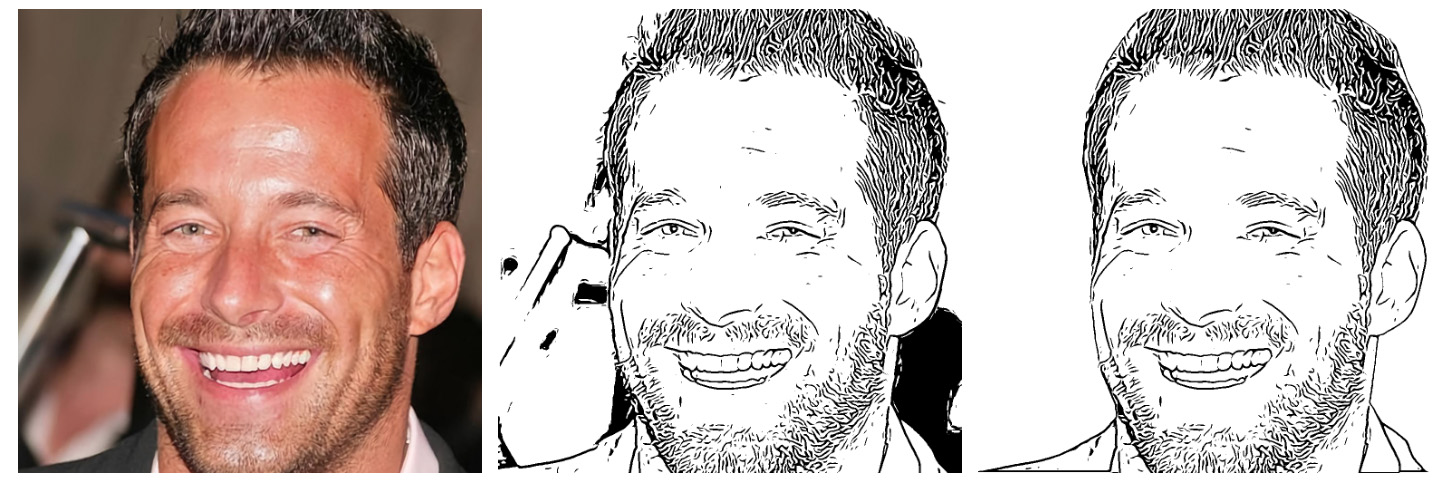}
    \includegraphics[width=0.75\textwidth]{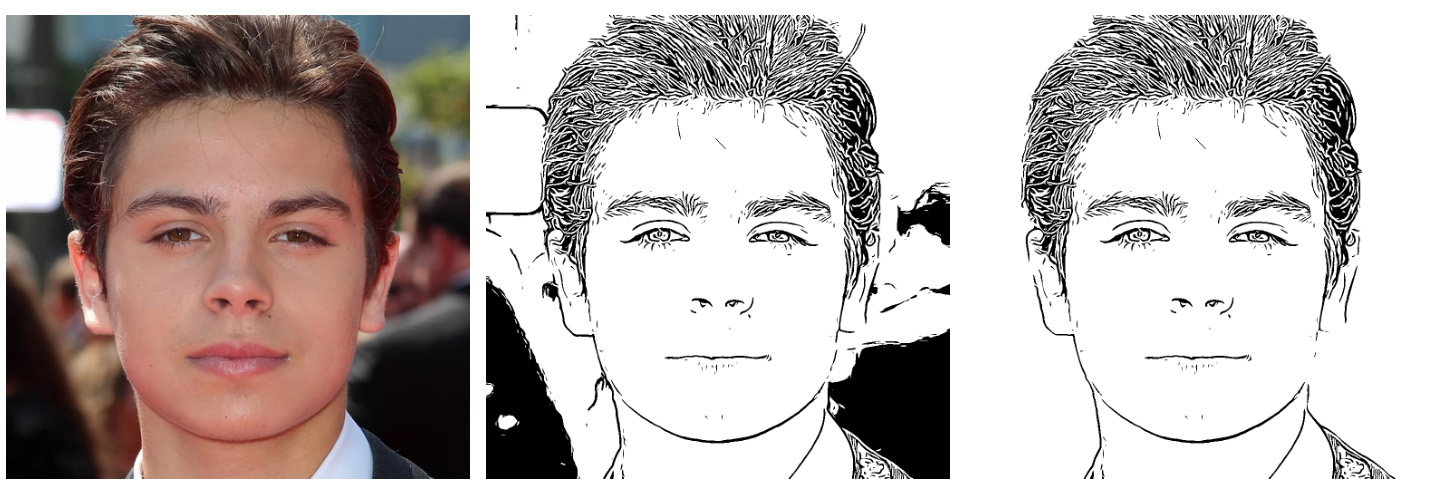}
    \caption{Background removal. The \ac{PPN} learns to reduce details and stroke-width for the \ac{XDoG} filter in such a way that background details seen in the default output (middle) are removed the output with locally adapted parameters (right).}\vspace{\figmargin}
    \label{fig:bgremoval}
\end{figure}

\FloatBarrier
\clearpage

\subsection{Style Transfer}\label{sup:sec:AdditionalResults:styletransfer}

\begin{figure}[!hb]\vspace{-0.5cm}
    \centering
\begin{subfigure}[b]{0.24\textwidth}
    \begin{minipage}[t][0.75\textwidth][b]{\textwidth}
        \rotatebox[origin=t]{90}{Style}\raggedleft\\[0.2\textwidth]
        Content\centering\vspace{0.5ex}
    \end{minipage}
    \includegraphics[width=\textwidth]{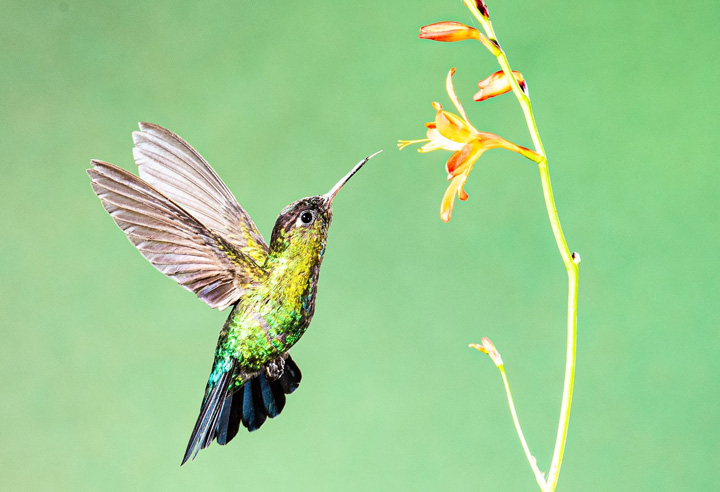}\\[0.3ex]
    \includegraphics[width=\textwidth]{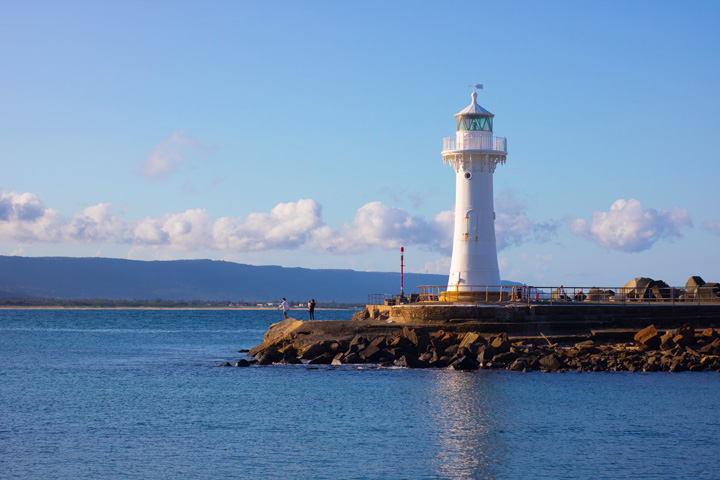}\\[0.3ex]
    \includegraphics[width=\textwidth]{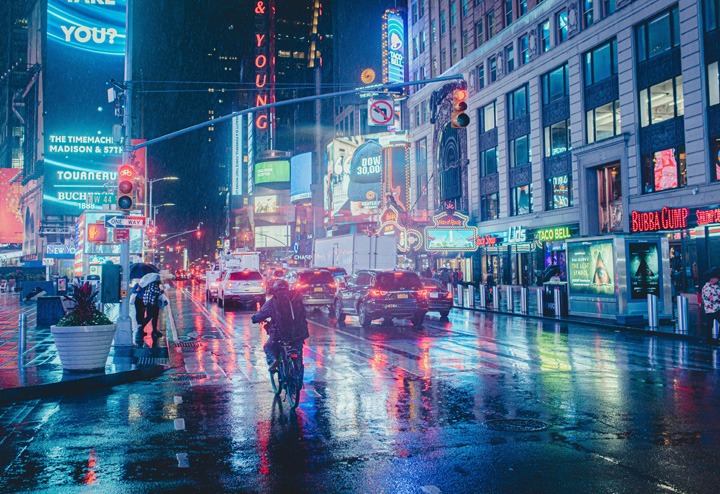}\\[0.3ex]
    \includegraphics[width=\textwidth]{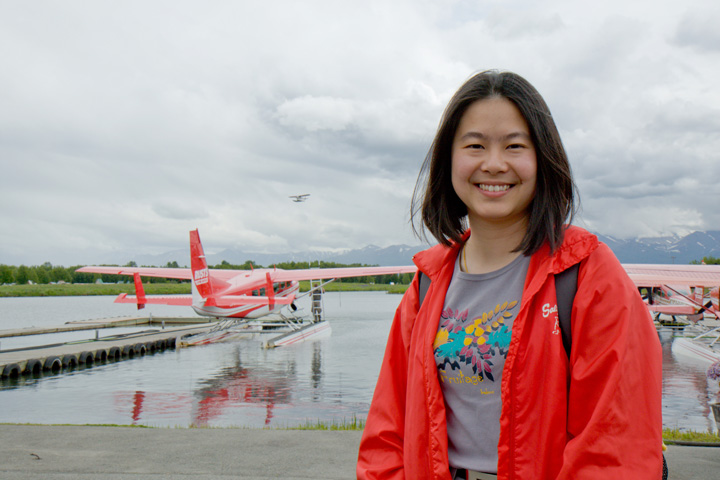}\\[0.3ex]
    \includegraphics[width=\textwidth]{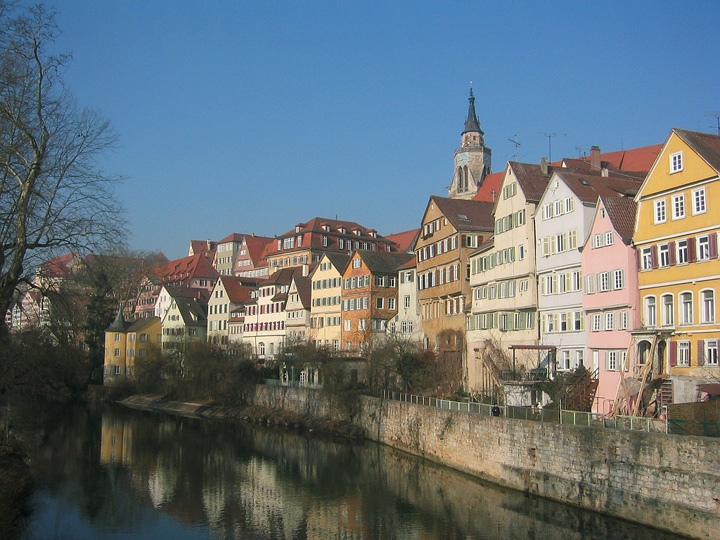}\\[0.3ex]
\end{subfigure}
\begin{subfigure}[b]{0.24\textwidth}
    \includegraphics[width=\textwidth,height=0.75\textwidth]{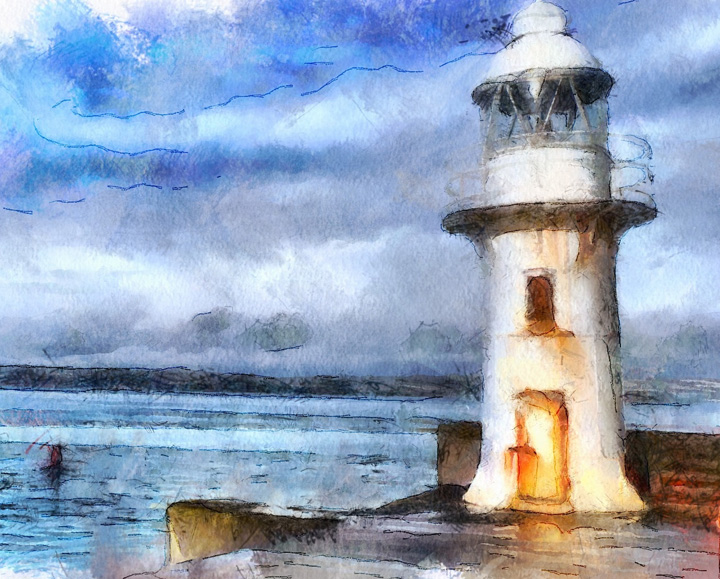}\\[0.3ex]
    \includegraphics[width=\textwidth]{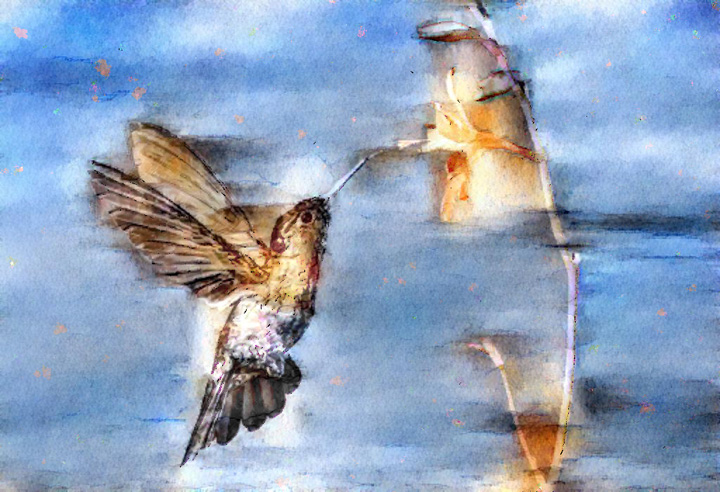}\\[0.3ex]
    \includegraphics[width=\textwidth]{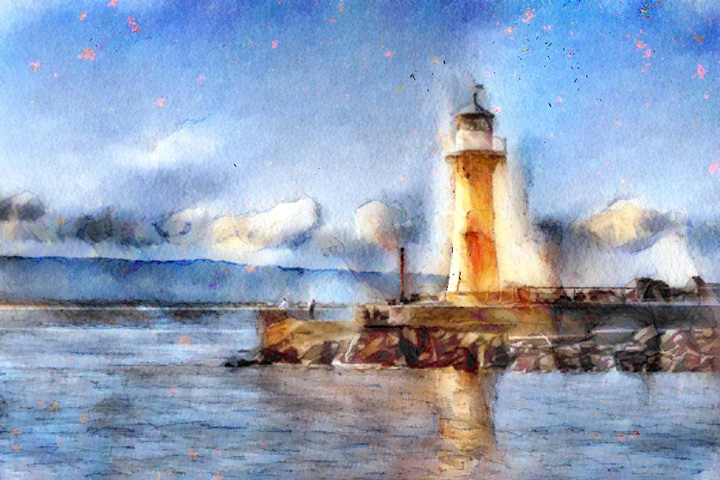}\\[0.3ex]
    \includegraphics[width=\textwidth]{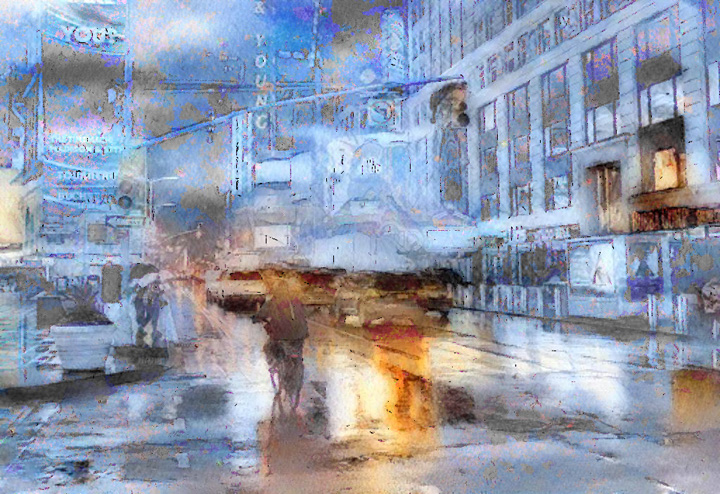}\\[0.3ex]
    \includegraphics[width=\textwidth]{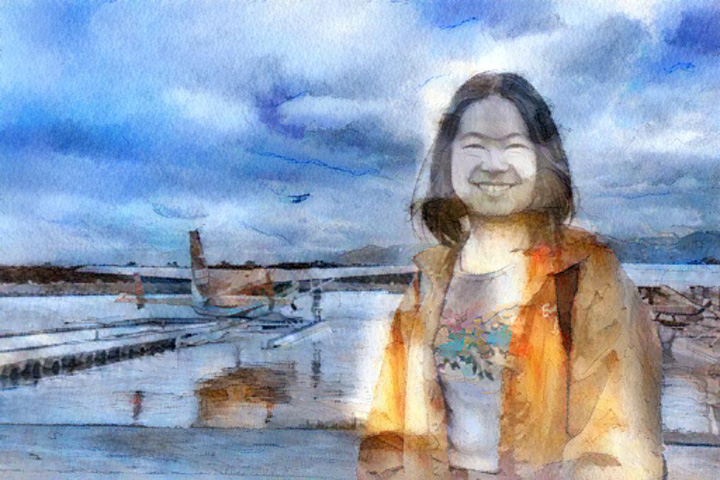}\\[0.3ex]
    \includegraphics[width=\textwidth]{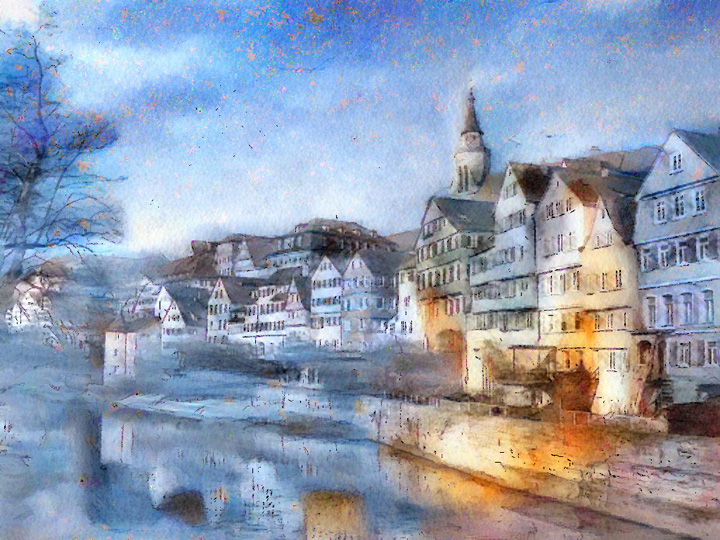}\\[0.3ex]
\end{subfigure}
\begin{subfigure}[b]{0.24\textwidth}
    \includegraphics[width=\textwidth,height=0.75\textwidth]{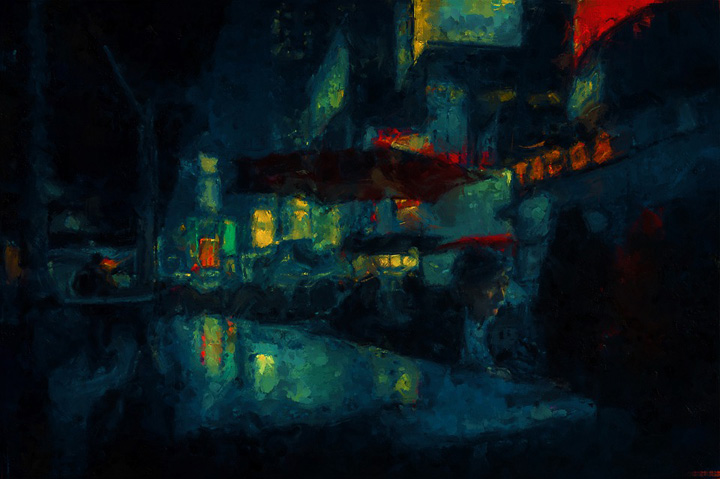}\\[0.3ex]
    \includegraphics[width=\textwidth]{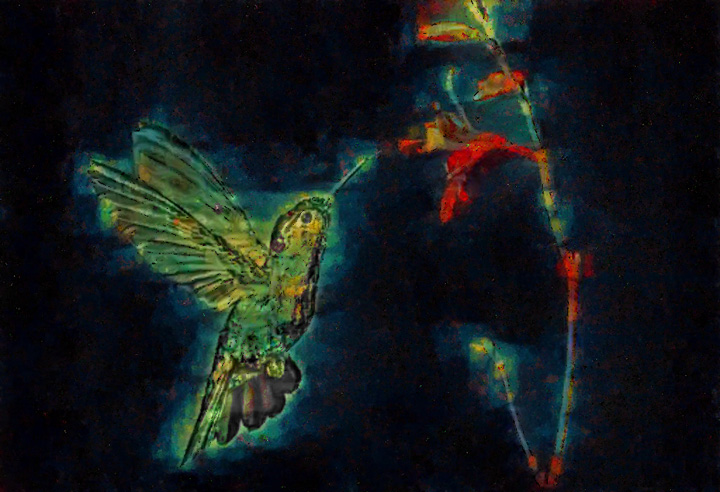}\\[0.3ex]
    \includegraphics[width=\textwidth]{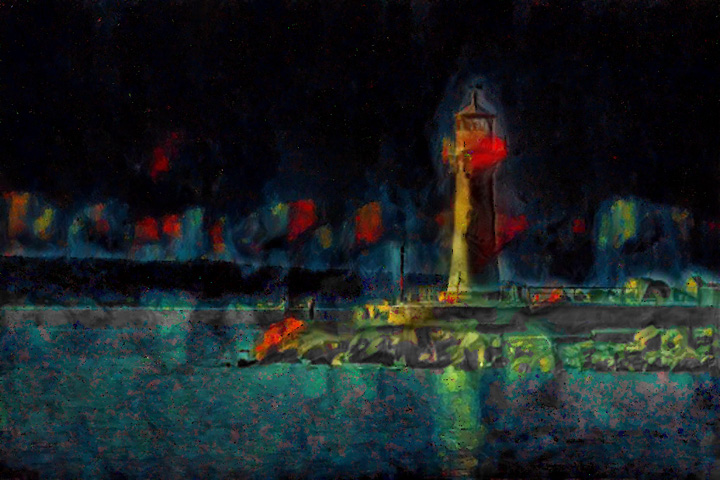}\\[0.3ex]
    \includegraphics[width=\textwidth]{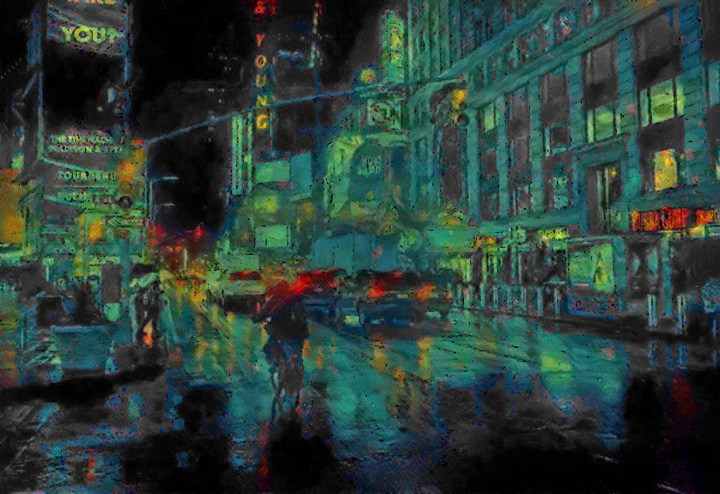}\\[0.3ex]
    \includegraphics[width=\textwidth]{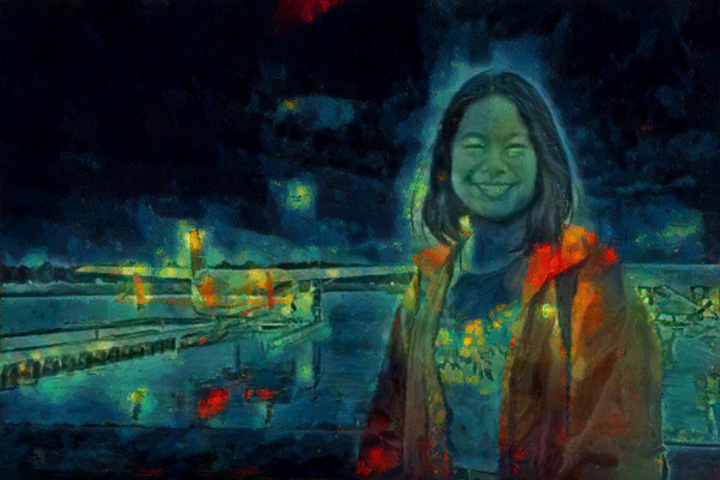}\\[0.3ex]
    \includegraphics[width=\textwidth]{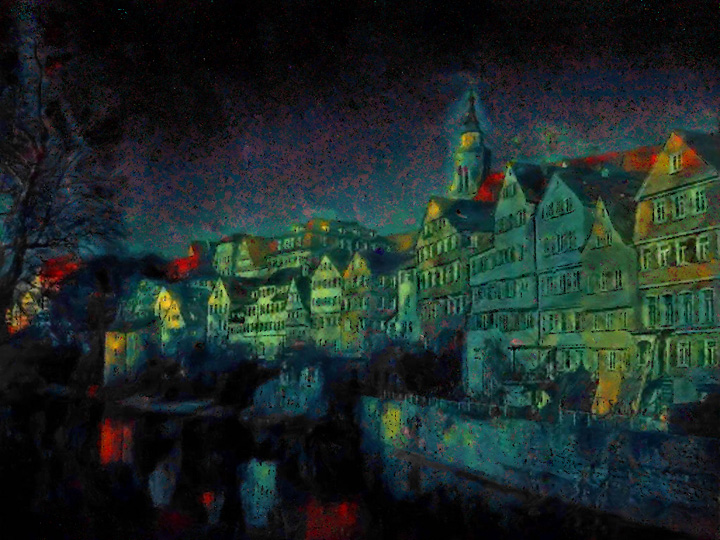}\\[0.3ex]
\end{subfigure}
\begin{subfigure}[b]{0.24\textwidth}
    \includegraphics[width=\textwidth,height=0.75\textwidth]{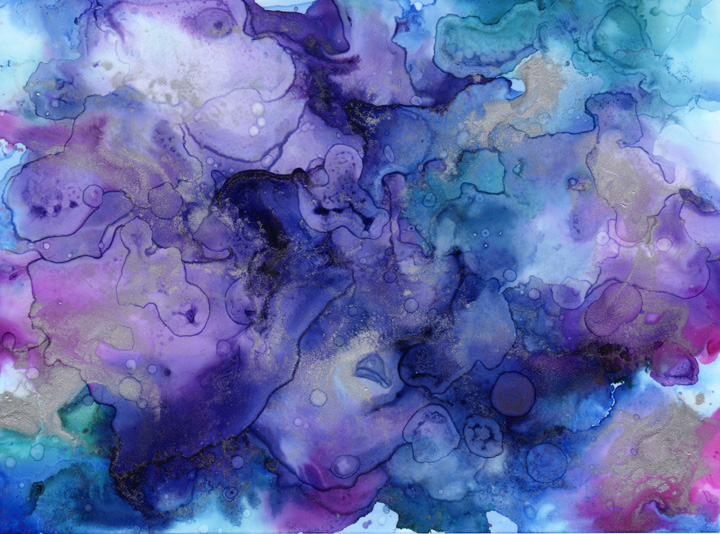}\\[0.3ex]
    \includegraphics[width=\textwidth]{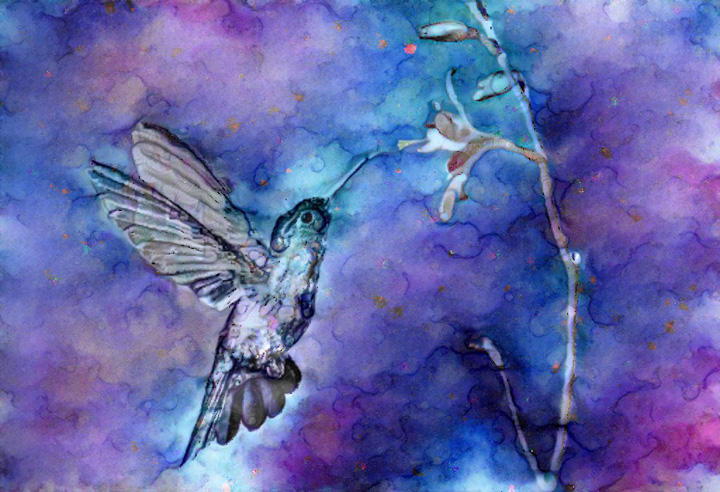}\\[0.3ex]
    \includegraphics[width=\textwidth]{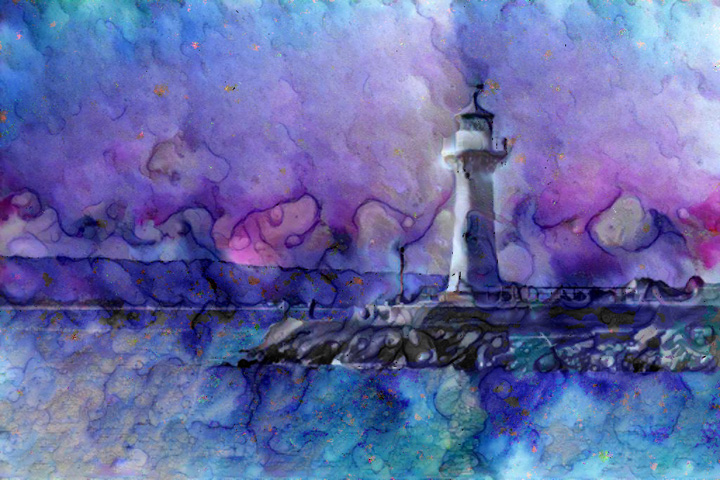}\\[0.3ex]
    \includegraphics[width=\textwidth]{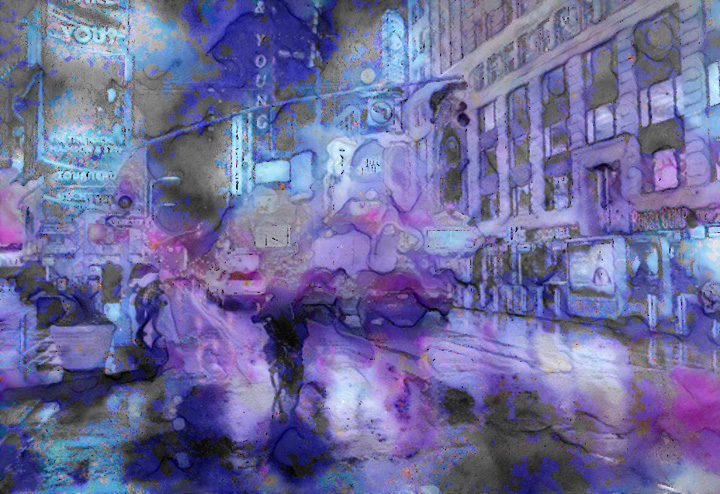}\\[0.3ex]
    \includegraphics[width=\textwidth]{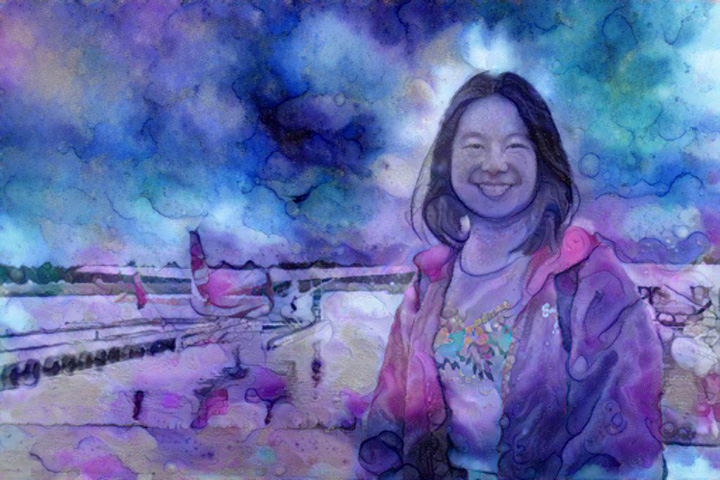}\\[0.3ex]
    \includegraphics[width=\textwidth]{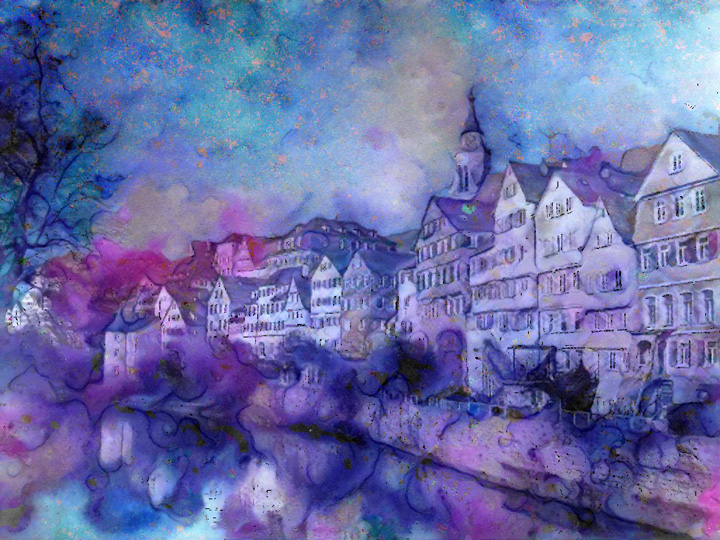}\\[0.3ex]
\end{subfigure}
\caption{\small{Additional results for style transfer optimization. Our parametric style transfer can match a wide variety of styles and remains editable after optimization (not performed here). While the presented parameter smoothing schedule removes most artefacts stemming from parameters falling into local optima, some can remain in the final image (best seen zoomed in). These can be easily removed in a final, manual parameter editing step, as shown in the supplemental video.}}
\end{figure}

\FloatBarrier
\clearpage

\subsection{Image Translation on APDrawing}\label{sup:sec:AdditionalResults:apdrawing}

\begin{figure*}\vspace{-0.5cm}
\centering

\begin{subfigure}[b]{0.23\textwidth}
    \includegraphics[width=\textwidth]{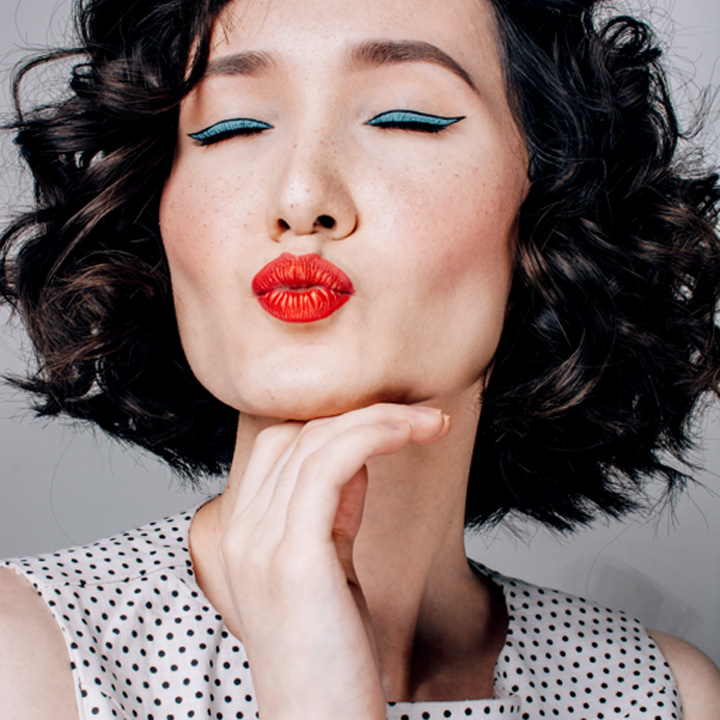}
    \includegraphics[width=\textwidth]{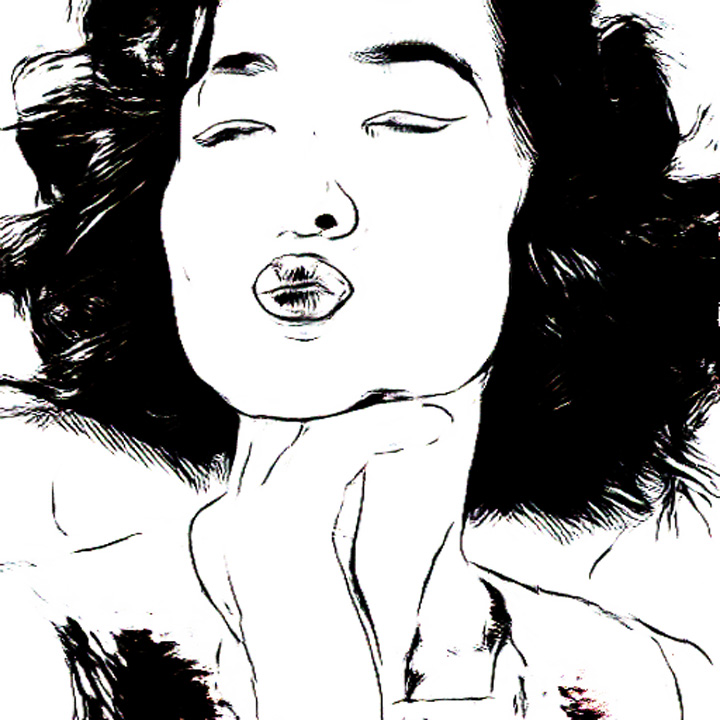}
\end{subfigure}
\begin{subfigure}[b]{0.23\textwidth}
    \includegraphics[width=\textwidth]{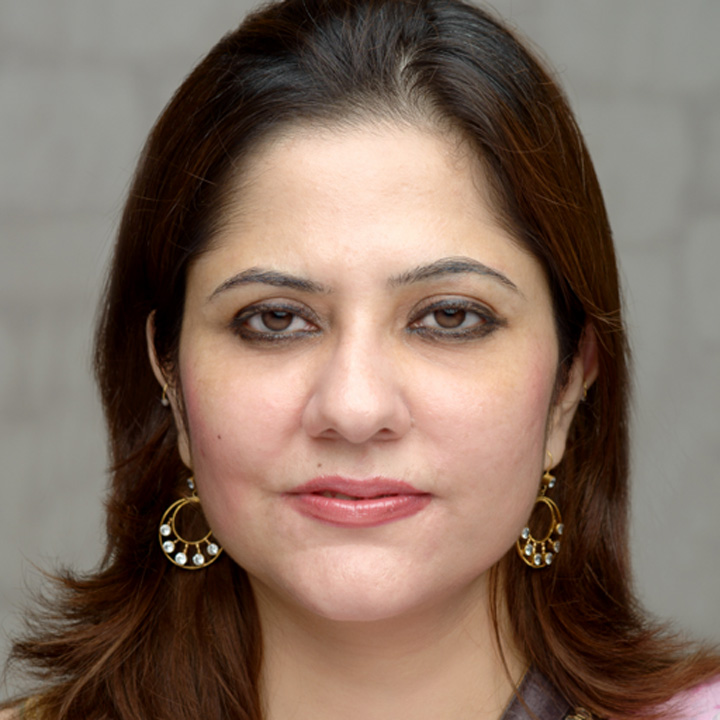}
    \includegraphics[width=\textwidth]{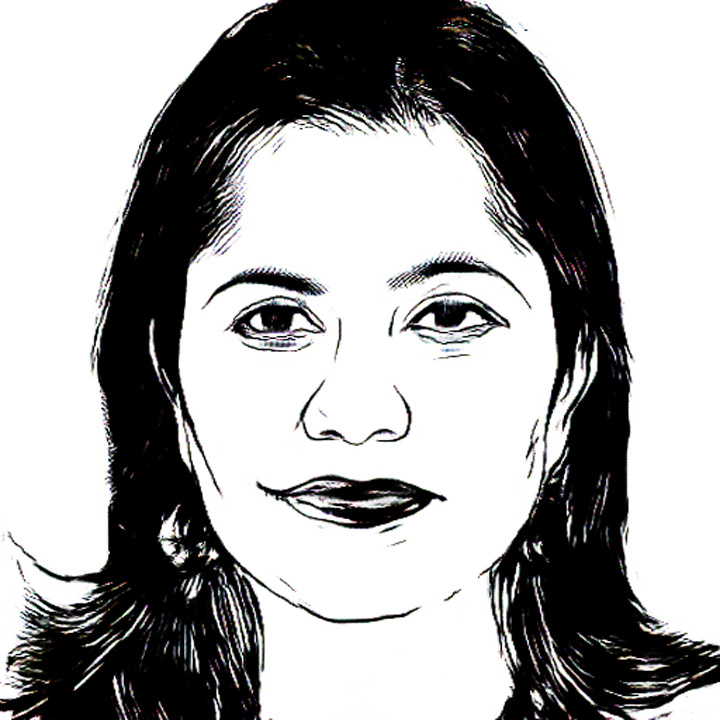}
\end{subfigure}
\begin{subfigure}[b]{0.23\textwidth}
    \includegraphics[width=\textwidth]{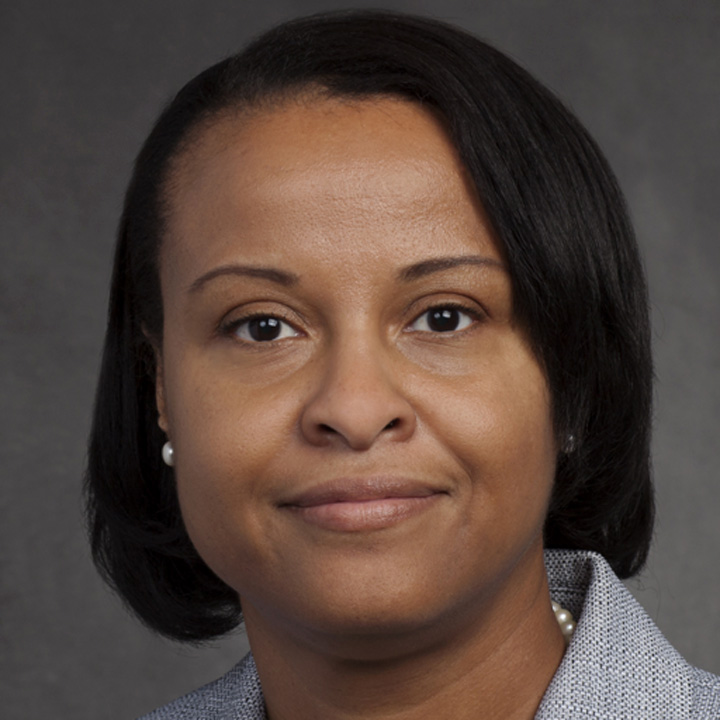}
    \includegraphics[width=\textwidth]{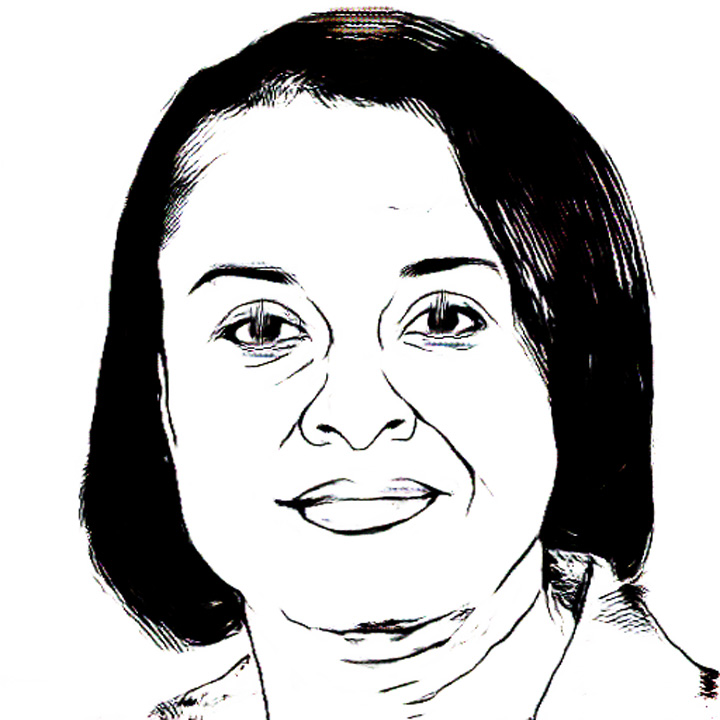}
\end{subfigure}
\begin{subfigure}[b]{0.23\textwidth}
    \includegraphics[width=\textwidth]{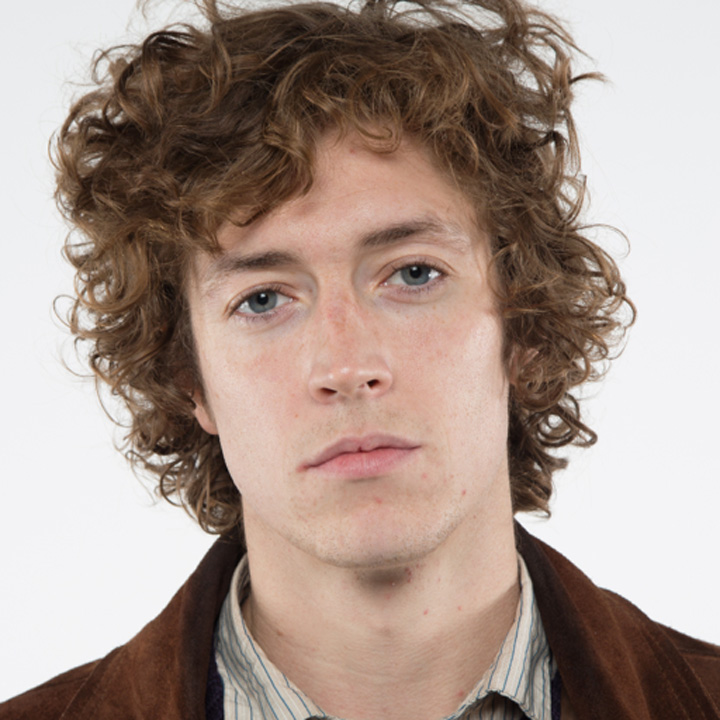}
    \includegraphics[width=\textwidth]{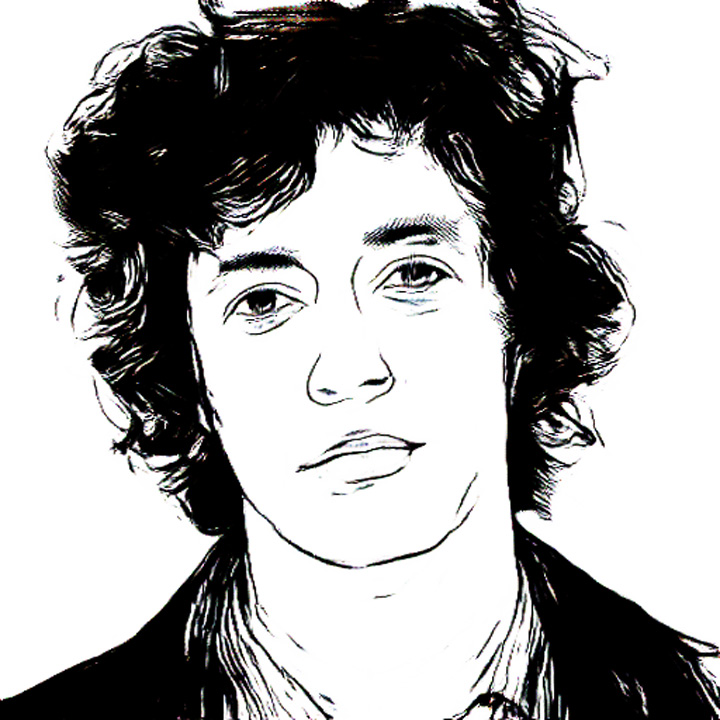}
\end{subfigure}\\[1ex]
\begin{subfigure}[b]{0.23\textwidth}
    \includegraphics[width=\textwidth]{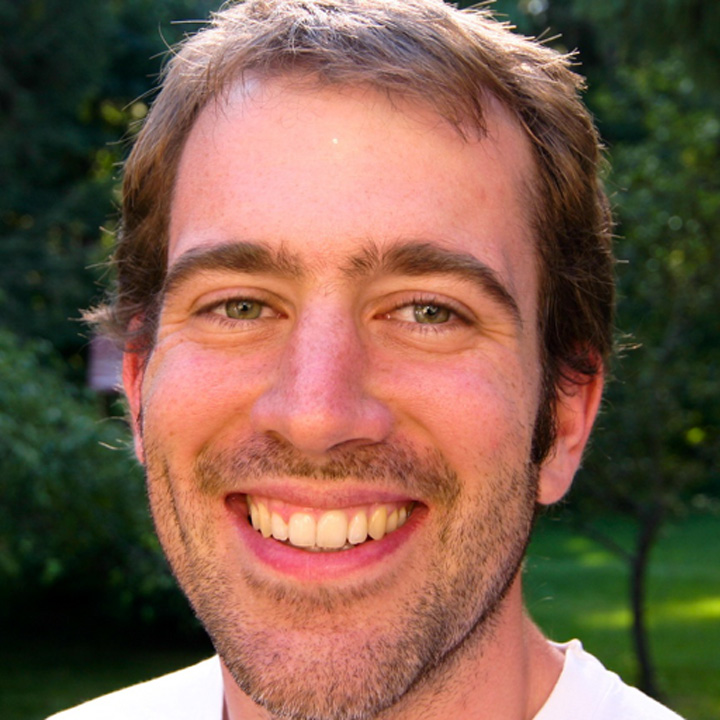}
    \includegraphics[width=\textwidth]{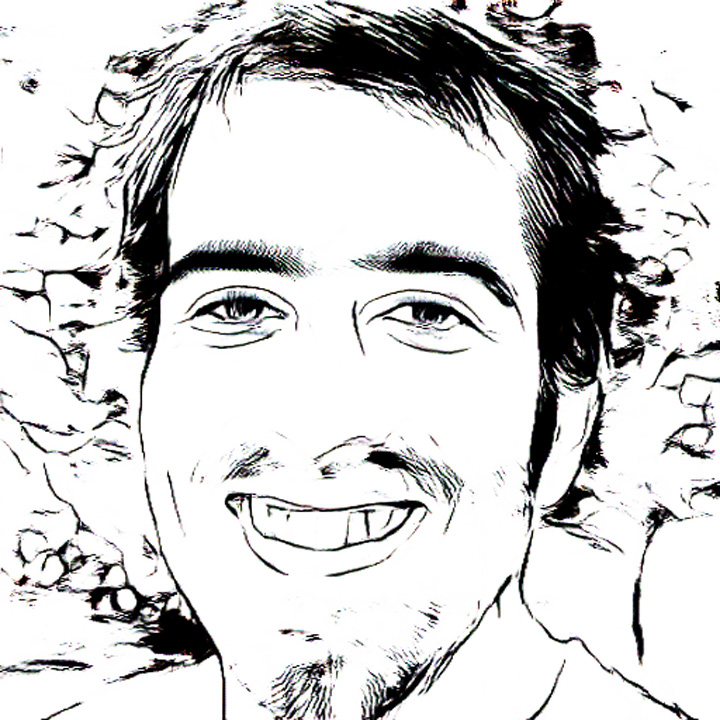}
\end{subfigure}
\begin{subfigure}[b]{0.23\textwidth}
    \includegraphics[width=\textwidth]{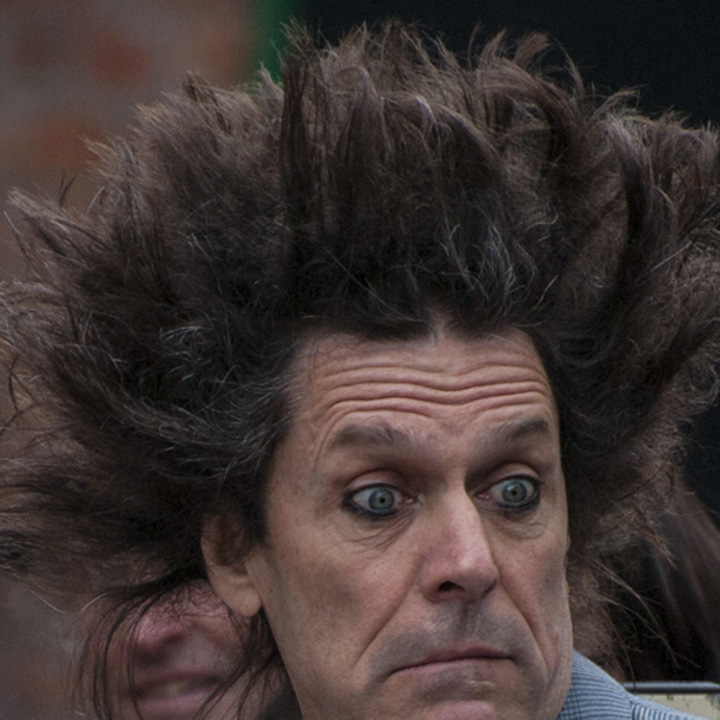}
    \includegraphics[width=\textwidth]{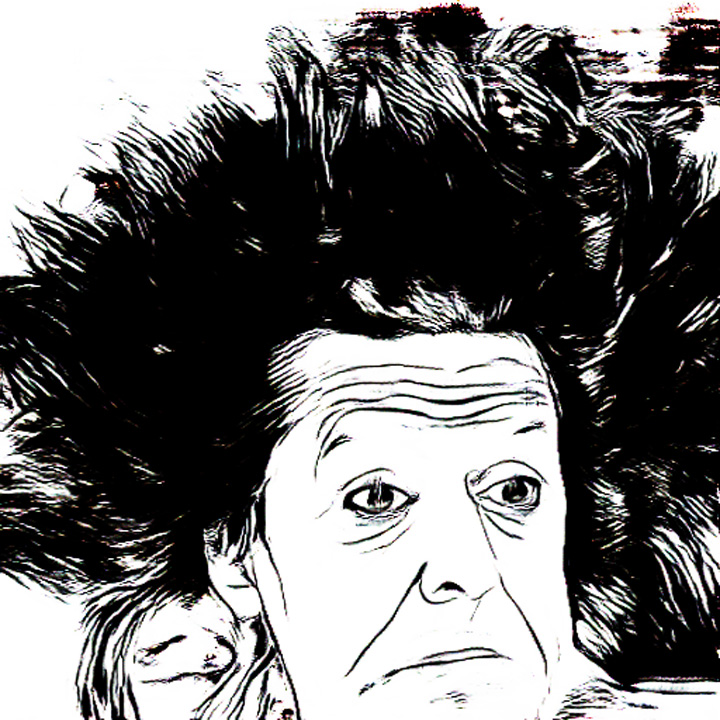}
\end{subfigure}
\begin{subfigure}[b]{0.23\textwidth}
    \includegraphics[width=\textwidth]{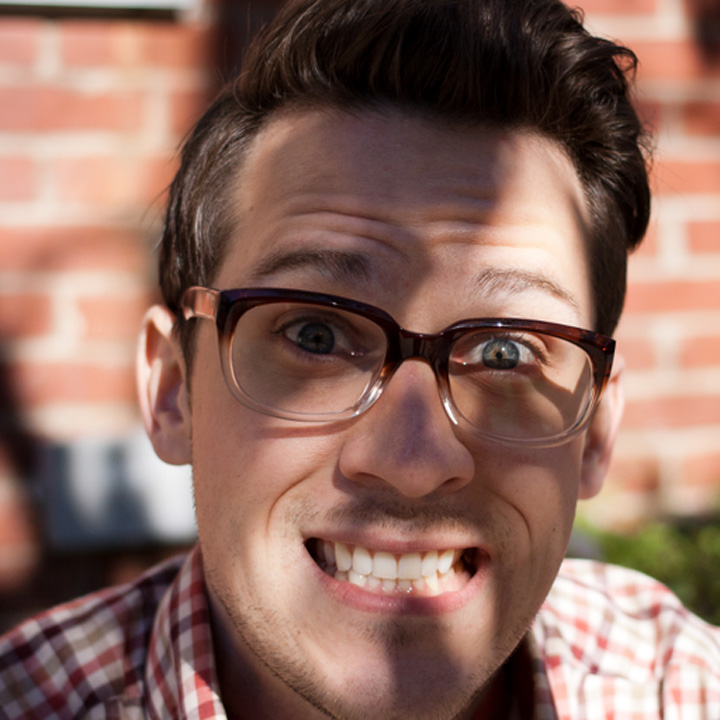}
    \includegraphics[width=\textwidth]{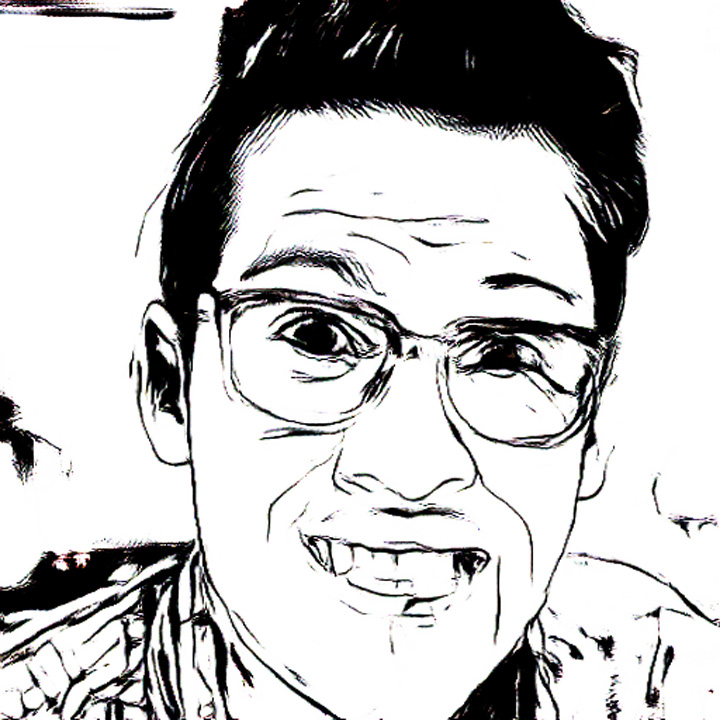}
\end{subfigure}
\begin{subfigure}[b]{0.23\textwidth}
    \includegraphics[width=\textwidth]{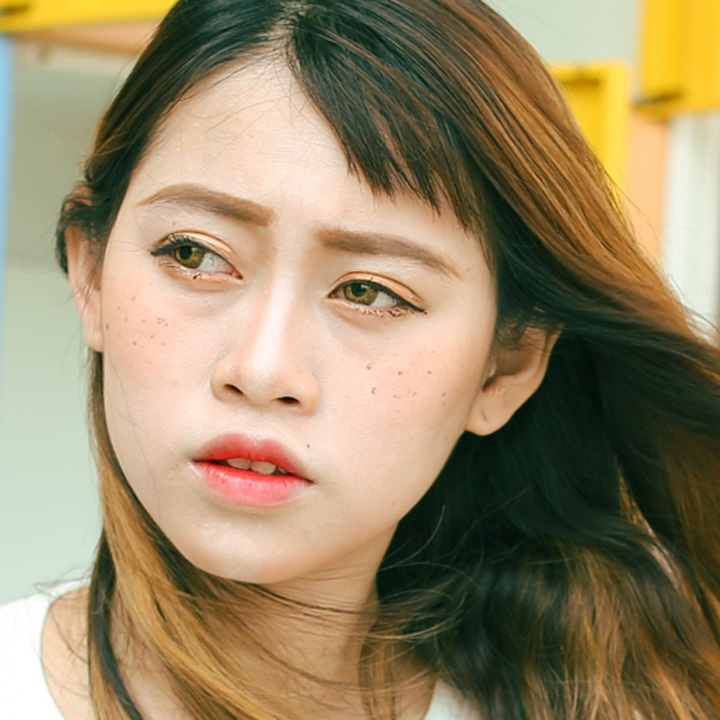}
    \includegraphics[width=\textwidth]{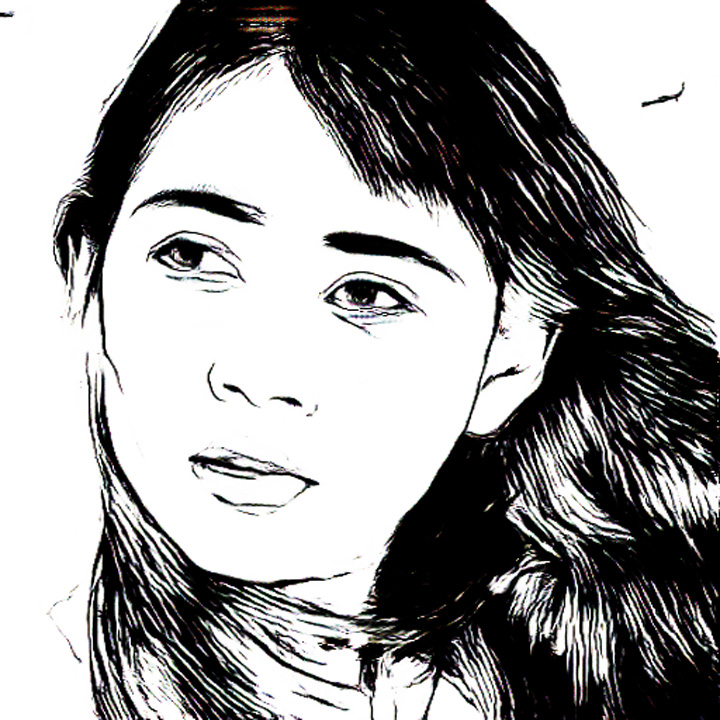}
\end{subfigure}
\vspace{0.2cm}
\caption{Our model generalizes to example images\protect\footnotemark drawn from the \ac{NPR} benchmark \cite{rosin2020nprportrait}. Our method retains only salient lines in the face, abstracting irrelevant details. Furthermore, lines generally flow consistently without unnatural discontinuities.  Our model has been trained on portraits with uniform background only, thus prediction on images with non-uniform backgrounds (\eg images in bottom row) may lead to background artefacts in the stylization.}
\label{sup:fig:apdrawing_npr}
\end{figure*}

\footnotetext{We show only images that are not part of APDrawing trainset}

\FloatBarrier
\clearpage

\subsection{Effect Variants}\label{sup:sec:AdditionalResults:effectpresets}

\begin{figure*}[!h]
\vspace{-5mm}\centering  
\begin{subfigure}[b]{0.28\textwidth}
    \includegraphics[width=\textwidth]{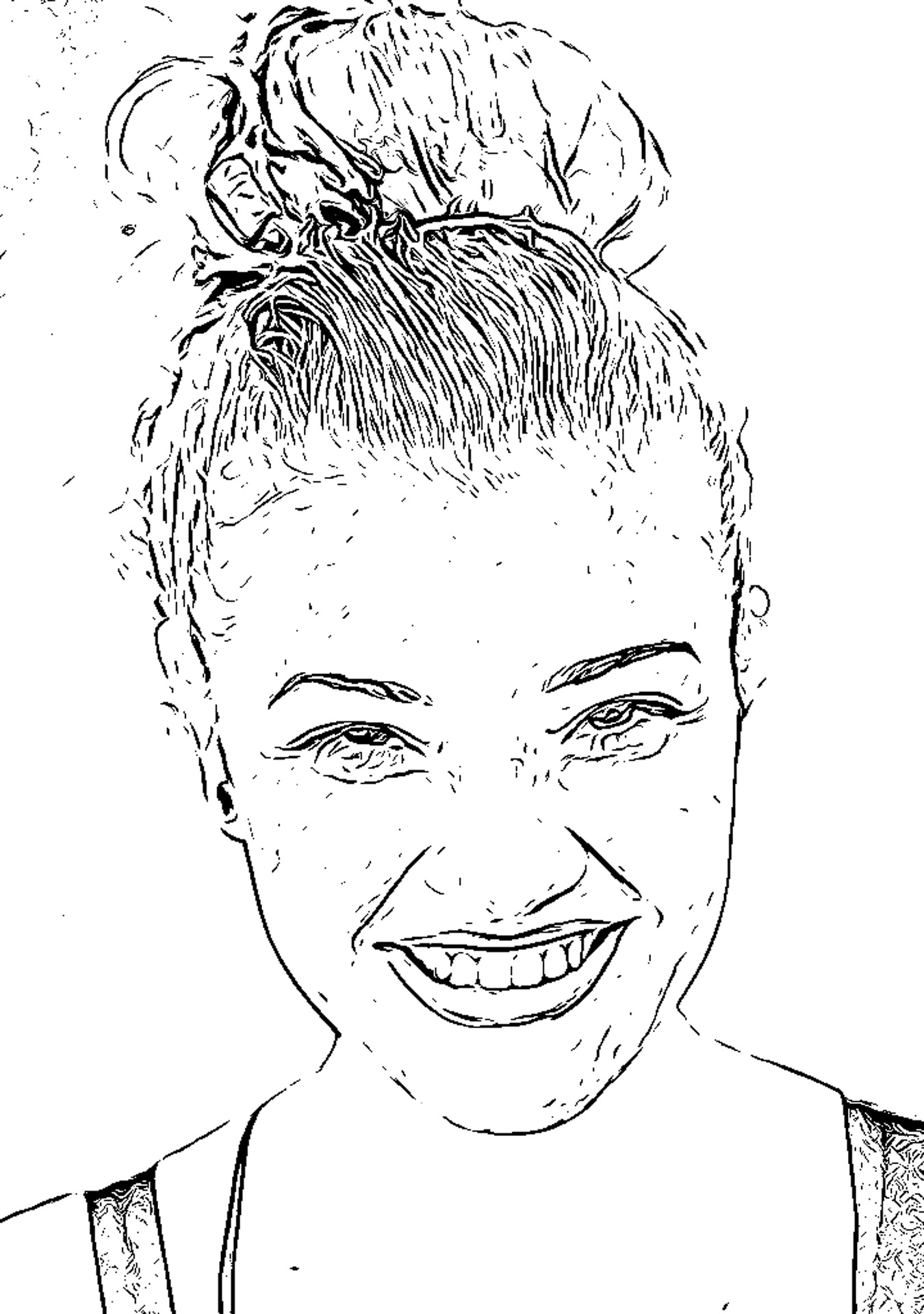}
    \caption{\ac{XDoG} variant A}
\end{subfigure}       
\begin{subfigure}[b]{0.28\textwidth}
    \includegraphics[width=\textwidth]{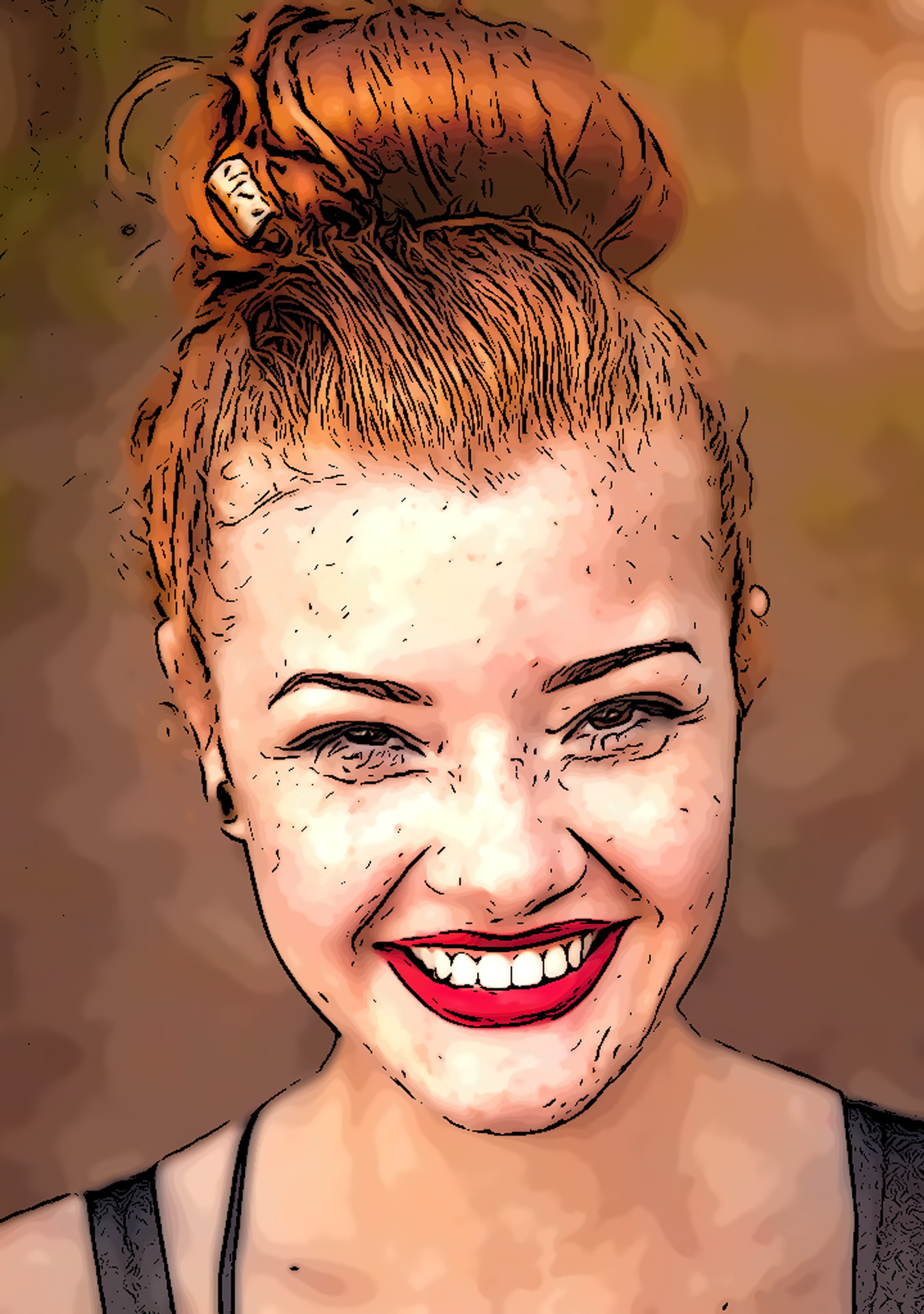}
    \caption{Cartoon variant A}
\end{subfigure}
\begin{subfigure}[b]{0.28\textwidth}
    \includegraphics[width=\textwidth]{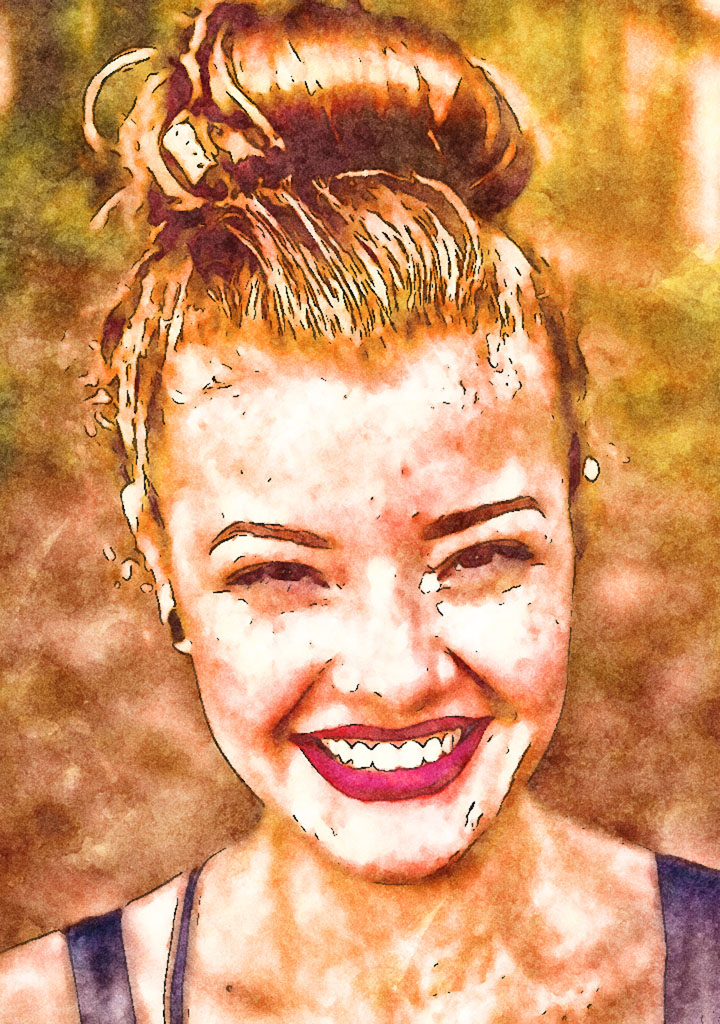}
    \caption{Watercolor variant A}
\end{subfigure}            

\begin{subfigure}[b]{0.28\textwidth}
    \includegraphics[width=\textwidth]{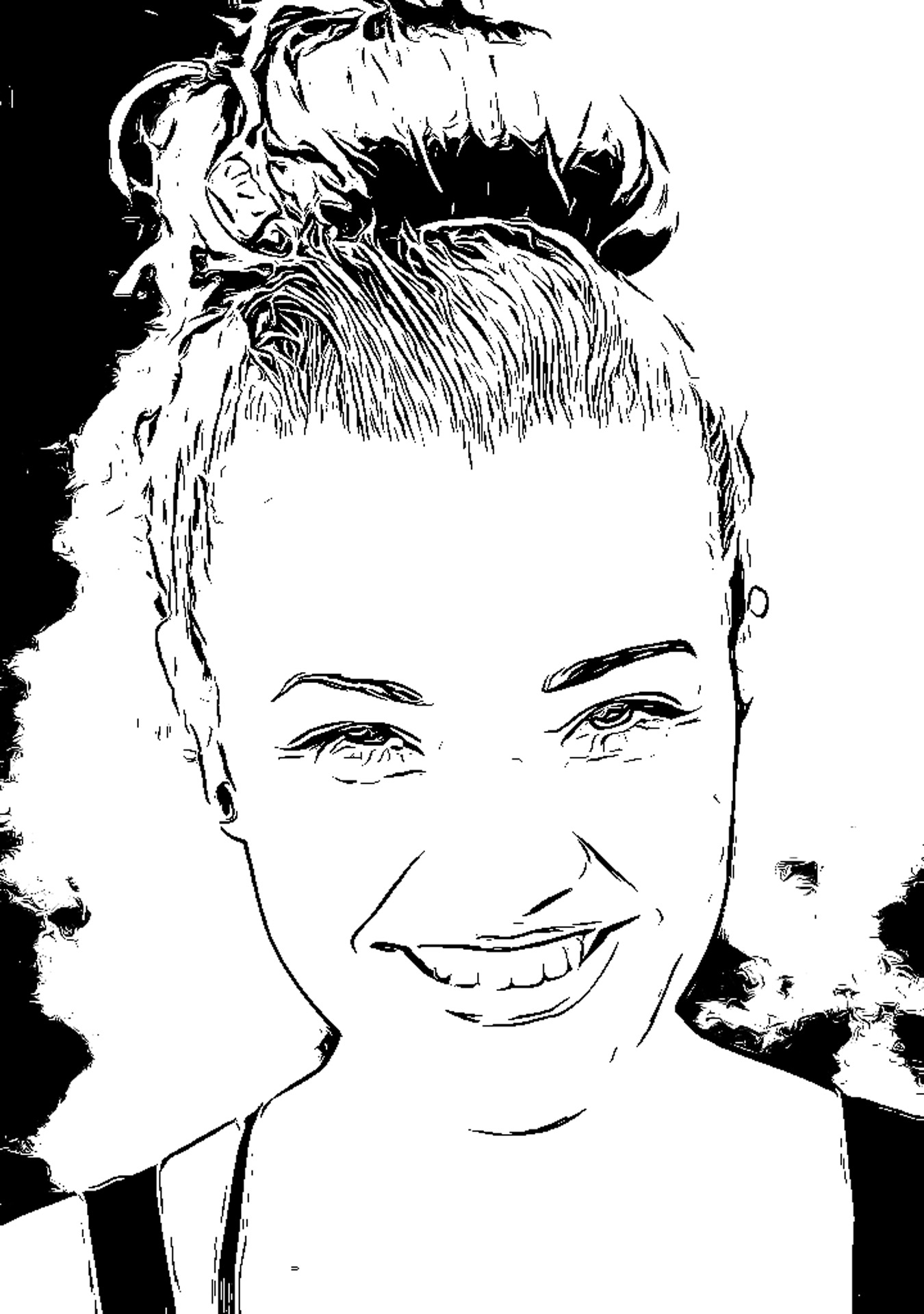}
    \caption{\ac{XDoG} variant B}
\end{subfigure}
\begin{subfigure}[b]{0.28\textwidth}
    \includegraphics[width=\textwidth]{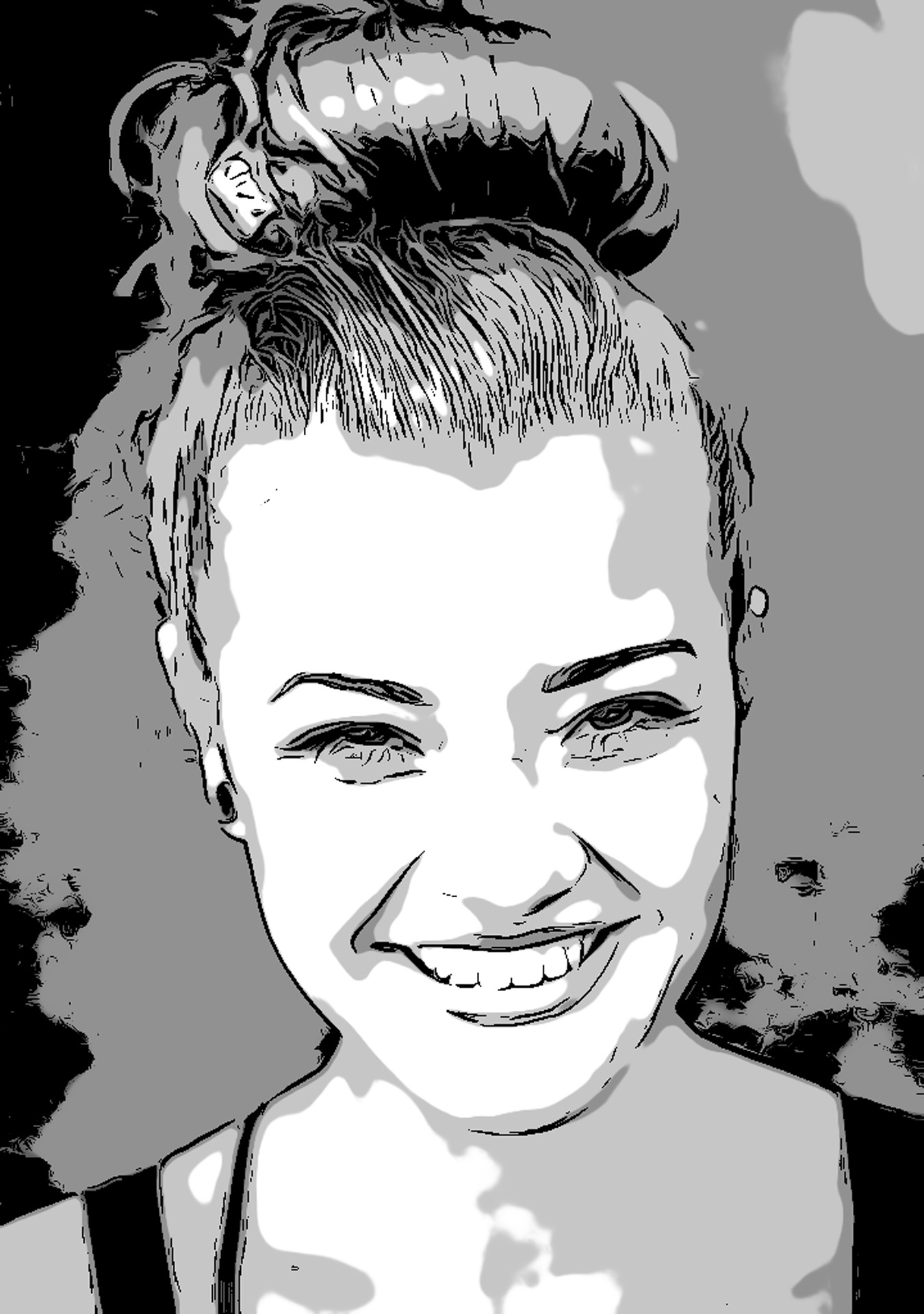}
    \caption{Cartoon variant B}
\end{subfigure}            
\begin{subfigure}[b]{0.28\textwidth}
    \includegraphics[width=\textwidth]{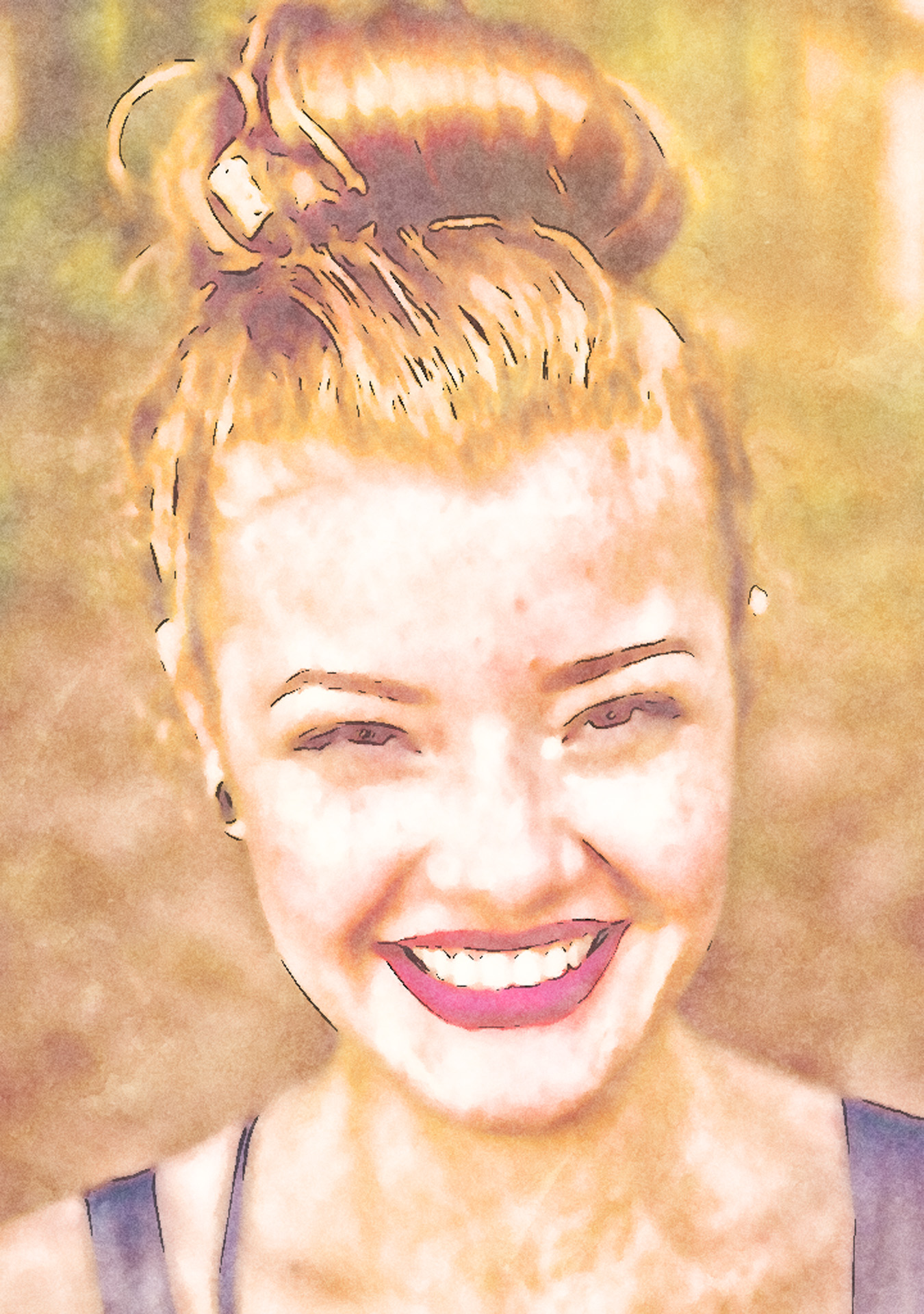}
    \caption{Watercolor variant B}
\end{subfigure}  

\begin{subfigure}[b]{0.28\textwidth}
    \includegraphics[width=\textwidth]{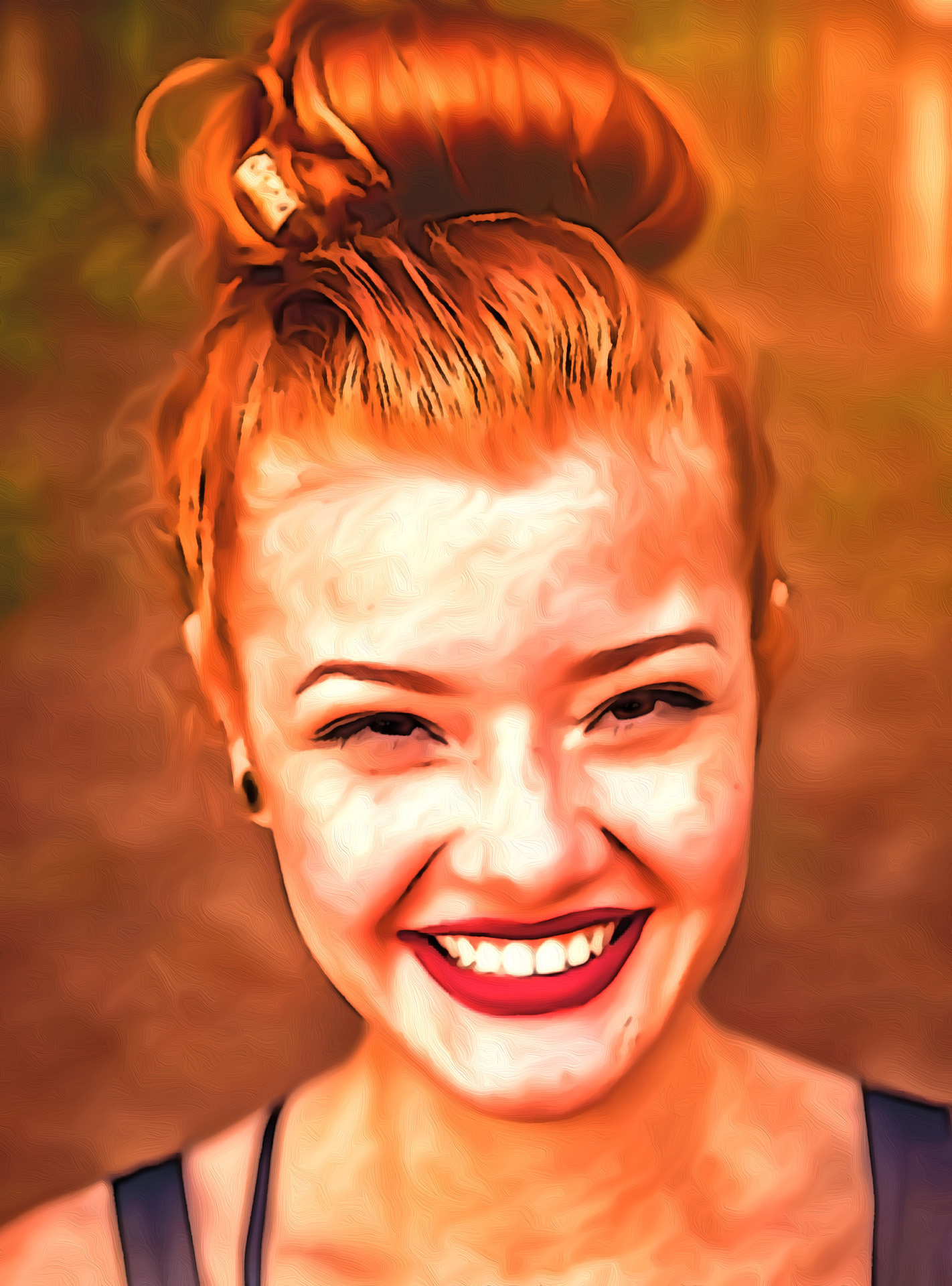}
    \caption{Oilpaint variant A}
\end{subfigure}
\begin{subfigure}[b]{0.28\textwidth}
    \includegraphics[width=\textwidth]{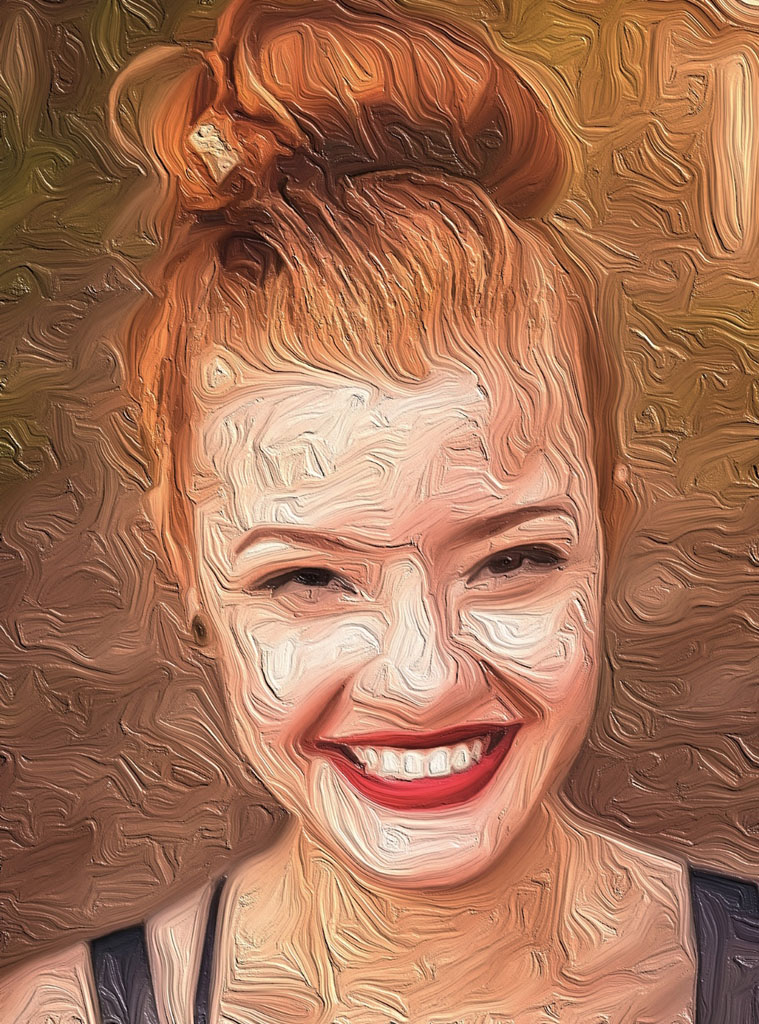}
    \caption{Oilpaint variant B}
\end{subfigure}
\caption{A selection of global parametrization variants of our differentiable effects are shown. These represent default states of the effect without any parameter learning applied.}
\label{sup:fig:results_differentiable_effects}
\end{figure*}

\end{document}